\documentclass[3p]{elsarticle}

\makeatletter
\def\ps@pprintTitle{%
	\let\@oddhead\@empty
	\let\@evenhead\@empty
	\def\@oddfoot{\centerline{\thepage}}%
	\let\@evenfoot\@oddfoot}
\makeatother

\usepackage{epsfig} 
\usepackage{hyperref}
\usepackage{amsmath}
\usepackage{amssymb}
\usepackage{float}
\usepackage{enumitem}
\usepackage[para,online,flushleft]{threeparttable}
\usepackage{algorithm}
\usepackage{algpseudocode}
\usepackage{dirtytalk}
\usepackage{mathtools}
\usepackage{xcolor}
\usepackage{bm}
\usepackage[normalem]{ulem}
\usepackage{booktabs}
\usepackage{dcolumn}
\usepackage{multirow}
\usepackage{graphicx}
\usepackage{cleveref}

\newcommand{\R}{\mathbb{R}}
\newcommand{\ssymbol}[1]{^{\@fnsymbol{#1}}}
\newcommand{\norm}[1]{\|#1\|}
\DeclareMathOperator*{\argminB}{argmin}   
\newtheorem{remark}{Remark}

\begin{document}
	
	\begin{frontmatter}
		
		\title{A Priori Denoising Strategies for Sparse Identification of Nonlinear Dynamical Systems: A Comparative Study}
		
		\author[colorado1]{Alexandre Cortiella}
		\ead{alexandre.cortiella@colorado.edu}
		\author[colorado1]{Kwang-Chun Park}
		\ead{kcpark@colorado.edu}
		\author[colorado1]{Alireza Doostan\corref{mycorrespondingauthor}}
		\cortext[mycorrespondingauthor]{Corresponding author}
		\ead{alireza.doostan@colorado.edu}
		\address[colorado1]{Smead Aerospace Engineering Sciences Department, University of Colorado, Boulder, CO 80309, USA}
		
	\begin{abstract}
		In recent years, identification of nonlinear dynamical systems from data has become increasingly popular. Sparse regression approaches, such as Sparse Identification of Nonlinear Dynamics (SINDy), fostered the development of novel governing equation identification algorithms assuming the state variables are known {\it a priori} and the governing equations lend themselves to sparse, linear expansions in a (nonlinear) basis of the state variables. In the context of the identification of governing equations of nonlinear dynamical systems, one faces the problem of identifiability of model parameters when state measurements are corrupted by noise. Measurement noise affects the stability of the recovery process yielding incorrect sparsity patterns and inaccurate estimation of coefficients of the governing equations. In this work, we investigate and compare the performance of several local and global smoothing techniques to {\it a priori} denoise the state measurements and numerically estimate the state time-derivatives to improve the accuracy and robustness of two sparse regression methods to recover governing equations: Sequentially Thresholded Least Squares (STLS) and Weighted Basis Pursuit Denoising (WBPDN) algorithms. We empirically show that, in general, global methods, which use the entire measurement data set, outperform local methods, which employ a neighboring data subset around a local point. We additionally compare Generalized Cross Validation (GCV) and Pareto curve criteria as model selection techniques to automatically estimate near optimal tuning parameters, and conclude that Pareto curves yield better results. The performance of the denoising strategies and sparse regression methods is empirically evaluated through well-known benchmark problems of nonlinear dynamical systems.
		
	\end{abstract}
	


\end{frontmatter}

	\section{Introduction}
	\label{sec:introduction}
	
	Engineers rely heavily on mathematical models of real systems. In the design process, engineers seek the best models that describe the physical system they seek to design and optimize. However, physical processes are not only highly complex and inadequately understood, but also the sources of noise are critically dependent upon the nature of nonlinearities. System identification allows for any system, however complex or obscure, to be modeled solely from measurements.
	
	Recent advances in data acquisition systems along with modern data science techniques have fostered the  development  of  accurate  data-driven  approaches~\cite{Fassois2007} for modeling of complex nonlinear dynamical systems with the aim of understanding their behavior and predicting future states. A current trend to identify dynamical system models relies on approximating the governing equations in an over-complete basis of the state variables and eliminate the expansion terms that do not {\color{black}carry significant signal energy}. Examples of this approach include polynomial NARMAX~\cite{Leontaritis1985}, symbolic polynomial regression~\cite{Schmidt81,Bongard2007}, and sparse (polynomial) regression~\cite{Wang2011,Brunton2016} dubbed SINDy in~\cite{Brunton2016}.
	
	System identification via sparse (polynomial) regression employs techniques from compressed sensing -- specifically regularization via sparsity promoting norms, such as $\ell_0$- and $\ell_1$-norms, to identify an {\it a priori} unknown subset of the basis describing the dynamics. The identified dynamics may then be further analyzed to understand the physical behavior of the dynamical system, and can be integrated in time to predict future states of the system. Stage-wise regression~\cite{goldberger1961stepwise}, Matching Pursuits~\cite{mallat1993matching,pati1993orthogonal,needell2009cosamp}, Least Absolute Shrinkage and Selection Operator (LASSO) \cite{tibshirani1996regression}, Least Angle Regression (LARS) \cite{Efron2004}, Sequentially Thresholded Least Squares (STLS)~\cite{Brunton2016}, and Basis Pursuit Denoising (BPDN) \cite{Chen2001} are some sparsity promoting algorithms that may be used for model recovery. For the interested reader, reference~\cite{bertsimas2020sparse} provides a thorough comparison of different state-of-the-art sparse regression algorithms. For high-dimensional systems and large data sets, sparse regression methods often suffer from the \textit{curse of dimensionality} since they can be prohibitively expensive in terms of computational cost and storage capacity. Even though this work explores tractable low-dimensional systems, several methods have been proposed to mitigate the curse of dimensionality for high-dimensional systems such as coordinate reduction via auto-encoders~\cite{champion2019data, fukami2021sparse}, or efficient array manipulation via low-rank tensor decompositions~\cite{gelss2019multidimensional}.
	
	In practice, analysts only have access to state measurements from real sensors that are invariably corrupted by noise. Although sparse regression methods perform well for low levels of measurement noise, the sparsity pattern and accuracy of the recovered coefficients are usually very sensitive to higher noise levels. {\color{black}Several techniques have been proposed to mitigate the effect of noise for recovering governing equations from state measurements. Denoising strategies for governing equation recovery can be divided into two categories: simultaneous denoising and equation recovery, and {\it a priori} state denoising. The former strategy generally involves nonlinear optimization procedures where the loss function usually contains a measurement data loss which measures mismatch between state measurements and state predictions,  and a state derivative error which measures the deviation between the state time derivatives and the estimated governing equation. Some of the simultaneous methods employ neural networks \cite{rudy2019deep} and automatic differentiation \cite{kaheman2020automatic} to decompose the noisy measurements into the deterministic and random noisy components to recover the governing equations more accurately. Others use smoothing splines~\cite{sun2021physics} and deep neural networks~\cite{chen2021physics} as a surrogate model to approximate the state variables for the data loss which then can be differentiated to obtain state time derivatives required for the state derivative error. However, simultaneous methods often require solving challenging nonlinear optimization problems and fine tuning of the hyperparameters of the algorithm. The recent {\it a priori} denoising strategy, adopted in this article,  focuses on estimating accurate and smooth state measurements and their derivatives as a pre-processing step for sparse regression algorithms to produce significant improvements in the accuracy of the identified dynamics. We adopt the SINDy framework, as proposed in \cite{Brunton2016}, and compare different approaches to denoise the state measurements and estimate accurate time-derivatives with no prior knowledge of the process that corrupted the measurements. Typically, numerical differentiation is carried out using finite differences. However, since data measured by digital equipment is inherently discrete and contains noise due to the device properties and quantization effects, naive finite differences produce inaccurate derivative estimates~\cite{cullum1971numerical,lu2006numerical,chartrand2011numerical}. A common approach to obtain derivative estimates is to approximate the underlying signal by a \textit{fitting} function from the measured data. One then hopes that the derivative found from this approximation is accurate enough for noise filtering.}
	
	There exists extensive work on nonparametric regression smoothing and numerical differentiation of noisy data \cite{green1993nonparametric, bowman1997applied, aydin2007comparison, chartrand2011numerical, knowles2014methods, ahnert2007numerical,lubansky2006general}, {\color{black} also referred to as denoising techniques in this article}. The underlying assumption of non-parametric smoothing techniques is that the signal is smooth and the derivatives up to certain order exist. Then, to filter out undesired noisy components from the signal, one poses the smoothing task as a trade-off between the \textit{accuracy} and \textit{smoothness} of the filtered solution. Denoising strategies can be split into local and global techniques. Local denoising fit a low-degree polynomial to the nearest neighbors of each data point with a weight proportional to inverse distance from the neighbor to the data point~\cite{Knowles2014}. Global denoising use the entire data points to estimate a denoised version of the original data. Some of the common denoising techniques include local kernel smoothers \cite{fan1996local, loader2006local}, smoothing splines \cite{de1978practical}, Tikhonov regularization \cite{tikhonov2013numerical,hodrick1997postwar}, total variation regularization \cite{rudin1992nonlinear, chambolle2004algorithm}, Fourier-based smoothing \cite{ahnert2007numerical}\cite[Chapter~5]{brunton2019data}, wavelet-based denoising \cite{mallat1989theory,donoho1995adapting}, or singular spectrum analysis \cite{elsner2013singular, golyandina2013singular}. More recent trends use deep neural networks for signal denoising~ \cite{zhu2019seismic,antczak2018deep,fan2020vibration}. However, these methods often require large amounts of data and several trajectory realizations, making them unsuitable for the present application, where we focus on a single trajectory. This article compares the performance of Savitzky-Golay filter and LOWESS as local methods; and, smoothing splines, Tikhonov smoother, and $\ell_1$-trend filtering as global smoothers.
	
	A key challenge of smoothing methods is the selection of the smoothing hyperparameter that controls the fidelity of the {\color{black}filtered solution, relative to the original, noisy measurements, and its smoothness after removing undesired (high-frequency) noise components.} In the statistics literature, the compromise between accuracy and smoothness is often referred as the \textit{bias-variance tradeoff} \cite{hastie2009elements}. In general, the filtered solution is highly sensitive to the smoothing hyperparameter, and a suitable selection criterion is of paramount importance. Several methods have been proposed to select appropriate smoothing hyperparameters such as AIC~\cite{akaike1974new} and BIC~\cite{schwarz1978estimating} criteria, Mallow's $C_p$~\cite{mallows2000some}, discrepancy principle~\cite{morozov1966solution, morozov2012methods}, cross-validation~\cite{stone1974cross,golub1979generalized} and {\color{black}the Pareto curve (also L-curve) criterion} ~\cite{Hansen1992Lcurve, Hansen1999}, among others. An ideal model selection criterion should approximate the solution as close as possible to the unknown dynamical process that generates the \textit{exact} (i.e. noiseless) data. Hence, the selection of both the smoothing algorithm and the model selection criterion are inherently linked and should be studied simultaneously to select a smoothing parameter that is \textit{near optimal} with respect to a suitable error measure. Due to their favorable properties, this article focuses on two robust model selection criteria: generalized cross-validation and Pareto curve criterion.
	\subsection*{Contribution of this work}
	\label{sec:contribution}
	The purpose of this article is to present an overview of major techniques for measurement denoising and numerical differentiation as a pre-processing step to recover the governing equations of nonlinear dynamical systems. Our aim is to provide an empirical assessment of several best-performing filtering techniques for system identification rather than delivering a thorough theoretical analysis.
	
	In this work, we study and assess the performance of local and global filtering regression techniques to denoise state measurements and accurately estimate state time-derivatives. To the best of our knowledge, no prior work has been done on both denoising and numerical differentiation tailored to data-driven governing equation recovery. We extend the findings in \cite{d1992comparison, ahnert2007numerical, chartrand2011numerical, knowles2014methods, stickel2010data} and apply {\color{black}smoothing techniques} to dynamical systems, where the state trajectories are assumed to be smooth on the \textit{Poincar\'{e} map}. One of the critical aspects of filtering and regularization techniques is the automatic selection of a suitable hyperparameter that controls the smoothness, for filtering, and sparsity, for equation recovery, of the solution. We also compare simple and robust model selection techniques to automatically select appropriate smoothing and regularization parameters. Although this article focuses on the differential form of sparse identification algorithms, the findings can also be used to any identification algorithm that requires accurate state measurement data and their time-derivatives. For example, the smoothed data could be fed into equation-error methods, typical in aircraft system identification~\cite{morelli2016aircraft, jategaonkar2006flight}.
	
	We begin, in the next section, by presenting a background on recovering dynamical system equations from state measurements using sparsity promoting regression techniques. In Section \ref{sec:filtering}, we present an overview of different local and global smoothing techniques to denoise state measurements and estimate accurate state time-derivatives. In Section \ref{sec:model_selection}, we introduce two hyperparameter selection techniques, namely generalized cross-validation and Pareto curve, to automatically select near optimal regularization parameters for smoothing and sparse regression methods. In Section \ref{sec:numerical_examples}, we compare the performance of the different smoothing, numerical differentiation and model selection techniques, and assess the performance of sparse regression methods to recover the governing equations of three well-known dynamical systems. Finally, in Section \ref{sec:conclusion}, we draw conclusions and discuss relevant aspects of the various smoothing methods, and offer directions for improvements of denoising techniques. 
	%
	
	\section{Problem statement and solution strategies}
	\label{sec:statement}
	
	Throughout this work, we assume that a dynamical system has the form
	\begin{equation}
		\dot{\mathbf{x}}(t) = \frac{\mathrm{d}\mathbf{x}(t)}{\mathrm{d}t} = \mathbf{f}(\mathbf{x}(t)), \quad \mathbf{x}(0) = \mathbf{x}_0,\label{eq:DynamicalSystem}
	\end{equation}
	where $\mathbf{x}(t)\in\R^n$ are the known and measurable state variables of the system at time $t \in [0,T]$ and $\mathbf{f}(\mathbf{x}(t))\colon \R^n \to \R^n$ is a state-dependent unknown  vector that describes the motion of the system. An important observation is that in many systems $\mathbf{f}(\mathbf{x}):=\mathbf{f}(\mathbf{x}(t))$ is a {\it simple} function of the state variables $\mathbf{x}:=\mathbf{x}(t)$ in that only a small set of state-dependent quantities contribute to the dynamics.
	
	Given that $\mathbf{f}(\mathbf{x})$ is unknown and following~\cite{Wang2011, Brunton2016}, we assume that each state dynamics $\dot{x}_j:=\dot{x}_j(t)$ or, equivalently, $f_j(\mathbf{x})$, $j = 1,\dots,n$, is spanned by a set of $p$ candidate nonlinear (in the state variables) basis functions $\phi_{i}(\mathbf{x})$ weighted by unknown coefficients $\xi_{ji}$, i.e.,
	\begin{equation}\label{eq:dynExpansion}
		\dot{x}_j = \sum_{i = 1}^p \xi_{ji} \phi_{i}(\mathbf{x}),\,\,\,\,\,j = 1,\dots,n.
	\end{equation}
	We assume the true dynamics may be described by only a subset of the the considered basis $\{\phi_{i}(\mathbf{x})\}$, and thus the coefficients $\xi_{ji}$ are sparse. Exploiting this sparsity in identifying the governing equations $\mathbf{f}(\mathbf{x})$ is the key idea behind SINDy algorithms~\cite{Brunton2016}. 
	
	{\color{black}To determine $\mathbf{f}(\mathbf{x})$ via Eqn.~(\ref{eq:dynExpansion}), we must numerically estimate the state time derivatives at each time-instance $t_i$ from discrete state observations $\mathbf{y}_j \in \R^m$. In practice, exact state measurements are not available and we only have access to discrete noisy state observations. In this article, we assume a \textit{signal plus noise} measurement model of the form  
		\begin{equation}
			\mathbf{y}_{j}(i) = x_j(t_i) + \epsilon_j,\,\,\,\,\,j = 1,\dots,n,\label{eq:measurement_model}
		\end{equation}
		where $i=1,\dots, m$, $m$ is the number of measurements sampled at a rate $\Delta t$, $x_j(t_i)$ is the $j$-th state variable evaluated at time instances $\{t_i\}_{i = 1}^{m}$, and we assume that the random perturbations $\epsilon_j$ are i.i.d. with zero mean $\mathbb{E}[\epsilon_j] = 0$ and constant variance $\mathbb{V}[\epsilon_j] = \sigma^2$. Hence, the discretized system given by Eqn.~(\ref{eq:dynExpansion}) may be written in matrix form as
		\begin{equation}\label{eq:matrix_main_system}
			\dot{\mathbf{y}}_j = \bm{\Phi}(\mathbf{y})\hat{\bm{\xi}}_j ,\,\,\,\,\,j = 1,\dots,n,
		\end{equation}
		where,
		\begin{gather*}
			\dot{\mathbf{y}}_j = 
			\begin{bmatrix}
				\dot{y}_j(t_1), & \dot{y}_j(t_2), & \hdots \,, &\dot{y}_j(t_m)
			\end{bmatrix}^T \in \R^{m};\\ \\
			\bm{\Phi}(\mathbf{y}) =
			\begin{bmatrix}
				\phi_{1}(\mathbf{y}(t_1)) & \phi_{2}(\mathbf{y}(t_1)) & \hdots & \phi_{p}(\mathbf{y}(t_1))\\
				\phi_{1}(\mathbf{y}(t_2)) & \phi_{2}(\mathbf{y}(t_2)) & \hdots & \phi_{p}(\mathbf{y}(t_2))\\
				\vdots         & \vdots         & \ddots & \vdots\\
				\phi_{1}(\mathbf{y}(t_m)) & \phi_{2}(\mathbf{y}(t_m)) & \hdots & \phi_{p}(\mathbf{y}(t_m))\\
			\end{bmatrix} \in \R^{m \times p};\,\,\,\, \text{and}\\
			\hat{\bm{\xi}}_j =
			\begin{bmatrix}
				\hat{\xi}_{1j}, & \hat{\xi}_{2j}, & \hdots \,, & \hat{\xi}_{pj}
			\end{bmatrix}^T \in \R^{p}.
		\end{gather*}
		Noise introduces perturbations on the basis functions $\{\phi_{i}(\mathbf{x})\}$ and complicates the numerical estimation of state time-derivatives. Figure~\ref{fig:noisy_basis_plot} illustrates the effect of measurement noise on the sparse linear system in Eqn.~(\ref{eq:matrix_main_system}) we aim to solve. First, noise is amplified in the state time-derivative vector $\dot{\mathbf{y}}_j$ if numerical differentiation is not carefully performed. Second, the basis functions in the linear expansion in Eqn.~(\ref{eq:matrix_main_system}) act as nonlinear transformations of the noise components. For example, the monomial basis function $\phi(\mathbf{x}) = x_1 x_2$,
		once evaluated at noisy measurements, yields $\phi(\mathbf{y}) = (x_1 + \epsilon_1)(x_2 + \epsilon_2) = x_1 x_2 + x_1\epsilon_2 + x_2 \epsilon_1 + \epsilon_1 \epsilon_2$, where the second and third terms are signal-dependent errors, and the last term stems from pure noise errors.
		\begin{figure}[H]
			\centering
			\includegraphics[trim = 0 0 0 0, clip,width=0.7\textwidth]{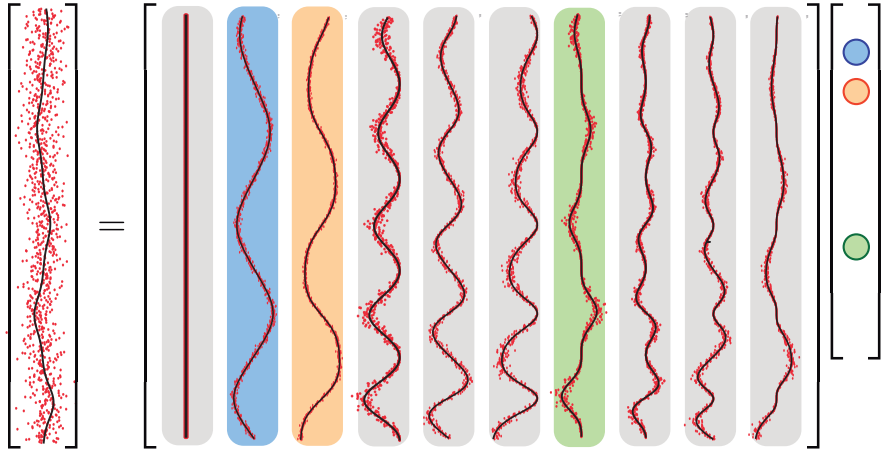}
			\caption{Schematic of the linear system in Eqn.~(\ref{eq:matrix_main_system}). Black solid lines represent exact signals, whereas red dots are their noisy counterparts. Measurement noise affects the state time-derivative accuracy and perturbs the space spanned by the basis functions represented by the columns of the measurement matrix $\bm{\Phi}$.}
			\label{fig:noisy_basis_plot}
		\end{figure}
		The linear system given in Eqn.~(\ref{eq:matrix_main_system}) can then be expressed in terms of the unavailable state variables as 
		\begin{equation}
			\dot{\mathbf{x}}_j + \mathbf{e}_j = [\bm{\Phi}(\mathbf{x}) + \mathbf{E}]\bm{\xi}_j,  
			\label{eq:matrix_main_system_errors}
		\end{equation}
		where $\mathbf{e}_j$ and $\mathbf{E}$ are implicitly the vector and matrix errors arising from the numerical estimation of state time-derivatives and the evaluation of the noisy terms through the basis matrix $\bm{\Phi}$, respectively. Noise perturbations change the correlation between the state-time derivative vector and the basis functions, making the recovery of the coefficient vector a challenging task. The aim of smoothing and numerical differentiation techniques is to reduce $\mathbf{e}_j$ and $\mathbf{E}$ as much as possible with the aim of achieving a more accurate and stable recovery of the sparse coefficient vector $\bm{\xi}_j$. Using {\it a priori} denoising methods, we replace $\dot{\mathbf{y}}_j$ and $\mathbf{y}_j$ for denoised versions of the state variables $\dot{\hat{\mathbf{x}}}_j$ and $\hat{\mathbf{x}}_j$, where the system in Eqn.~(\ref{eq:matrix_main_system}) now yields
		\begin{equation}\label{eq:matrix_main_system_denoised}
			\dot{\hat{\mathbf{x}}}_j = \bm{\Phi}(\hat{\mathbf{x}})\hat{\bm{\xi}}_j ,\,\,\,\,\,j = 1,\dots,n.
		\end{equation}
	\begin{remark}\label{rem:remark1}
		Problem in Eqn.~(\ref{eq:matrix_main_system_errors}) can be seen as an \textit{errors-in-variables} model where perturbations are present in both the dependent and independent variables. There are specialized methods to solve this problem, such as \textit{Total Least Squares}~\cite{golub1980analysis,van1991total}. Instead, this work focuses on reducing the size of the perturbations via filtering, where no specific structure is assumed in $\mathbf{e}_j$ and $\mathbf{E}$. 
	\end{remark}
	
	\noindent{\bf Notation.} {\it For the interest of a simpler notation, we henceforth drop the subscript $j$ from $\dot{\mathbf{x}}_j$ and $\bm{\xi}_j$ in Eqn.~(\ref{eq:matrix_main_system}). Unless otherwise stated, $\dot{\mathbf{x}}$ refers to the measurements of $\dot{x}_j$ and not the dynamics $\dot{\mathbf{x}}$ in Eqn.~(\ref{eq:DynamicalSystem}).}\\[-.3cm]
	
	\subsection{Sparse regression}
	\label{subsec:sparse_regression}
	
	{\color{black} Dynamical systems arising from physics often contain forcing terms, such as damping, viscous, elastic, gravitational forces, to name a few, that can be expressed in a basis where only a few terms are active (i.e. non-zero). Other physical systems that satisfy invariances and symmetries are governed by dynamics that also contain few terms of a specific basis. Sparse regression seeks a solution to a linear system of equations where only a few of the components of the solution vector are non-zero, avoiding overfitting and worsening the prediction performance of the identified dynamical model.}
	The sparse regression problem can be mathematically stated as
	\begin{equation}\label{eq:l0_minimization}
		\min_{\bm{\xi}} \norm{\bm{\xi}}_0\quad \text{subject to}\quad \norm{\bm{\Phi}(\mathbf{x})\bm{\xi} - \dot{\mathbf{x}}}_2 \leq \varrho,
	\end{equation}
	where $\norm{\bm{\xi}}_0 = \#\{i: \xi_i \neq 0, i = 1,...,p \}$ is the $\ell_0$-pseudonorm, which counts the number of non-zero elements in $\bm{\xi}$, and $\varrho$ is a tolerance that controls the size of the residual. The optimization problem in Eqn.~(\ref{eq:l0_minimization}) is known to be a NP-hard combinatorial problem~\cite{natarajan1995sparse}. Several alternatives have been proposed to approximate the solution of Eqn.~(\ref{eq:l0_minimization}) in a computationally tractable manner: from greedy algorithms to convex optimization procedures (see \cite{eldar2012compressed, bertsimas2020sparse} for an overview). In general, the accuracy of the recovered coefficients may strongly depend on the choice of the optimization strategy, especially in the large measurement noise regimes, \cite{cortiella2021sparse,fukami2021sparse}. In this article, we focus on Sequentialy Thresholded Least Squares (STLS), used in the original SINDy algorithm \cite{Brunton2016}, and Weighted Basis Pursuit Denoising (WBPDN) \cite{Candes08c,cortiella2021sparse}. 
	
	\subsection{Sequential Thresholded Least Squares (STLS)}
	\label{subsec:STLS}
	
	Sequentially Thresholded Least Squares (STLS) algorithm~\cite{Brunton2016} iteratively solves a least squares regression problem and hard-thresholds the coefficients to promote sparsity and thereby regularize the regression problem. The procedure is repeated on the non-zero entries of $\bm{\mathbf{\xi}}$ until the solution converges or the algorithm reaches a maximum number of iterations. Specifically, let $\mathcal{S}(\bm\xi):=\{i:\ \xi_i\neq 0\}$ denote the support of an instance of $\bm\xi$. At the $(k+1)$th iteration of STLS, $\bm\xi^{(k+1)}$ is computed from a least squares problem over $\mathcal{S}(\bm\xi^{(k)})$ and its components smaller than some threshold parameter $\gamma>0$ are set to zero,
	%
	\begin{equation}\label{eq:STLS}
		\begin{split}
		\text{(STLS)} \qquad    &\bm\xi^{(k+1)}\longleftarrow\underset{\bm\xi}{\mathrm{argmin}} \{\norm{\bm{\Phi}(\mathbf{x})\bm{\xi} - \dot{\mathbf{x}}}_2^2\\
		&\text{subject to} \quad \mathcal{S}(\bm\xi) = \mathcal{S}(\bm\xi^{(k)})\}\\
		&\bm\xi^{(k+1)}\longleftarrow \mathcal{T}(\bm\xi^{(k+1)};\gamma),
		\end{split}
	\end{equation}
	where the thresholding operator $\mathcal{T}(\cdot;\gamma)$ is defined as
	\begin{equation}\label{eq: threshold_operator}
		\mathcal{T}_i(\bm{\xi};\gamma) = 
		\begin{cases}
			\xi_i& \text{if }\ |\xi_i| > \gamma\\
			0              & \text{otherwise}
		\end{cases}
		,\quad i = 1,\dots,p. 
	\end{equation}
	
	The sufficient conditions for general convergence, rate of convergence, conditions for one-step recovery, and a recovery result with respect to the condition number and noise are addressed in~\cite{zhang2019convergence}. Algorithm \ref{alg:STLS} summarizes a practical implementation of STLS.
	
	\begin{algorithm}[H]
		\caption{Sequentially Thresholded Least Squares (STLS)~\cite{Brunton2016}} \label{alg:STLS}
		\begin{algorithmic}[1]
			\Procedure {STLS}{$\bm{\Phi}(\mathbf{x})$, $\dot{\mathbf{x}}$, $\gamma$}
			
			\State Solve $\bm{\Phi}(\mathbf{x})\bm\xi^{(0)} = \dot{\mathbf{x}}$ for $\hat{\bm\xi}^{(0)}$ using least squares.
			\State Apply thresholding $\hat{\bm\xi}^{(0)}\longleftarrow \mathcal{T}(\hat{\bm\xi}^{(0)};\gamma)\nonumber$.
			
			\While{\text{not converged} \text{or} $k < k_{\max}$}
			
			\State Delete the columns of $\bm{\Phi}(\mathbf{x})$ whose corresponding $\bm\xi^{(k)}$ component is 0, obtaining $\bm{\Phi}'(\mathbf{x})$.
			
			\State Solve $\bm{\Phi}'(\mathbf{x})\bm\xi^{(k)} = \dot{\mathbf{x}}$ for $\hat{\bm\xi}^{(k)}$ using least squares.
			
			\State Apply thresholding $\hat{\bm\xi}^{(k)}\longleftarrow \mathcal{T}(\hat{\bm\xi}^{(k)};\gamma)\nonumber$.
			
			\State $k = k + 1$.
			\EndWhile
			\EndProcedure
		\end{algorithmic}
	\end{algorithm}
	
	\subsection{Weighted Basis Pursuit Denoising}
	\label{subsec:WBPDN}
	Weighted Basis Pursuit Denoising (WBPDN) is an iterative reweighted version of the Basis Pursuit Denoising (BPDN) \cite{Wang2011} where the solution at the current iteration is weighted by solution-dependent weights from the previous iteration. To improve the robustness of SINDy with respect to the state and state derivative noise, WBPDN regularizes the regression problem involving Eqn.~(\ref{eq:matrix_main_system}) via weighted $\ell_1$-norm of $\bm\xi$,
	\begin{equation}
		\Vert\mathbf{W}\bm\xi\Vert_1 = \sum_{i=1}^p w_i\vert\xi_i\vert.    
	\end{equation}
	Here, $\mathbf{W}\in\mathbb{R}^{p\times p}$ is a diagonal matrix with diagonal entries $w_i>0$, $i=1,\dots,p$. WBPDN is inspired by the work in~\cite{zou2006adaptive, Candes2008,Yang13,Peng14,adcock2017infinite}
	from the statistics, compressed sensing, and function approximation literature, where weighted $\ell_1$-norm has been shown to outperform the standard $\ell_1$-norm in promoting sparsity, especially in the case of noisy measurements or when the solution of interest is not truly sparse, i.e., many entries of $\bm\xi$ are near zero~\cite{Candes2008,Yang13,Peng14,adcock2017infinite}. 
	
	More specifically, we solve the weighted variant of the BPDN problem,
	\begin{equation}\label{eq:WBPDN}
		(\text{WBPDN})\qquad    \min_{\bm{\xi}} \norm{\bm{\Phi}(\mathbf{x})\bm{\xi} - \dot{\mathbf{x}}}_2^2 + \lambda \norm{\mathbf{W}\bm{\xi}}_1,
	\end{equation}
	which coincides with the adaptive LASSO approach of~\cite{zou2006adaptive}. Depending on the choice of $\mathbf{W}$, $\Vert\mathbf{W}\bm\xi\Vert_1$ gives a closer approximation to the $\ell_0$-norm of $\bm\xi$ than the $\ell_1$-norm, and thus better enforces sparsity in $\bm\xi$. The main goal of using a weighted $\ell_1$-norm -- instead of its standard counterpart -- is to place a stronger penalty on the coefficients $\xi_i$ that are anticipated to be small (or zero). As proposed in \cite{zou2006adaptive, Candes2008}, the weights are set according to
	\begin{equation}\label{eq:WL1weigths}
		w_i^{(k+1)} = \frac{1}{|\xi_i^{(k)}|^q + \upsilon},
	\end{equation}
	where $q>0$ represents the strength of the penalization and $\upsilon$ is a small parameter to prevent numerical issues when $\xi_i$ is zero. In our numerical experiments, we set $q=2$ and $\upsilon = 10^{-4}$ as they robustly produce accurate coefficient recovery.
	The problem in Eqn.~(\ref{eq:WBPDN}) may be solved via BPDN solvers for standard $\ell_1$-minimization{, \color{black} such as \textit{SparseLab}~\cite{Donoho2009sparselab}, \textit{SPGL1}~\cite{spgl1site} or \textit{CVX}~\cite{grant2014cvx, diamond2016cvxpy} to name a few,} with the simple transformations $\tilde{\bm\xi}:=\mathbf{W}\bm\xi$ and $\tilde{\bm{\Phi}}(\mathbf{x}) = \bm{\Phi}(\mathbf{x})\mathbf{W}^{-1}$, i.e.,
	\begin{equation}\label{eq:transformed-BPDN}
		\min_{\tilde{\bm{\xi}}} \norm{\tilde{\bm{\Phi}}(\mathbf{x})\bm{\xi} - \dot{\mathbf{x}}}_2^2 + \lambda \norm{\tilde{\bm{\xi}}}_1.\nonumber
	\end{equation}
	Given the solution $\tilde{\bm\xi}$ to Eqn.~(\ref{eq:transformed-BPDN}), $\bm\xi$ is then computed from $\bm\xi = \mathbf{W}^{-1}\tilde{\bm\xi}$. Algorithm \ref{alg:WBPDN}, adopted from \cite{Candes2008}, outlines the steps involved in WBPDN.
	
	\begin{algorithm}[H]
		\caption{Iteratively Reweighted Weighted Basis Pursuit Denoising (WBPDN) adopted from \cite{Candes2008}}\label{alg:WBPDN}
		\begin{algorithmic}[1]
			\Procedure {WBPDN}{$\bm{\Phi}(\mathbf{x})$, $\dot{\mathbf{x}}$, $\lambda$, $q$, $\upsilon$}
			
			\State Set the iteration counter to $k = 0$ and $w_i^{(0)} = 1, \quad i = 1,\dots,p$.
			\While{\text{not converged} \text{or} $k < k_{\max}$}
			
			\State Solve the WBPDN problem in Eqn.~(\ref{eq:WBPDN}) for a specific $\lambda$
			\begin{equation*}
				\bm{\xi}^{(k)} = \underset{\bm\xi}{\mathrm{argmin}}\left\{\norm{\bm{\Phi}(\mathbf{x})\bm{\xi} - \dot{\mathbf{x}}}_2^2 + \lambda \norm{\mathbf{W}^{(k)}\bm{\xi}}_1\right\}.
			\end{equation*}
			
			\State Update the weights for $i = 1,\dots,p$
			
			\begin{equation*}
				w_i^{(k+1)} = \frac{1}{|\xi_i^{(k)}|^q + \upsilon}.
			\end{equation*}
			\State $k = k + 1$.
			\EndWhile
			\EndProcedure
		\end{algorithmic}
	\end{algorithm}
	
	\section{Measurement denoising and derivative estimation}
	\label{sec:filtering}
	
	In the present work, the aim of denoising techniques is to obtain an approximate model of the state variables with maximum accuracy with respect to an appropriate error measure as well as the smoothest solution possible. The derivatives are then computed by differentiating the approximate model. The assumption  in  all  the  methods  is to  approximate  the state variables $\mathbf{x}(t)$ to then estimate the time derivatives by differentiating the state variable approximations. {\color{black} The assumption is that if the approximation of the state variables $\hat{\mathbf{x}}(t)$ is close to the original, unknown variables $\mathbf{x}(t)$, then the estimated time derivatives $\dot{\hat{\mathbf{x}}}(t)$ are also  close  to $\dot{\mathbf{x}}(t)$. Mathematically, this can be expressed as $\norm{\dot{\hat{\mathbf{x}}}(t) - \dot{\mathbf{x}}(t)} \leq C \norm{\hat{\mathbf{x}}(t) - \mathbf{x}(t)}$ for some positive constant {\color{black}$C \sim O(\Delta t)$, where $\Delta t$ is the sampling rate}, and a suitable norm, usually the $\ell_2$-norm.}
	
	One can divide smoothing techniques into local and global methods. On the one hand, local methods fit a regression function locally in time using neighboring measurements around a specific data point, and ignores data outside the neighborhood. On the other hand, global methods employ all the data available to estimate a smooth approximation over the entire time domain. Inspired by the findings in \cite{d1992comparison, ahnert2007numerical, chartrand2011numerical, knowles2014methods, stickel2010data}, this article compares several best-performing local and global techniques for data denoising and numerical time differentiation.
	\subsection{Local methods}
	\label{sec:local_methods}
	Local smoothing methods fit a model locally around a neighborhood of each query point. That is, for any given $t$, local methods fit a smooth regression function $x(t)$ using the data points within a local neighborhood with a specific width. The local time derivative can be estimated by differentiating the approximation $\hat{x}(t)$ analytically. The width of the neighborhood, $h(t)$, called the \textit{window size} or \textit{bandwidth}, localizes the data around $t$. Usually, local regression methods approximate $x(t)$ using a low-degree polynomial, where the local coefficients are estimated via least squares. The benefit of local polynomial regression is that it denoises the data and estimates the derivative simultaneously. {\color{black} Alternative local smoothing methods employed for smooth surface reconstruction from scattered data points exist in the literature that can also be applied. Examples of these methods include moving least squares~\cite{lancaster1981surfaces, sober2019manifold, sober2021approximation} or radial basis function smoothing~\cite{carr2003smooth}.}
	\begin{figure}[H]
		\centering
		\includegraphics[trim = 0 0 0 0, clip,width=0.8\textwidth]{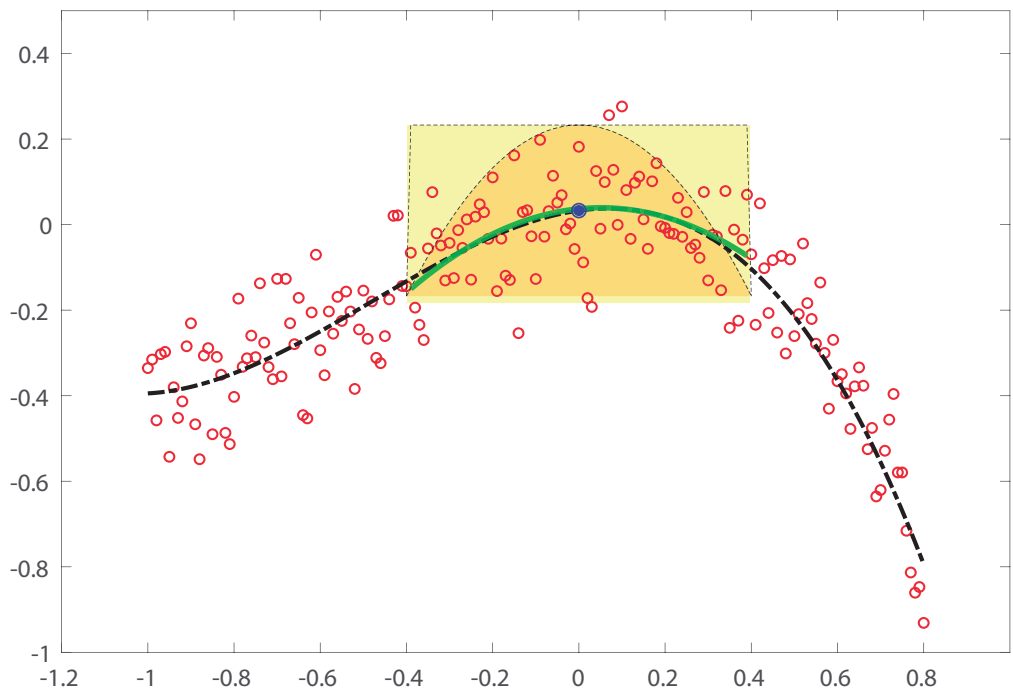}
		\caption{Illustration of the local polynomial regression method using a local quadratic polynomial (green solid line) over a window of size 30\% of the entire data range. The shaded regions correspond to the rectangular (yellow) and Epanechnikov~\cite{epanechnikov1969non} (orange) kernel functions.}
		\label{fig:local_smoothing}
	\end{figure}
	The local polynomial of degree $d$ at $t_0$ is given by a local basis expansion of the form
	\begin{equation}\label{eq:local_polynomial}
		x(t;t_0) = \sum_{k=1}^{d+1}\theta_k(t_0) b_k(t),
	\end{equation}
	where $\{\theta_k(t_0)\}_{k=1}^{d+1}$ are the local polynomial coefficients at $t_0$ and the local basis functions $b_k(t) = (t - t_0)^{k-1},\,\,k = 1,...,d+1$, are monomials of increasing degree. Local polynomial regression estimates the local coefficients via weighted least squares as
	\begin{equation}\label{eq:local_regression_problem}
		\hat{\bm{\theta}}(t_0) = \argminB_{\bm{\theta}\in \R^{d+1}}\,\, (\mathbf{y} - \mathbf{B}\bm{\theta})^T \mathbf{W}(t_0) (\mathbf{y} - \mathbf{B}\bm{\theta}),  
	\end{equation}
	where $\mathbf{B} \in \R^{m \times (d+1)}$ is the basis matrix such that $\{\mathbf{B}\}_{ik} = b_k(t_i)$, $\bm{\theta} = [\theta_0, \theta_1,..., \theta_d]^T \in \R^{d+1}$ is the coefficient vector, and $\mathbf{W}(t_0) \in \R^{m \times m}$ is the diagonal weight matrix defined as $\{\mathbf{W}(t_0)\}_{ii} = K_h(t_0, t_i)$, $K_h$ being the kernel function. The kernel function localizes the data around point $t_0$ by weighting its neighbors according to a distance metric. For example, the popular tricube kernel is given by~\cite{cleveland1979robust},
	\begin{equation}\label{eq:kernel_equation}
		K_h(t_0, t) = D\bigg(\frac{|t - t_0|}{h}\bigg),
	\end{equation}
	with
	\begin{equation}
		D(u) = \begin{cases}
			(1 - |u|^3)^3 &\text{if}\,\,\,\,|u| \leq 1;\\
			0 &\text{otherwise.}
		\end{cases}
	\end{equation}
	Those data points whose distance with respect to $t_0$ exceed the bandwidth $h$ receive a zero weight, meaning that they are excluded from the local regression. Figure~\ref{fig:local_smoothing} shows a schematic of local polynomial regression at a specific data point with two different kernels, represented by the shaded regions.
	The solution to Eqn.~(\ref{eq:local_regression_problem}) at $t_0$ is given by
	\begin{equation}
		\hat{\bm{\theta}}(t_0) = (\mathbf{B}^T\mathbf{W}(t_0)\mathbf{B})^{-1}\mathbf{B}^T\mathbf{W}(t_0)\mathbf{y},
	\end{equation}
	and the estimate and derivatives at $t_0$ are given by $x(t_0) = \hat{\theta}_0(t_0)$ and $\dot{x}(t_0) = \hat{\theta}_1(t_0)$. 
	
	The quality of the local estimates is highly dependent on the local bandwidth $h(t)$ and the degree of the polynomial. Large bandwidths tend to under-parametrize the regression function and increase the estimate bias because of curvature effects~\cite{fan1995data}. Small bandwidths, on the other hand, over-parametrize the unknown function, increase the variance and produce noisy estimates. A similar trade-off occurs with the degree of the polynomial, although the resulting estimates are less sensitive as compared to the bandwidth~\cite{fan1996local}. For a fixed bandwidth $h$, a large degree $d$ reduces the bias but increases the variance, whereas a low order reduces the variance and increases the bias due to curvature effects. As recommended in~\cite{fan1995adaptive}, and since we are interested in first-order derivatives, we use a fixed polynomial degree $d = 2$ in the numerical experiments of this work. How to select an appropriate bandwidth $h$ will be discussed in Section \ref{sec:model_selection}. In this article, we focus on the Savitzky-Golay (S-G) filter~\cite{savitzky1964smoothing} and a locally weighted polynomial regression~\cite{cleveland1988locally}. 
	
	\begin{remark}\label{remark:remak1}
		Reference~\cite{fan1995adaptive} suggests a polynomial degree $d = \nu +1$, where $\nu$ is the order of the derivative ($\nu = 0$ is the regression function estimate). Since we expect higher error on the derivative estimates than on the function estimates, we focus on first-order derivatives and set $d = 2$ for computing the function and derivative estimates.
	\end{remark}
	
	\subsubsection{Savitzky–Golay (S-G) filter}
	
	Savitzky and Golay~\cite{savitzky1964smoothing} proposed a simplified polynomial least squares fit convolution for smoothing and estimating derivatives of a set of sequential data. The convolution can be understood as a weighted moving average filter with weighting given as a low-degree polynomial within the filter window (see \cite{schafer2011savitzky} for a detailed analysis). The S-G filter was designed to preserve higher moments within the data and to reduce the bias.  In certain fields, such as econometrics and statistics, S-G filter is often referred as \textit{Local Estimated Scatterplot Smoothing (LOESS)} when the bandwidth and spacing are constant. The S-G filter belongs to the class of local polynomial regression framework whose kernel is the rectangular function with compact support given by Eqn.~(\ref{eq:kernel_equation}) with
	\begin{equation}\label{eq:rect_kernel}
		D(u) = \begin{cases}
			1 &\text{if}\,\,\,\,|u| \leq 1;\\
			0 &\text{otherwise.}
		\end{cases}
	\end{equation}
	The kernel in Eqn.~(\ref{eq:rect_kernel}) acts as a window centered at $t_0$ where data points outside of it are discarded.
	
	\subsubsection{Locally weighted polynomial regression}
	Locally weighted polynomial regression~\cite{cleveland1988locally}, often termed \textit{Locally Weighted Scatterplot Smoothing (LOWESS)},  is a generalization of S-G filter with different kernel functions, and possibly variable bandwidth and non-uniform spacing. Following the extensive work by~\cite{fan1995data,fan1995adaptive, fan1996local, fan1996study}, we use the \textit{Epanechnikov kernel} function of the form Eqn.~(\ref{eq:kernel_equation}) with
	\begin{equation}\label{eq:Epanechnikov_kernel}
		D(u) = \begin{cases}
			\frac{3}{4}(1 - u^2) &\text{if}\,\,\,\,|u| \leq 1;\\
			0 &\text{otherwise.}
		\end{cases}
	\end{equation}
	The Epanechnikov kernel has compact support and is the optimal kernel function in the sense that it minimizes the mean squared error (MSE) in the interior of the data domain \cite{epanechnikov1969non}.
	\subsection{Global methods}
	
	Global methods yield an estimate over the entire data domain by imposing a smoothing constraint to the optimization problem. In this work, we consider smoothing algorithms whose global loss function is of the form
	\begin{equation}\label{eq:loss_global_denoising}
		\mathcal{L}(\mathbf{x}) = SSE(\mathbf{x}) + \lambda \mathcal{R}(\mathbf{x}),
	\end{equation}
	where the first term is the sum of the squared errors given by $SSE(\mathbf{x}) = \norm{\mathbf{y} - \mathbf{x}}_2^2$, and the second term promotes smoothness of the solution via a regularization function $\mathcal{R}$, and $\lambda \geq 0$ is the regularization -- also smoothing -- parameter. The latter controls the trade-off between the closeness to the data and the smoothness of the function estimates. Figure~\ref{fig:global_smoothing} illustrates the behavior of the global smoothing methods as the regularization parameter is varied. {\color{black}When there is no regularization, i.e., $\lambda$ = 0, the estimates interpolate the data, whereas when the regularization parameter is gradually increased, the resulting estimate tends to the best linear approximation~\cite{hastie2009elements}.} The \textit{optimal} $\lambda$ resides between these two extreme cases and must be conveniently selected. In Section~\ref{sec:model_selection}, we present methods to automatically determine near optimal regularization parameters. This article considers three well-known global smoothing algorithms: smoothing splines, Tikhonov smoother (a.k.a. H-P filter), and $\ell_1$-trend filtering. 
	\begin{figure}[H]
		\centering
		\includegraphics[trim = 0 0 0 0, clip,width=0.8\textwidth]{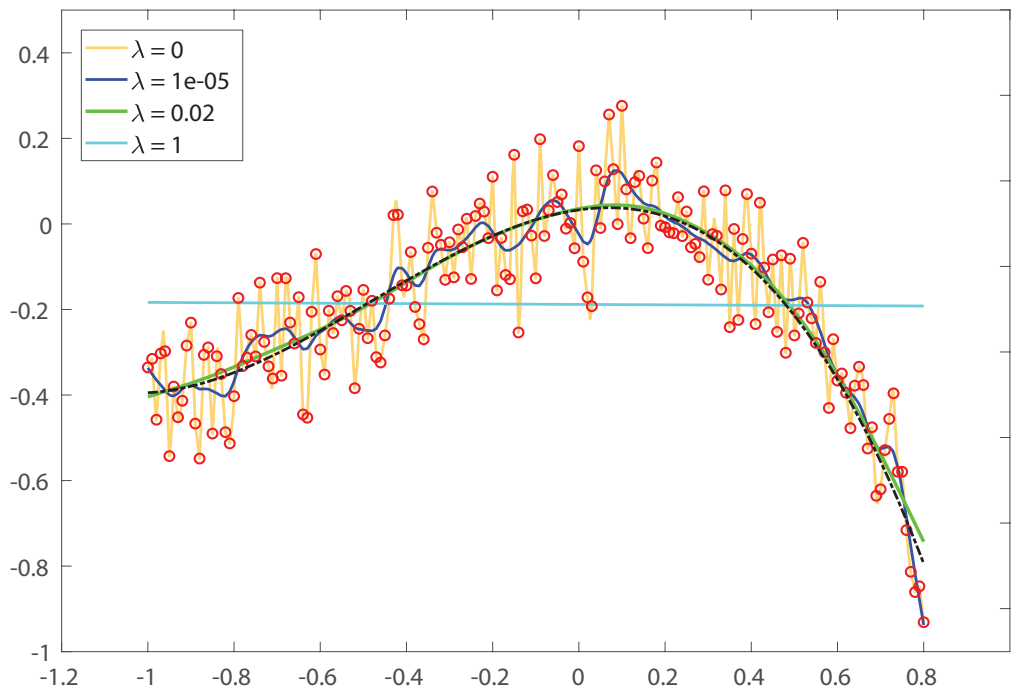}
		\caption{Schematic of data filtering via global smoothing methods. The regularization parameter can be varied from $\lambda = 0$, where data is interpolated, to infinity, which yields the best linear fit. The optimal $\lambda$ lives within the two extreme cases. The $\lambda$ parameter was normalized in the $[0,1]$ range.}
		\label{fig:global_smoothing}
	\end{figure}
	\subsubsection{Smoothing splines}
	Smoothing splines is a variant of regression splines~\cite{hastie2009elements} for function approximation based on truncated piece-wise polynomial basis where a regularization term is added to penalize against high curvatures, and hence a smooth estimate. Smoothing splines uses a basis expansion to approximate the function given by
	\begin{equation}\label{eq: sspline_continuous}
		x(t) = \sum_{k=1}^p N_k(t)\theta_k,
	\end{equation}
	where $\{N_j(t)\}_{k=1}^p$ is the $p$-dimensional natural spline basis over the knots $t_1,...,t_m$, and $\{\theta_j\}_{k=1}^p$ are the parameters weighting each basis function. Denoting $\hat{\mathbf{x}} = \mathbf{N}\bm{\theta}$ the vector of $m$ fitted values at specific $t_i$, the smoothing splines problem can be defined as 
	\begin{equation}\label{eq:ssplines_problem}
		\hat{\bm{\theta}} = \argminB_{\bm{\theta}\in \R^{p}} \norm{\mathbf{y} - \mathbf{N}\bm{\theta}}_2^2 + \lambda \bm{\theta}^T \bm{\Omega}_N \bm{\theta},
	\end{equation}
	where $\mathbf{N} \in \R^{m \times p}$ is the basis matrix of natural splines, $\bm{\theta} \in \R^{p}$ is the vector of parameters, $\bm{\Omega}_N\ = \int (\mathbf{N}^{''}(\tau))^T\mathbf{N}^{''}(\tau) d\tau \in \R^{p \times p}$ is the matrix that penalizes the curvature. The unique minimizer of Eqn.~(\ref{eq:ssplines_problem}) is a natural cubic spline with knots at $\{t_i\}_{i=1}^m$ where regularization is introduced through the penalty term $\bm{\Omega}_N$ by weighting the basis coefficients $\bm{\theta}$ \cite{green1993nonparametric}. The limit when $\lambda \rightarrow 0$ corresponds to an estimate $\hat{\bm{\theta}}$ that interpolates the data, whereas as $\lambda \rightarrow \infty$ the estimate tends to the best linear fit since no second-order derivative is permitted~\cite{hastie2009elements}. Problem in Eqn.~(\ref{eq:ssplines_problem}) admits a closed-form solution given by
	\begin{equation}\label{eq:ssplines_solution}
		\hat{\bm{\theta}} = (\mathbf{N}^T\mathbf{N} + \lambda \bm{\Omega}_N)^{-1}\mathbf{N}^T \mathbf{y}.
	\end{equation}
	Once the parameters $\hat{\bm{\theta}}$ are estimated, the function estimates yield
	\begin{equation}\label{eq: sspline_approximation}
		\hat{x}(t) = \sum_{j=1}^p N_j(t)\hat{\theta}_j,
	\end{equation}
	and one can easily obtain derivatives by differentiating Eqn.~(\ref{eq: sspline_approximation}). Smoothing splines can be thought as a linear filter that projects the input measurement vector $\mathbf{y}$ to a subspace spanned by the columns of a smoother matrix $\mathbf{S}_{\lambda}$ as 
	\begin{equation}\label{eq: sspline_projection}
		\hat{\mathbf{x}} = \mathbf{S}_{\lambda}\mathbf{y},
	\end{equation}
	where $\mathbf{S}_{\lambda} = \mathbf{N}(\mathbf{N}^T\mathbf{N} + \lambda \bm{\Omega}_N)^{-1}\mathbf{N}^T$. By choosing $B$-spline basis functions, the smoothing splines estimate $\hat{\mathbf{x}}$ in Eqn.~(\ref{eq: sspline_continuous}) can be computed in $O(m)$ arithmetic operations (see \cite{de1978practical} for an efficient implementation).
	\subsubsection{Tikhonov smoother}
	Tikhonov smoother, often referred as \textit{Hodrick-Prescott (H-P) filter}~\cite{hodrick1997postwar} when used as a data-smoother, incorporates an $\ell_2$-norm regularization term to the loss function of a least squares problem to control the regularity of the solution. Specifically, for any given regularization parameter $\lambda \geq 0$, the solution of Tikhonov smoother is given by solving the convex optimization problem
	
	\begin{equation}\label{eq:tikhonov_filter_problem}
		\hat{\mathbf{x}} = \argminB_{\mathbf{x} \in \R^m} \norm{\mathbf{y} - \mathbf{x}}_2^2 + \lambda \norm{\mathbf{D}_{(2)}\mathbf{x}}_2^2,
	\end{equation}
	where $\mathbf{D}_{(2)} \in \R^{(m-2) \times m}$ is the second-order difference matrix (see Appendix A). The term $\norm{\mathbf{D}_{(2)}\mathbf{x}}_2^2$ penalizes the curvature of $\mathbf{x}$, yielding smooth approximations. For a fixed $\lambda$, problem in Eqn.~(\ref{eq:tikhonov_filter_problem}) admits the closed-form unique solution
	\begin{equation}\label{eq:tikhonov_solution}
		\hat{\mathbf{x}} = (\mathbf{I} + \lambda \mathbf{D}_{(2)}^T\mathbf{D}_{(2)})^{-1}\mathbf{y}.
	\end{equation}
	The relative fitting error of Tikhonov smoother satisfies the following property \cite{kim2009ell_1}
	\begin{equation}
		\frac{\norm{\mathbf{y} - \hat{\mathbf{x}}}_2}{\norm{\mathbf{y}}_2} \leq \frac{32 \lambda}{1 + 32\lambda}.
	\end{equation}
	Tikhonov smoother offers linear computational complexity and the solution in Eqn.~(\ref{eq:tikhonov_solution}) can be computed in $O(m)$ arithmetic operations~\cite{Kim2007}. After computing the state estimates $\hat{\mathbf{x}}$, we use natural cubic splines to fit the state estimates continuously and approximate the state time-derivatives via differentiation.
	As in smoothing splines, we can project the input data onto the subspace of basis functions weighted by $\lambda$ via $\hat{\mathbf{x}} = \mathbf{S}_{\lambda}\mathbf{y}$, where $\mathbf{S}_{\lambda} = (\mathbf{I} + \lambda \mathbf{D}_{(2)}^T\mathbf{D}_{(2)})^{-1}$ is the smoother matrix.
	
	\subsubsection{$\ell_1$-trend filtering}
	
	The $\ell_1$-trend filtering is a variation of the Hodrick-Prescott filter where the $\ell_2$-norm of the regularization term is substituted for the $\ell_1$-norm to penalize the variations of the estimated trend. As opposed to smoothing splines and Tikhonov smoother, $\ell_1$-trend filtering does not possess a closed-form solution, and optimization algorithms, such as interior-point methods, must be employed to solve the problem. Originally, $\ell_1$-trend filtering was proposed to produce piece-wise linear trend estimates~\cite{kim2009ell_1}. However, more recent work by~\cite{tibshirani2014adaptive} extended $\ell_1$-trend filtering accounting for piece-wise polynomials of any degree, leading to smoother estimates. They empirically observed that $\ell_1$-trend filtering estimates achieve {\color{black}better local adaptivity} than smoothing splines and offered similar performance to locally adaptive smoothing splines \cite{mammen1997locally} at reduced cost. A $k$th-order $\ell_1$-trend filtering fits a piece-wise polynomial of degree $k$ to the data. In the specific case of $k = 0$, the $\ell_1$-trend filtering reduces to the well-known one-dimensional total variation regularization~\cite{rudin1992nonlinear}, or the one-dimensional fussed LASSO~\cite{tibshirani2005sparsity}. 
	For a given integer $k\geq 0$, the $k$th-order $\ell_1$-trend filtering estimate is defined as
	\begin{equation}\label{eq:trendfilter_problem}
		\hat{\mathbf{x}} = \argminB_{\mathbf{x} \in \R^m} \frac{1}{2} \norm{\mathbf{y} - \mathbf{x}}_2^2 + \lambda \norm{\mathbf{D}_{(k+1)}\mathbf{x}}_1,
	\end{equation}
	where $\lambda$ is the regularization parameter and $\mathbf{D}_{(k+1)} \in \R^{(m-k-1) \times (m-k)}$ is the difference matrix of order $k+1$ (see Appendix A). For any order $k \geq 0$ and a pre-choosen $\lambda$, the $\ell_1$-trend filtering estimate $\hat{\mathbf{x}}$ is a unique minimizer of Eqn.~(\ref{eq:trendfilter_problem}) since the problem is strictly convex. 
	
	The $\ell_1$-trend filtering estimates are not linear functions of the input data. Thus, there does not exist a smoother matrix that projects the input data onto its column space. Similar to smoothing splines and Tikhonov smoother, the $\ell_1$-trend filtering possesses interesting convergence properties, as studied in \cite{kim2009ell_1,tibshirani2011solution}. In particular, the maximum fitting error is bounded as
	\begin{equation}\label{eq:trendfilter_error}
		\norm{\mathbf{y} - \hat{\mathbf{x}}}_{\infty} \leq 4\lambda,
	\end{equation}
	where $\|\cdot\|_{\infty} \coloneqq \max|\cdot|$ is the infinity norm. The bound in Eqn.~(\ref{eq:trendfilter_error}) indicates that as $\lambda \rightarrow 0$ the estimate $\hat{\mathbf{x}}$ tends to the original input data. On the other extreme, as $\lambda \rightarrow \infty$ the estimate tends to the best linear fit. However, there exists a finite value $\lambda_{\text{max}} = \norm{(\mathbf{D}_{(k+1)}^T\mathbf{D}_{(k+1)})^{-1}\mathbf{D}_{(k+1)}\mathbf{y}}_{\infty}$ for which that is achieved. This is, for $\lambda \geq \lambda_{\text{max}}$, the estimate $\hat{\mathbf{x}}$  corresponds to the same optimal linear fit. The computational complexity of $\ell_1$-trend filtering is $O(m)$ per iteration and can be solved efficiently via specialized primal-dual interior point methods~\cite{tibshirani2014adaptive}. If speed and robustness are a concern, \cite{ramdas2016fast} proposes an enhanced and more efficient $\ell_1$-trend filtering solving strategy.
	
	We note that one can directly estimate derivatives by reformulating $\ell_1$-trend filtering in terms of an integral operator and regularizing the derivative estimates using the $\ell_1$-norm. Similar to Tikhonov, we did not perceive significant differences on the results compared to differentiation via spline interpolation on the denoised states.
	
	\section{Hyperparameter selection methods}
	\label{sec:model_selection}
	Model selection is the task of optimally selecting its hyperparameters from a set of candidates, given a set of data. In the context of state variable denoising the selection criterion aims to optimize the predictive quality of the model, i.e., denoised states, by selecting the most appropriate one that balances the closeness of fit and the complexity. All the methods described in Sections~\ref{subsec:sparse_regression} and ~\ref{sec:filtering} contain a sparse regularization parameter, in case of sparse regression algorithms, and smoothing parameter, bandwidth $h$ for local smoothing and $\lambda$ for global smoothing, that controls this trade-off. A considerable number of methods have been proposed, following different philosophies and exhibiting varying performances; see \cite{ding2018model, hansen2010discrete, hastie2009elements} for a review. In this work, we focus on two data-driven selection methods widely used in regularized regression without prior knowledge of the noise process: generalized cross-validation (GCV) and Pareto curve criterion.
	\subsection{Generalized cross-validation}
	Cross-validation (CV) is a general approach for choosing \textit{near optimal} regularization parameters $\lambda$ for estimating the prediction error in regression problems. CV separates the data samples into training and validation sets, and use the training set to compute the solution which is then used to predict the elements of the validation set. By repeating this procedure for different regularization parameters, one can select the regularizer that gives the minimum estimated mean squared error. \textit{$K$-fold cross-validation} splits the data into $K$ roughly equal-sized portions. For the $k$th portion, the model is fitted to the other $K-1$ parts of the samples, and the prediction error of the fitted model is estimated using the $k$th part (see \cite{hastie2009elements} for more details).
	The case $K = m$ is known as \textit{leave-one-out} cross-validation and all the data is used to fit the model except one sample. Leave-one-out CV can be computationally expensive since a model must be fit $m$ times for each regularizer. \textit{Generalized cross-validation} (GCV) approximates the leave-one-out cross-validation in a tractable and convenient manner. For smoothing splines, GCV is more reliable than ordinary cross-validation in the sense that it has less tendency to under-smooth the data~\cite{ramsay2004functional,gu2013smoothing}. The GCV function is defined as
	\begin{equation}\label{eq:gcv_function}
		\text{GCV}(\lambda) = \frac{m\,\norm{\mathbf{y} - \hat{\mathbf{x}}}_2^2}{(m - \text{df}_{\lambda})^2},
	\end{equation}
	where $\hat{\mathbf{x}}$ is the estimate of $\mathbf{x}$ for a given regularization parameter $\lambda$, and $\text{df}_{\lambda}$ are the \textit{degrees of freedom} of the regularized regression method. For linear smoothers, the degrees of freedom are defined as the trace of the smoother matrix, $\text{df}_{\lambda} = \text{trace}(\mathbf{S}_{\lambda})$,~\cite{hastie2009elements}. The degrees of freedom measure the number of effective model parameters and therefore, the complexity of the resulting model.
	
	In case of $\ell_1$-regularized methods, which do not admit a closed-form solution, the definition of the degrees of freedom depends on the problem. Specifically, for LASSO-type problems the number of non-zero coefficients of the solution $\bm{\xi}$ is an unbiased estimate of the degrees of freedom, i.e., $\text{df}_{\lambda} = \norm{\bm{\xi}}_0$,~\cite{zou2007degrees}, whereas for $\ell_1$-trend filtering the degrees of freedom are defined as $\text{df}_{\lambda} = \norm{\mathbf{D}_{(k+1)}\hat{\mathbf{x}}}_0+k+1$~\cite{tibshirani2014adaptive}. The regularization parameter $\lambda$ is chosen as the minimizer of the GCV function  
	\begin{equation}\label{eq:lambda_gcv}
		\lambda_{GCV} = \argminB_{\lambda} \text{GCV}(\lambda).
	\end{equation}
	GCV is generally an accurate and robust method to compute estimates with minimum prediction error. However, there are some limitations. For $\ell_2$-regularization problems, GCV may suffer from severe under-smoothing \cite{hansen2010discrete}, and the region around the minimum of the GCV function is often very flat. Depending on the resolution, the minimum may be difficult to locate numerically. Additionally, GCV may fail to select near optimal regularization parameters when the errors are highly correlated. According to \cite{wahba1990spline}, GCV may confuse correlated errors {\color{black}to be part of the exact trajectory}, and therefore tends towards regularization parameters  that  only  filter  out  the  uncorrelated component of the noise. For $\ell_1$-regularization problems, GCV may tend towards zero for zero regularization parameters and the global minimum may hide the local minimum that yields optimal prediction errors \cite{jansen2015generalized}.
	
	For local methods, the GCV functions are defined in terms of the kernel bandwidths $h$. There are two alternatives to select suitable bandwidths using GCV: varying the bandwidth locally or globally. In the former case, the local bandwidth is varied within a range $h(t_0) \in [h_{min}(t_0), h_{max}(t_0)]$ for each data point $t_0$, and the one that minimizes the local GCV is selected. In the latter case, a local regression for all data points with a fixed bandwidth is performed and repeated with different bandwidths within a range $h \in [h_{min}, h_{max}]$, and select the one that minimizes the global GCV function. In our numerical experiments, we did not observe significant differences in performance between the two approaches. We therefore only present the global bandwidth approach in Section~\ref{sec:numerical_examples}.
	\subsection{Pareto curve criterion}
	The Pareto curve is a graph that traces the trade-off between the residual, i.e. $SSE$ in Eqn.~(\ref{eq:loss_global_denoising}), and the regularization constraint by varying the regularization parameter $\lambda$. Typically, the Pareto curve is an L-shaped graph generated by plotting the norm of the regularization constraint, i.e. $\mathcal{R}$ in Eqn.~(\ref{eq:loss_global_denoising}), versus the norm of the residual in a log-log scale for different values of $\lambda$. The Pareto curve we are referring to in this work is often called the L-curve for $\ell_2$-regularization~\cite{Hansen1992Lcurve} and Pareto frontier for $\ell_1$-regularization~\cite{van2009probing}. We will refer to the Pareto curve when the $\ell_2$-norm is employed for the residual and either the $\ell_1$- or $\ell_2$-norms are used for the regularization constraint, and we will be specific in cases it creates confusion. The Pareto curve often has a corner located where the solution changes from being dominated by the regularization error, corresponding to the steepest part of the curve, to being dominated by the residual error, where the curve becomes flat. The Pareto curve criterion employs the heuristic {\color{black}that makes use of the observation:} the optimal value of $\lambda$ corresponds to the curve’s corner such that a good balance of the regularization and residual errors is achieved.
	
	{\color{black}For continuous Pareto curves, defined for all $\lambda \in [0,\infty)$,}~\cite{Hansen1992Lcurve} suggests defining the corner point as the point with maximum curvature. For discrete curves, however, it becomes more complicated to define the location of the corner point. Hansen et al.~\cite{Hansen2007} highlights the difficulties in computing the corner point from discrete L-curves built using a finite number of $\lambda$ values. The discrete curves may contain small local corners other than the global one that may give sub-optimal regularization parameters. To alleviate this issue, they propose an adaptive pruning algorithm, where the best corner point is computed by using candidate corners from curves at different resolutions. Since the L-curves must be sampled at different scales, the pruning algorithm can be computationally expensive. We instead compute the corner points using a simpler method proposed in Cultrera et al.~\cite{cultrera2020simple}. The algorithm iteratively employs the Menger curvature~\cite{leger1999menger} of a circumcircle and the golden section search method to efficiently locate the point on the curve with maximum curvature. 
	
	As compared to the GCV criterion, the Pareto curve criterion for $\ell_2$-regularization is often more robust to correlated errors since it combines information about the residual norm and the norm of the solution, whereas GCV only uses information about the residual norm~\cite{hansen1992analysis}.
	The Pareto curve criterion also presents some limitations. {\color{black}The L-curve is likely to fail for problems with very smooth solutions~\cite{hanke1996limitations}, and non-convergence may occur when the generalized Fourier coefficients decay at the  same  rate  or  less  rapidly  than  the  singular  values  of  the  basis matrix, $\bm{\Phi}$ in this article (see~\cite{vogel1996non} for details).} Additionally, the location of the corner is dependent on the scale in which the Pareto curve is considered, and it may not appear in certain scales~\cite{Reginska1996}.
	In the numerical examples section, we compare the performance and discuss limitations of GCV and Pareto curve for the smoothing and sparse regression methods considered in this article.
	%
	\section{Numerical examples}
	\label{sec:numerical_examples}
	In this section, we present numerical examples to assess the performance of the smoothing and numerical differentiation methods presented in Section~\ref{sec:filtering}, as well as model selection methods  in Section~\ref{sec:model_selection}. We then compare the accuracy of the sparse regression algorithms in Section~\ref{sec:statement}, specifically WBPDN and STLS, to recover governing equations of three nonlinear dynamical systems using the pre-processed data. In all cases, we assume no prior knowledge of the noise values and dynamics that generated the data, except that the governing equations can be sparsely expressed in a multivariate monomial basis in the state variables. We only have access to the noisy state measurements at discrete times represented by the measurement vector $\mathbf{y}_j \in \R^m, j = 1,...,n$, sampled every $\Delta t$ units of time. The exact state variables are computed by integrating the nonlinear ODEs using the fourth-order explicit Runge-Kutta (RK4) integrator with a tolerance of $10^{-10}$. In this work, we assume that the state variables are contaminated with independent zero-mean additive noise with variance $\sigma^2$. Specifically, for all the numerical examples, we employ white Gaussian noise model  given by $\epsilon_j \sim \mathcal{N}(0,\sigma^2),\quad j = 1,...,n$, and vary $\sigma$ to study different noise levels. We additionally study the effect of Gaussian distributed colored noise generated with a power law spectrum~\cite{timmer1995generating}.
	
	While here we work with a Gaussian white noise model, we note that more general anisotropic and  correlated noise can be easily incorporated through the generalized least squares formulation of the smoothing algorithms. We refer the interested reader to~\cite{kariya2004generalized} for a thorough discussion on the generalized least squares, and Section 4.6.2. of~\cite{ramsay2004functional} to estimate the weighting matrix involved in the method. To measure the signal magnitude relative to the noise level, we provide, in Appendix B, the signal-to-noise ratio (SNR) for each state $j$. 
	
	The sampling time is fixed to $\Delta t = 0.01$, and the number of samples is fixed to $201$, which are enough samples to capture the essential behavior of each dynamical system. We run 100 realizations for the training set for each of the numerical examples, and report the mean and variance of the errors. The state time-derivatives are computed over an interval that is 5\% (from each side) larger than the intended training time span, but only the data over the original training time is retained to compute $\hat{\bm\xi}$. To show the quality of the resulting state and derivative estimates and state predictions, we report the relative errors
	\begin{equation}\label{eq:relative_filter_errors}
		e_{X} = \frac{\|\hat{\mathbf{X}} - \mathbf{X}^* \|_F}{\|\mathbf{X}^*\|_F},\quad e_{\dot{X}} = \frac{\|\dot{\hat{\mathbf{X}}} - \dot{\mathbf{X}}^*\|_F}{\|\dot{\mathbf{X}}^*\|_F},
	\end{equation}
	where $\mathbf{X} = [\mathbf{x}(t_1), \mathbf{x}(t_2), ..., \mathbf{x}(t_m)] \in \R^{n \times m}$, $\norm{\cdot}_F$ is the Frobenius norm, and $\hat{(\cdot)}$ and $(\cdot)^*$ represent the estimated and true values, respectively.
	
	We approximate the governing equations of each example by a multivariate monomial basis of total degree $d$ in $n$ state variables. That is, $\phi_i(\mathbf{x})$ in Eqn.~(\ref{eq:dynExpansion}) are given by
	\begin{equation}
		\phi_i(\mathbf{x})=\prod_{j=1}^{n}x_j^{i_j} \quad \text{such that}\quad \sum_{j=1}^n i_j \le d, \quad i_j \in \mathbb{N}\cup\{0\},
	\end{equation}
	where the non-negative integer $i_j$ denotes the degree of the monomial in state variable $x_j$. The size of the approximation basis is then $p = \binom{n + d}{n}$. For all test cases, we set the total degree of basis $d$ to one more than the highest total degree monomial present in the governing equations. We run the WBPDN optimization procedure using solveBP routine in MATLAB from the open-source SparseLab 2.1 package~\cite{Donoho2009sparselab,donoho2007sparselab}, and employ the publicly available MATLAB codes for STLS\footnote{MATLAB routines for STLS can be found at \url{https://faculty.washington.edu/kutz/}}. We report the relative solution error defined as
	\begin{equation}\label{eq:relsolerror}
		e_{\xi} = \frac{\|\hat{\bm{\xi}} - \bm{\xi}^*\|_2}{\|\bm{\xi}^*\|_2},
	\end{equation}
	where $\hat{\bm{\xi}}$ and $\bm{\xi}^*$ are the approximate and exact solution vectors for each state variables, respectively.
	
	The considered nonlinear dynamical systems are the Lorenz 63 system as a base model for identifying chaotic dynamics, and the Duffing and Van der Pol oscillators as nonlinear stiffness and damping models. These dynamical systems are well-known reference problems and have been extensively studied in the literature.
	
	We provide in Tab.~\ref{table:filtering_codes} a compendium of available codes in MATLAB, Python and R for each of the smoothing methods. We highlighted in bold the codes we used in this article for smoothing splines and $\ell_1$-trend filtering. For local methods and Tikhonov smoother, we used our own implementation.
	
	\begin{table*}[h]
		\centering
		\caption{\label{table:filtering_codes} Compendium of filtering software codes and packages available in MATLAB, Python and R. We provide routine names and packages/toolboxes (enclosed in parentheses).}
		\begin{center}
		\begin{threeparttable}
		\begin{tabular}{lccc}
			\toprule
			Filter & \multicolumn{1}{c}{MATLAB} & \multicolumn{1}{c}{Python} & \multicolumn{1}{c}{R} \\
			\midrule
			\addlinespace[10pt]
			Savitzky-Golay & sgolayfilt (signal processing) & savgol\textunderscore filter (scipy) &  sgolayfilt (signal) \\
			LOWESS & lowess (curve fitting) &  lowess (lowess) &  lowess (stats) \\
			Tikhonov & hpfilter (econometrics) &  hpfilter (statsmodels) &  hpfilter (mFilter) \\ Ssplines & csaps (curve fitting) &  UnivariateSpline (scipy) &  \textbf{smooth.spline} (stats) \\
			$\ell_1$-trend filter & l1\textunderscore tf\tnote{1} &  $\ell_1$-trend filtering (CVXPY)\tnote{2} &  \textbf{trendfilter} (genlasso)\\ 
			\bottomrule
		\end{tabular}
		\begin{tablenotes}
			\fontsize{8}{12}\selectfont \item[1] Code available at: \url{https://stanford.edu/~boyd/l1\textunderscore tf/}\\
			\item[2] Implementation available at: \url{https://www.cvxpy.org/examples/applications/l1_trend_filter.html}
		\end{tablenotes}
		\end{threeparttable}
		\end{center}
	\end{table*}
	%
	
	
	\subsection{Lorenz 63 system}
	The Lorenz 63 system is a canonical model for nonlinear chaotic dynamics that was developed by Lorenz  as a simplified model for atmospheric convection \cite{Lorenz1963}. This system of nonlinear ODEs has been fundamental for the study of chaotic systems, wherein the future states of the system are highly sensitive to initial conditions. The state trajectories are three-dimensional, chaotic, deterministic, non-periodic and confined within a butterfly-shaped attractor, making them hard to predict. The Lorenz 63 model is given by the following system of first-order equations
	\begin{subequations}\label{eq:Lorenz63}
		\begin{alignat}{3}
			\dot{x}_1 = \gamma(x_2 - x_1), &\quad x_1(0) = x_{1,0},\label{eq:Lorenz63_a}\\
			\dot{x}_2 = x_1(\rho - x_3) - x_2, &\quad x_2(0) = x_{2,0},\label{eq:Lorenz63_b}\\
			\dot{x}_3 = x_1 x_2 - \beta x_3, &\quad x_3(0) = x_{3,0},\label{eq:Lorenz63_c}
		\end{alignat}
	\end{subequations}
	where the parameter values are set to $\gamma = 10$, $\rho = 28$ and $\beta = 8/3$, and the initial condition is $(x_{1,0},x_{2,0},x_{3,0}) = (-8,7,27)$. Assuming a total degree $d=3$ basis, for the right-hand-side of Eqn.~(\ref{eq:Lorenz63}) is described exactly by seven of the $p = 20$ monomials. We simulated the Lorenz 63 system from $t=0$ to $t=2.2$ time units to obtain the state trajectories. We then sampled the exact state variables at $\Delta t = 0.01$ resulting in $m=221$ samples, and perturbed them with 100 random noise realizations. We repeated the process for four different noise levels $\sigma$.
	
	Figure~\ref{fig:Lorenz63_filter_comparison} compares the relative state (top row) and state time-derivative (bottom row) errors defined in Eqn.~(\ref{eq:relative_filter_errors}) after running each of the local and global smoothing methods using GCV (left column) and Pareto curve criterion (right column) to automatically select the regularization parameter $\lambda$.
	\begin{figure}[H]
		\centering
		\includegraphics[trim = 0 0 00 0, clip,width=0.48\textwidth]{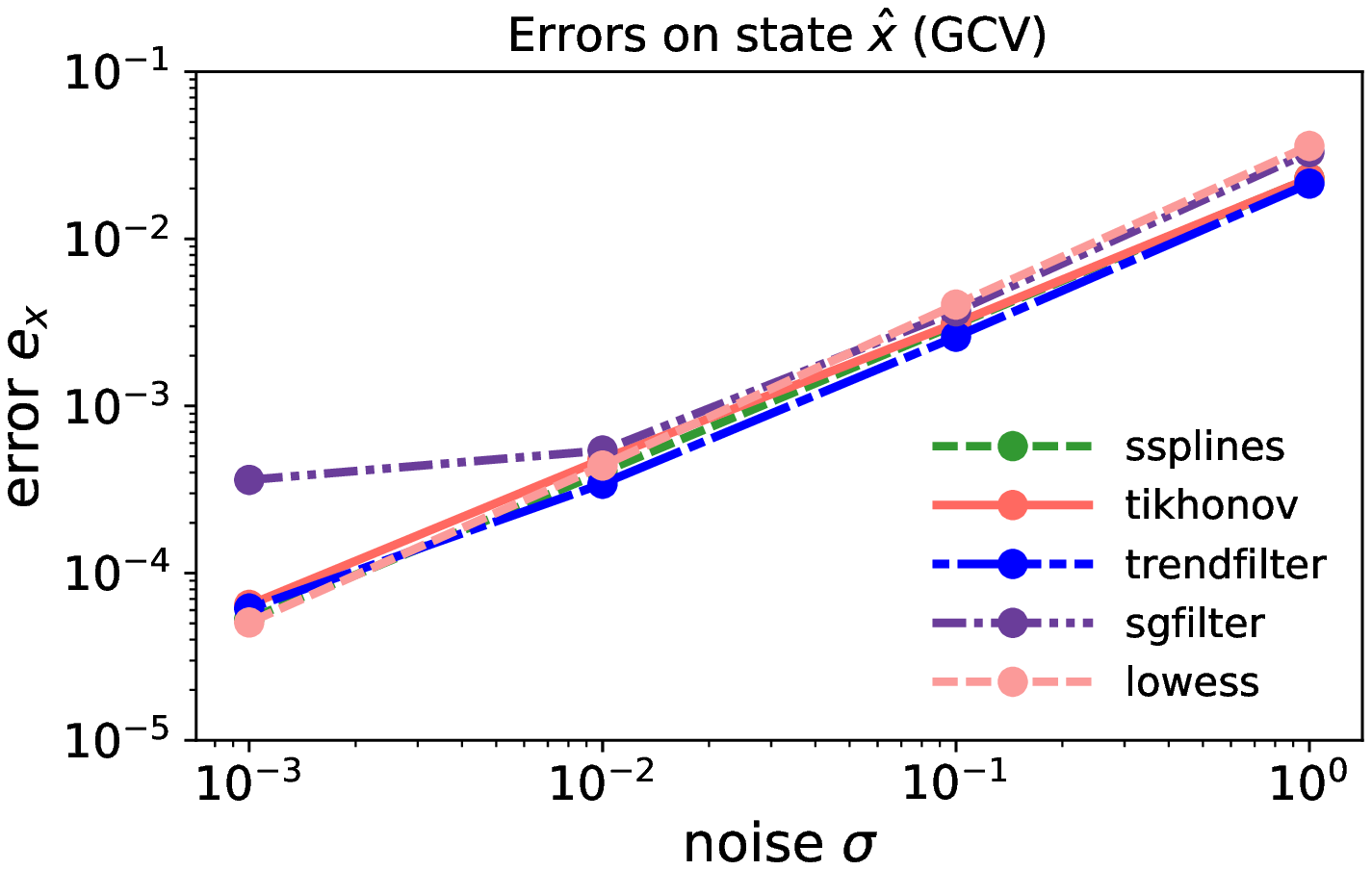}
		\includegraphics[trim = 0 0 00 0,
		clip,width=0.49\textwidth]{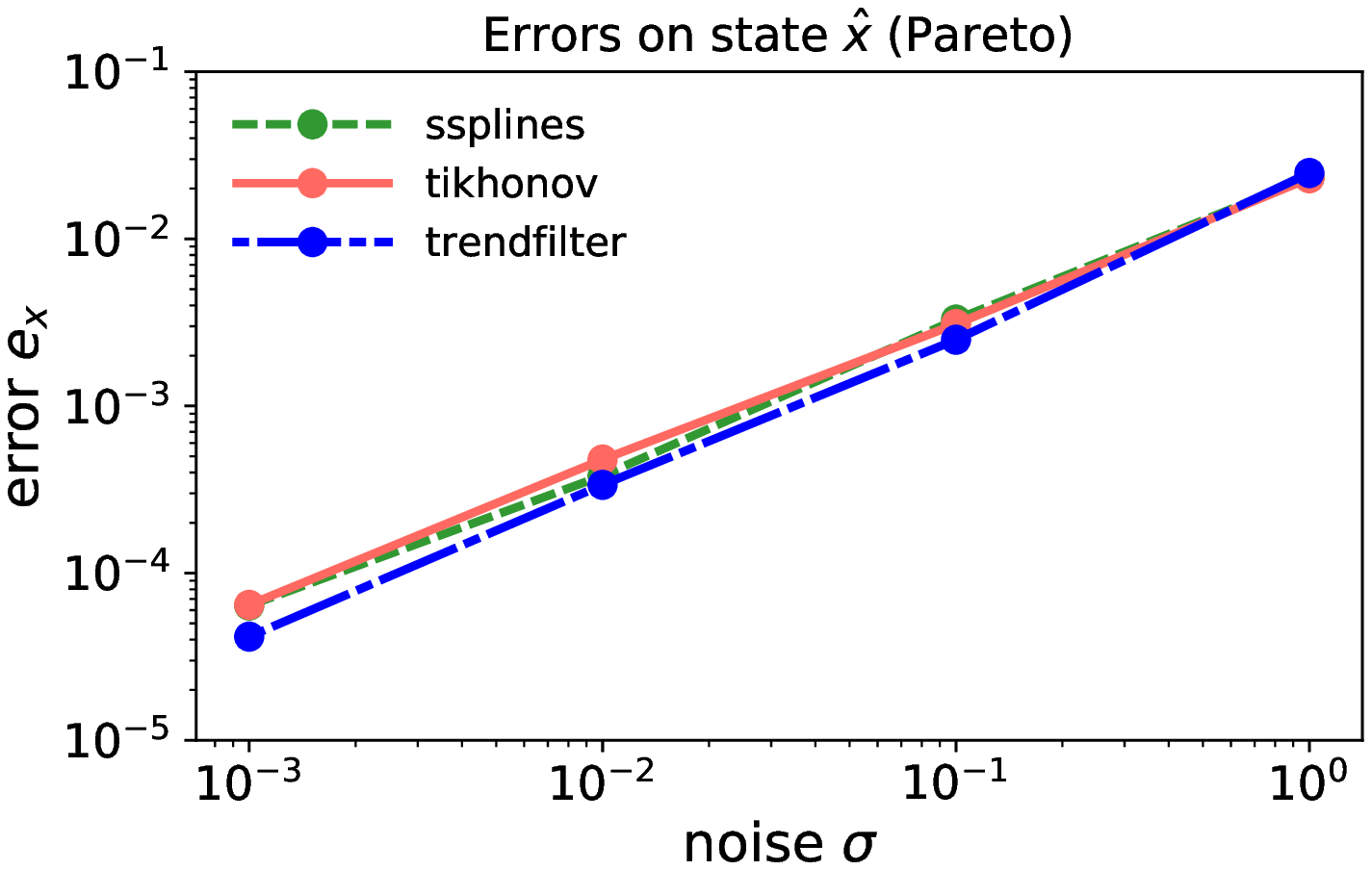}
		\includegraphics[trim = 0 0 00 0, clip,width=0.48\textwidth]{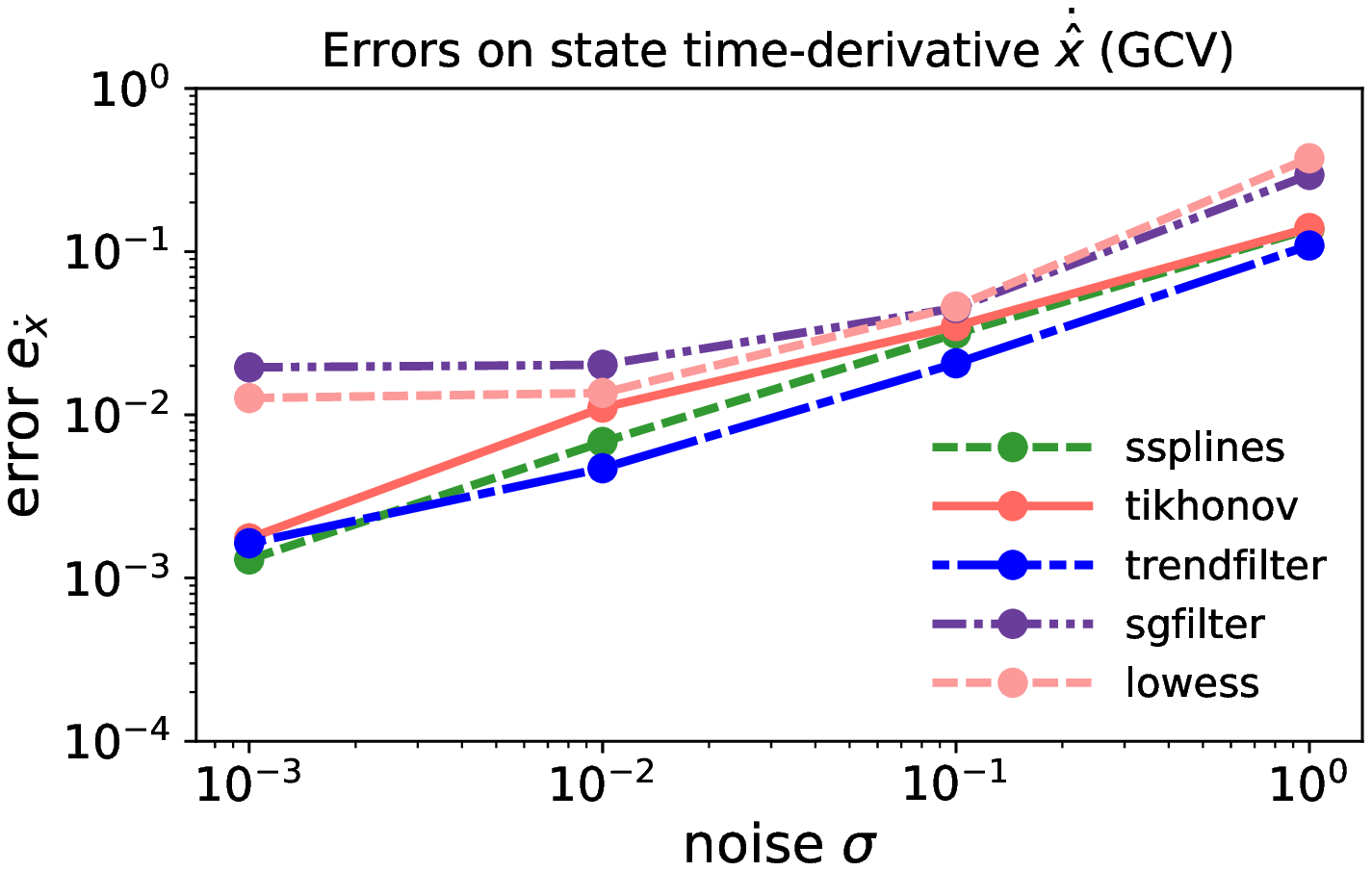}
		\includegraphics[trim = 0 0 00 0,
		clip,width=0.49\textwidth]{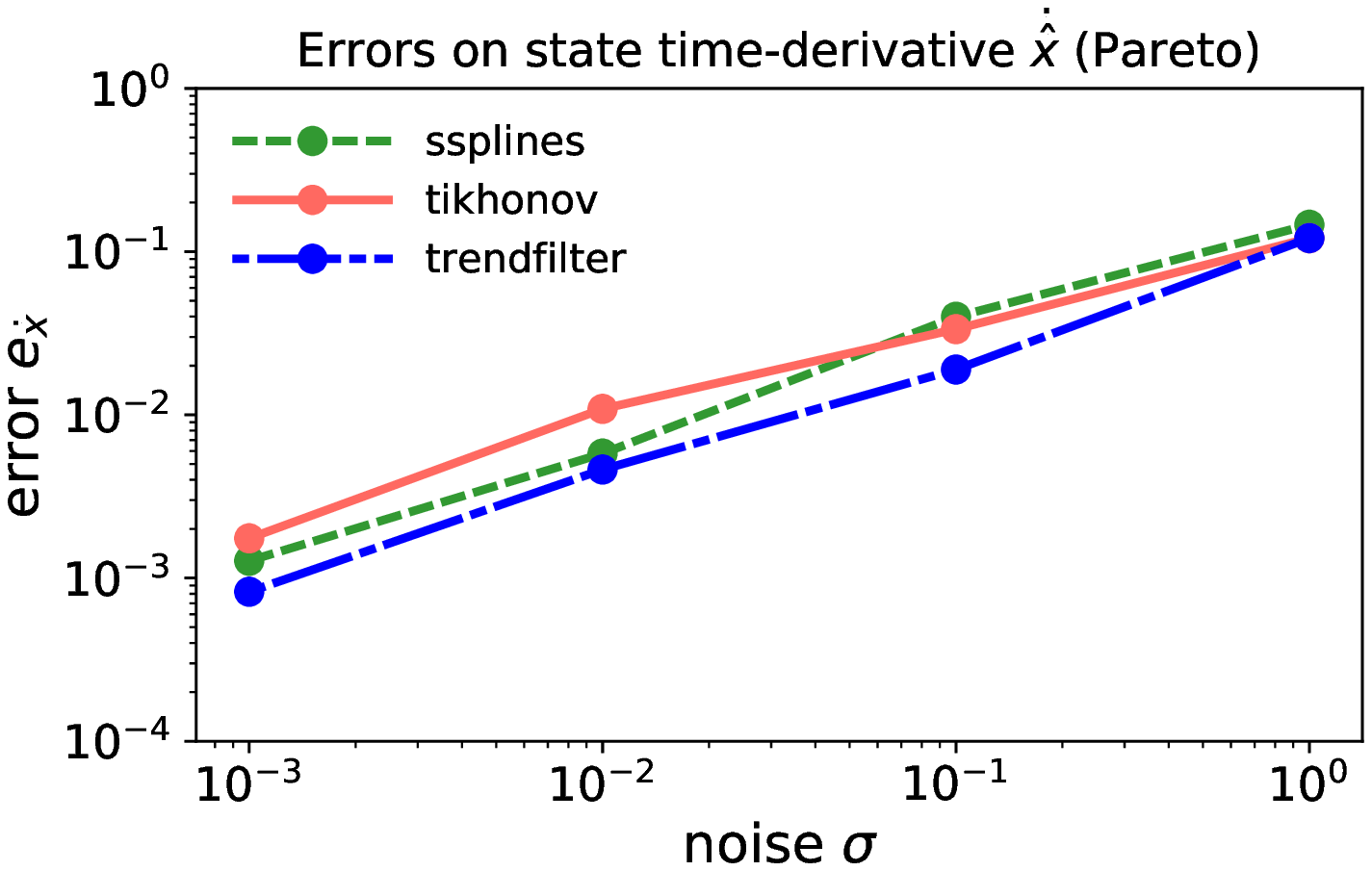}
		\caption{Comparison of the relative state (top row) and state time-derivative (bottom row) errors for the Lorenz 63 system using local and global smoothers. The regularization parameters were computed using GCV (left column) and the Pareto curve criterion (right column).}
		\label{fig:Lorenz63_filter_comparison}
	\end{figure}
	The error trend with respect to different noise levels behaves as expected, with a smooth error increase as noise level grows. We notice that, in general, global methods outperform local methods for all noise levels. This observation is aligned with the conclusions drawn in \cite{ahnert2007numerical}. Specifically, LOWESS with the optimal Epanechnikov kernel is superior to the Savitzky-Golay filter with a uniform kernel. For the estimation of state variables, the difference in accuracy between local and global methods is not as evident as in the the case of state time-derivative estimates. We also observe that more accurate state estimates yield more accurate state time derivatives for the different filters presented. In this example, global methods perform similarly, with $\ell_1$-trend filtering yielding the best results in general. As discussed in \cite{tibshirani2014adaptive}, the local adaptativity of $\ell_1$-trend filtering produces better estimates as compared to cubic smoothing splines. However, we do not notice significant superiority in terms of overall accuracy.
	
	Figure~\ref{fig:gcv_Lorenz63_SG_and_LOWESS} compares the GCV function defined in Eqn.~(\ref{eq:gcv_function}) for the Savisnky-Golay filter and LOWESS as a function of the global bandwidth $h$ at different noise levels. The green circles and black crosses correspond to the minimum error -- i.e. optimal bandwidth -- and the minimum of the GCV function, respectively. The general trend for both filters are consistent with noise levels: the bandwidth $h$ increases as the noise level is increased. The interpretation is that as noise levels increase, the local features of the signal are progressively lost and a larger neighborhood is needed to estimate the signal. Reference~\cite{little2017multiscale} provides an explanation of this phenomenon from a geometric viewpoint. However, GCV tends to select larger bandwidths than the optimal ones, thus yielding over-smoothed estimates.
	
	\begin{figure}[H]
		\centering
		\includegraphics[trim = 10 0 10 0, clip,width=0.34\textwidth]{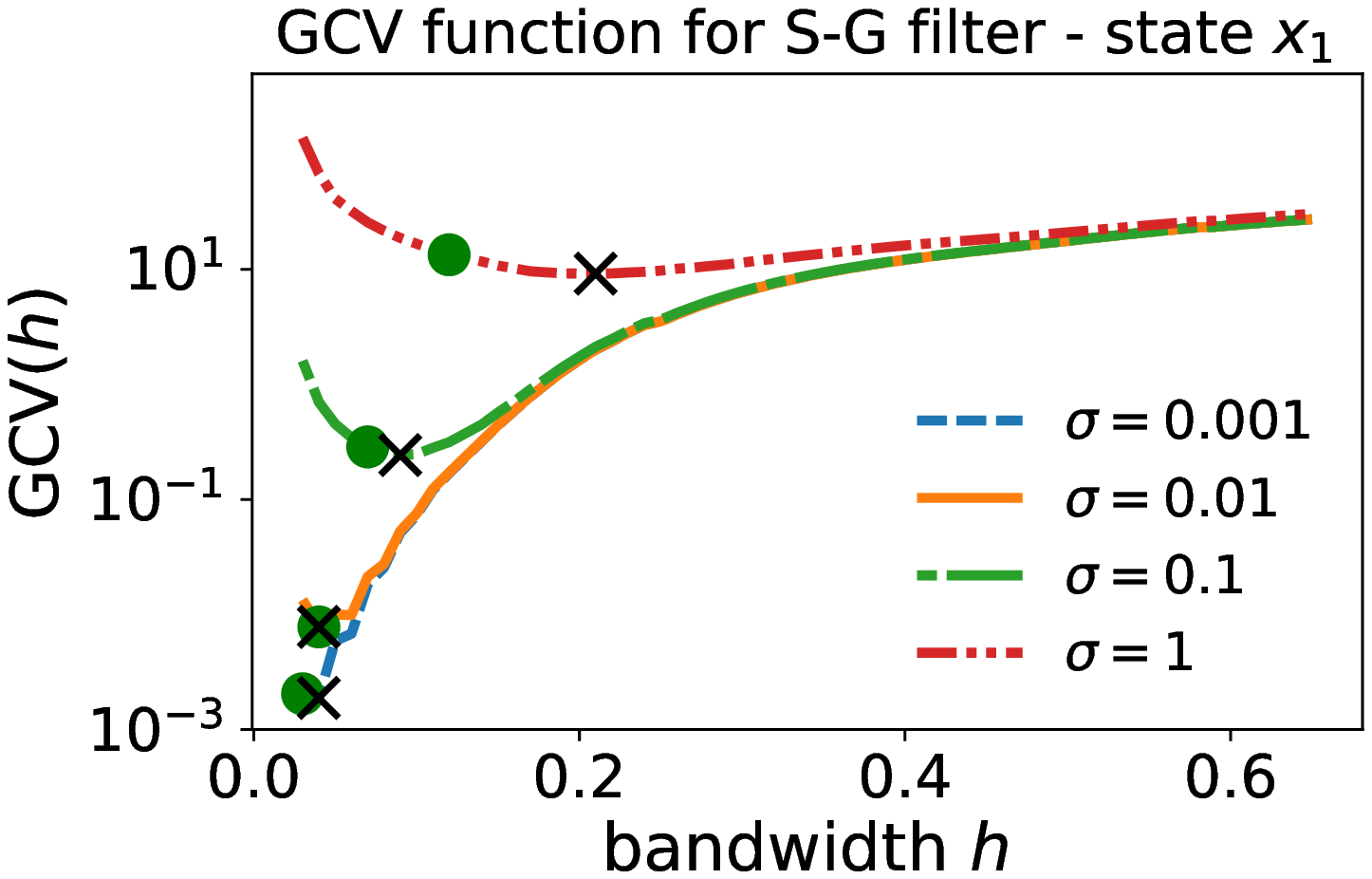}
		\includegraphics[trim = 30 0 10 0, clip,width=0.32\textwidth]{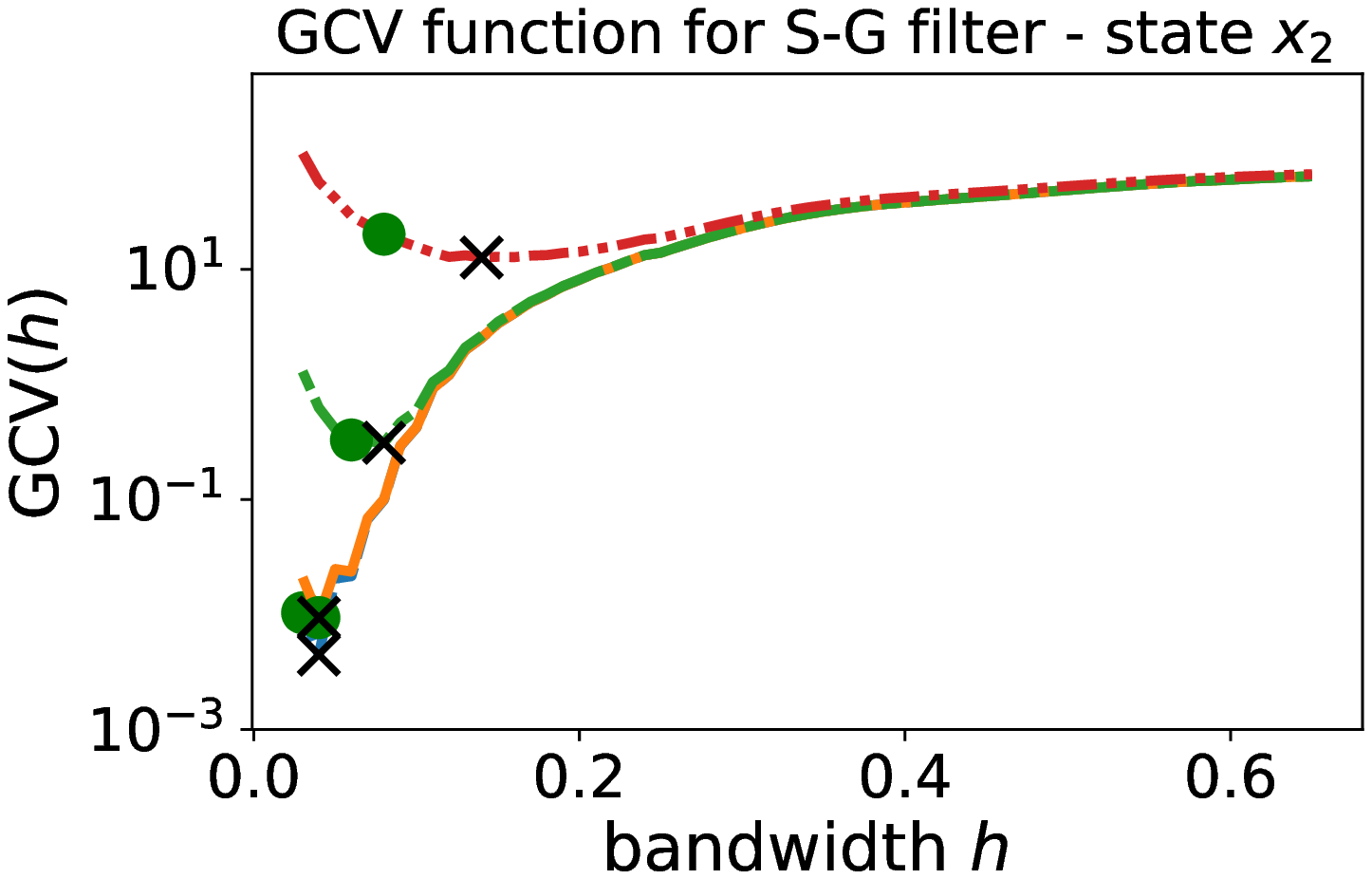}
		\includegraphics[trim = 30 0 10 0, clip,width=0.32\textwidth]{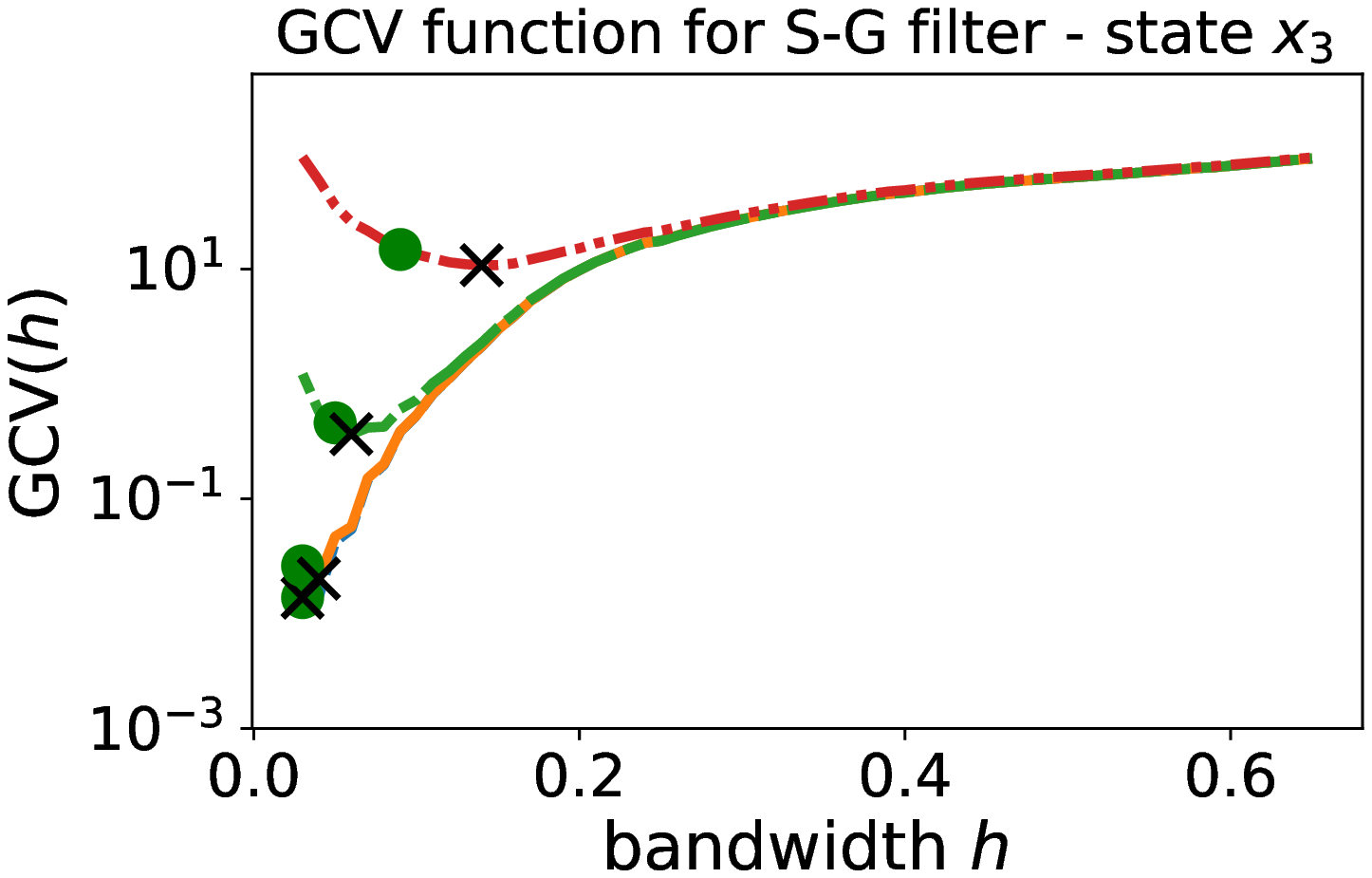}
		\includegraphics[trim = 10 0 10 0, clip,width=0.34\textwidth]{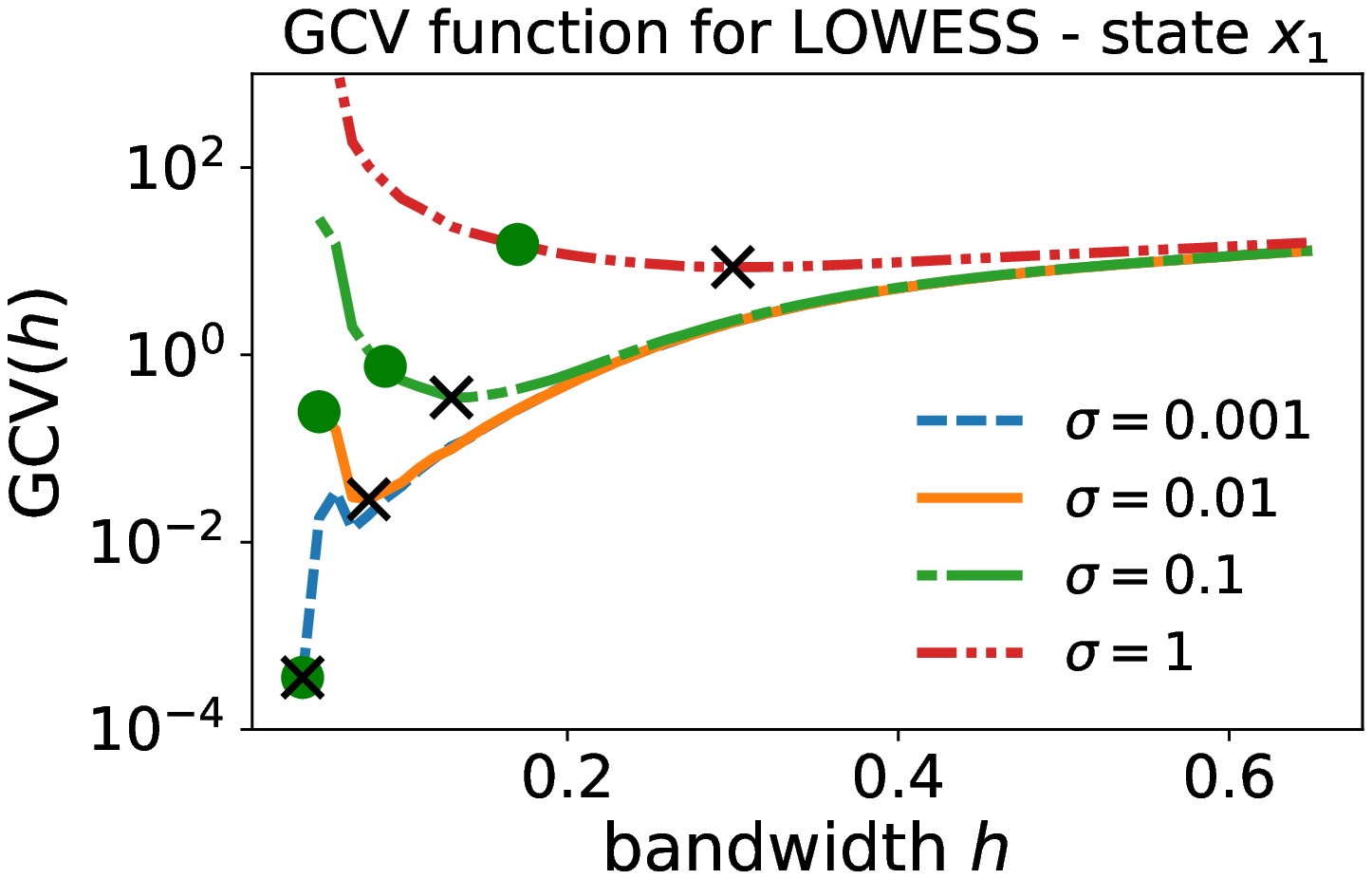}
		\includegraphics[trim = 30 0 10 0, clip,width=0.32\textwidth]{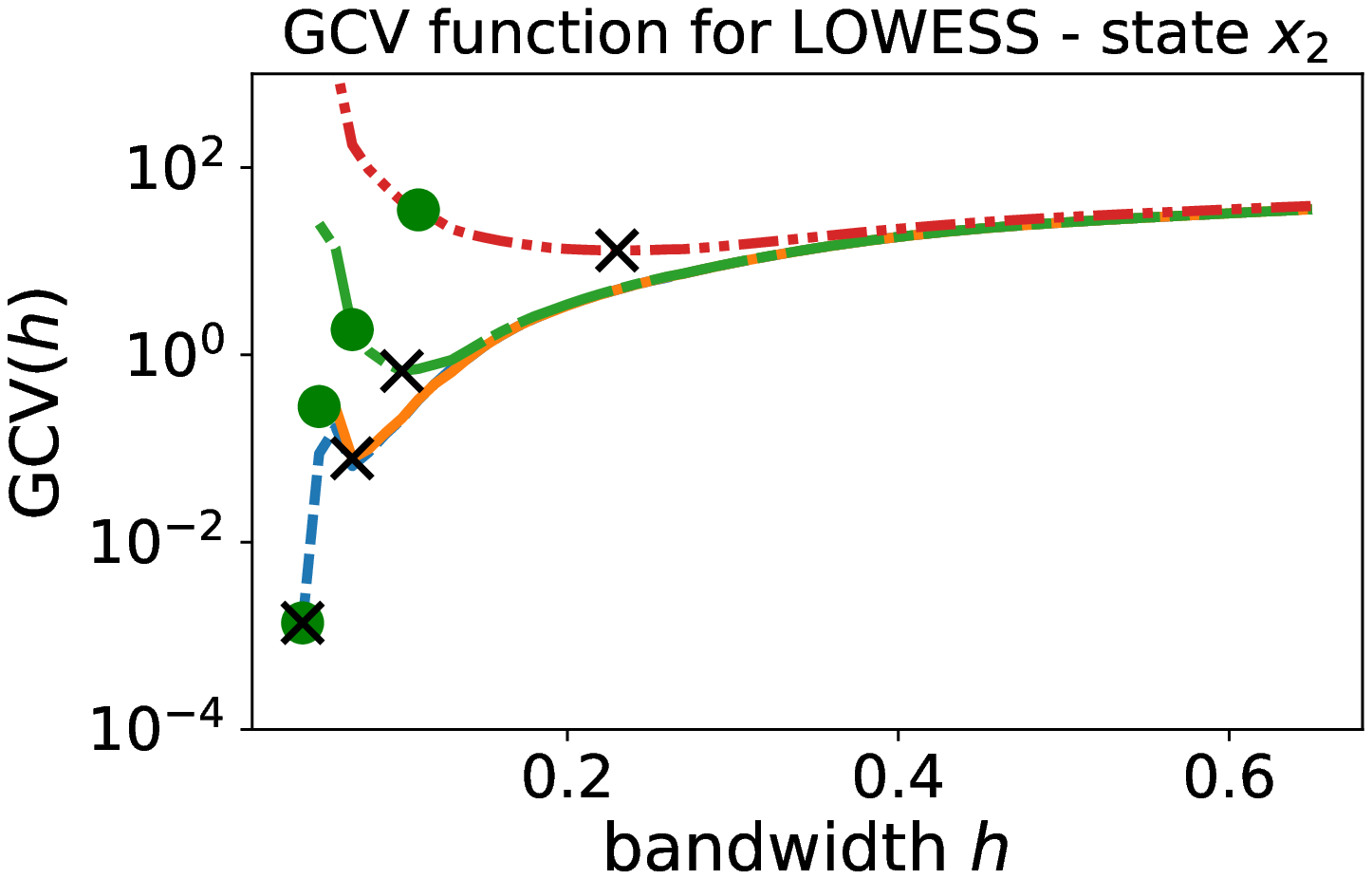}
		\includegraphics[trim = 30 0 10 0, clip,width=0.32\textwidth]{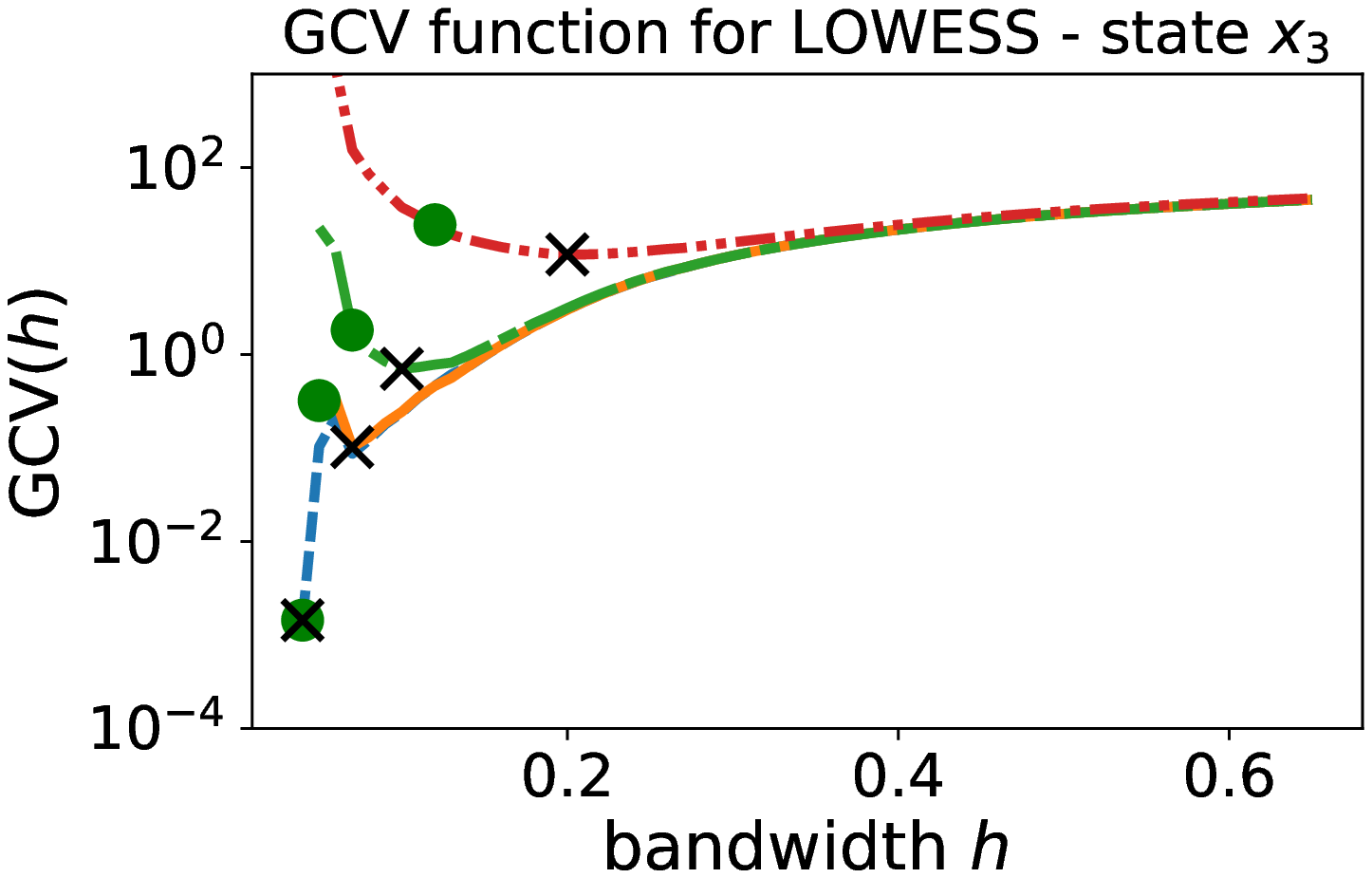}
		\caption{GCV functions for each state variable of Lorenz 63 system using Savitzky-Golay and LOWESS filters at different noise levels for an arbitrary realization. The green circles represent the global bandwidth $h$ that yield the minimum estimation error defined in Eqn.~(\ref{eq:relative_filter_errors}). Black crosses represent the  minima of the GCV functions for the GCV selection strategy.}
		\label{fig:gcv_Lorenz63_SG_and_LOWESS}
	\end{figure}
	
	For the global methods, we also illustrate the behavior of GCV and Pareto curves to select near optimal smoothing parameters $\lambda$. Figure~\ref{fig:ssplines_Lorenz63_Pareto_and_GCV} shows the Pareto curves (top panels) and GCV functions (bottom panels) for smoothing splines at different noise levels for one state trajectory realization. {\color{black} In the case of Pareto curve criterion, the crosses represent the corners of the L-shaped curves found via the algorithm in~\cite{cultrera2020simple}. In the case of GCV criterion, the crosses represent the minimum of the GCV function computed via Eqn.~(\ref{eq:gcv_function}). In both criteria, the green dots represent the $\lambda$ that yield the lowest relative coefficient error $e_{\xi}$}. The first observation is that the Pareto curves consistently select larger $\lambda$ -- smoothing increases as we move along the Pareto curve from top to bottom -- as we increase the noise levels. We noticed that the L-shaped structure becomes clearer and sharper in regions between low and high noise level extremes. Overall, in all noise cases, the $\lambda$ corresponding to the corner points of the Pareto curves almost coincide with the optimal ones. In the GCV case, we observe a flat region, where the minimum is located, and a steep region, where over-smoothing errors become more noticeable. As with Pareto curves, GCV also results in a good model selection strategy for selecting the smoothing parameter $\lambda$. We omit the plots for Tikhonov smoother since it yields similar trends and observations. 
	\begin{figure}[H]
		\centering
		\includegraphics[trim = 10 0 10 0, clip,width=0.34\textwidth]{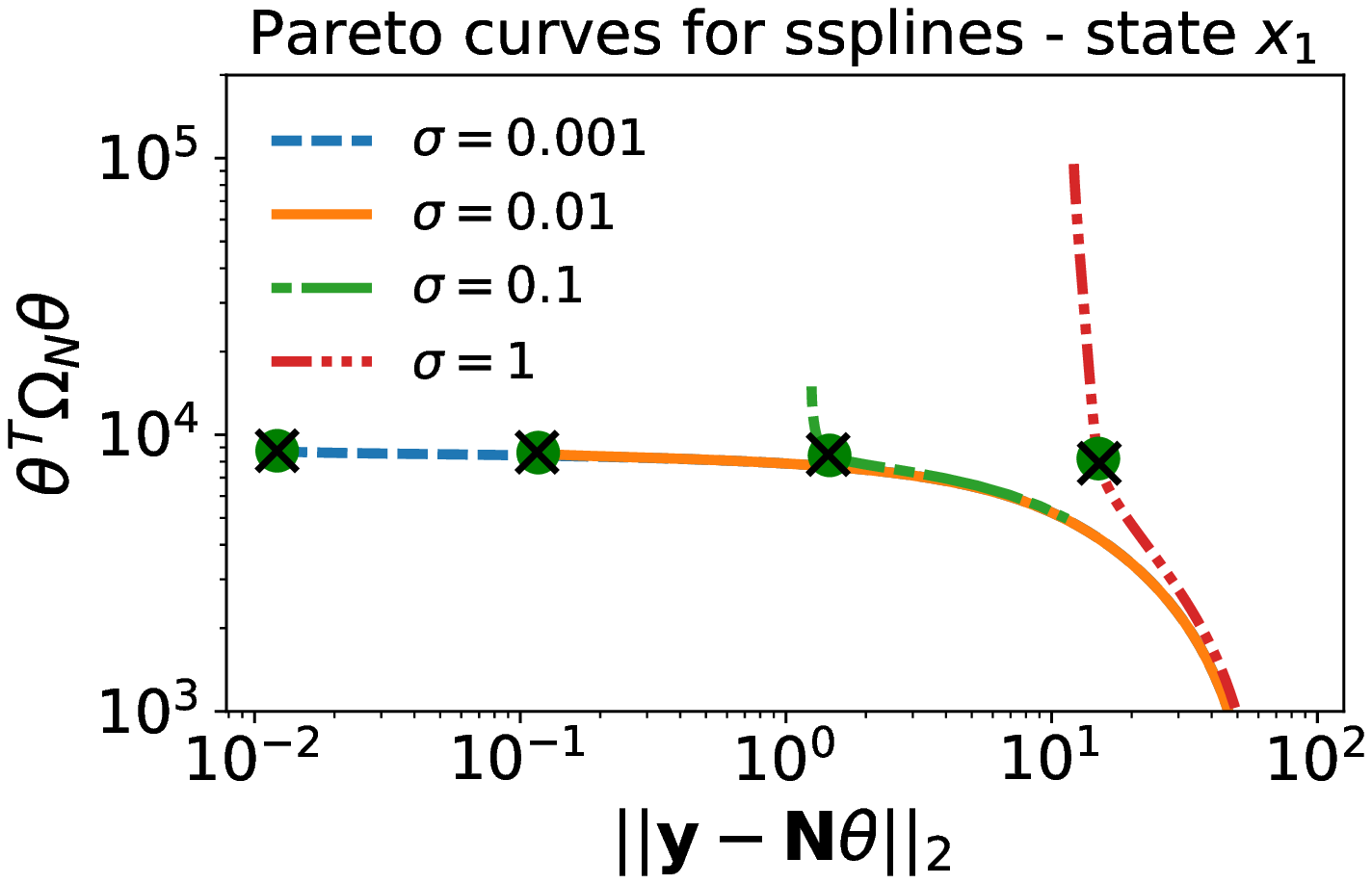}
		\includegraphics[trim = 35 0 10 0, clip,width=0.32\textwidth]{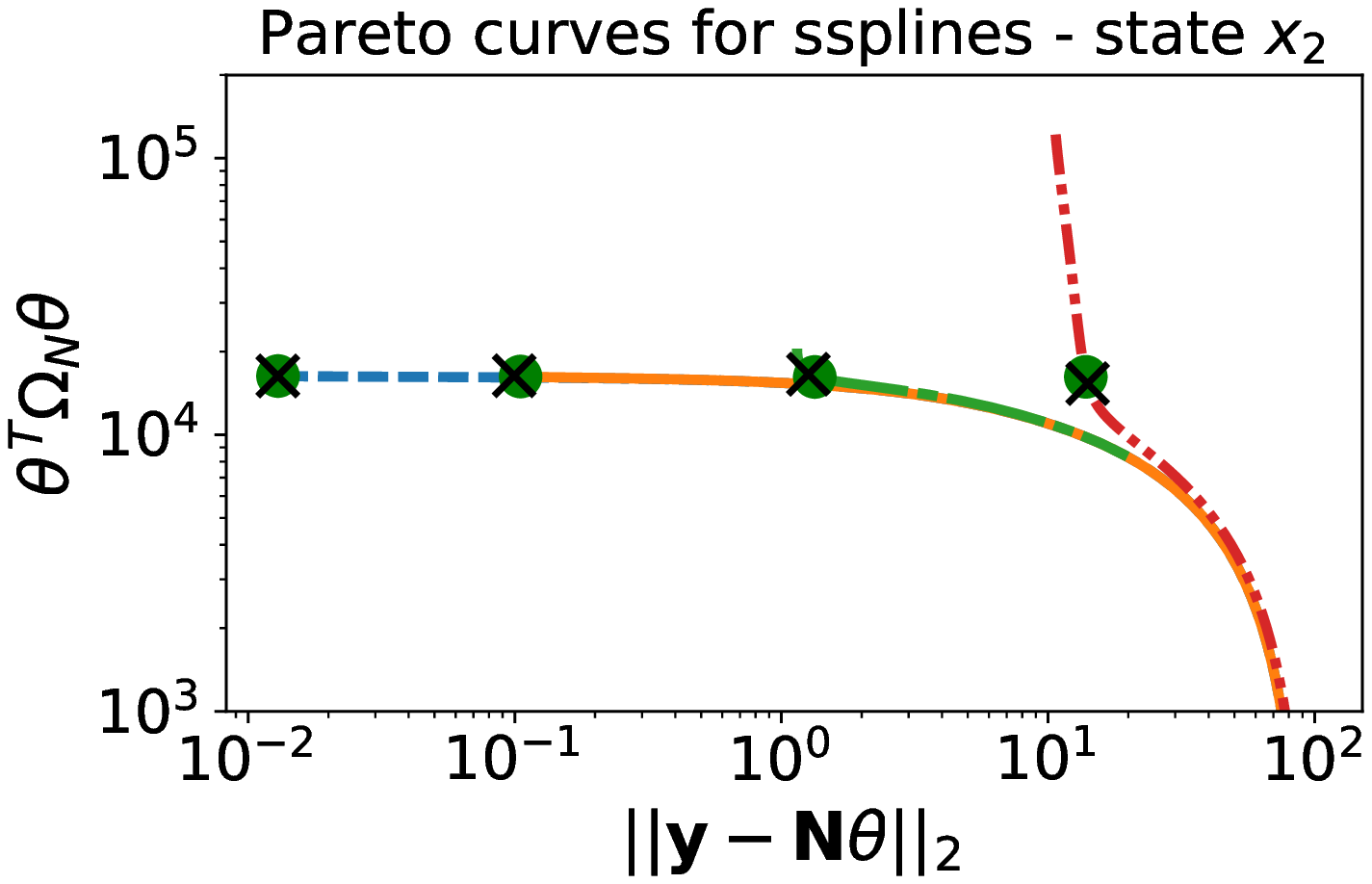}
		\includegraphics[trim = 35 0 10 0, clip,width=0.32\textwidth]{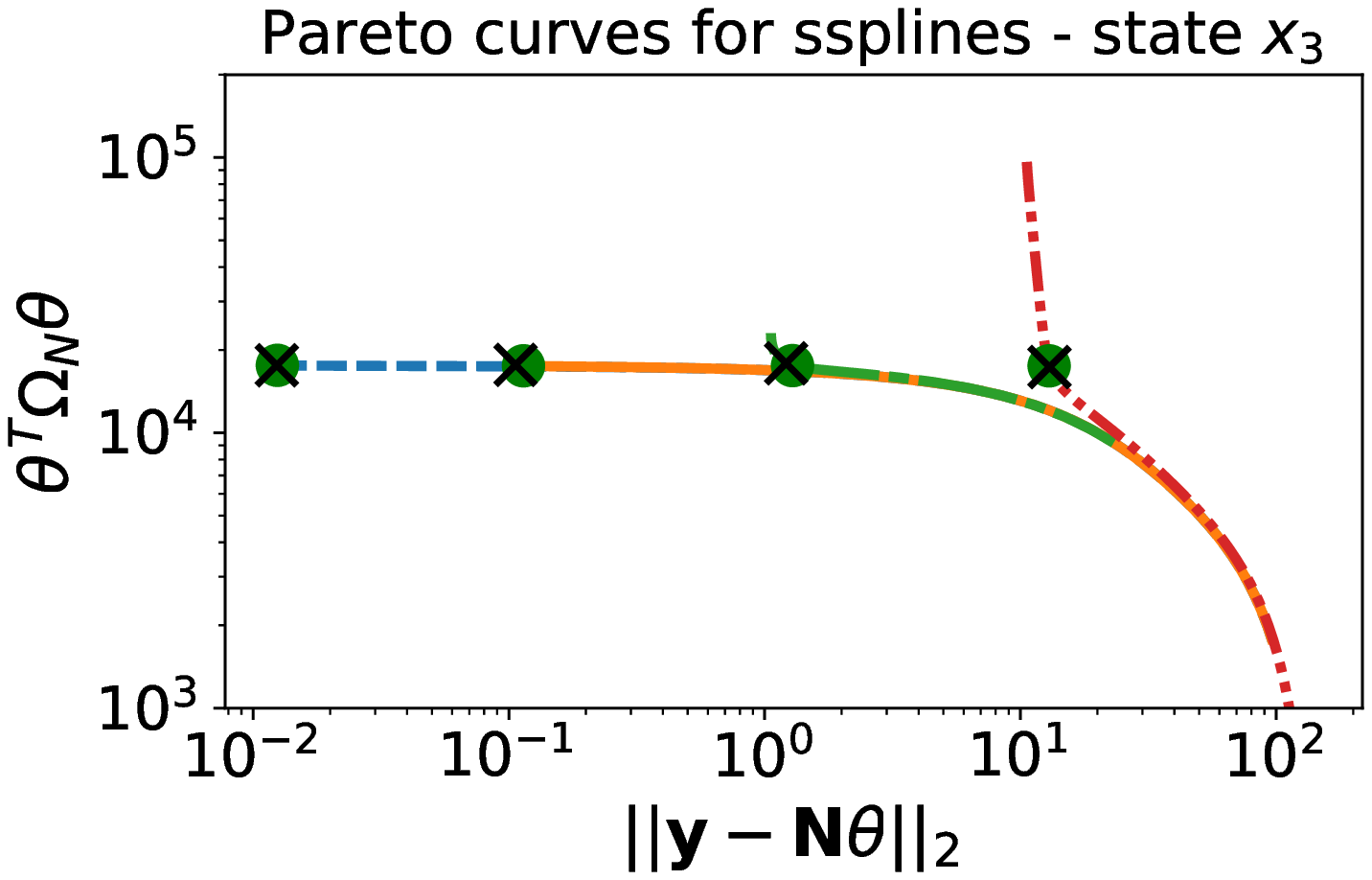}
		\includegraphics[trim = 10 0 10 0, clip,width=0.34\textwidth]{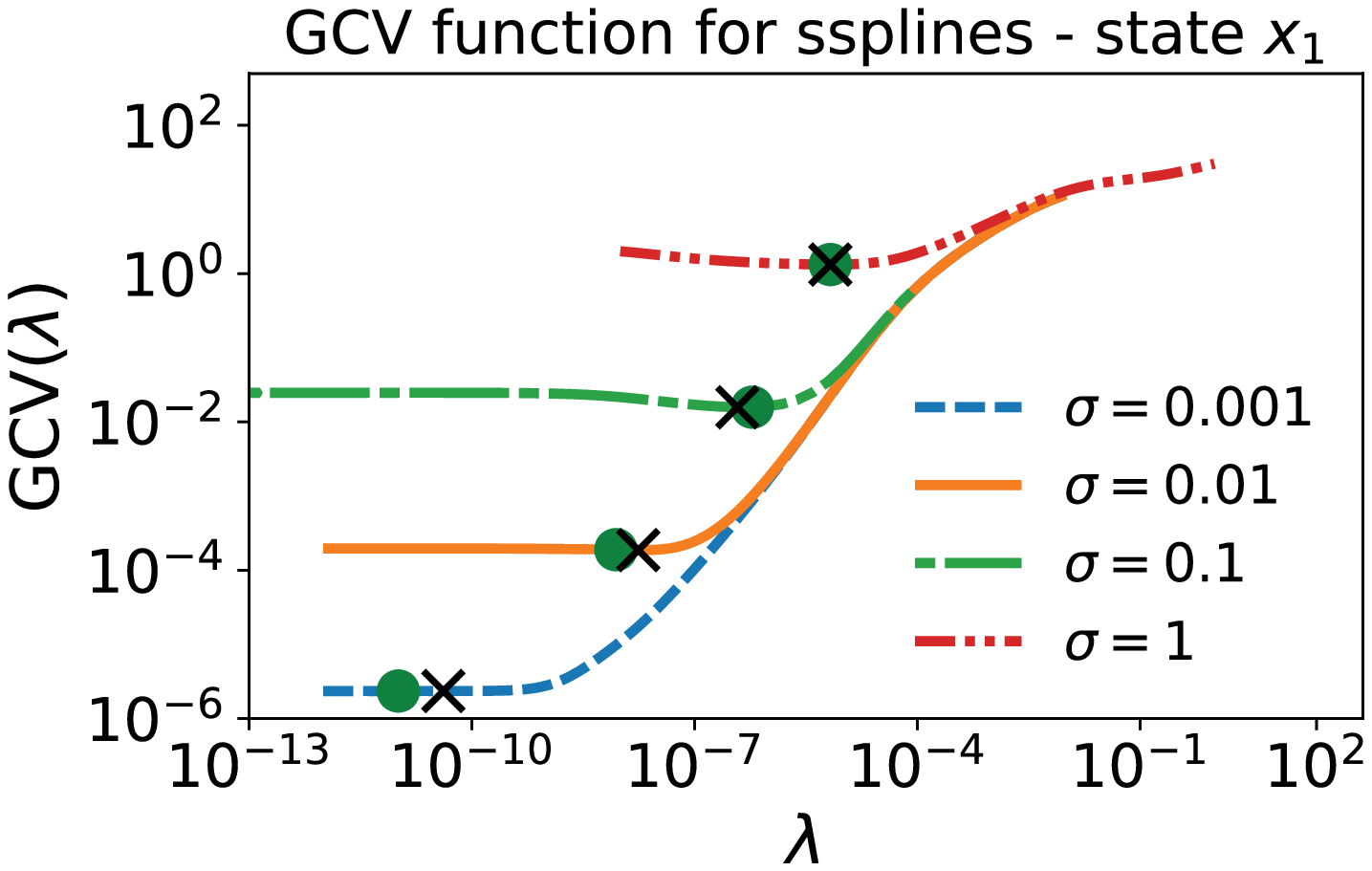}
		\includegraphics[trim = 35 0 10 0, clip,width=0.32\textwidth]{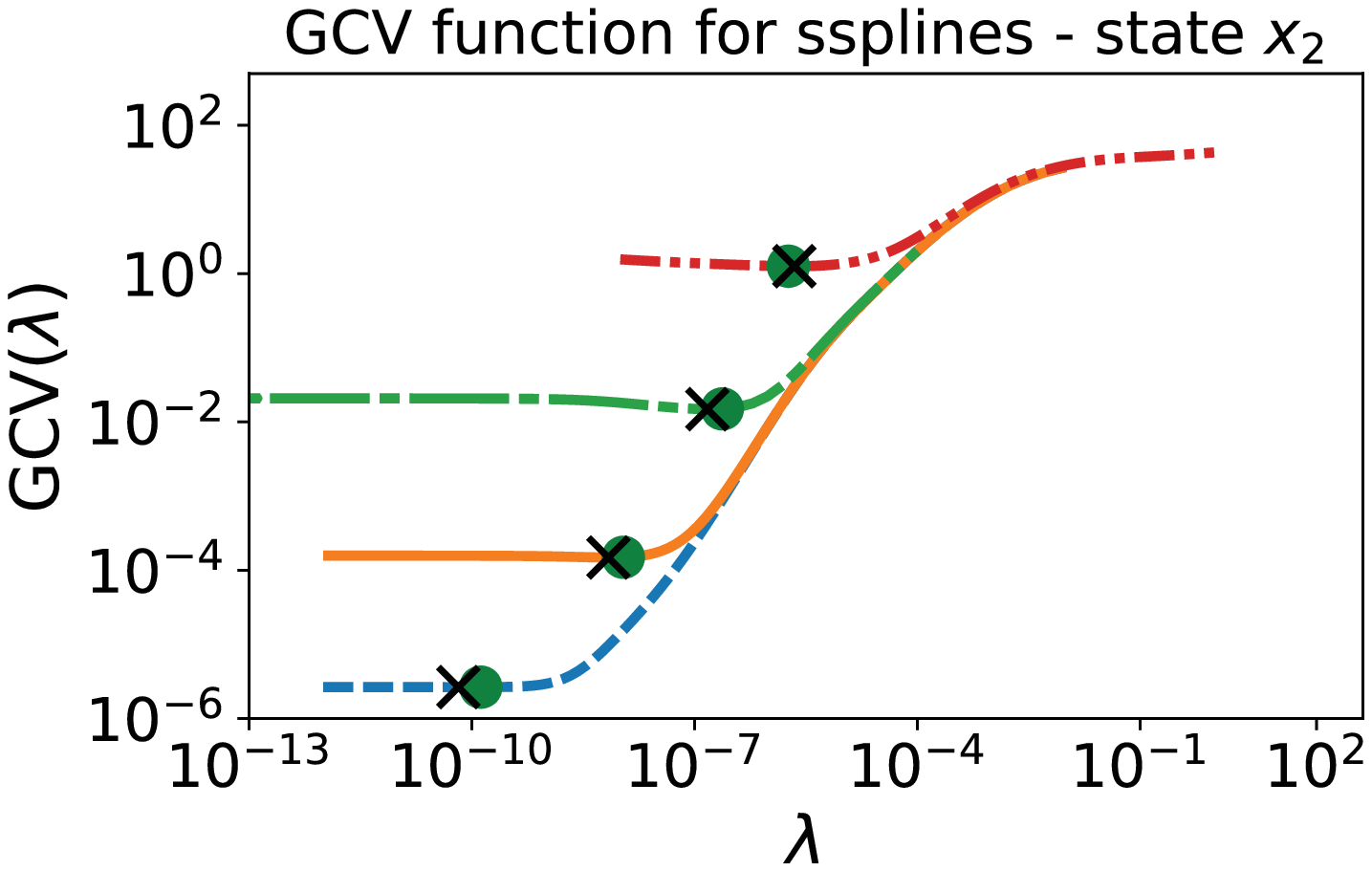}
		\includegraphics[trim = 35 0 10 0, clip,width=0.32\textwidth]{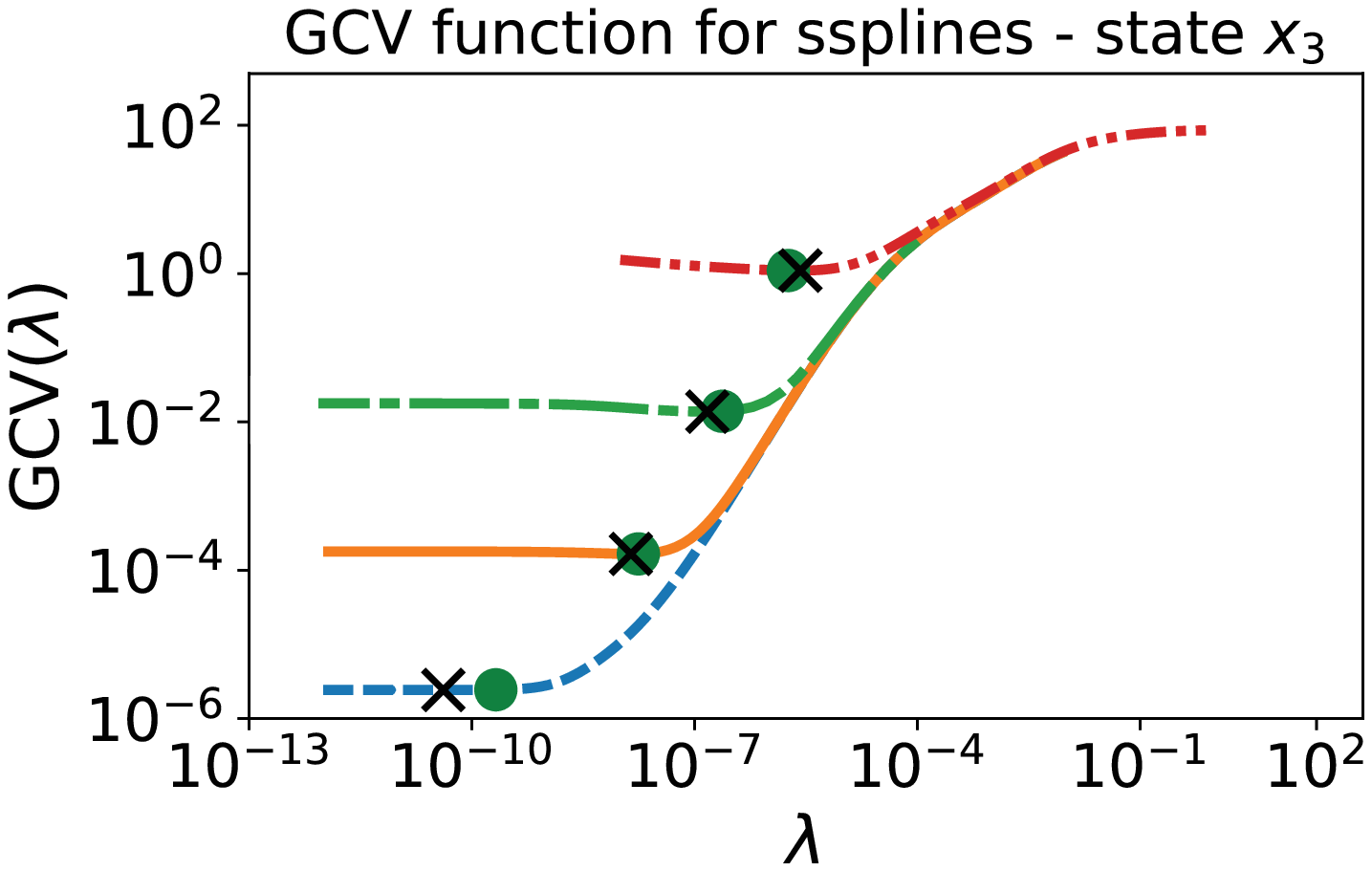}
		\caption{Pareto curves (top panels) and GCV functions (bottom panels) for each state variable of Lorenz 63 systems using smoothing splines at different noise levels for an arbitrary realization. The green circles represent the regularization parameters $\lambda$ that yield the minimum estimation error defined in Eqn.~(\ref{eq:relative_filter_errors}). Black crosses represent the converged corner points for Pareto curves (using the iterative algorithm in \cite{cultrera2020simple}) and the minima of the GCV functions for GCV. Note that the corner on Pareto curves for small noise levels is not visible because of the scale of the plot; however, the corner exists if one amplifies the plot around that region.}
		\label{fig:ssplines_Lorenz63_Pareto_and_GCV}
	\end{figure}
	The Pareto curves for $\ell_1$-trend filtering, as shown in Fig.~\ref{fig:trendfilter_Lorenz63_Pareto_and_GCV} (top panels), become better defined than the smoothing splines and Tikhonov smoother cases, yielding smoothing parameters  closer to the optimal ones. However, in the case of GCV, we notice that the behavior of the GCV function around the region of small $\lambda$ is inconsistent and may yield minima resulting in under-smoothed state estimates. We also observed that by removing the small $\lambda$ region corresponding to the noisy GCV behavior, the $\lambda$ corresponding to the new minima matched the optimal ones well (see, for example, the $\sigma = 0.1$ case for state $x_1$ in Fig.~\ref{fig:trendfilter_Lorenz63_Pareto_and_GCV}). The results reported for $\ell_1$-trend filtering using GCV in Fig.~\ref{fig:Lorenz63_filter_comparison} were obtained after truncating the region of small $\lambda$ with irregularities in the GCV function.
	\begin{figure}[H]
		\centering
		\includegraphics[trim = 10 0 10 0, clip,width=0.34\textwidth]{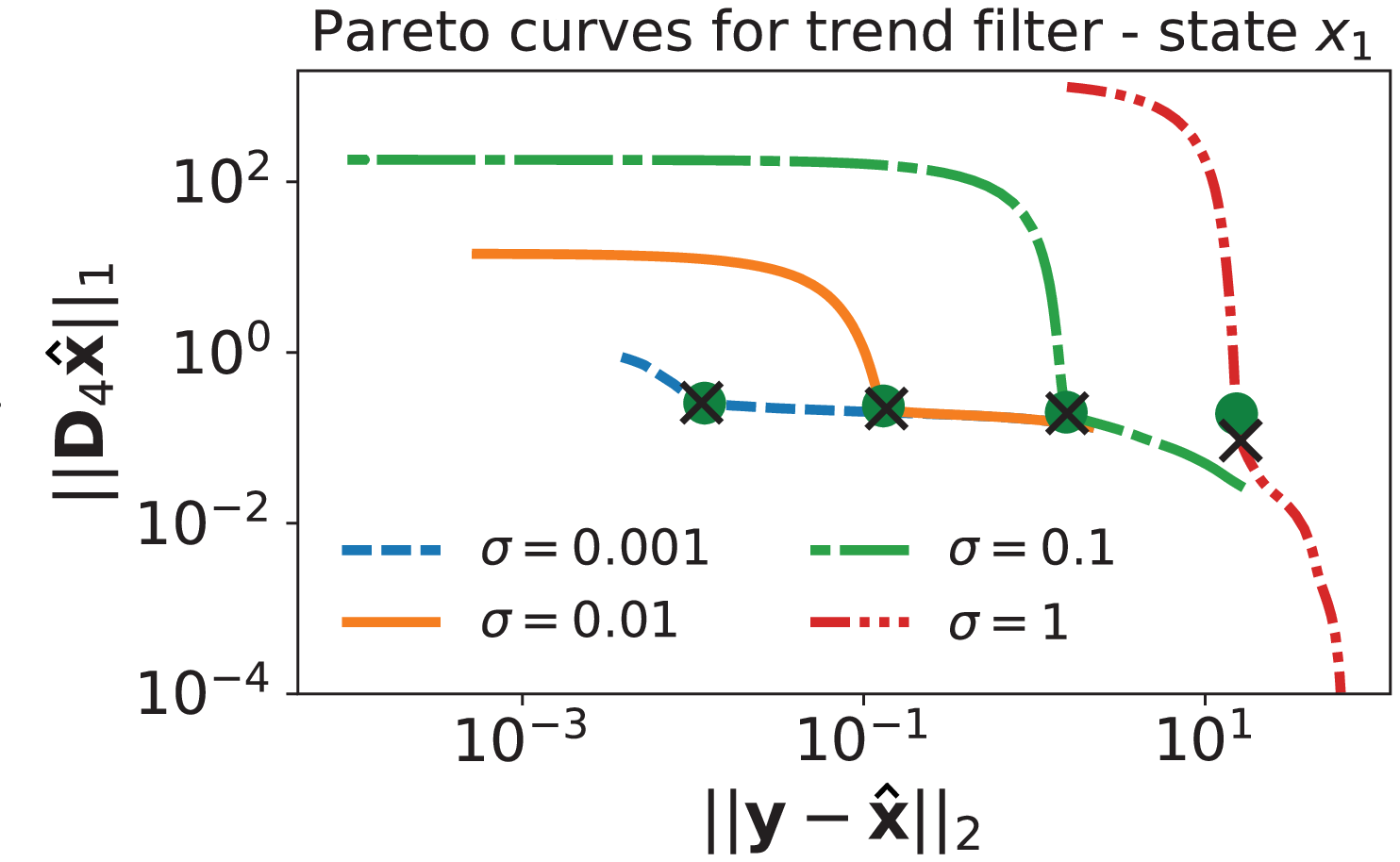}
		\includegraphics[trim = 35 0 10 0, clip,width=0.32\textwidth]{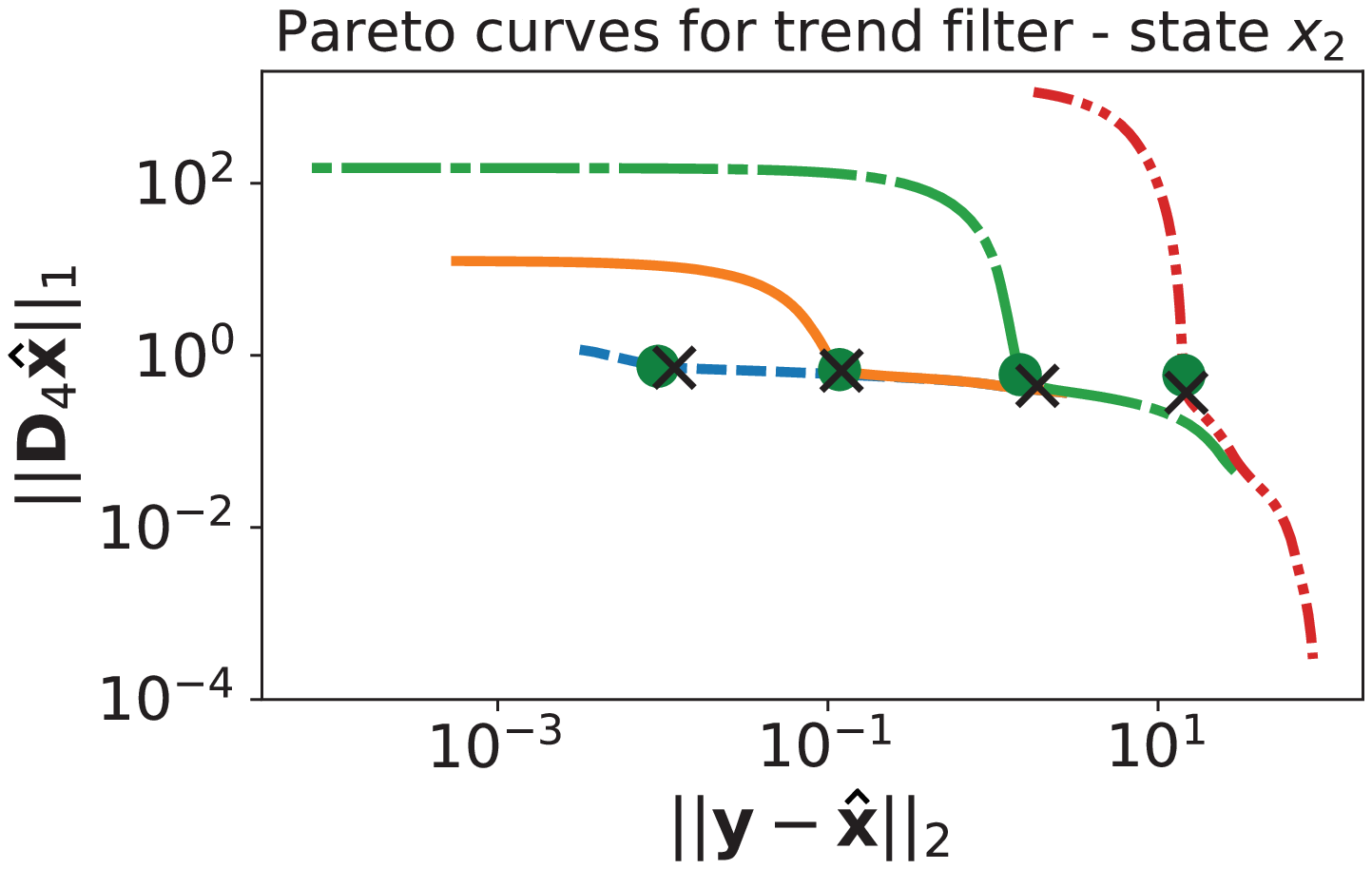}
		\includegraphics[trim = 35 0 10 0, clip,width=0.32\textwidth]{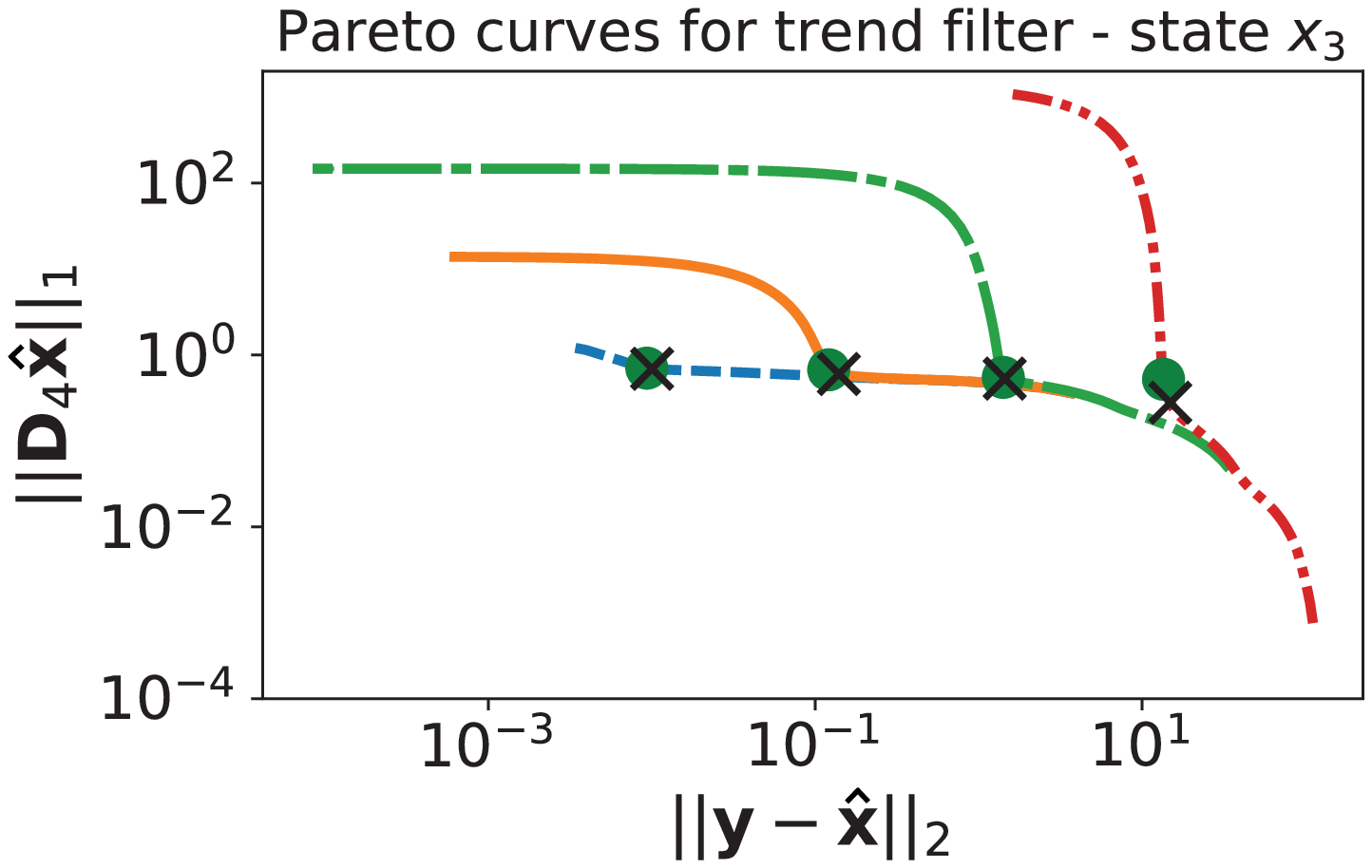}
		\includegraphics[trim = 10 0 10 0, clip,width=0.34\textwidth]{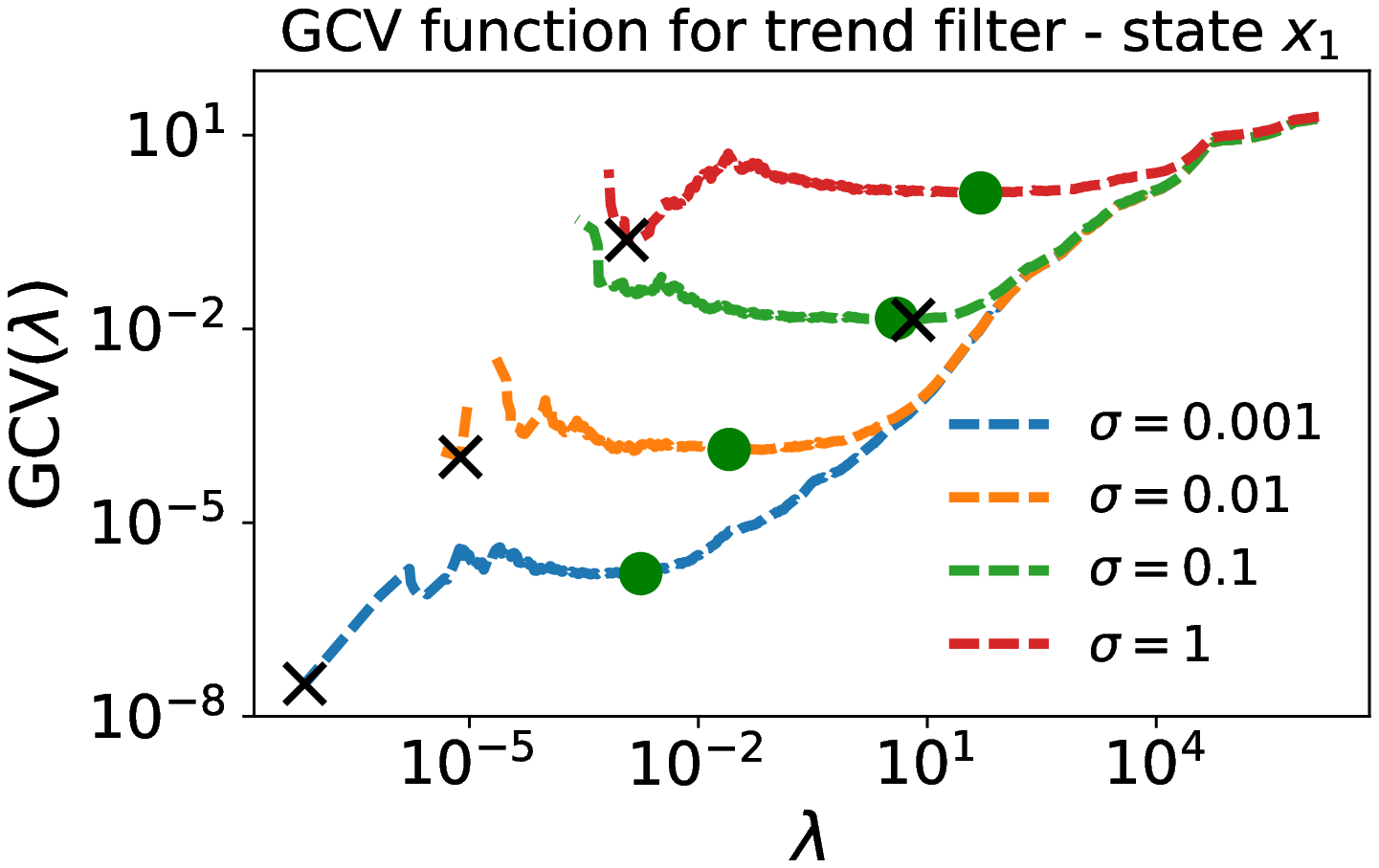}
		\includegraphics[trim = 35 0 10 0, clip,width=0.32\textwidth]{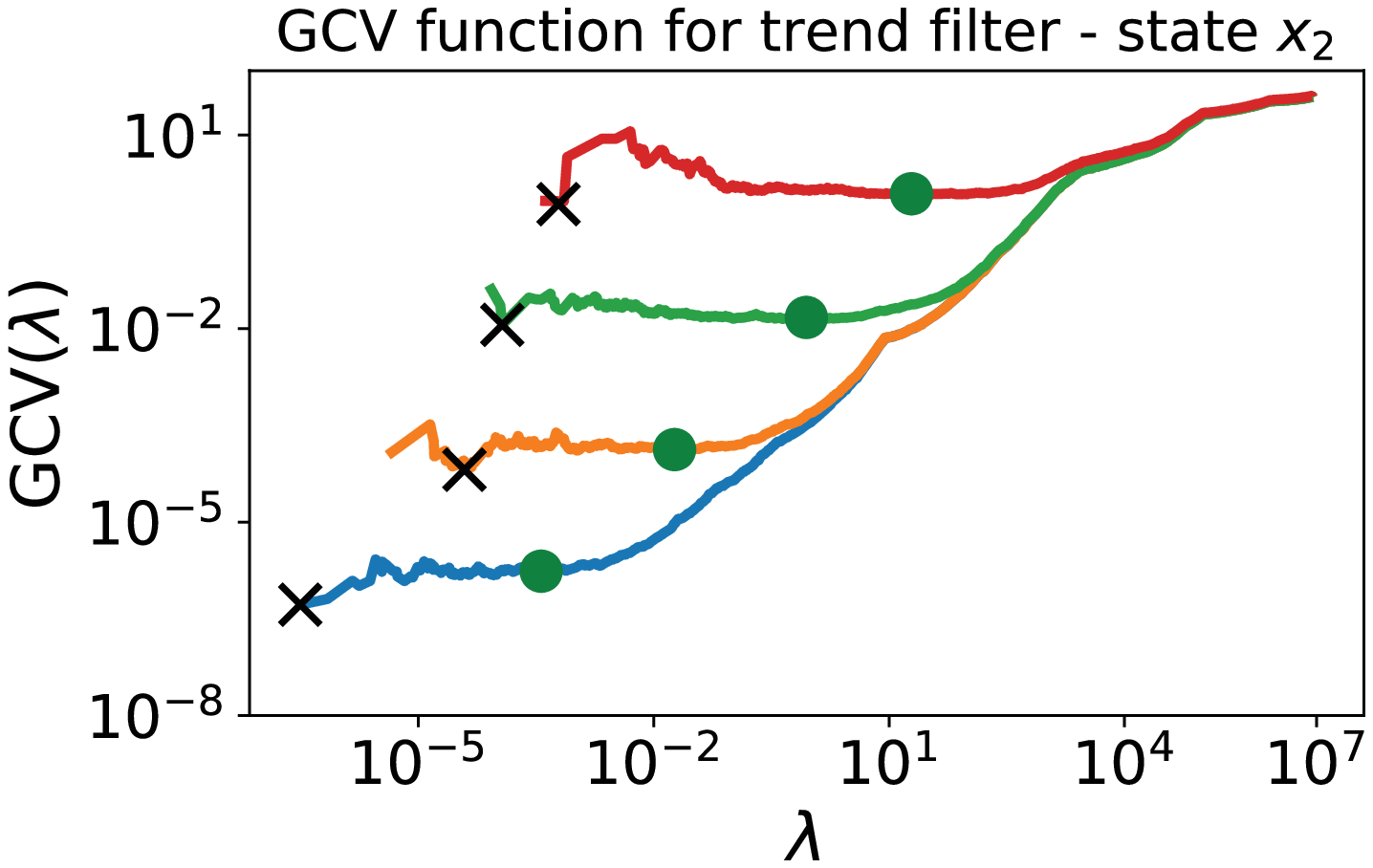}
		\includegraphics[trim = 35 0 10 0, clip,width=0.32\textwidth]{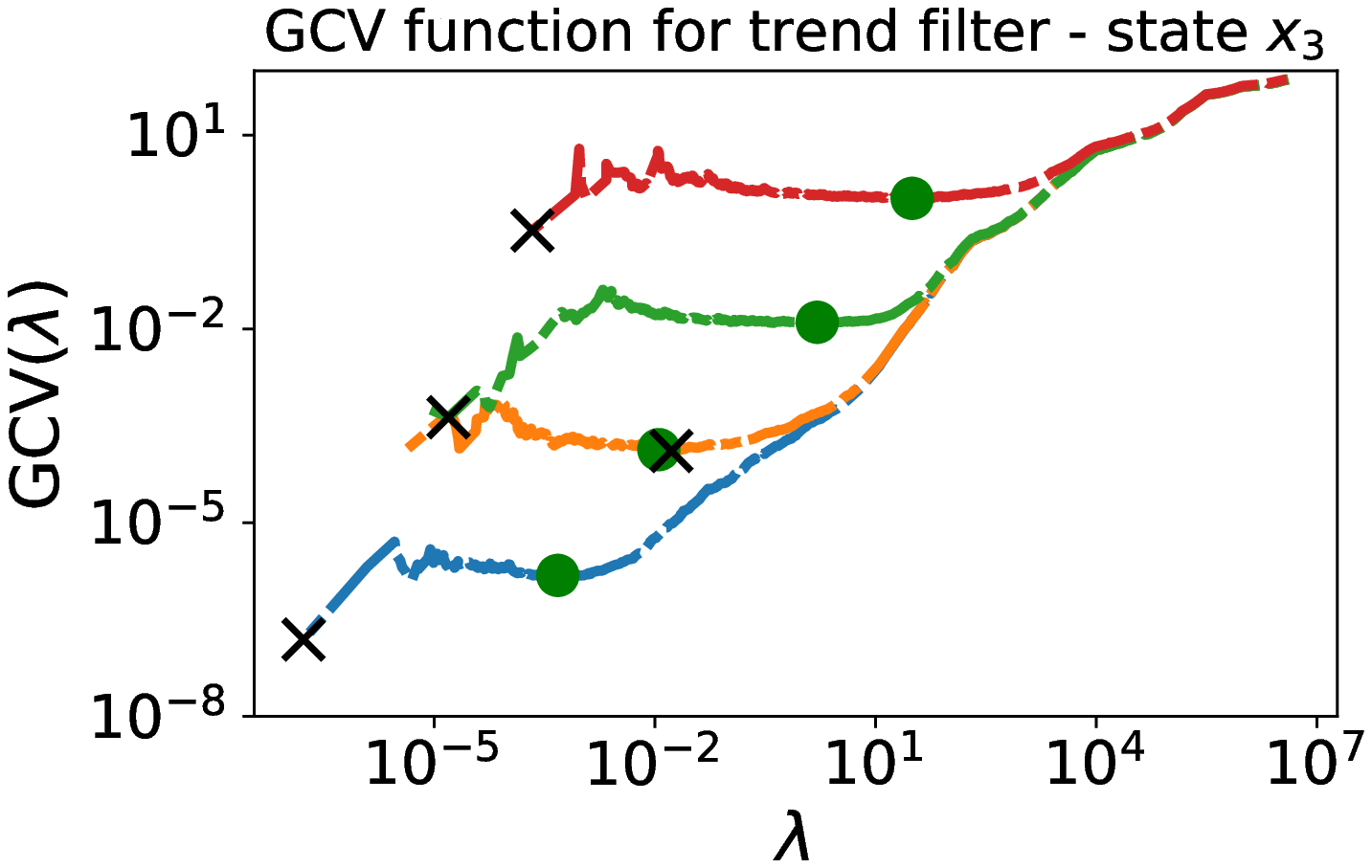}
		\caption{Pareto curves (top panels) and GCV functions (bottom panels) for each state variable of Lorenz 63 systems using $\ell_1$-trend filtering at different noise levels for an arbitrary realization. The green circles represent the regularization parameters $\lambda$ that yield the minimum estimation error defined in Eqn.~(\ref{eq:relative_filter_errors}). Black crosses represent the converged corner points for Pareto curves and the minima of the GCV functions for GCV.}
		\label{fig:trendfilter_Lorenz63_Pareto_and_GCV}
	\end{figure}
	Next, we present the performance of WBPDN and STLS for estimating the governing equation coefficients $\hat{\bm{\xi}}_j, j = 1,...,n$, using the filtered data by each of the global smoothing methods. For each method, we used the filtered data via Pareto curve since it slightly outperformed GCV. We omit the analysis of local methods since we observed that global methods generally offer better accuracy for the state and state time-derivative estimates in the examples of this article.  Figure~\ref{fig:wbpdn_and_stls__Lorenz63_filters} compares the relative solution error, given in Eqn.~(\ref{eq:relsolerror}), for WBPDN (top panels) and STLS (bottom panels) at different noise levels, and using GCV (dashed lines) and Pareto curves (solid lines) as automatic hyperparameter selection techniques. Generally, as noted in~\cite{cortiella2021sparse}, WBPDN significantly outperforms STLS for the hyperparameter selection methods presented in this work. The error trend consistently increases as we increase the noise level for both sparse regression methods. Overall, WBPDN is more robust to noise and yields more accurate coefficient estimates. 
	\begin{figure}[H]
		\centering
		\includegraphics[trim = 10 0 10 0, clip,width=0.34\textwidth]{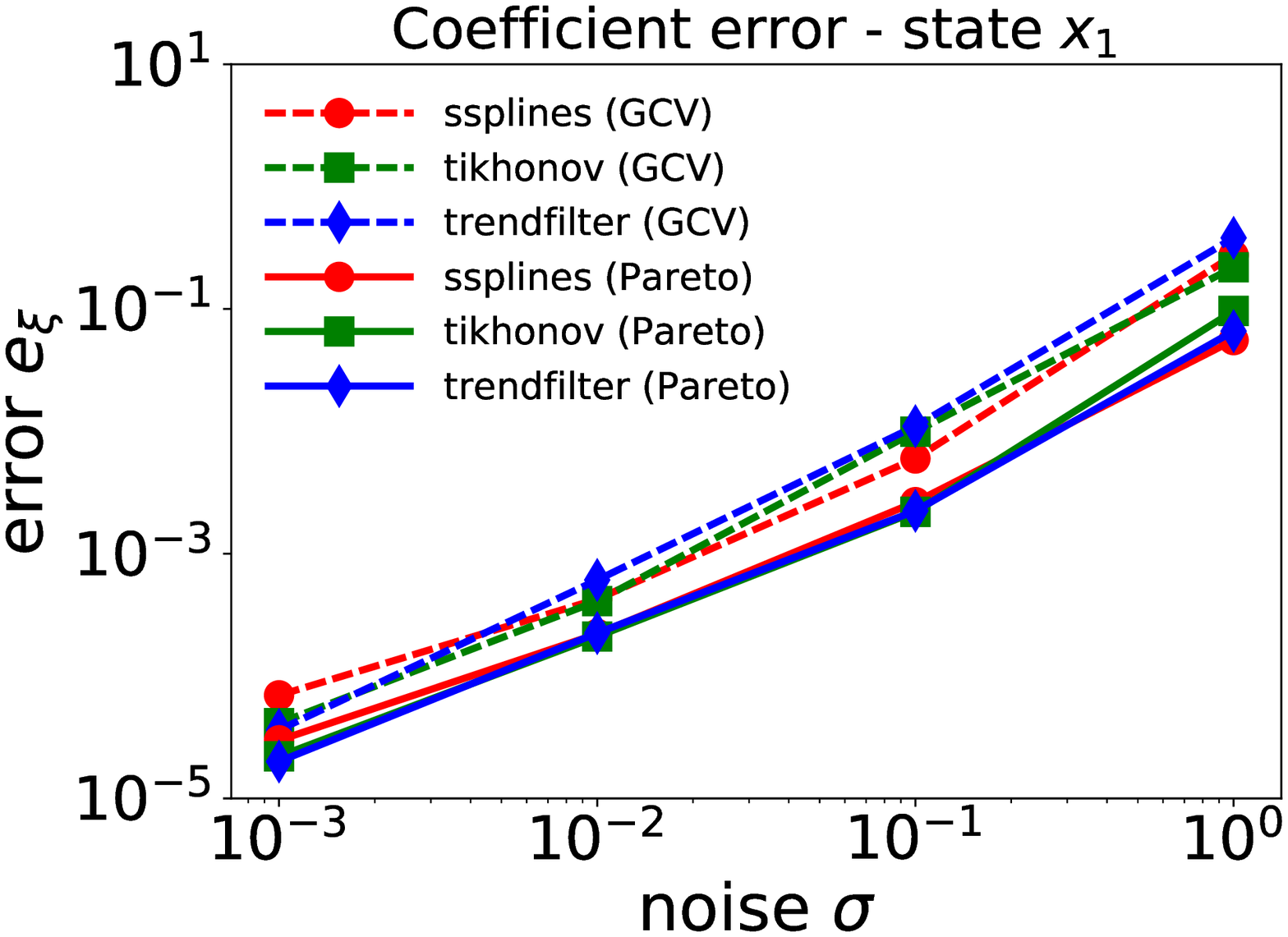}
		\includegraphics[trim = 40 0 10 0, clip,width=0.32\textwidth]{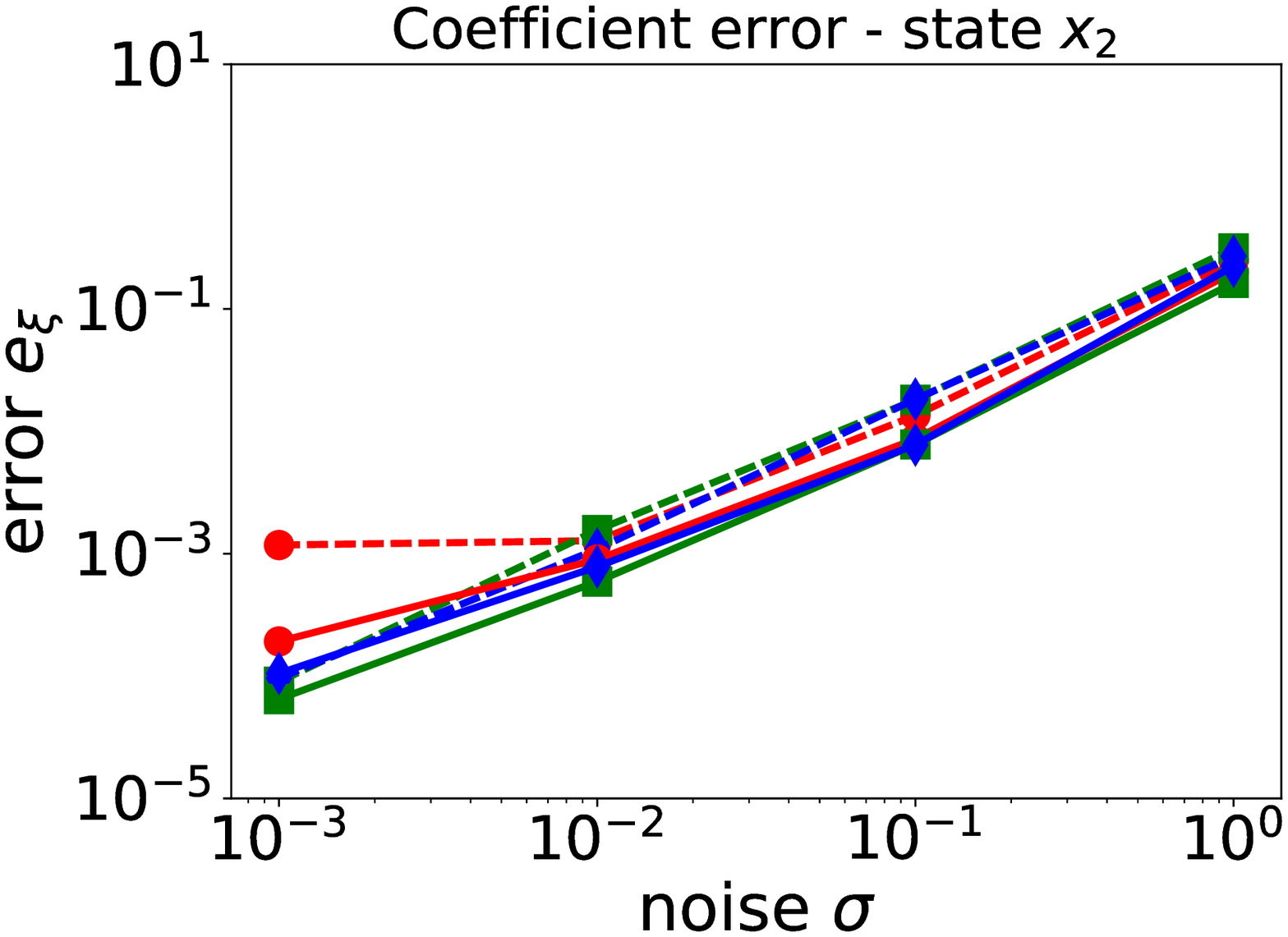}
		\includegraphics[trim = 40 0 10 0, clip,width=0.32\textwidth]{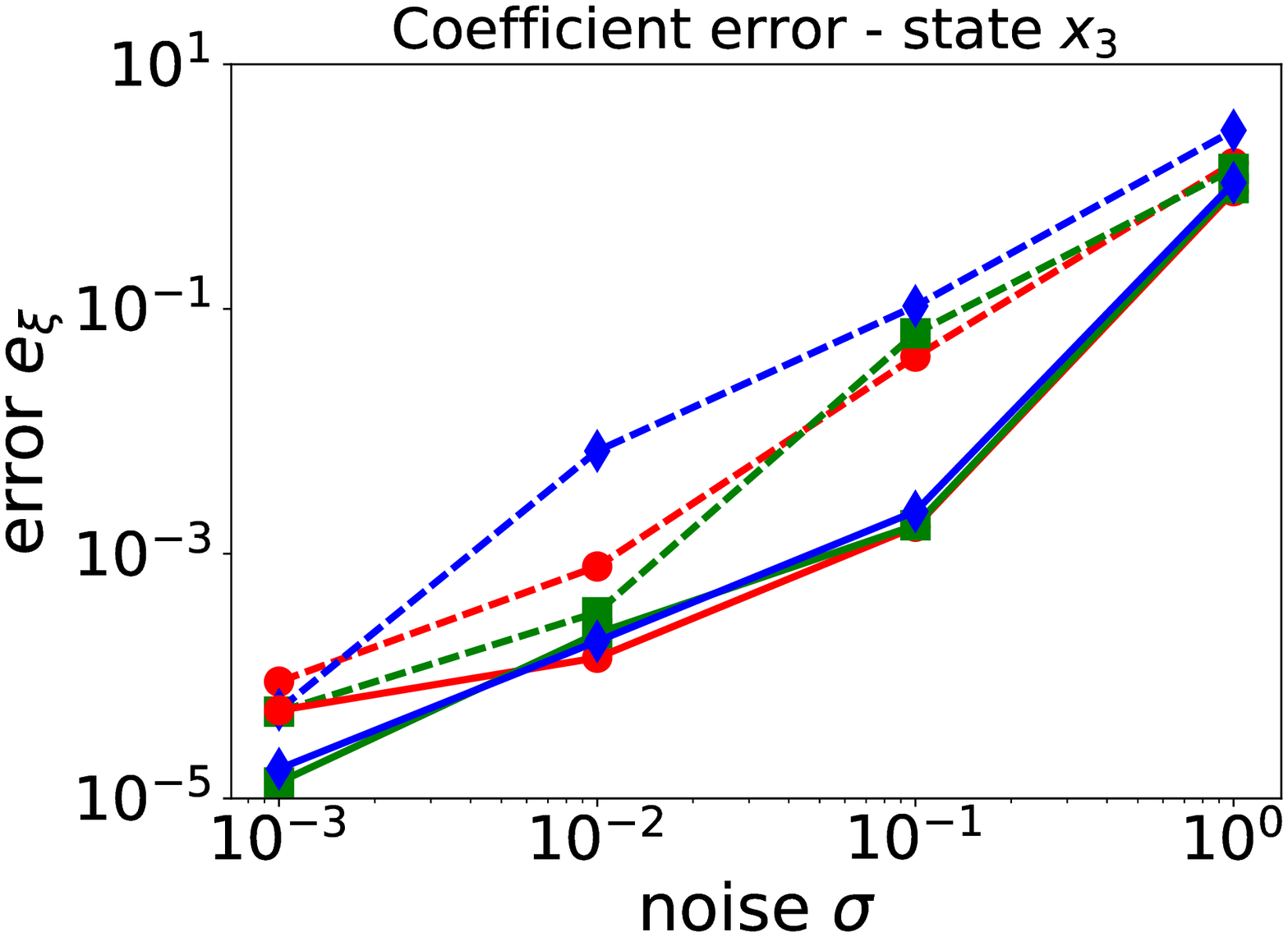}
		\includegraphics[trim = 10 0 10 0, clip,width=0.34\textwidth]{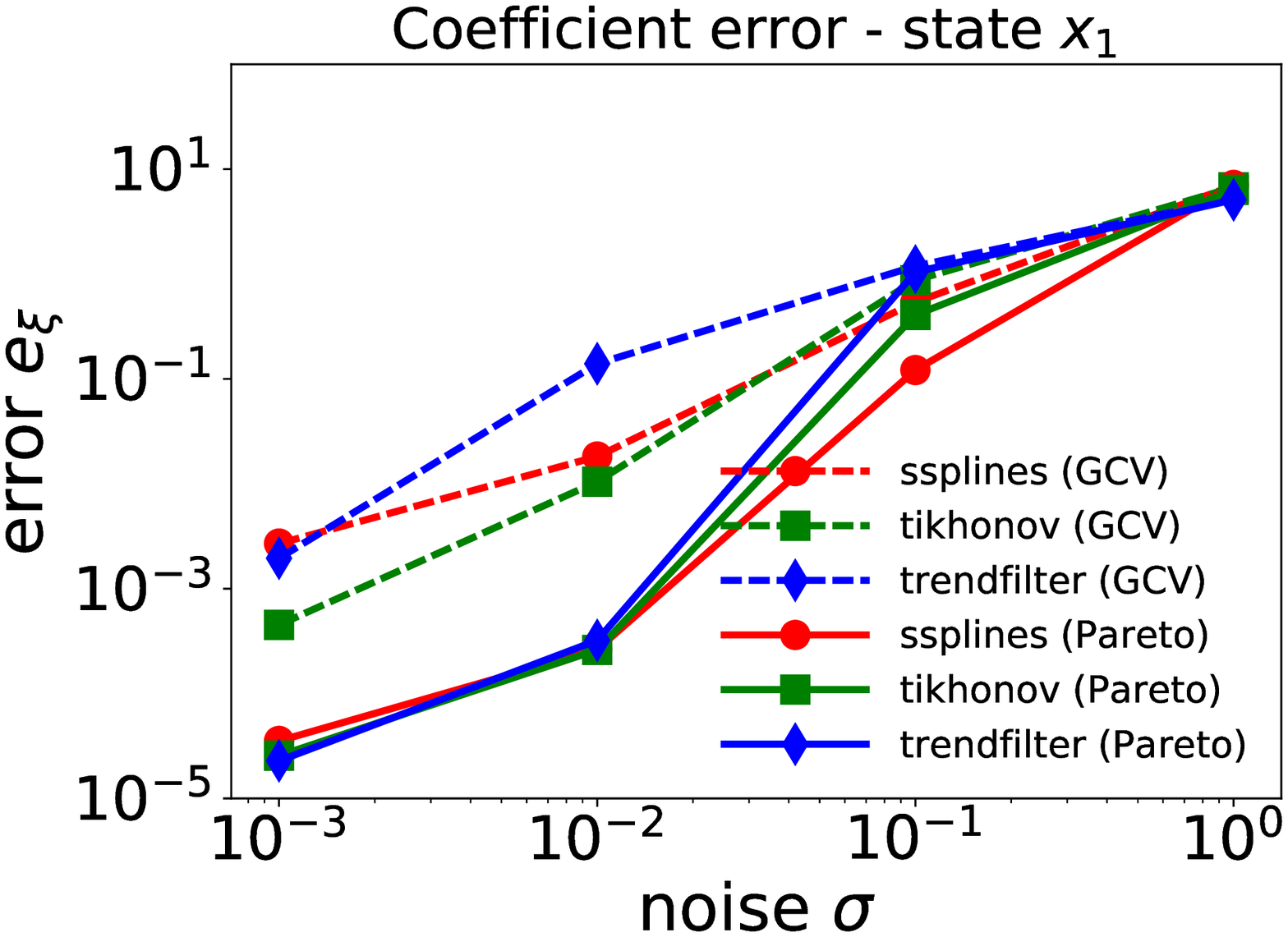}
		\includegraphics[trim = 40 0 10 0, clip,width=0.32\textwidth]{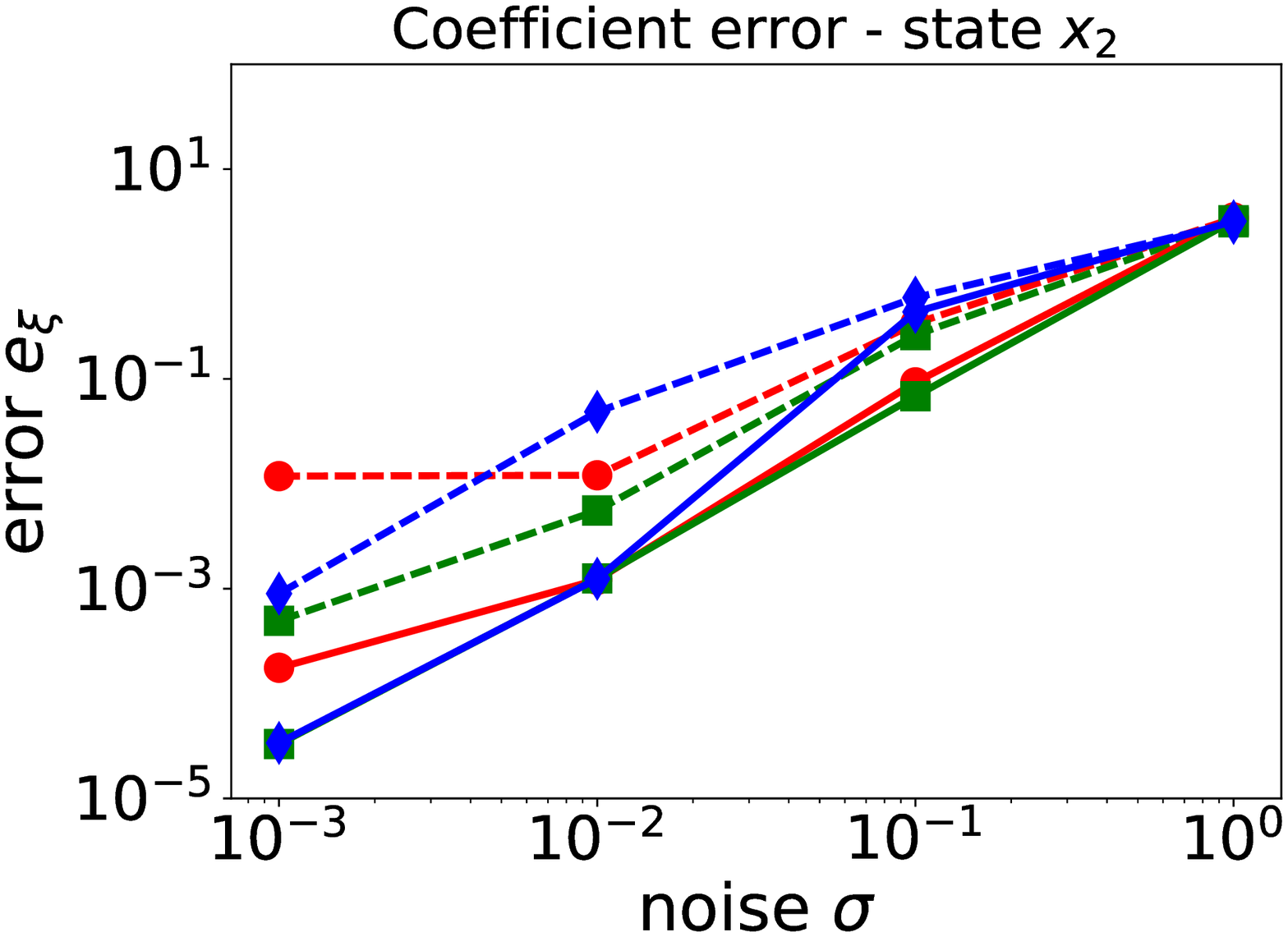}
		\includegraphics[trim = 40 0 10 0, clip,width=0.32\textwidth]{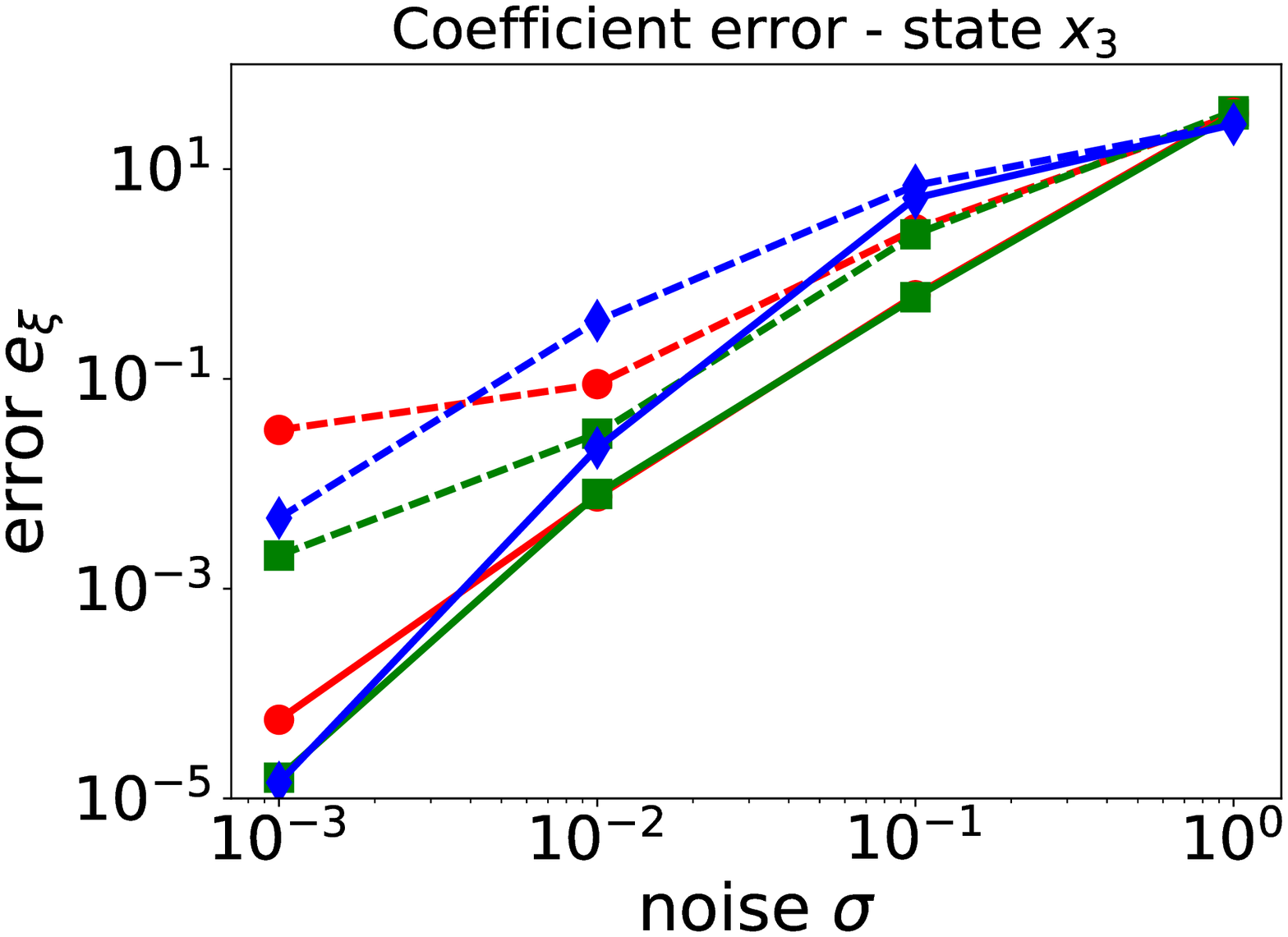}
		\caption{Relative  coefficient  errors  of  Lorenz  63 for WBPDN (top panels) and STLS (bottom panels)  with  respect  to  different  noise  levels $\sigma$ with measurements pre-processed via smoothing splines, Tikhonov smoother and $\ell_1$-trend filtering. The dashed and solid lines correspond to the solution using GCV and Pareto curves for selecting $\lambda$, respectively.}
		\label{fig:wbpdn_and_stls__Lorenz63_filters}
	\end{figure}
	Similar to the filtering stage, the Pareto curve criterion results in superior performance as compared to GCV for the sparse regression algorithms presented in this article. We also note that even though $\ell_1$-trend filtering yields slightly better state and state time-derivative estimates than smoothing splines and Tikhonov smoother, that difference is not noticeable of the error on the coefficient estimates. Thus, the global smoothing methods along with the Pareto curve criterion give similar results for the recovery of the coefficients. Figure~\ref{fig:WBPDN_trendfilter_Lorenz63_Pareto_and_GCV} shows the Pareto curves and GCV functions for the 0th iteration  -- i.e. no weighting -- of WBPDN and data filtered with $\ell_1$-trend filtering. We emphasize that good coefficient estimates for the 0th iteration of WBPDN are crucial for the subsequent iterations to enhance sparsity, reduce the error and converge. As can be observed, the corner point criterion of Pareto curves results in accurate estimation of near optimal regularization parameters $\lambda$, whereas the minima of the GCV functions generally tend towards smaller $\lambda$ from the optimal ones. Similar results were observed for Tikhonov smoother and smoothing splines.
	\begin{figure}[H]
		\centering
		\includegraphics[trim = 10 0 10 0, clip,width=0.34\textwidth]{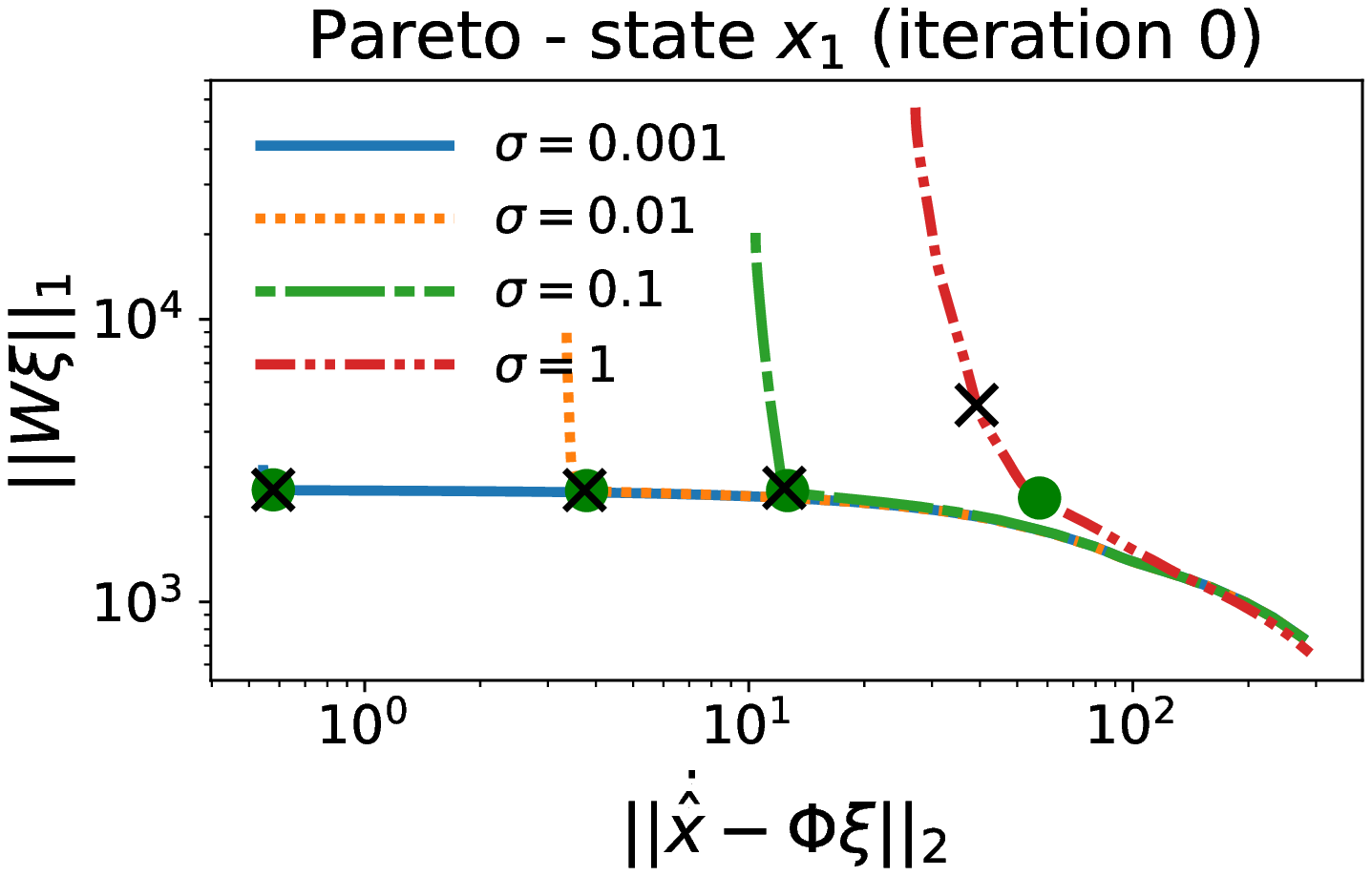}
		\includegraphics[trim = 35 0 10 0, clip,width=0.32\textwidth]{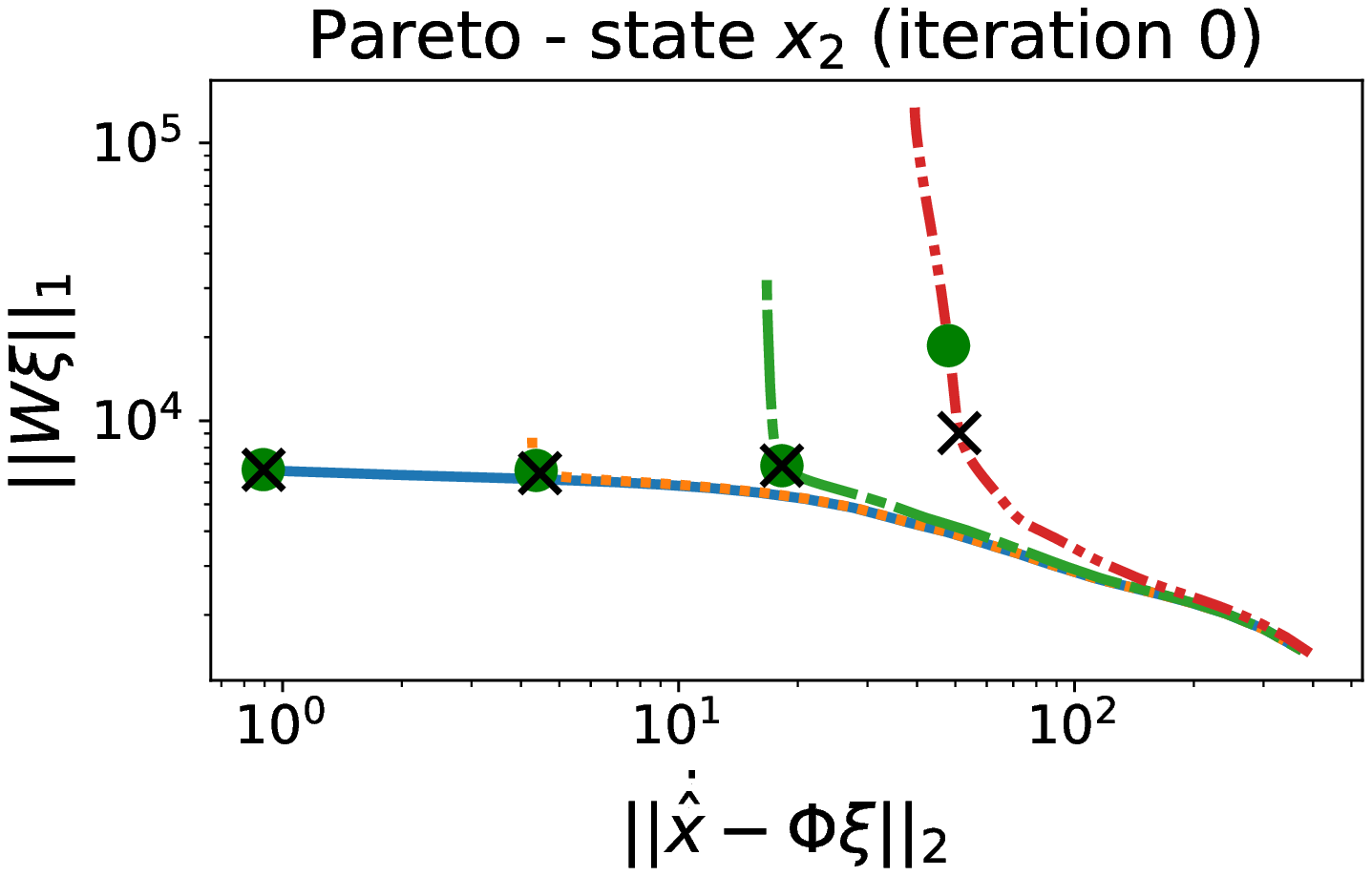}
		\includegraphics[trim = 35 0 10 0, clip,width=0.32\textwidth]{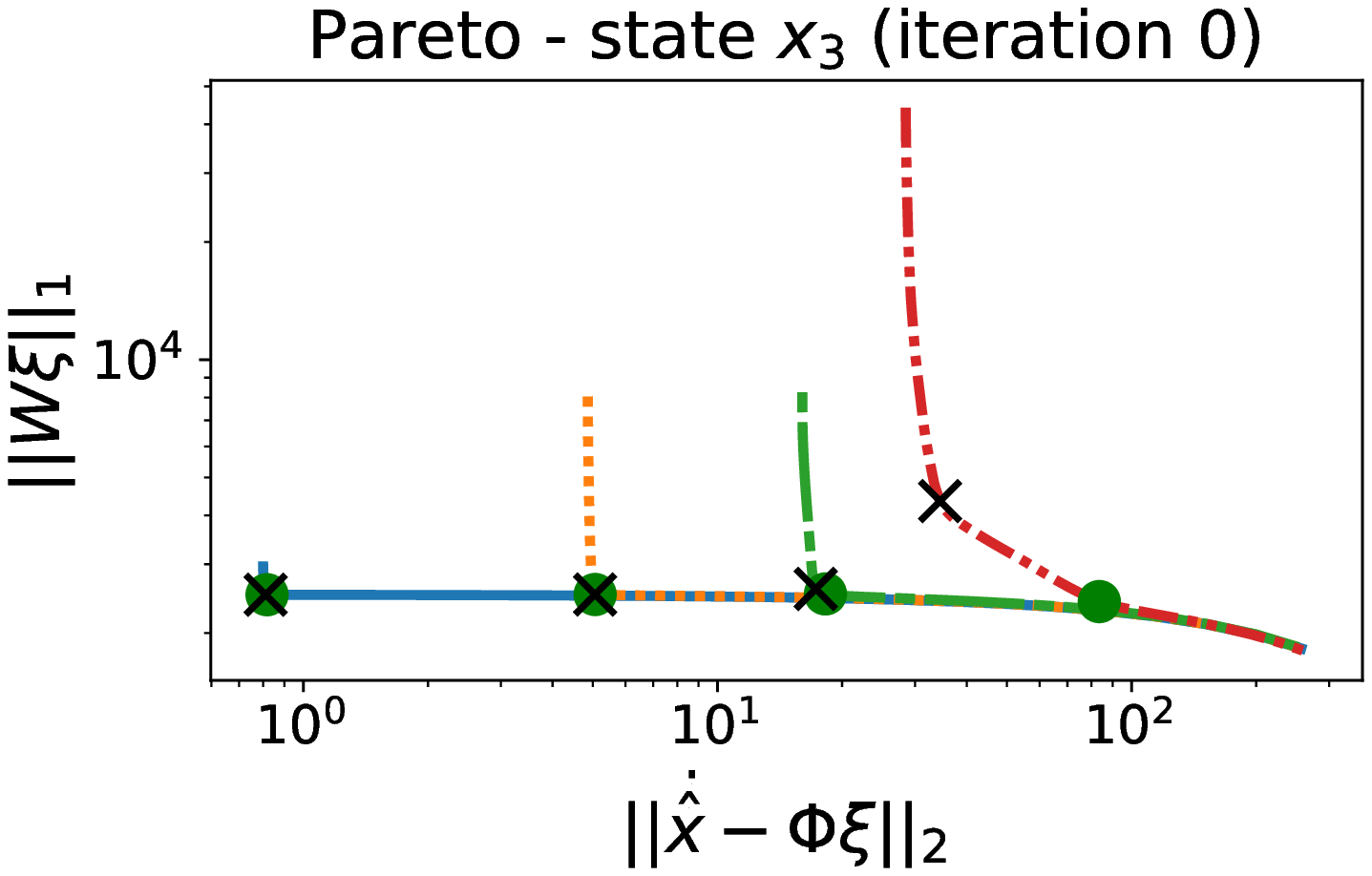}
		\includegraphics[trim = 10 0 10 0, clip,width=0.34\textwidth]{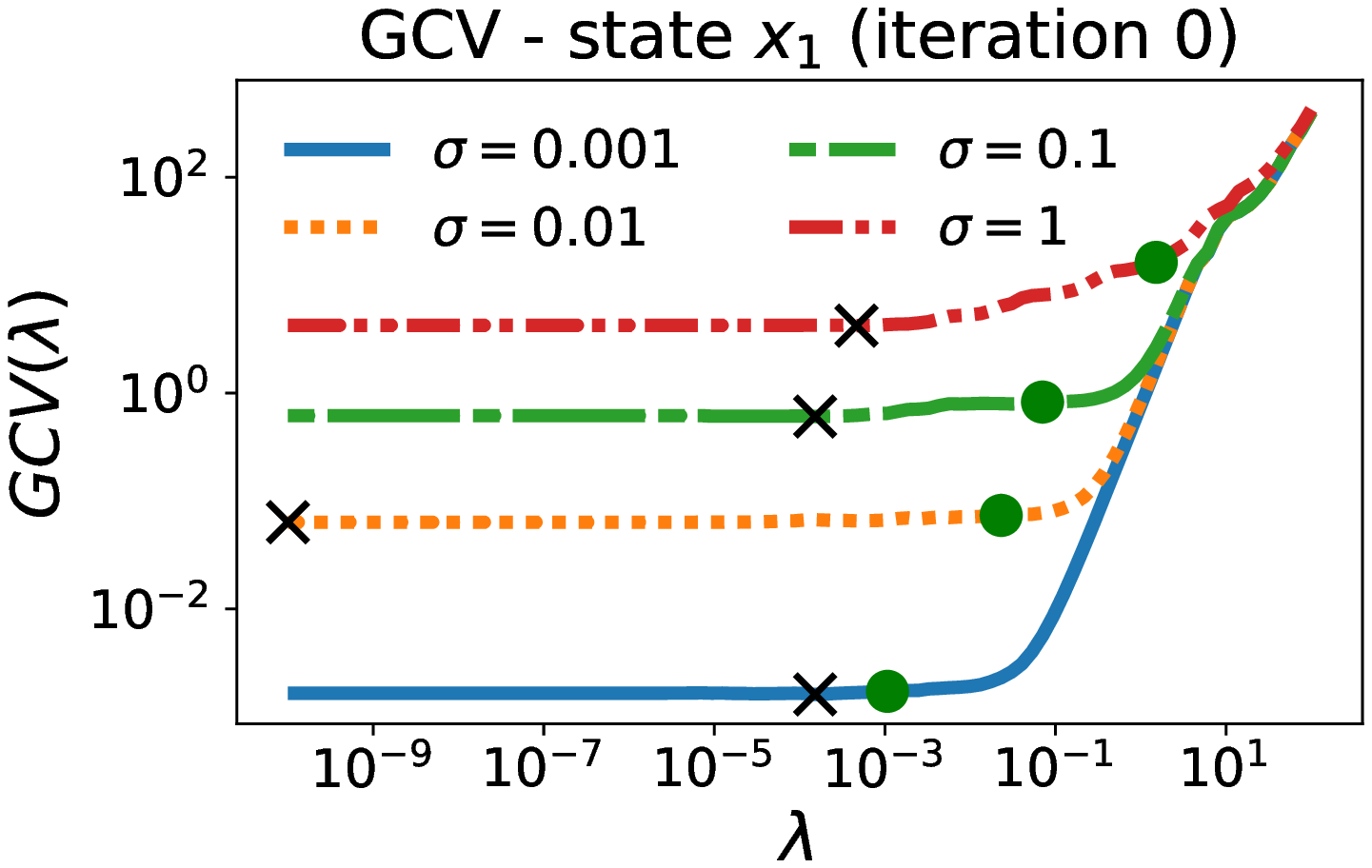}
		\includegraphics[trim = 35 0 10 0, clip,width=0.32\textwidth]{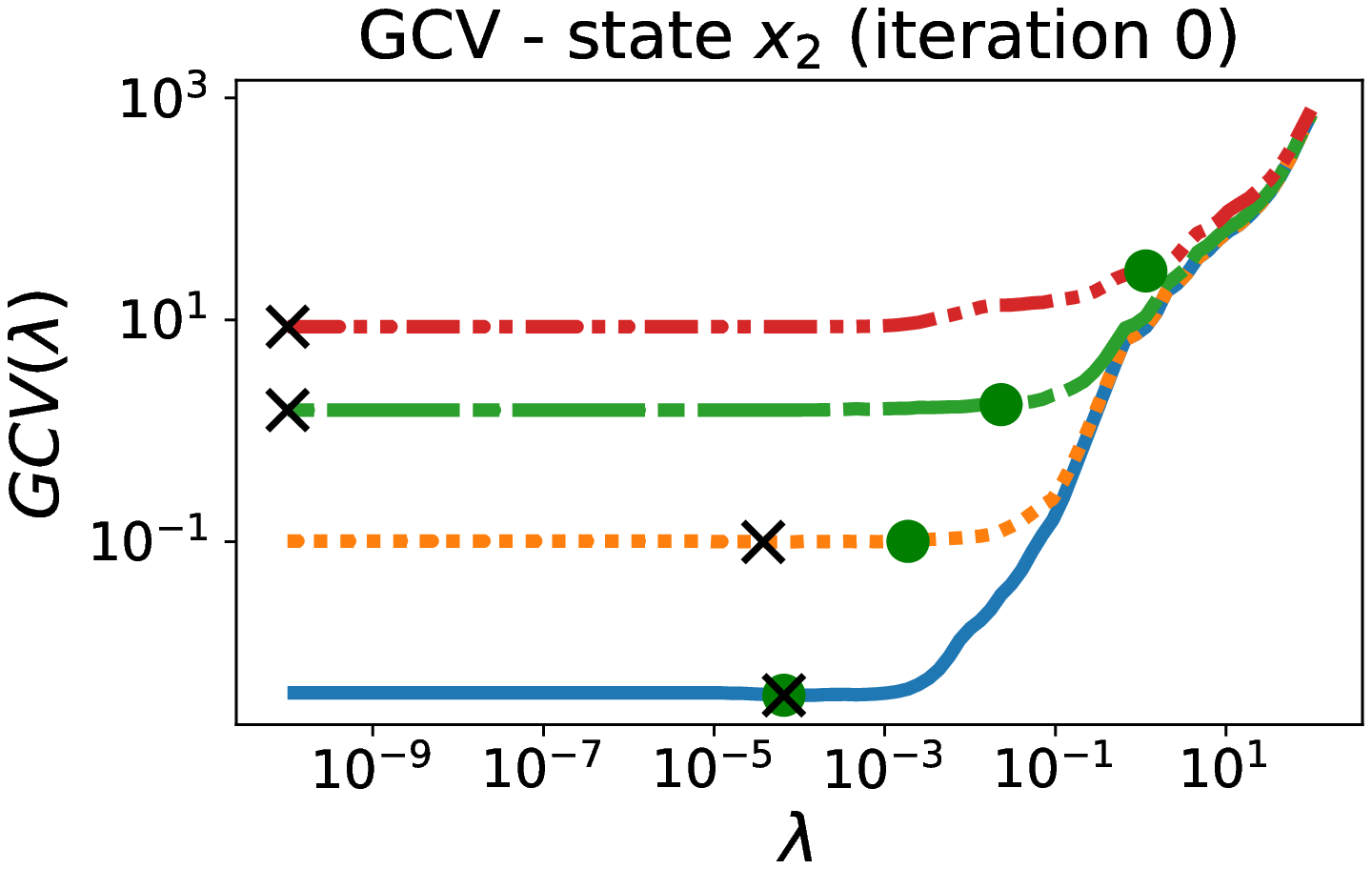}
		\includegraphics[trim = 35 0 10 0, clip,width=0.32\textwidth]{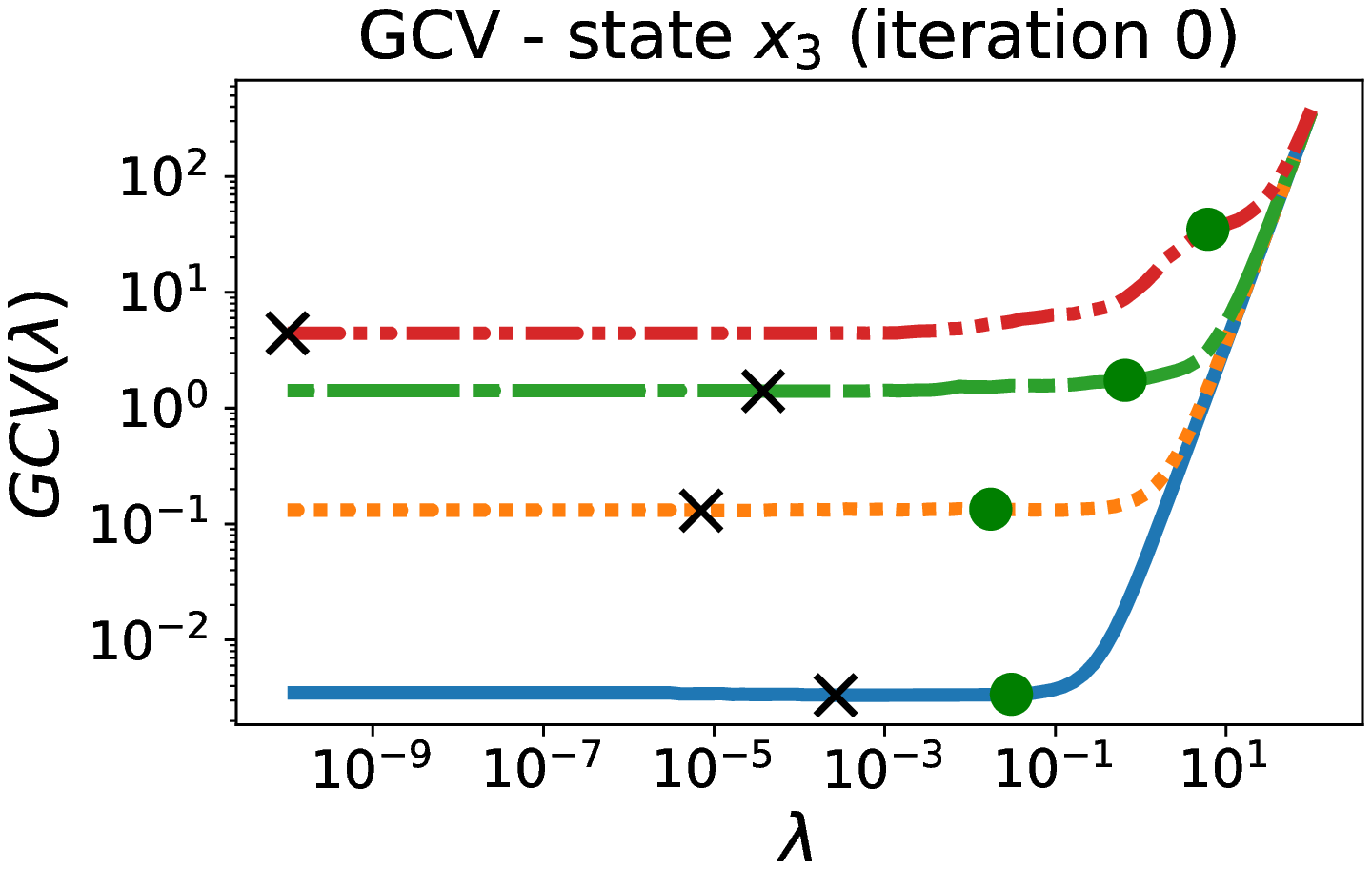}
		\caption{Pareto curves (top panels) and GCV functions (bottom panels) for each state variable of Lorenz 63 system for WBPDN using $\ell_1$-trend filtering at different noise levels for an arbitrary realization. The green circles represent the regularization parameters $\lambda$ that yield the minimum estimation error defined in Eqn.~(\ref{eq:relative_filter_errors}). Black crosses represent the converged corner points for Pareto curves and the minima of the GCV functions for GCV. Similar trends were observed for smoothing splines and Tikhonov smoother.}
		\label{fig:WBPDN_trendfilter_Lorenz63_Pareto_and_GCV}
	\end{figure}
	For STLS, we noticed that both GCV and Pareto curves were non-smooth as compared to the WBPDN case, possibly yielding sub-optimal $\lambda$ estimates.  We also observed that measuring the complexity of the model -- i.e. vertical axis of the Pareto curve log-log plots -- using the $\ell_0$- and $\ell_1$-norms for STLS did not produce well-defined L-shaped curves and corners, resulting in poor estimates of optimal $\lambda$.  Instead, we used $1 / \lambda$. Figure~\ref{fig:STLS_tikhonov_Lorenz63_Pareto_and_GCV} shows the GCV functions and Pareto curves for STLS with data filtered using Tikhonov smoother. As seen, the $\ell_0$-Pareto curve (top panels) is inconsistent with the typical L shape of Pareto curves. Similar results were observed for the $\ell_1$-Pareto curve.  Both $1/\lambda$-Pareto curves and GCV functions present discontinuities for STLS, which may be caused by the hard-thresholding step of the algorithm.
	\begin{figure}[H]
		\centering
		\includegraphics[trim = 10 0 10 0, clip,width=0.34\textwidth]{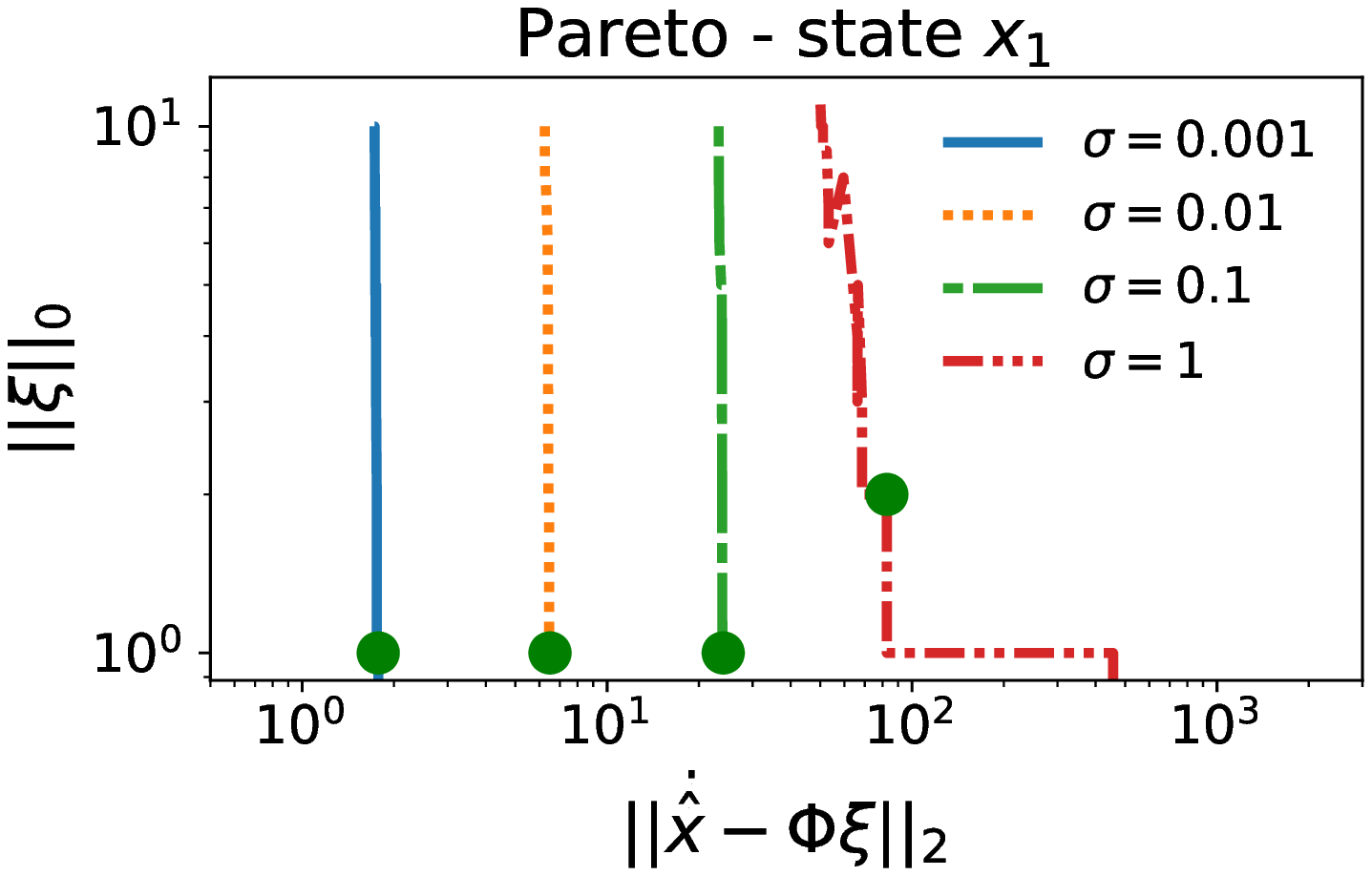}
		\includegraphics[trim = 35 0 10 0, clip,width=0.32\textwidth]{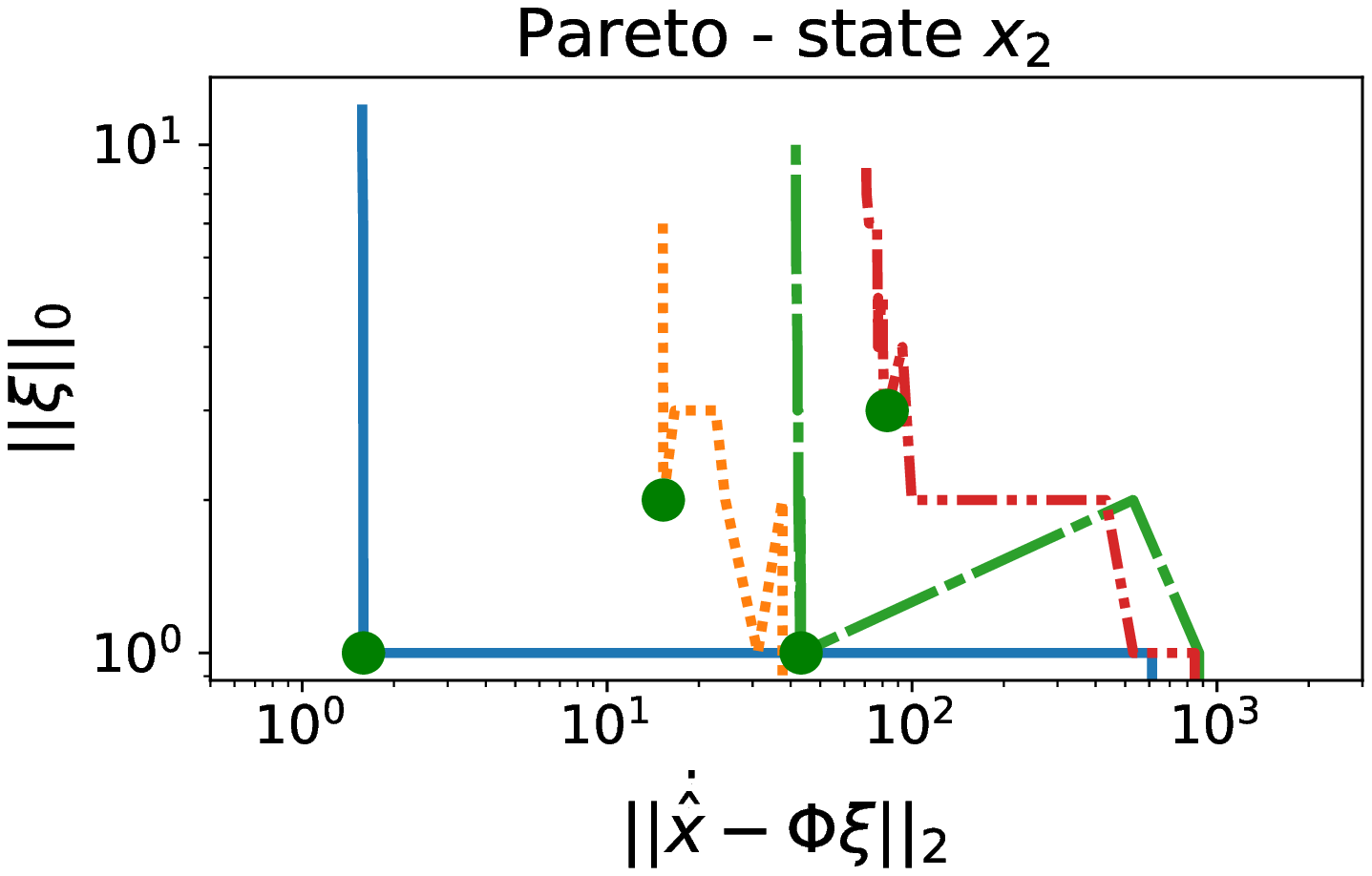}
		\includegraphics[trim = 35 0 10 0, clip,width=0.32\textwidth]{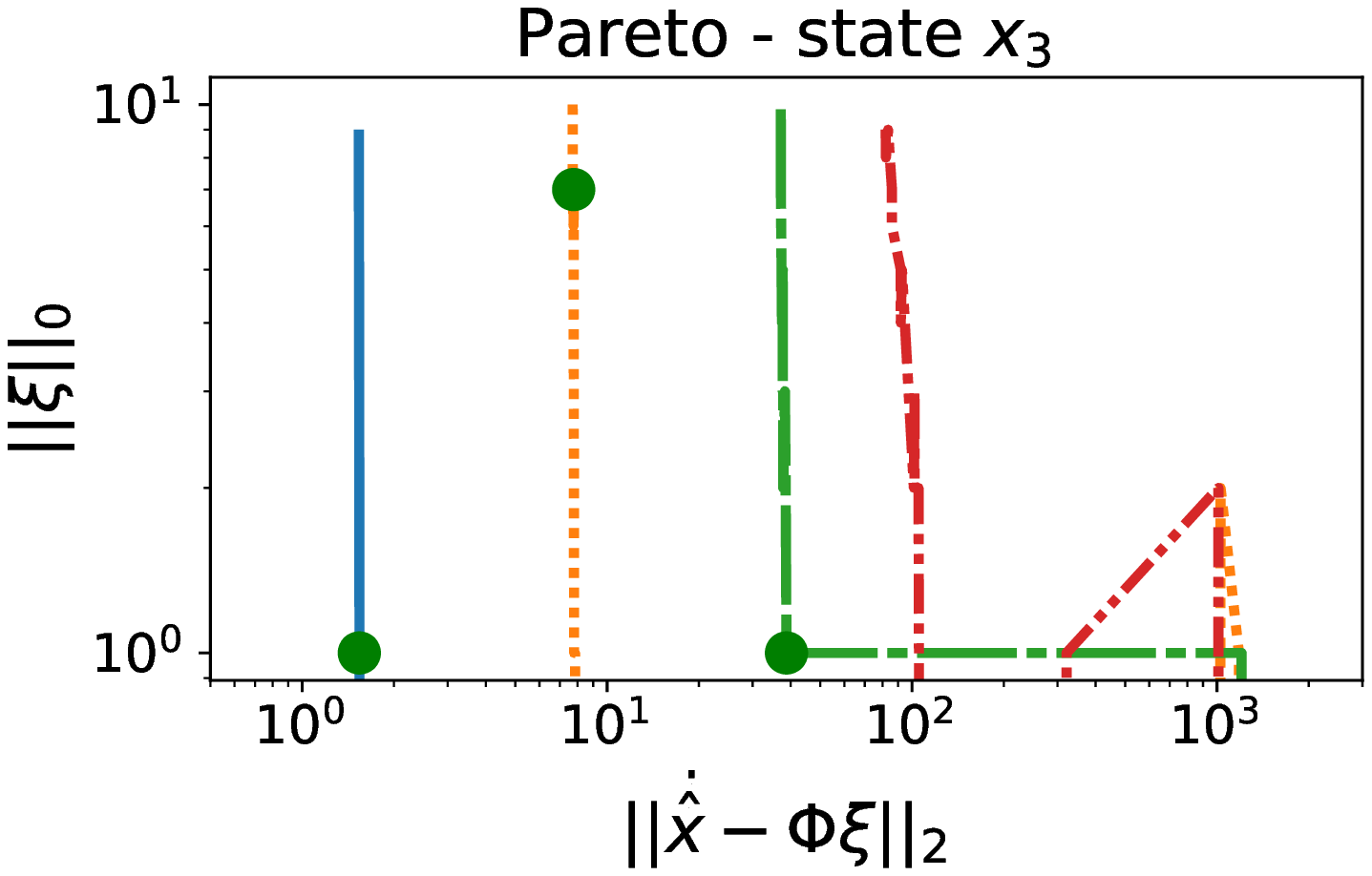}
		\includegraphics[trim = 10 0 10 0, clip,width=0.34\textwidth]{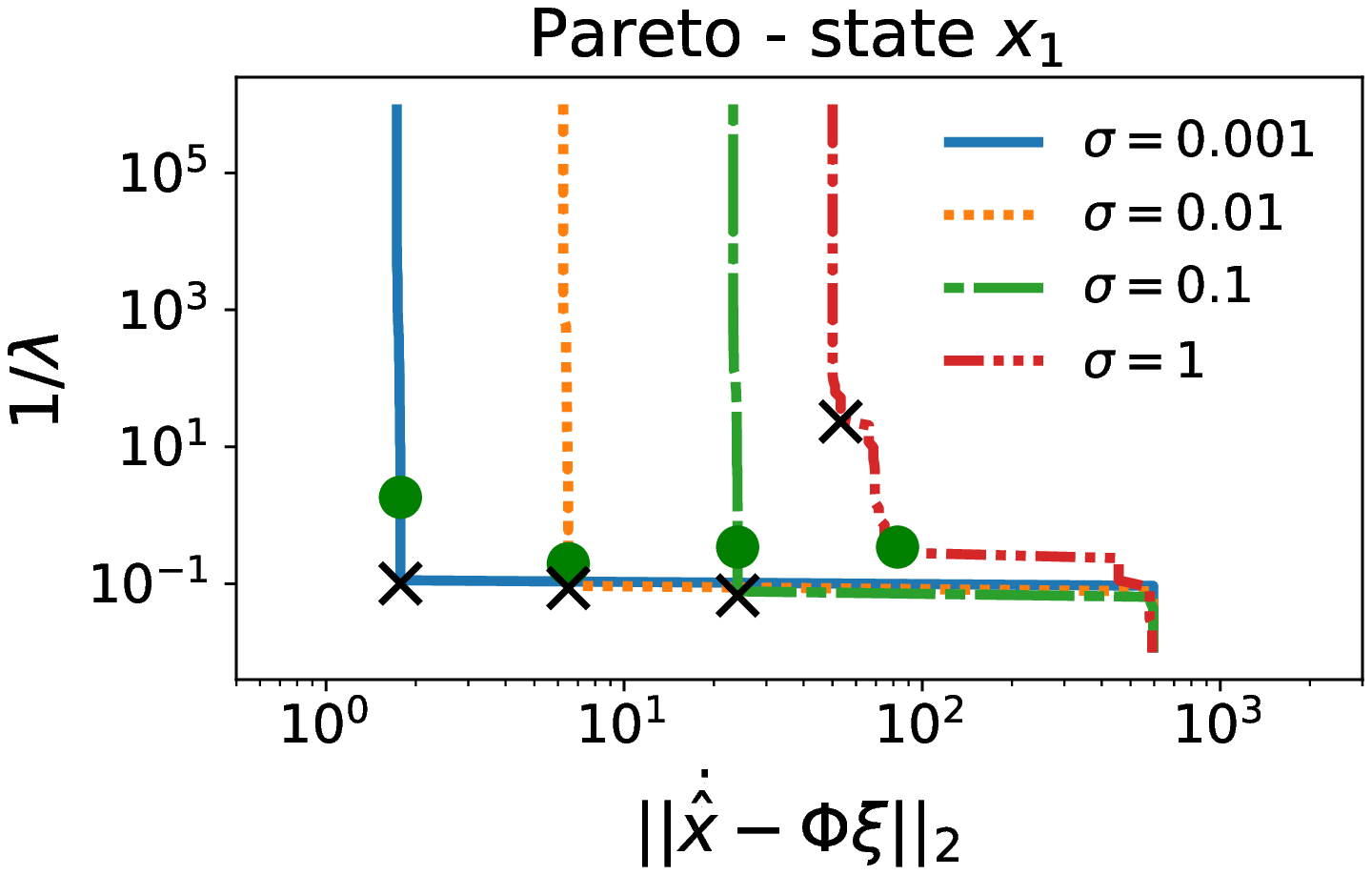}
		\includegraphics[trim = 35 0 10 0, clip,width=0.32\textwidth]{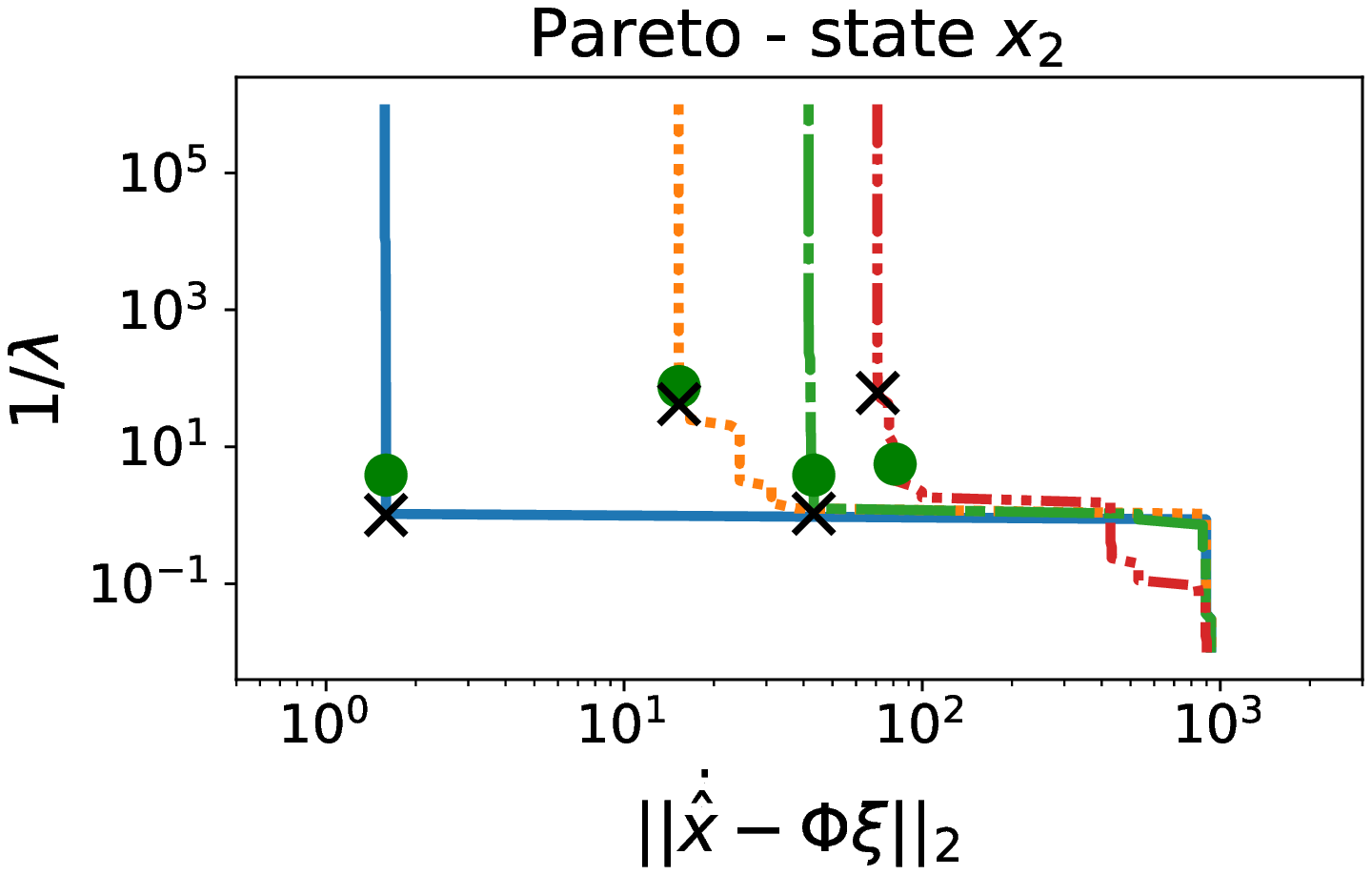}
		\includegraphics[trim = 35 0 10 0, clip,width=0.32\textwidth]{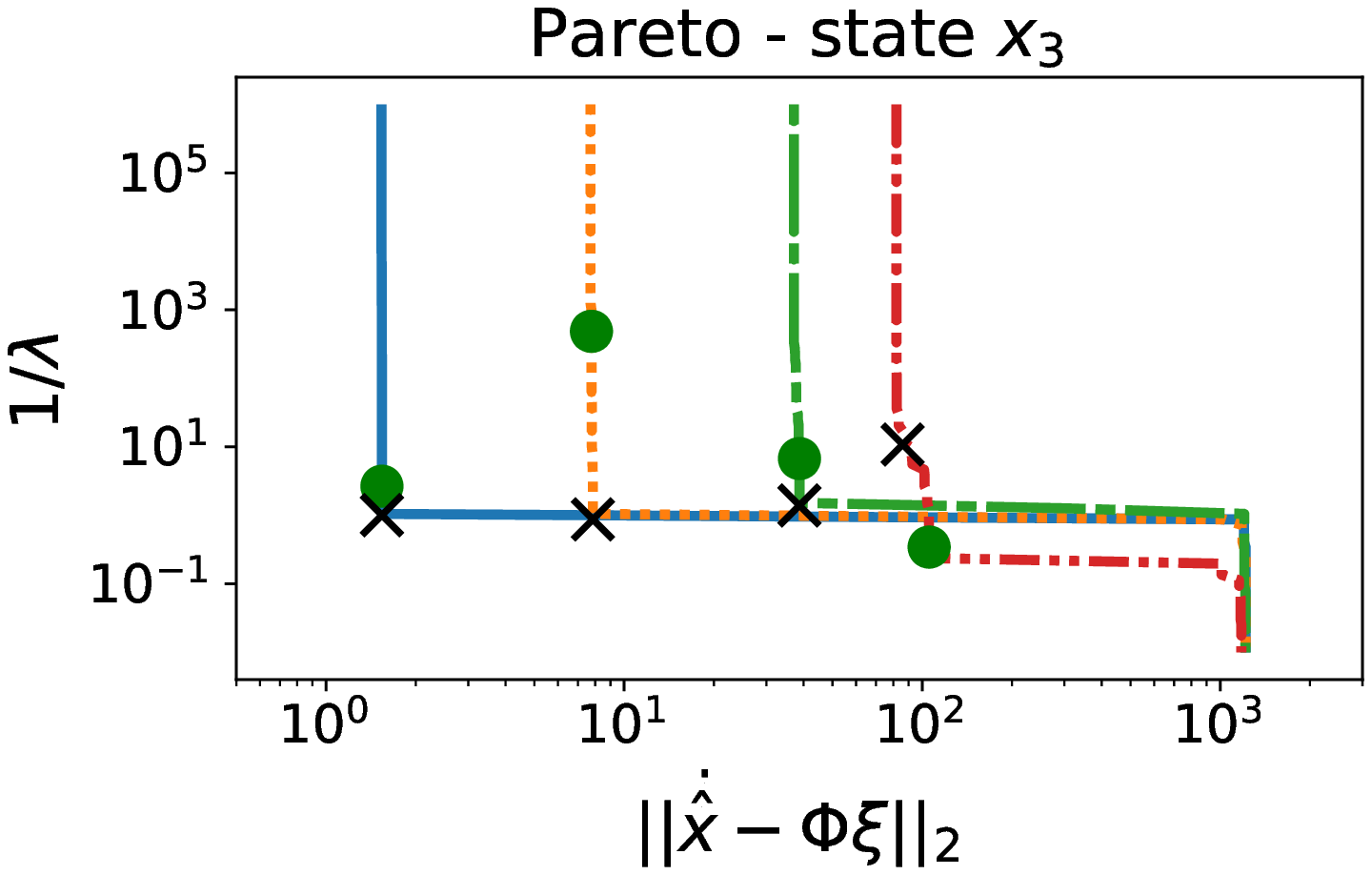}
		\includegraphics[trim = 10 0 10 0, clip,width=0.34\textwidth]{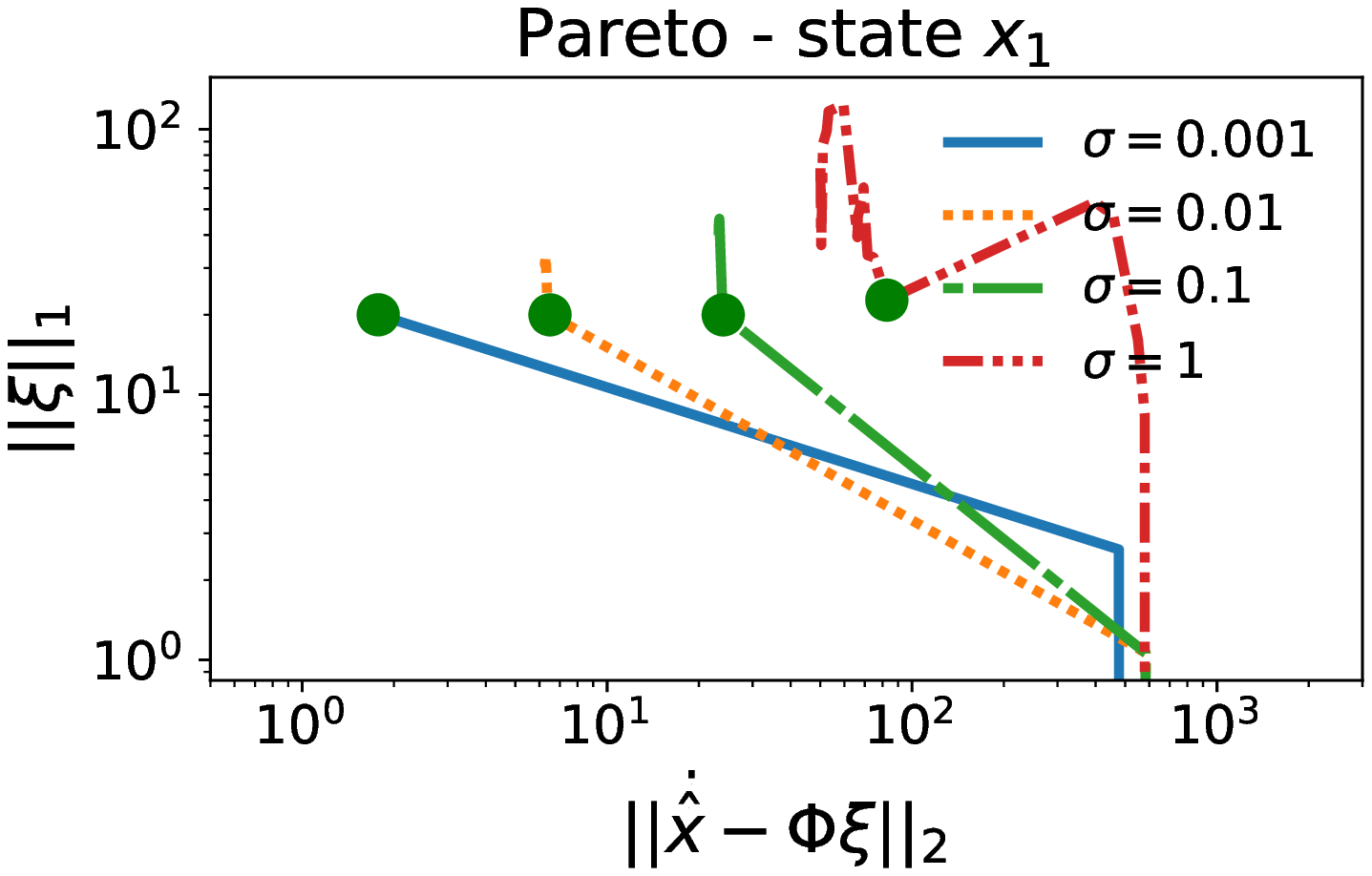}
		\includegraphics[trim = 35 0 10 0, clip,width=0.32\textwidth]{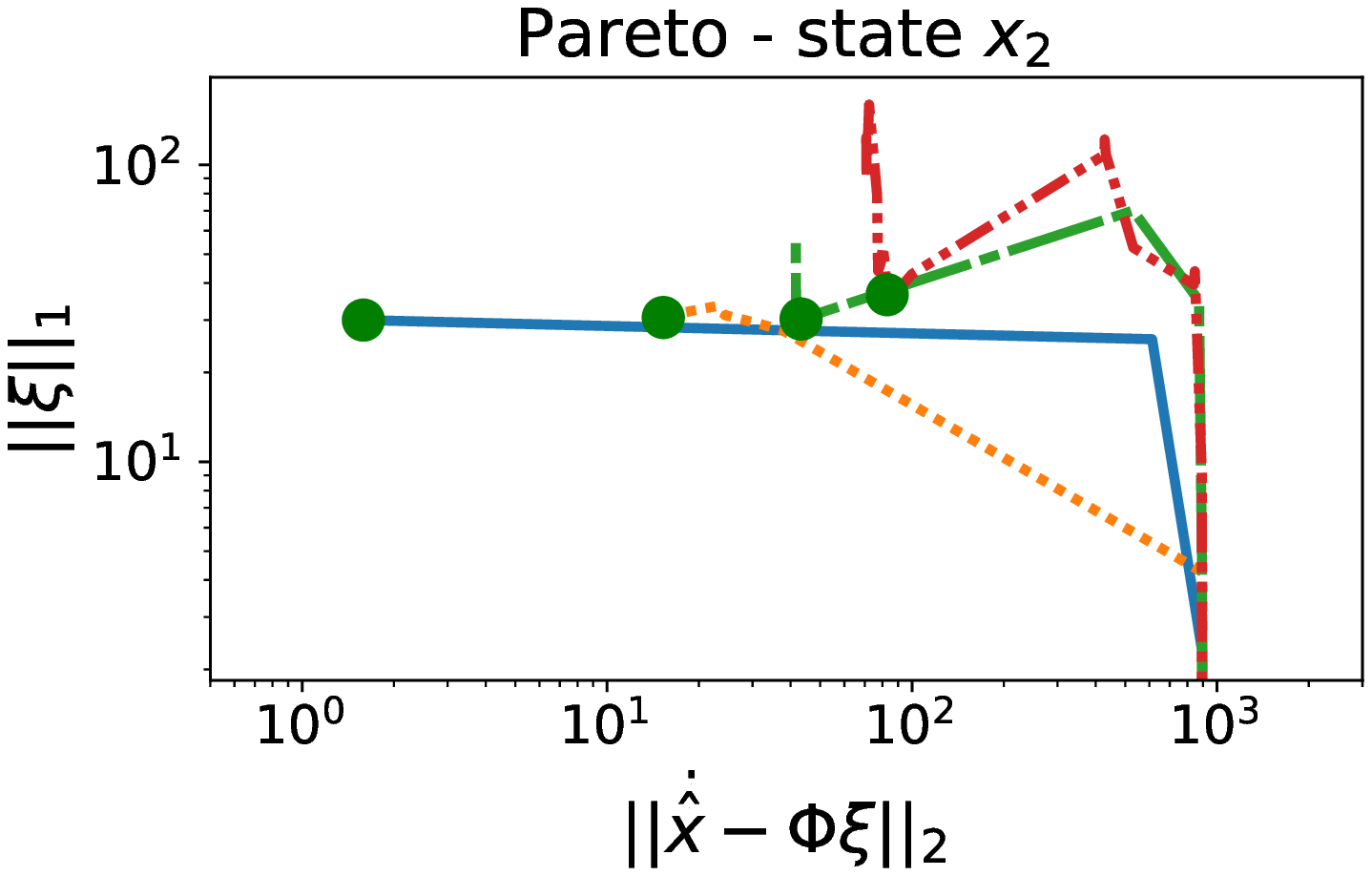}
		\includegraphics[trim = 35 0 10 0, clip,width=0.32\textwidth]{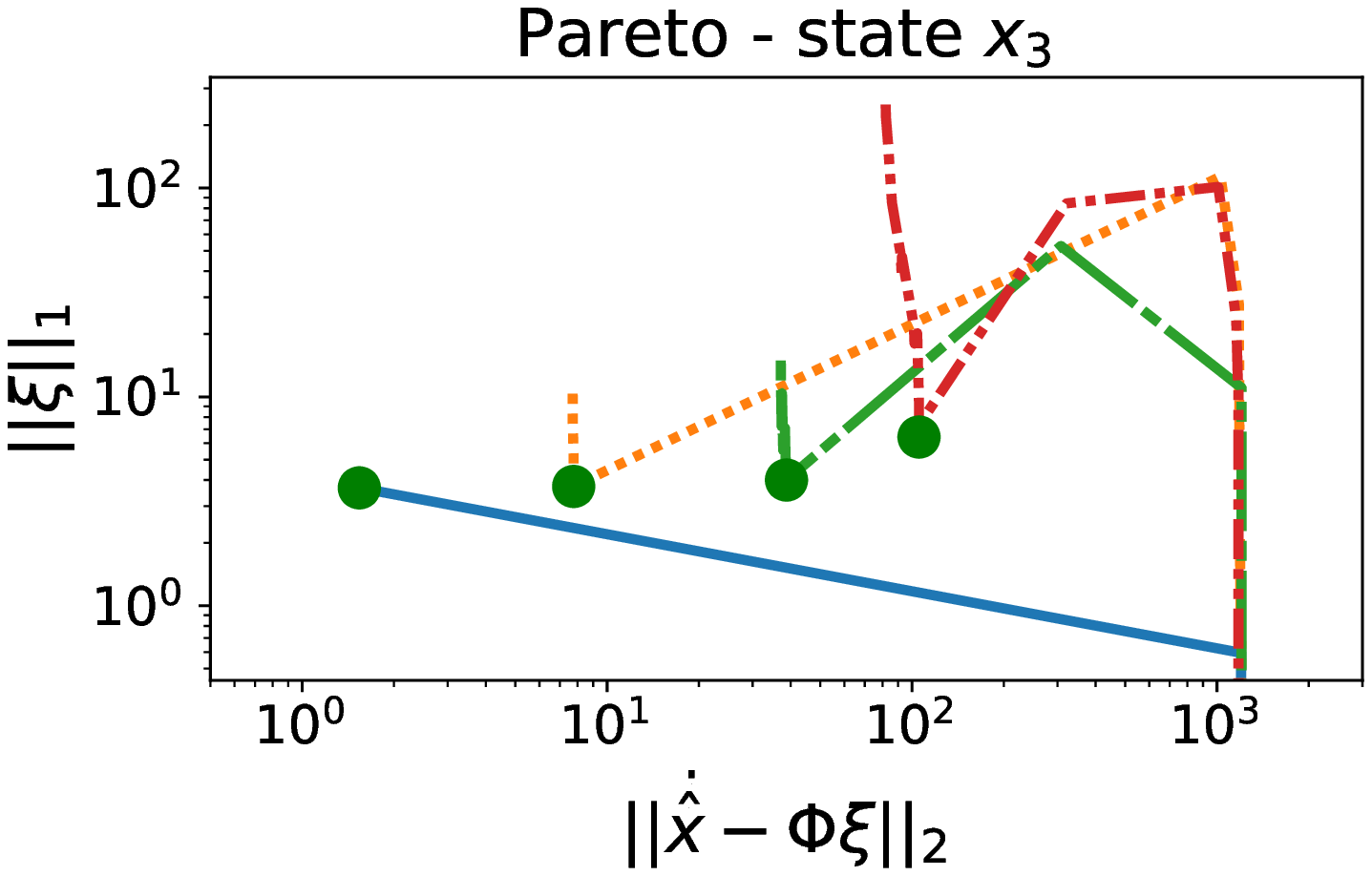}
		\includegraphics[trim = 10 0 10 0, clip,width=0.34\textwidth]{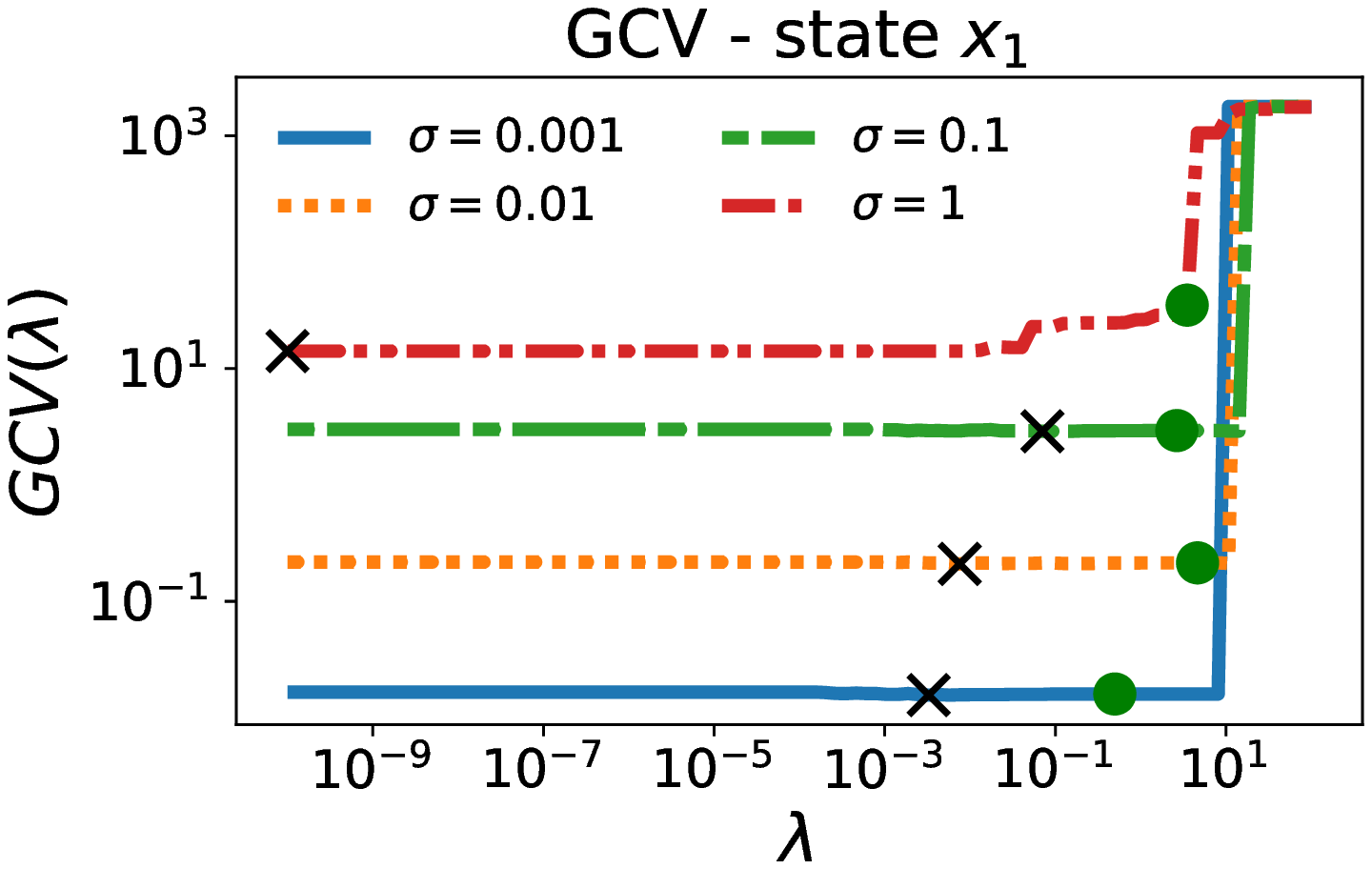}
		\includegraphics[trim = 35 0 10 0, clip,width=0.32\textwidth]{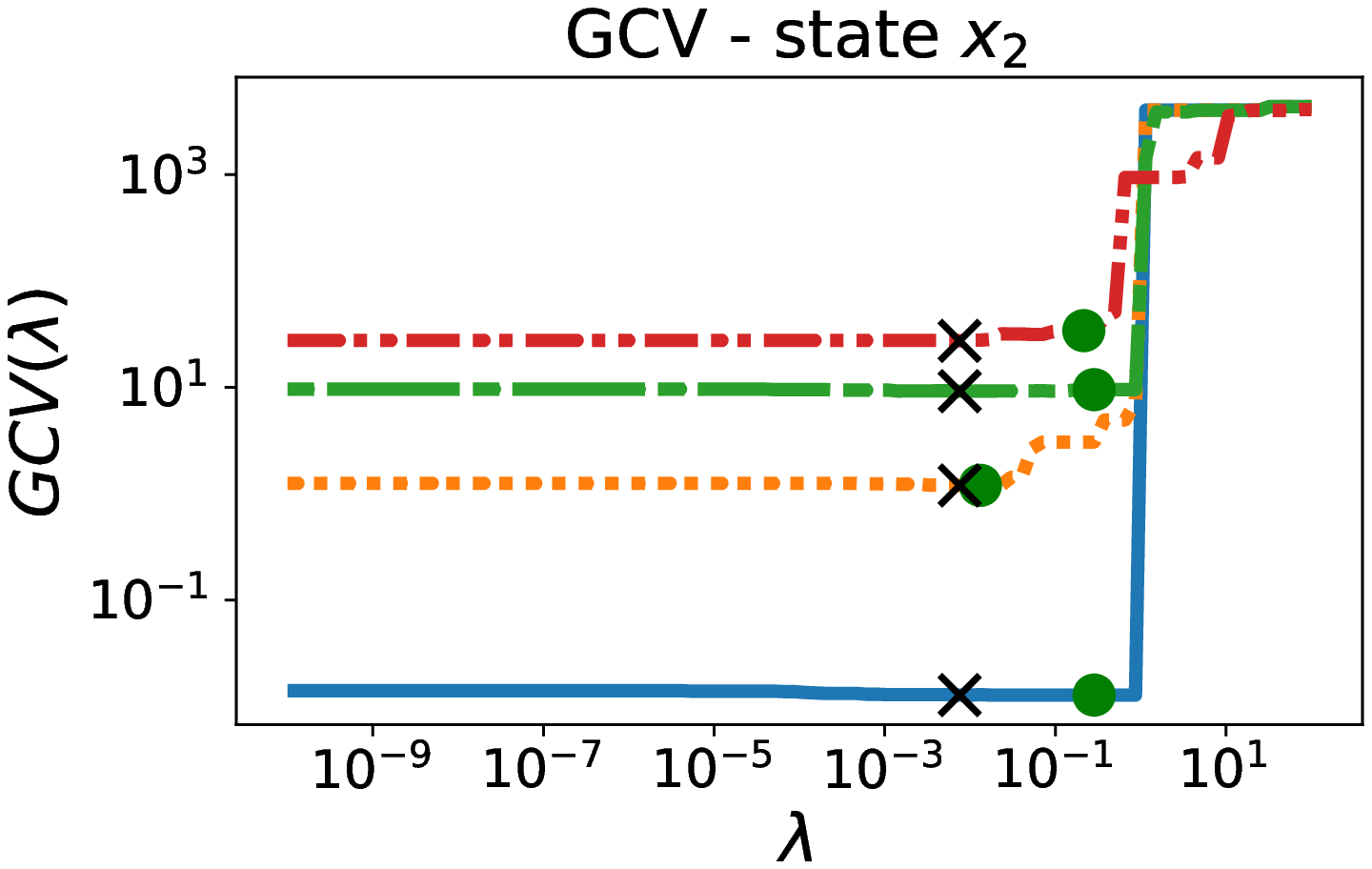}
		\includegraphics[trim = 35 0 10 0, clip,width=0.32\textwidth]{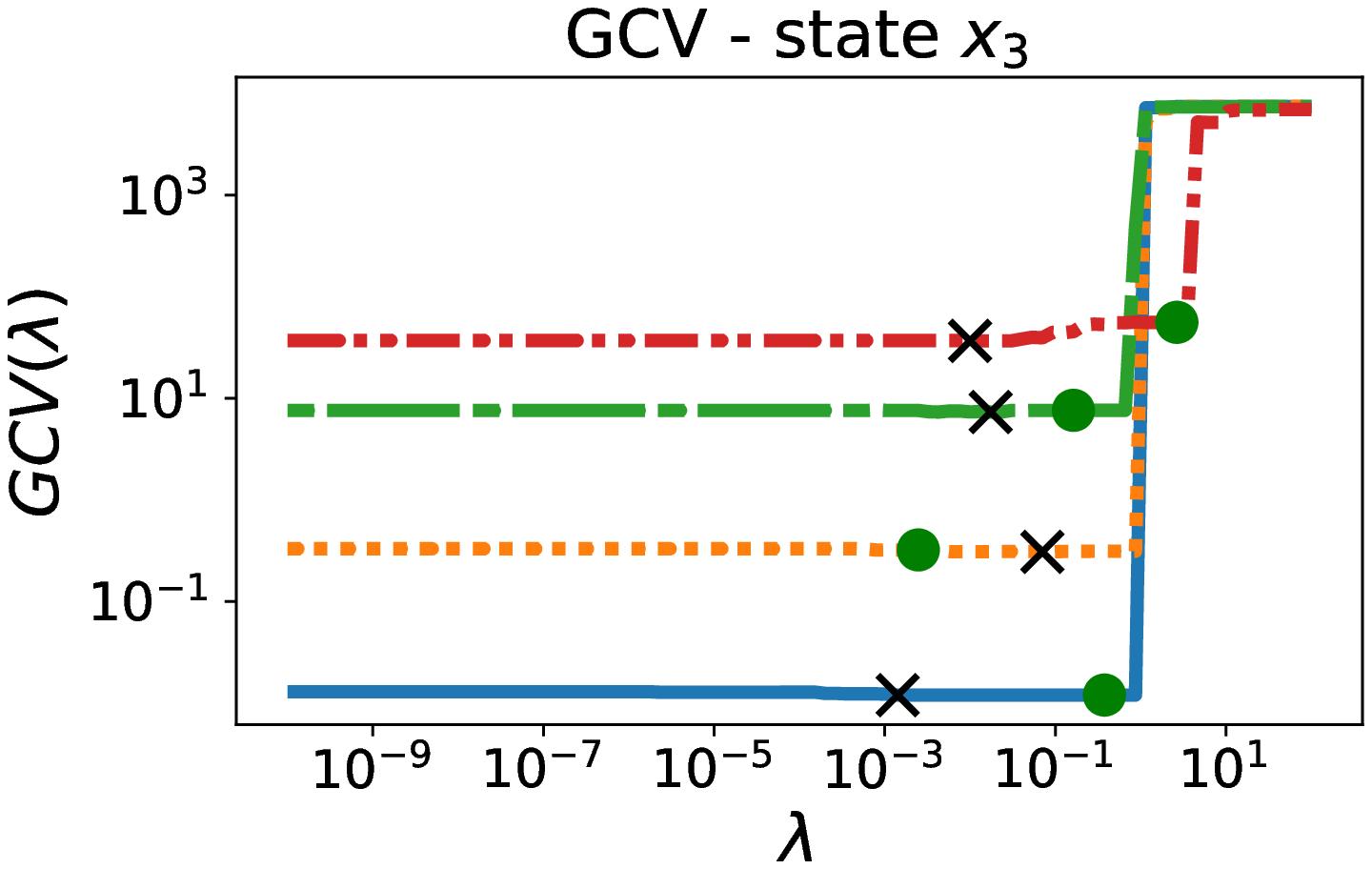}
		\caption{$\ell_0$- and $1/\lambda$-Pareto curves (top and central panels) and GCV functions (bottom panels) for each state variable of Lorenz 63 system for STLS using Tikhonov smoother at different noise levels for an arbitrary realization. The green circles represent the regularization parameters $\lambda$ that yield the minimum estimation error defined in Eqn.~(\ref{eq:relative_filter_errors}). Black crosses represent the converged corner points for Pareto curves and the minima of the GCV functions for GCV. Similar trends were observed for smoothing splines and $\ell_1$-trend filtering.}
		\label{fig:STLS_tikhonov_Lorenz63_Pareto_and_GCV}
	\end{figure}
	Next, we assess the prediction accuracy of the identified governing equations for the best performing procedure -- i.e. $\ell_1$-trend filtering for denoising the data and estimating derivatives and WBPDN for recovering the Lorenz 63 system dynamics. For the lowest noise case $\sigma = 0.001$, even though the coefficients were recovered with high accuracy, we noticed an abrupt change in the predicted state trajectory from the exact one at around 8 time units. Therefore, we only report the prediction of the identified dynamics up to 8 time units for all noise cases. Figure~\ref{fig:Lorenz63_samples_predictions} illustrates the measurements used for training (left panel) and the state predictions of the mean recovered model (right panel). {\color{black} We generated 100 state trajectory realizations of training measurements and computed the mean of the estimated coefficients for each realization}. We can see that the butterfly-shaped behavior of the Lorenz 63 system is captured for all noise cases. However, the $\sigma = 1$ case results in poor prediction performance, as the state trajectory diverges significantly form the exact one. Table~\ref{table:prediction_error_Lorenz} provides the quantitative assessment of the prediction accuracy up to 8 time units for all noise cases. 
	\begin{remark}\label{remark:remark4}
		In the $\sigma = 0.1$ and $\sigma = 1$ cases, specially the latter, we noticed that the coefficients recovered for some of the 100 realizations yielded unstable dynamical models that could not be integrated for state prediction. We removed those cases for computing the average and the standard deviation to generate the plot in Fig.~\ref{fig:Lorenz63_samples_predictions} (right panel) and assess the prediction accuracy in Tab.~\ref{table:prediction_error_Lorenz}.
	\end{remark}
	\begin{figure}[H]
		\centering
		\includegraphics[trim = 10 0 10 0, clip,width=0.49\textwidth]{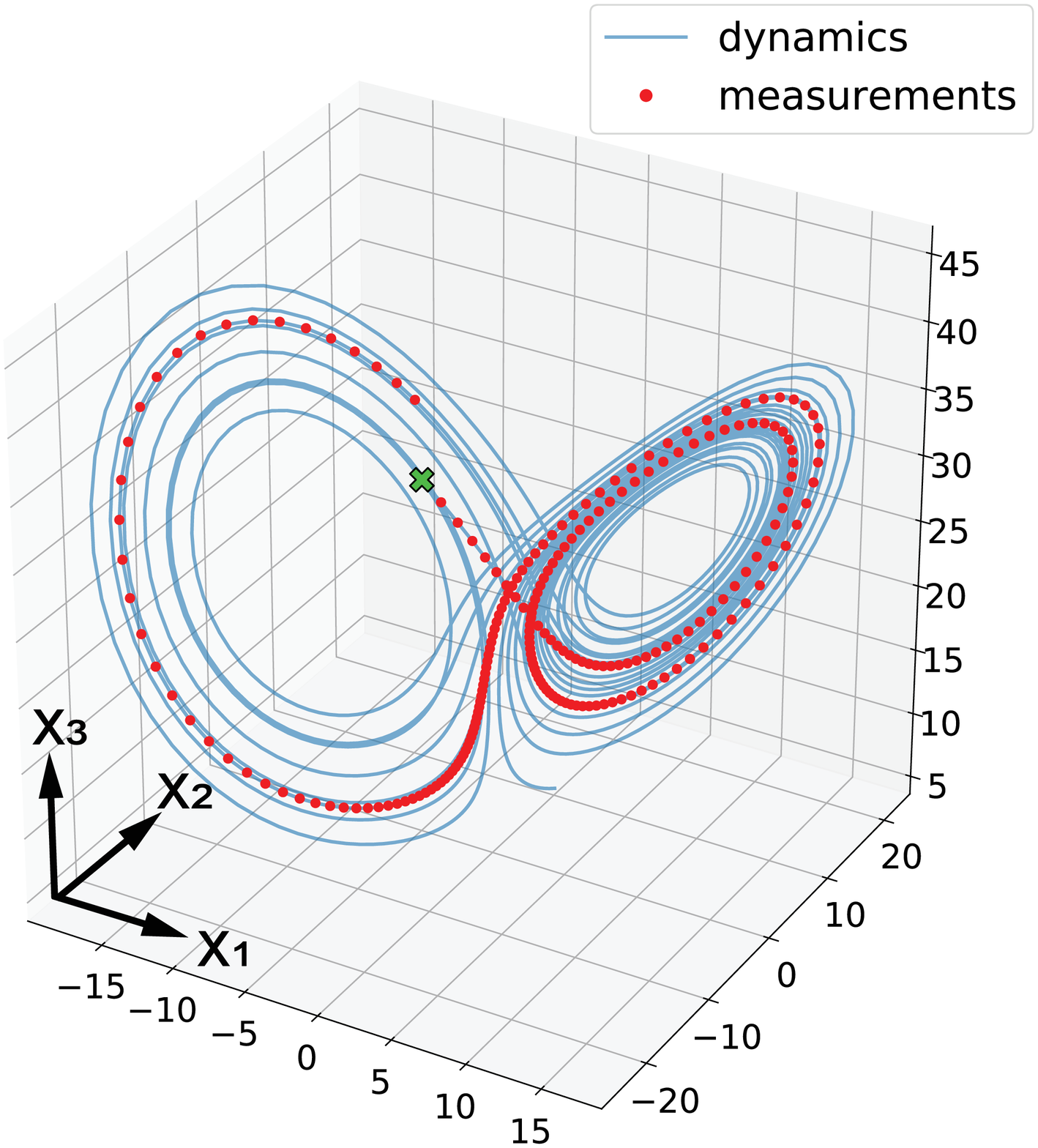}
		\includegraphics[trim = 10 0 10 0, clip,width=0.49\textwidth]{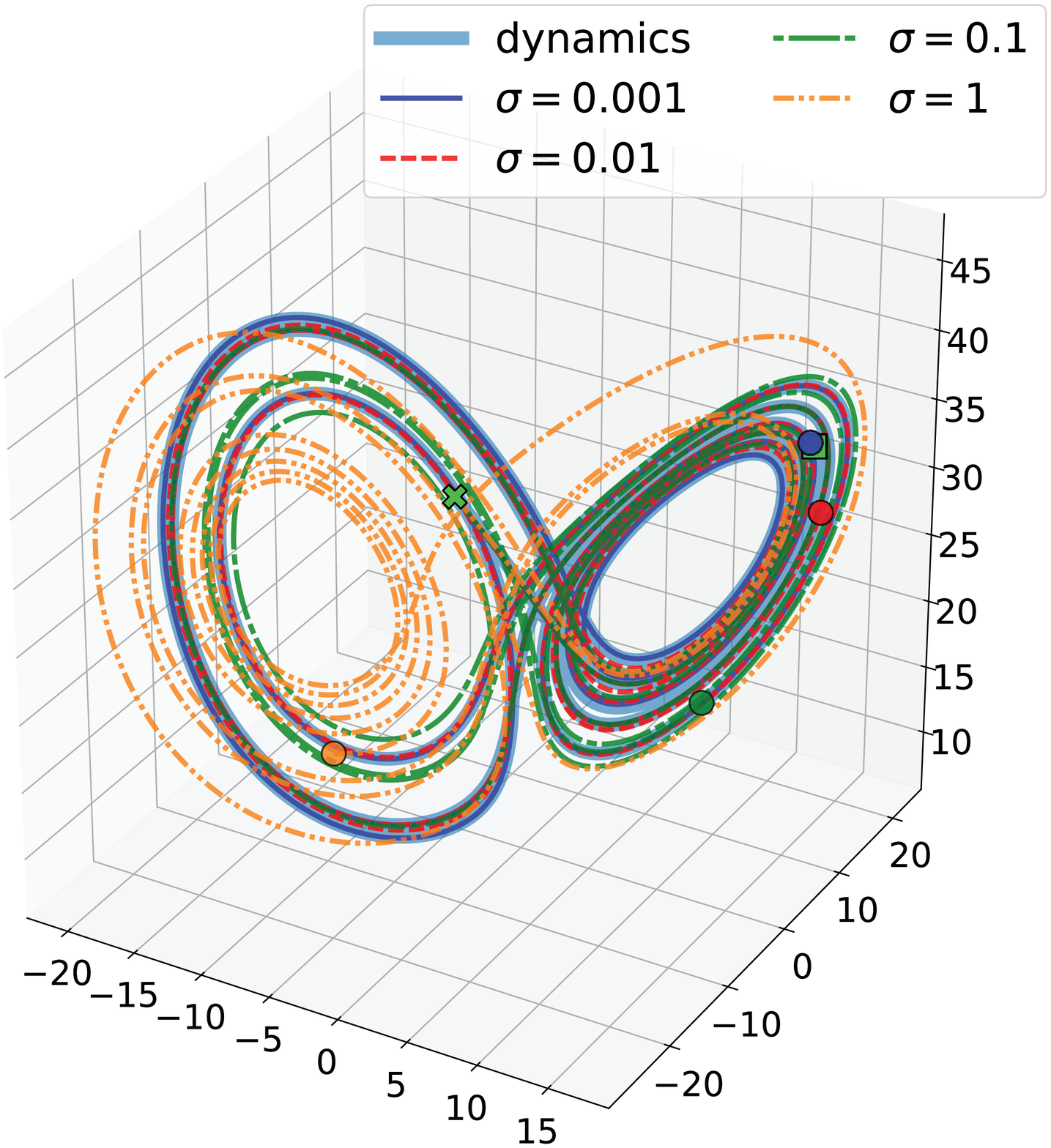}
		\caption{Predicted states for Lorenz 63 from the recovered governing equations for different noise levels using $\ell_1$-trend filter/Pareto for filtering and WBPDN/Pareto for sparse regression. Left: The true dynamics are highlighted in blue, and discrete observations used for training are represented by the red dots. The green cross indicates the initial condition. Right: State predictions for different noise levels up to 8 time units.}
		\label{fig:Lorenz63_samples_predictions}
	\end{figure}
	\begin{table*}
		\centering
		\caption{Mean prediction errors and their standard deviations (enclosed in parenthesis) up to 8 time units for Lorenz 63  for different noise levels using $\ell_1$-trend filter/Pareto for filtering and WBPDN/Pareto for sparse regression. For each of the recovered models from the 100 noisy trajectory realizations, we predicted the state trajectories via integrating the identified governing equations. We then computed the mean and standard deviation of the prediction errors defined in Eqn.~(\ref{eq:relative_filter_errors}).}
		\label{table:prediction_error_Lorenz}
		\resizebox{\linewidth}{!}{
			\begin{tabular}{lcccc}
				\toprule
				Noise &  \multicolumn{1}{c}{$\sigma = 0.001$} & \multicolumn{1}{c}{$\sigma = 0.01$} & \multicolumn{1}{c}{$\sigma = 0.1$} & \multicolumn{1}{c}{$\sigma = 1$} \\
				\midrule
				\addlinespace[10pt]
				$e_{x_1}$    & 3.46$\times 10^{-2}$ (3.05$\times 10^{-2}$) & 4.24$\times 10^{-1}$ (3.90$\times 10^{-1}$) & 8.47$\times 10^{-1}$ (2.69$\times 10^{-1}$) & 1.27$\times 10^{0}$ (1.89$\times 10^{-1}$)\\
				$e_{x_2}$    & 4.60$\times 10^{-2}$ (4.07$\times 10^{-2}$) & 4.67$\times 10^{-1}$ (3.85$\times 10^{-1}$) & 8.68$\times 10^{-1}$ (2.58$\times 10^{-1}$) & 1.25$\times 10^{0}$  (1.54$\times 10^{-1}$)\\
				$e_{x_3}$    & 2.19$\times 10^{-2}$ (1.89$\times 10^{-2}$) & 1.18$\times 10^{-1}$ (7.88$\times 10^{-2}$) & 2.39$\times 10^{-1}$ (7.71$\times 10^{-2}$) & 4.03$\times 10^{-1}$ (7.66$\times 10^{-2}$)\\
				\midrule
				$e_X$      & 2.19$\times 10^{-2}$ (1.93$\times 10^{-2}$) & 2.18$\times 10^{-1}$ (1.78$\times 10^{-1}$) & 4.10$\times 10^{-1}$ (1.19$\times 10^{-1}$) & 5.93$\times 10^{0}$ (7.87$\times 10^{-2}$)\\
				\bottomrule
			\end{tabular}
		}
	\end{table*}
	
	Finally, we study the performance of local and global smoothing techniques for additive Gaussian colored noise. We generated noise $\epsilon_j$ for each state variable $j$ and different noise levels using a power spectral density (PSD) per unit of bandwidth proportional to $1/|f|^{d}$, where $f$ denotes frequency. Specifically, we simulated pink, blue, and brown noise corresponding to exponents $d = 1, -1, -2$, respectively, and added it to the true state trajectory.
		
	Figures~\ref{fig:filter_comparison_x_colored_noise} show the relative state error performance against noise level for all local and global smoothers using GCV and Pareto curves as hyperparameter selection methods. As illustrated, the color of noise does not affect the performance of the smoothing methods yielding almost identical results. The same conclusion as in the white noise case can be drawn: global smoothers outperform local smoothers, and $\ell_1$-trend filtering is slightly more accurate than the rest of the global methods. For completeness, the error performance on the time-derivatives can be found in Appendix~D. We omit the analysis of the governing equation recovery performance since the smoothing results with colored noise are similar to the white noise case (Fig.~\ref{fig:Lorenz63_filter_comparison}).
	\begin{figure}[H]
		\centering
		\includegraphics[trim = 10 0 10 0, clip,width=0.32\textwidth]{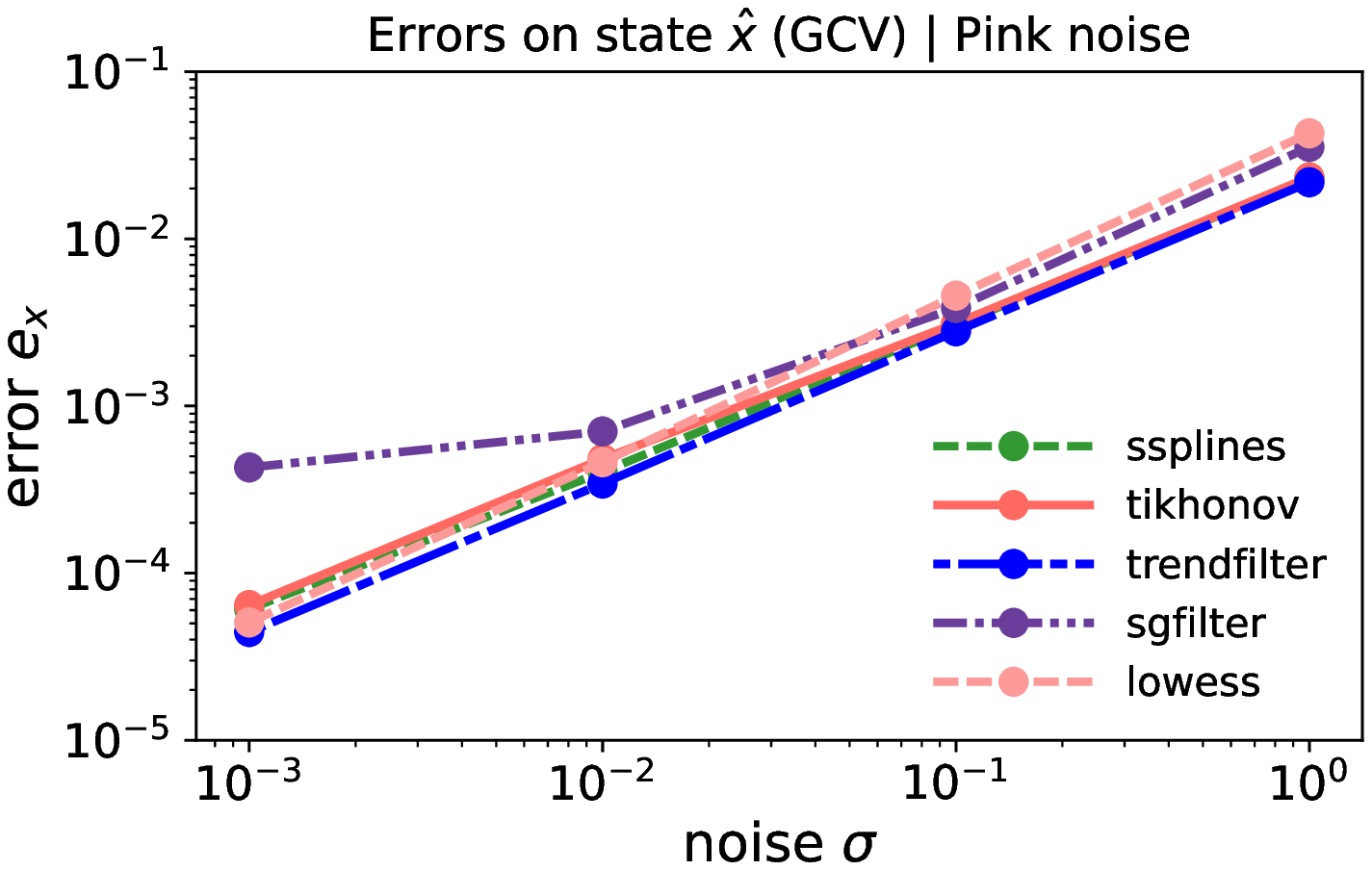}
		\includegraphics[trim = 10 0 10 0, clip,width=0.32\textwidth]{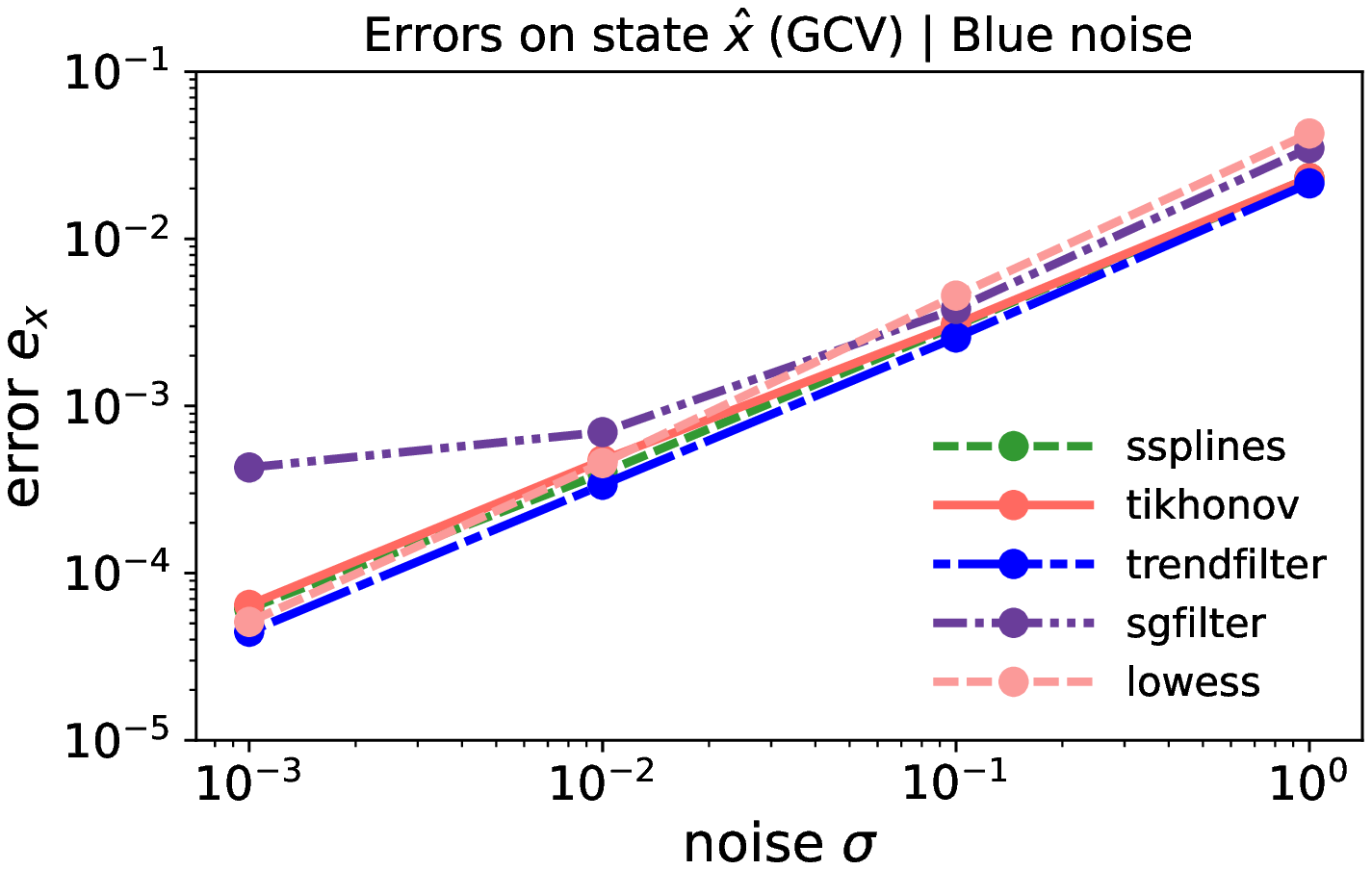}
		\includegraphics[trim = 10 0 10 0, clip,width=0.32\textwidth]{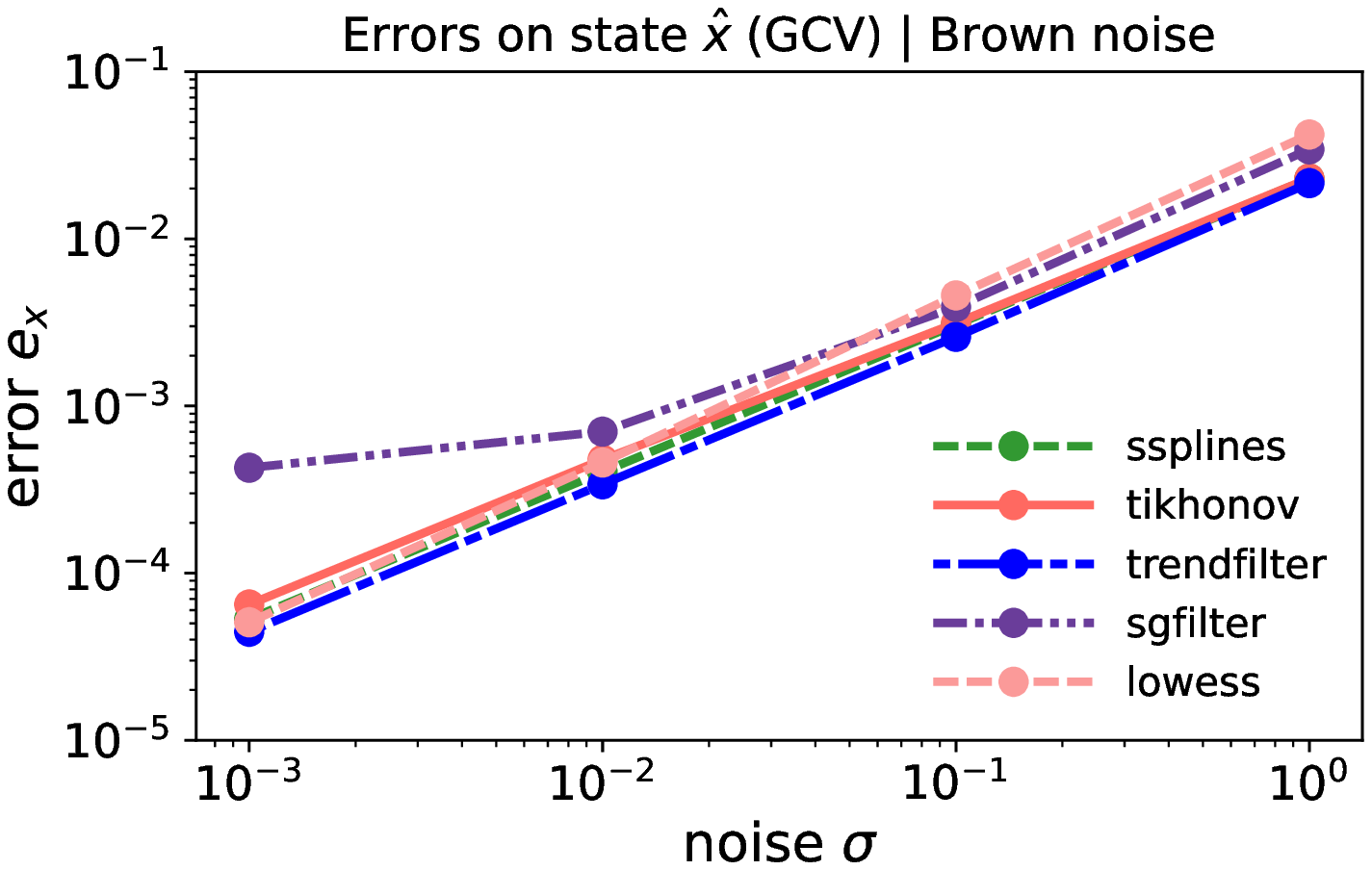}
		\includegraphics[trim = 10 0 10 0, clip,width=0.32\textwidth]{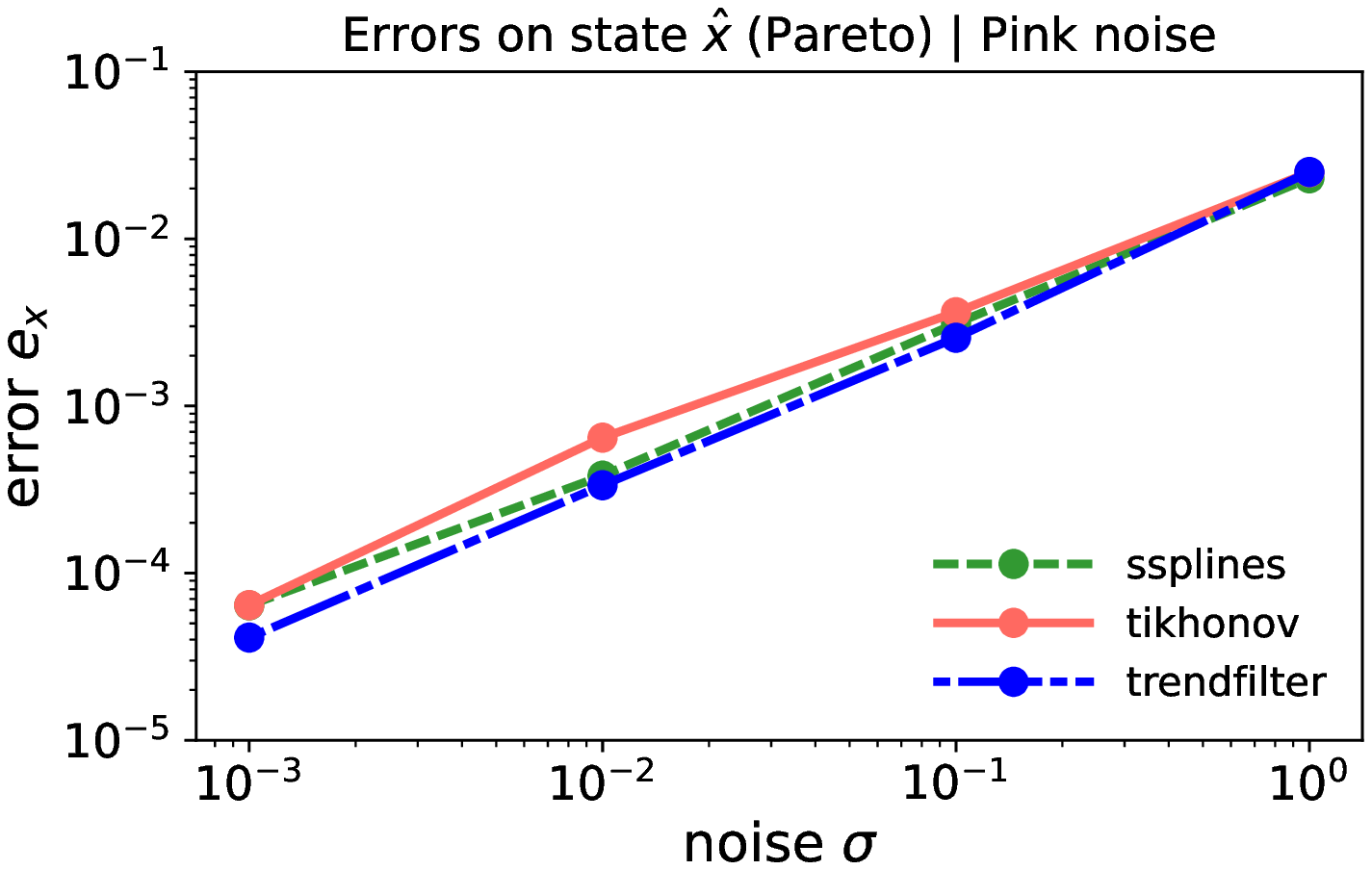}
		\includegraphics[trim = 10 0 10 0, clip,width=0.32\textwidth]{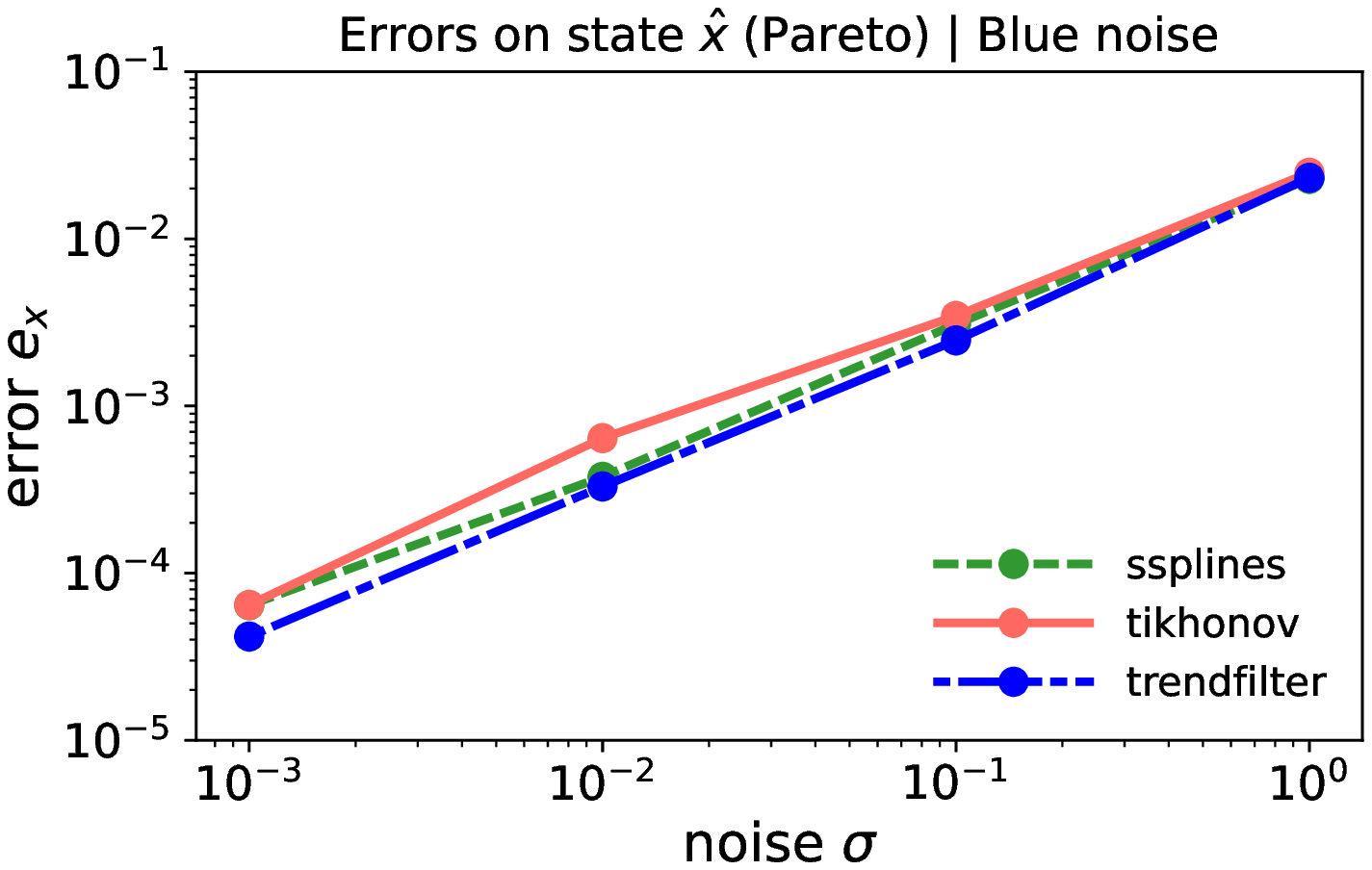}
		\includegraphics[trim = 10 0 10 0, clip,width=0.32\textwidth]{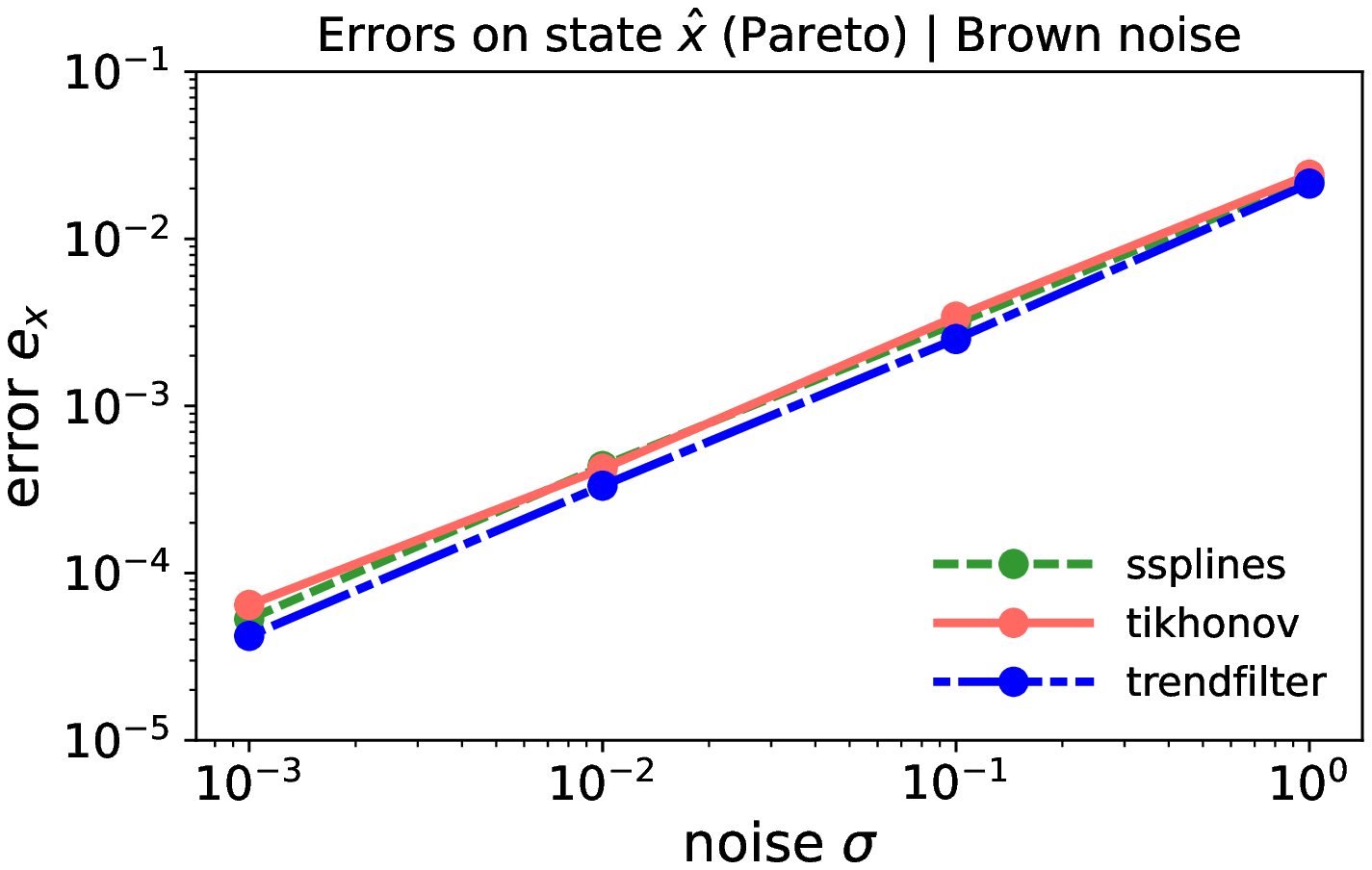}
		\caption{Comparison of the relative state errors for the Lorenz 63 system using local and global smoothers with pink, blue and brown noise (from left to right). The regularization parameters were computed using GCV (top row) and the Pareto curve criterion (bottom row).}
		\label{fig:filter_comparison_x_colored_noise}
	\end{figure}
	%
	\subsection{Duffing and Van der Pol oscillators}
	The Duffing~\cite{duffing1918forced} and Van der Pol~\cite{VanDerPol1920} oscillators are classical benchmark problems in nonlinear dynamics which feature a cubic nonlinearity. The equation describing the dynamics of the unforced oscillators is given by
	\begin{equation}\label{eq:nonlinear_oscillator}
		\ddot{\zeta} + D(\zeta)\dot{\zeta} + K(\zeta)\zeta  = 0,
	\end{equation}
	where $D(\zeta)$ and $K(\zeta)$ are the nonlinear damping and stiffness functions. The second-order system in Eqn.~(\ref{eq:nonlinear_oscillator}) can be transformed into first-order by setting $x_1 = \zeta$ and $x_2 = \dot{\zeta}$, giving
	\begin{subequations}
		\begin{alignat}{2}
			\dot{x}_1 = x_2, &\quad x_1(0) = x_{1,0},\label{eq:DuffingSys_a}\\
			\dot{x}_2 = - D(x_1)x_2 - K(x_1)x_1, &\quad x_2(0) = x_{2,0}.\label{eq:DuffingSys_b}
		\end{alignat}
	\end{subequations}
	The Duffing oscillator physically models a spring-damper-mass system with a spring whose stiffness is $K(x_1) = \kappa + \varepsilon x_1^2$, and constant damping $D(x_1) = \delta$. We focus on the non-chaotic, stable spiral case around the equilibrium $(x_{1,e},x_{2,e}) = (0,0)$, where the parameters of the system are set to $\kappa = 1$, $\delta = 0.1$ and $\varepsilon = 5$, and the initial condition to $(x_{1,0},x_{2,0}) = (1,0)$.
	The Van der Pol system is a second-order oscillator with a nonlinear damping term of the form $D(x_1) = -\gamma + \mu x_1^2$, and constant stiffness $K(x_1) = 1$. It was originally proposed by Van der Pol as a model to describe the oscillation of a triode in an electrical circuit. The Van der Pol oscillator exhibits a limit cycle behavior around the origin $(x_{1,e},x_{2,e}) = (0,0)$. For this case, we set the parameters to $\gamma = 2$ and $\mu = 2$, and the initial condition to $(x_{1,0},x_{2,0}) = (0,1)$.
	For both nonlinear oscillators, the number of state variables is $n = 2$ and the degree of the polynomial basis is set to $d = 4$, yielding $p = 15$ monomial terms. Out of these, only 4 describe the dynamics. In this case, the displacement $x$ and velocity $y$ are measured and we used 201 samples over 2 time units, from $t$ = 0.1 to $t$ = 2.1, to recover the system.
	
	We present the performance of the different filtering strategies to denoise the state and estimate time-derivatives in Fig.~\ref{fig:Duffing_filter_comparison} for the Duffing and Fig.~\ref{fig:VanderPol_filter_comparison} for the Van der Pol oscillators. Similar to the Lorenz 63 example, global methods outperform local methods for the model selection techniques used in this article.
	Again, $\ell_1$-trend filtering yields slightly better results than smoothing splines and Tikhonov smoother, specially when estimating derivatives. Both GCV and Pareto curves robustly select suitable $\lambda$ and perform similarly for the global methods.
	\begin{figure}[H]
		\centering
		\includegraphics[trim = 0 0 0 0, clip,width=0.48\textwidth]{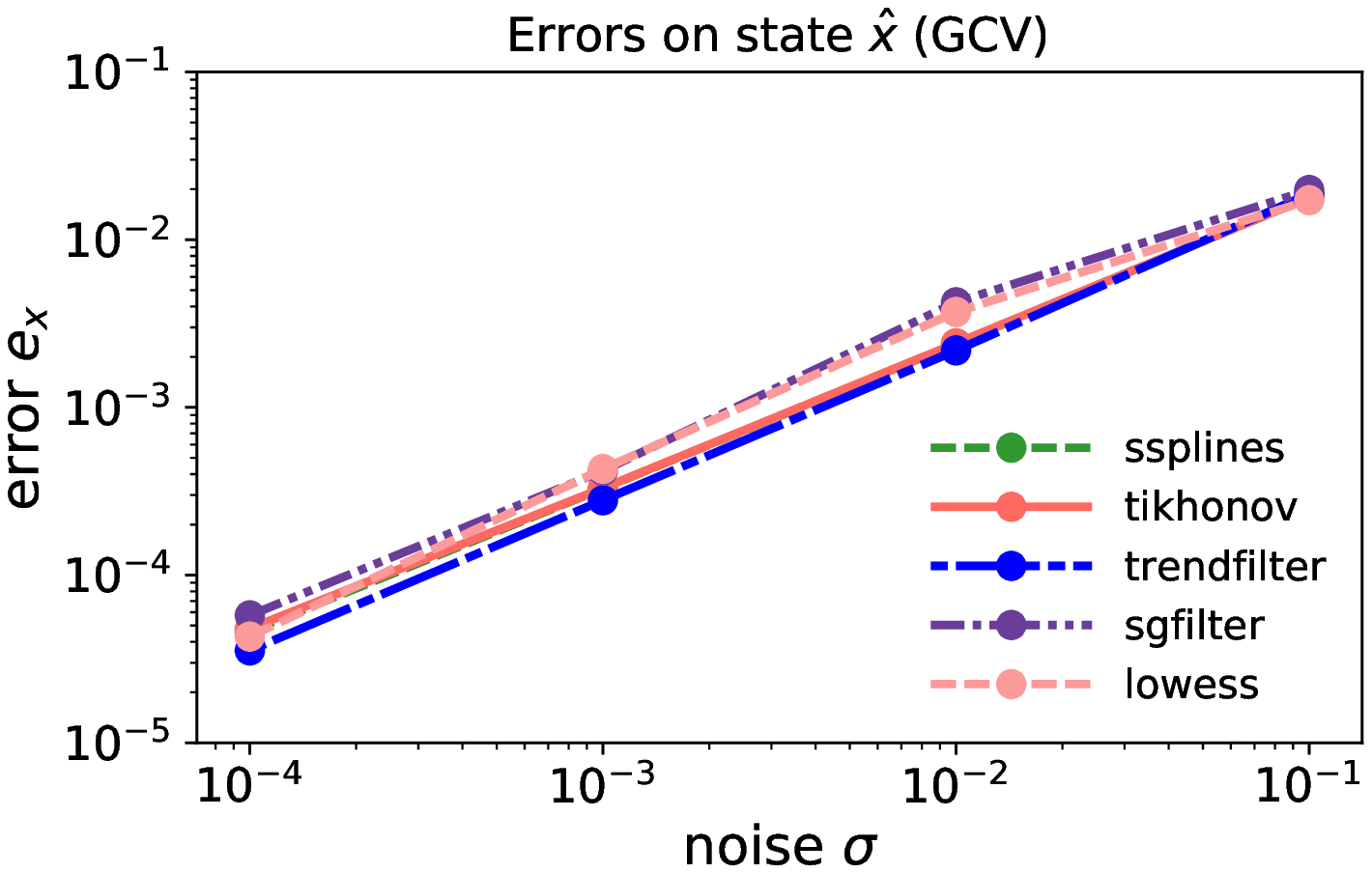}
		\includegraphics[trim = 0 0 0 0,
		clip,width=0.49\textwidth]{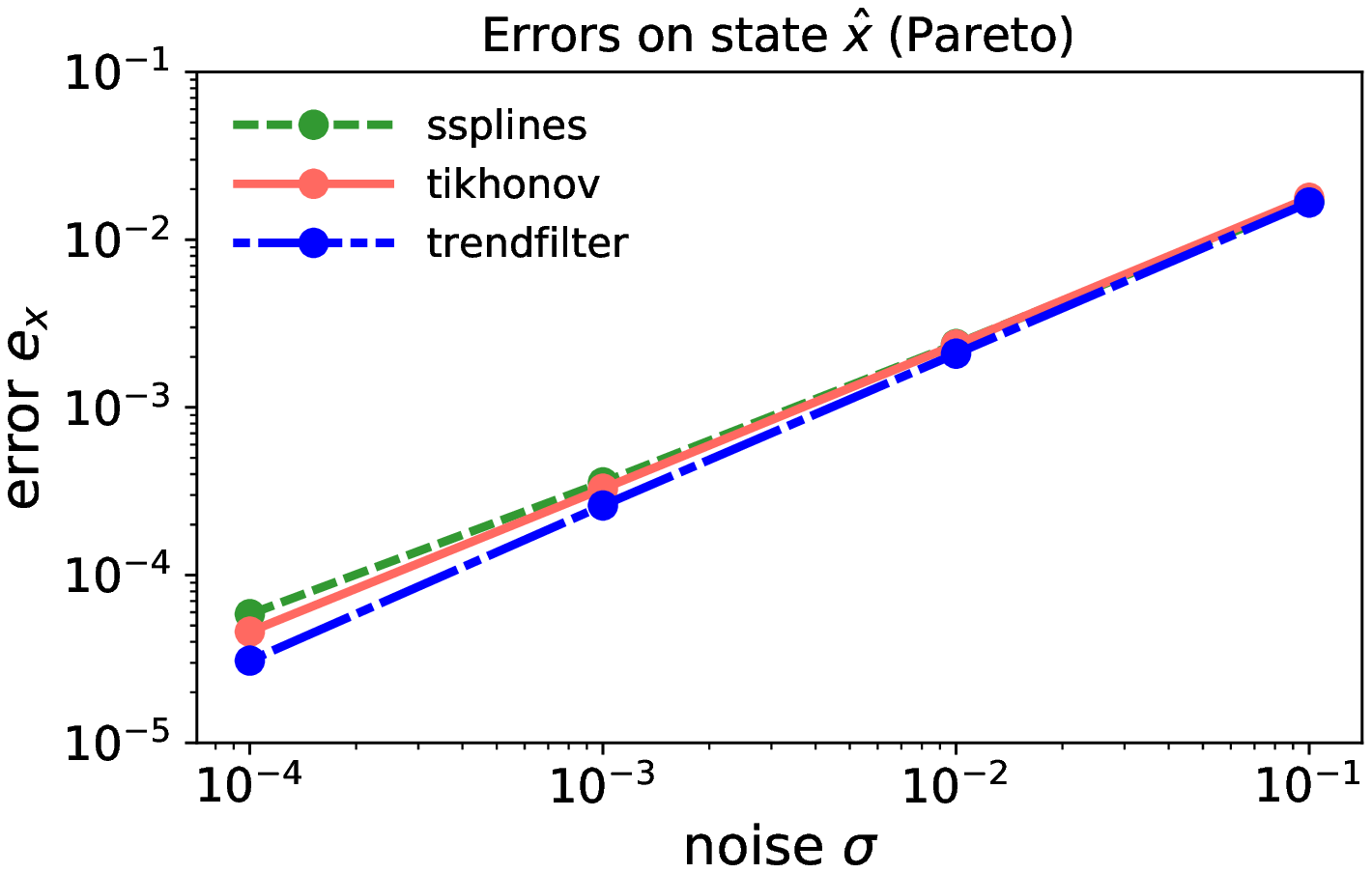}
		\includegraphics[trim = 0 0 0 0, clip,width=0.48\textwidth]{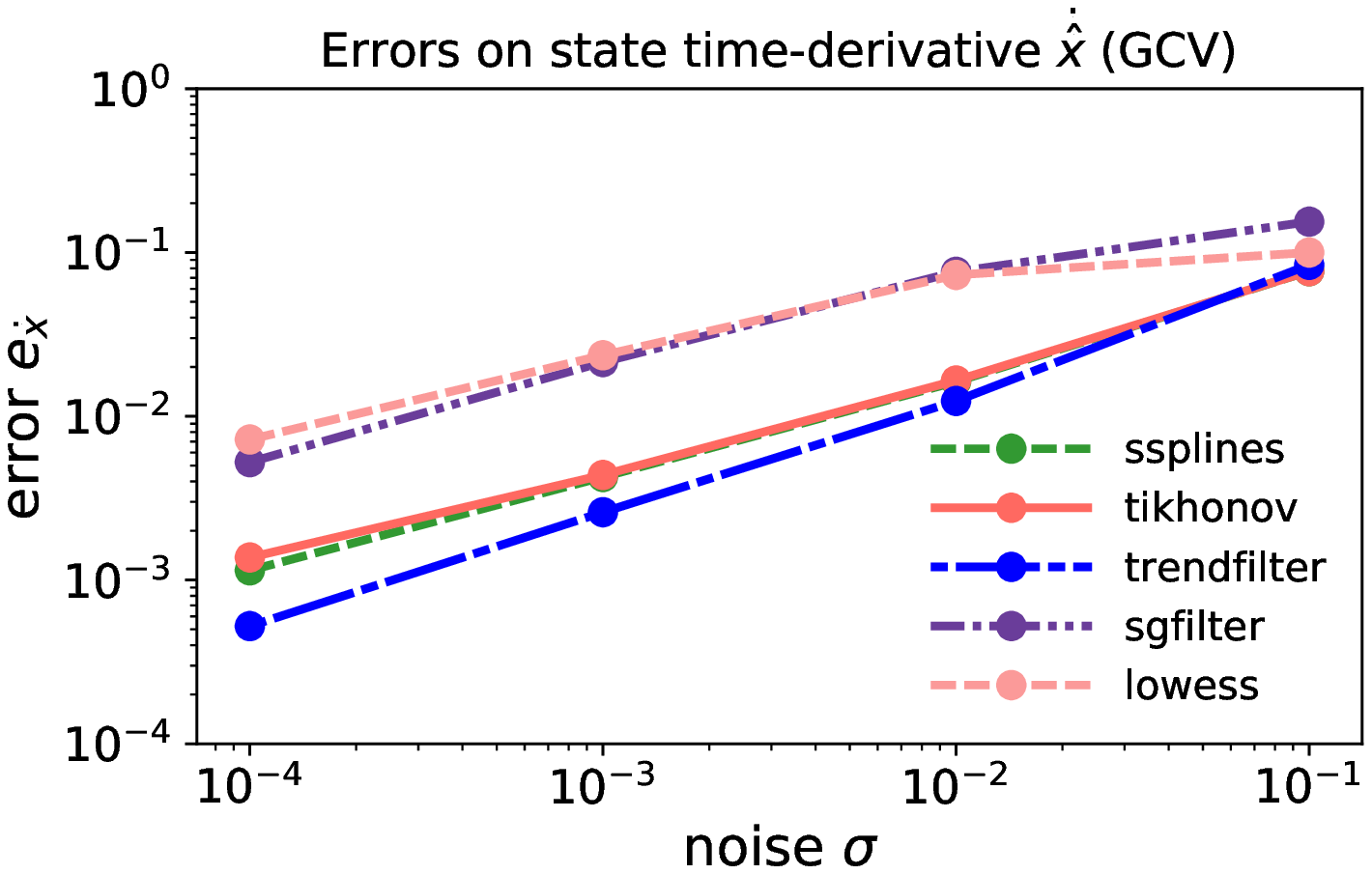}
		\includegraphics[trim = 0 0 0 0,
		clip,width=0.49\textwidth]{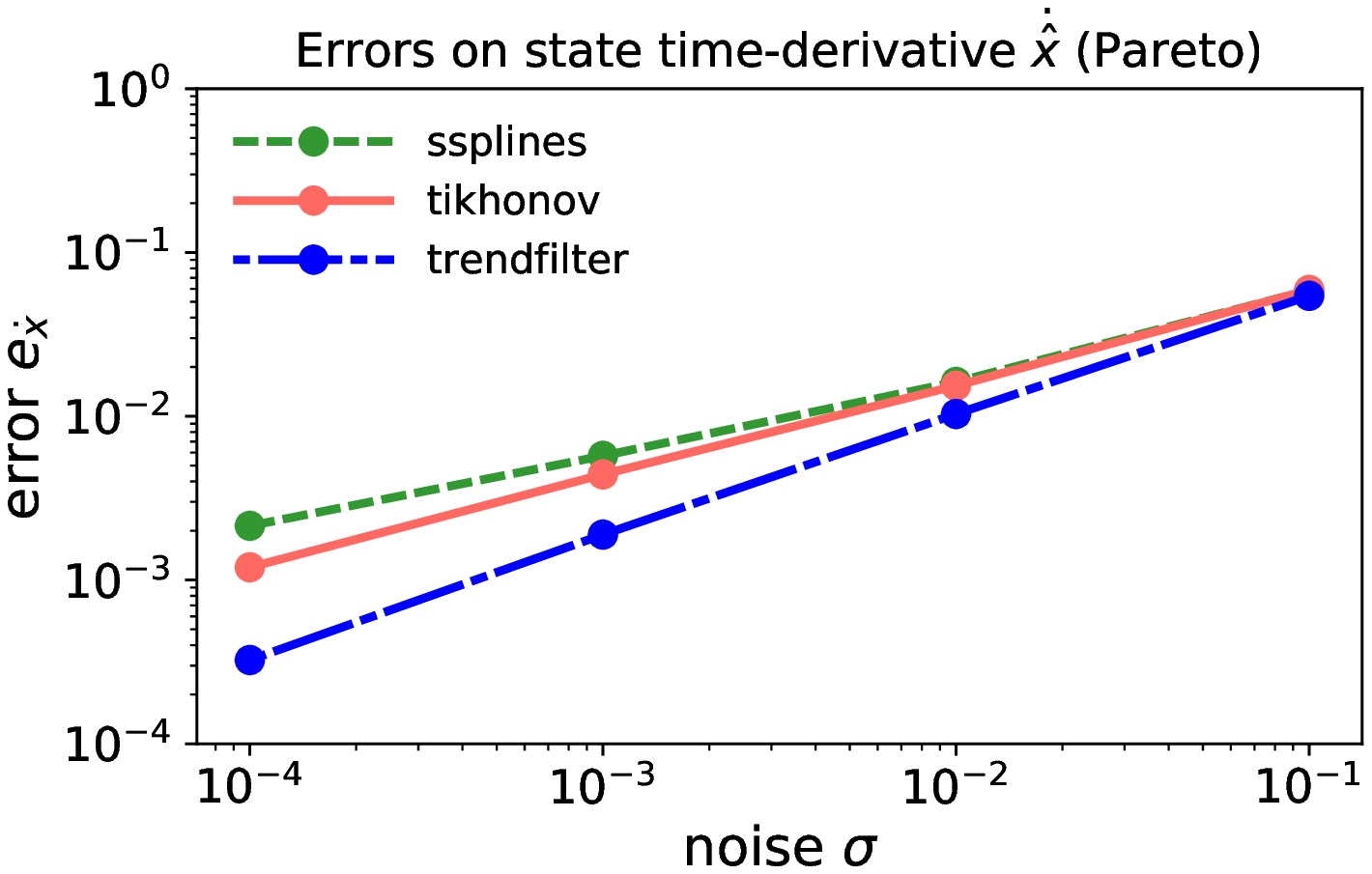}
		\caption{Comparison of the relative state (top row) and state time-derivative (bottom row) errors for the Duffing oscillator using local and global smoothers. The regularization parameters were computed using GCV (left column) and the Pareto curve criterion (right column).}
		\label{fig:Duffing_filter_comparison}
	\end{figure}
	\begin{figure}[H]
		\centering
		\includegraphics[trim = 0 0 0 0, clip,width=0.48\textwidth]{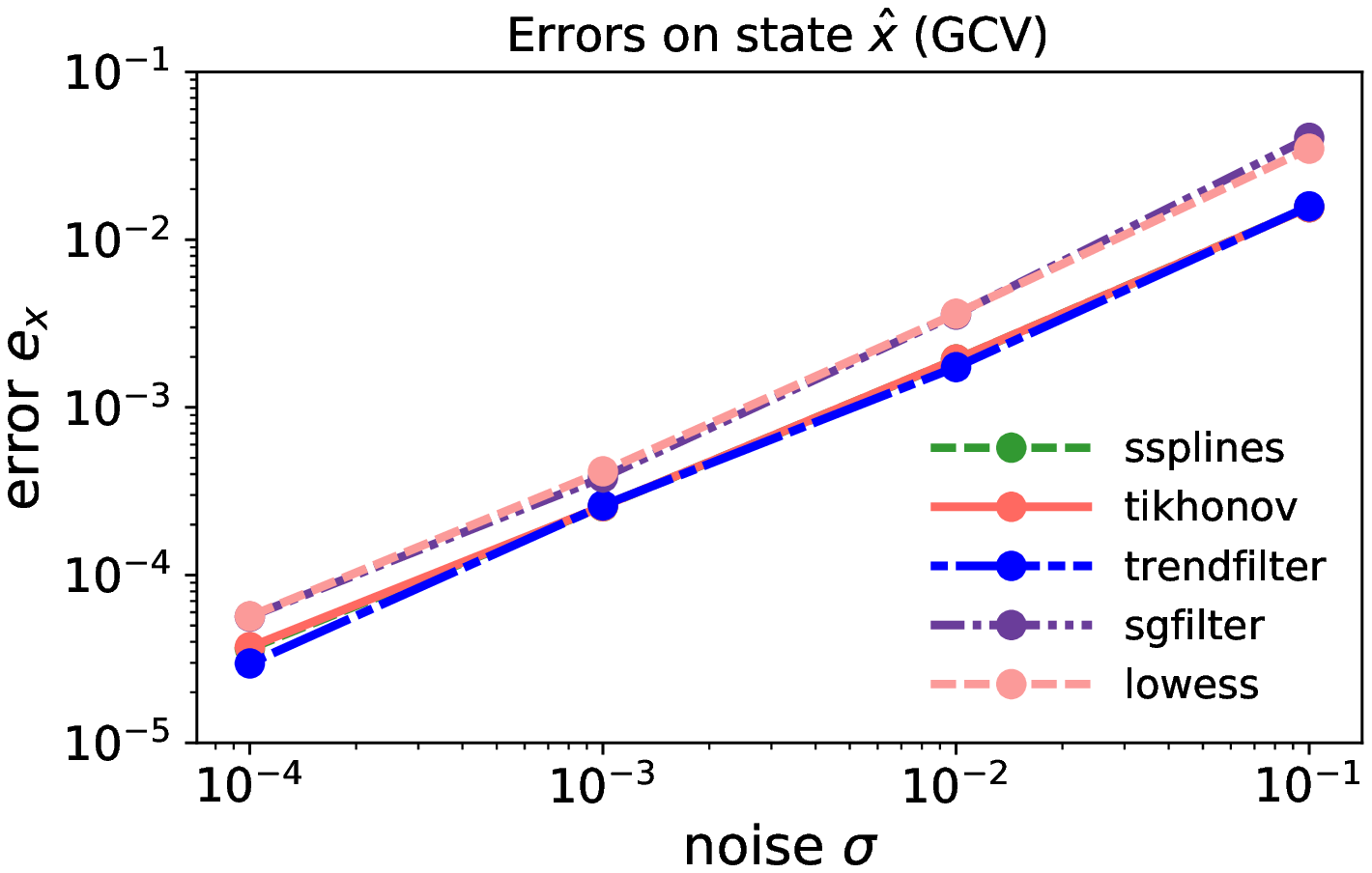}
		\includegraphics[trim = 0 0 0 0,
		clip,width=0.48\textwidth]{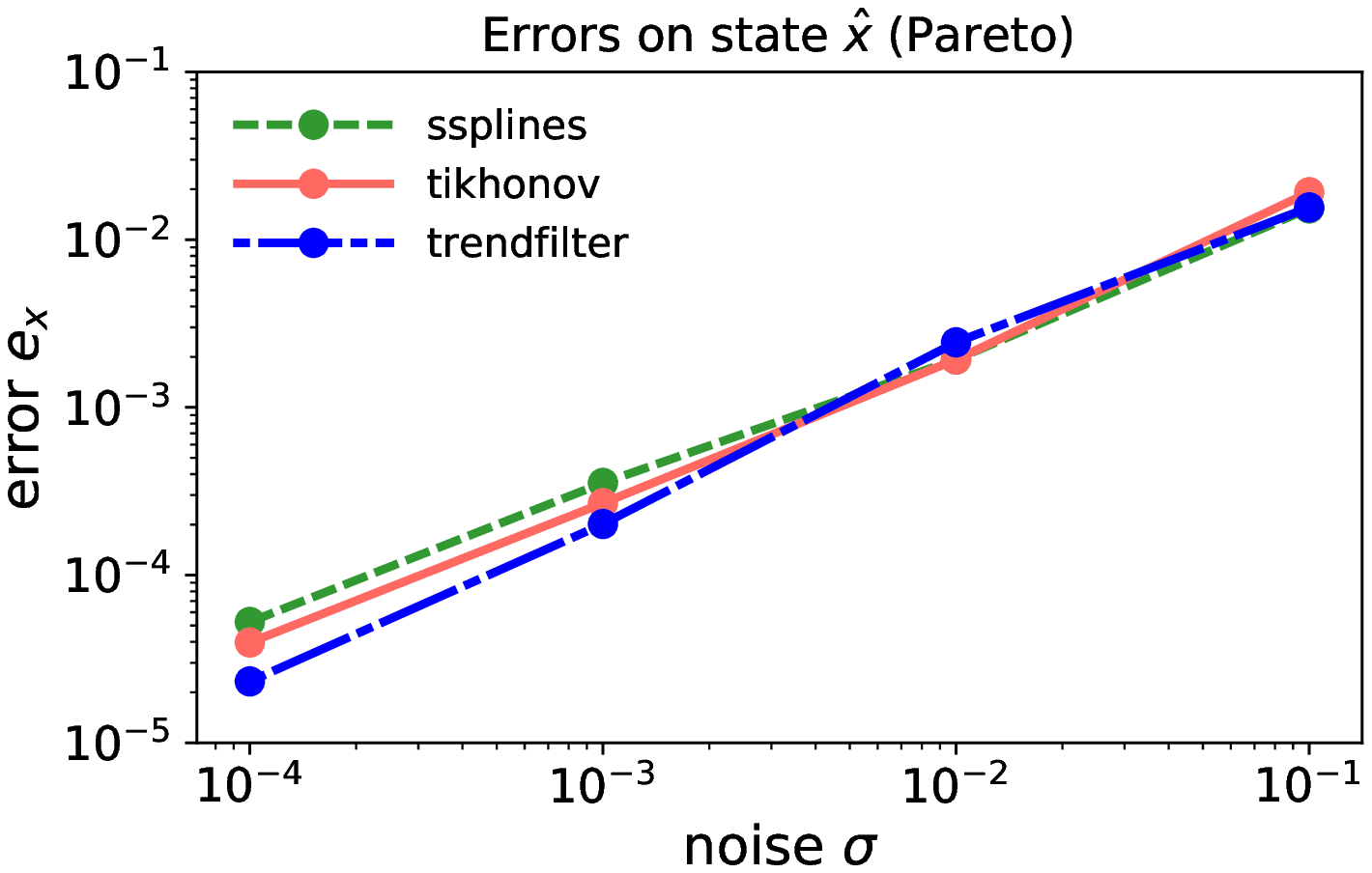}
		\includegraphics[trim = 0 0 0 0, clip,width=0.48\textwidth]{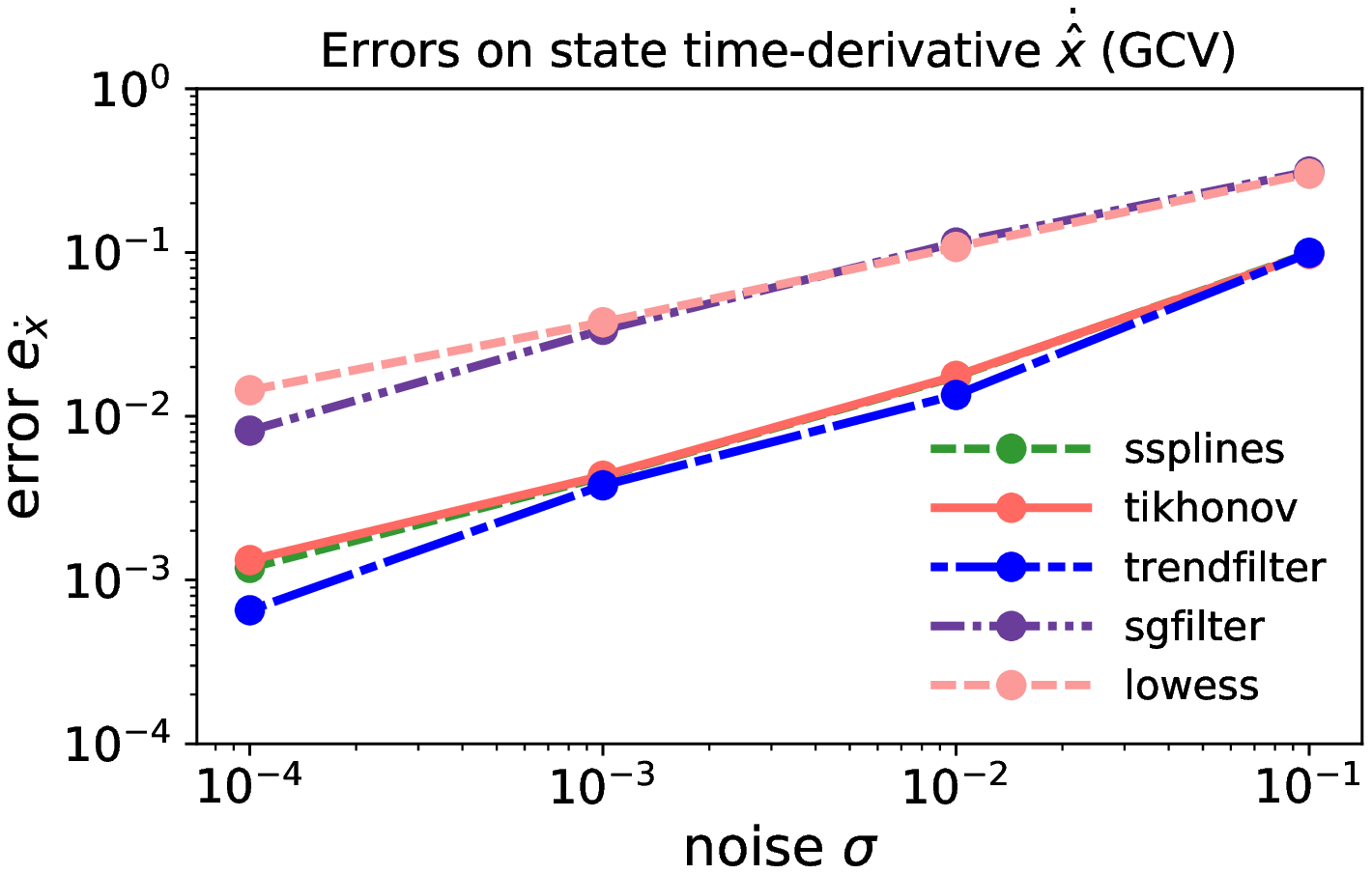}
		\includegraphics[trim = 0 0 0 0,
		clip,width=0.48\textwidth]{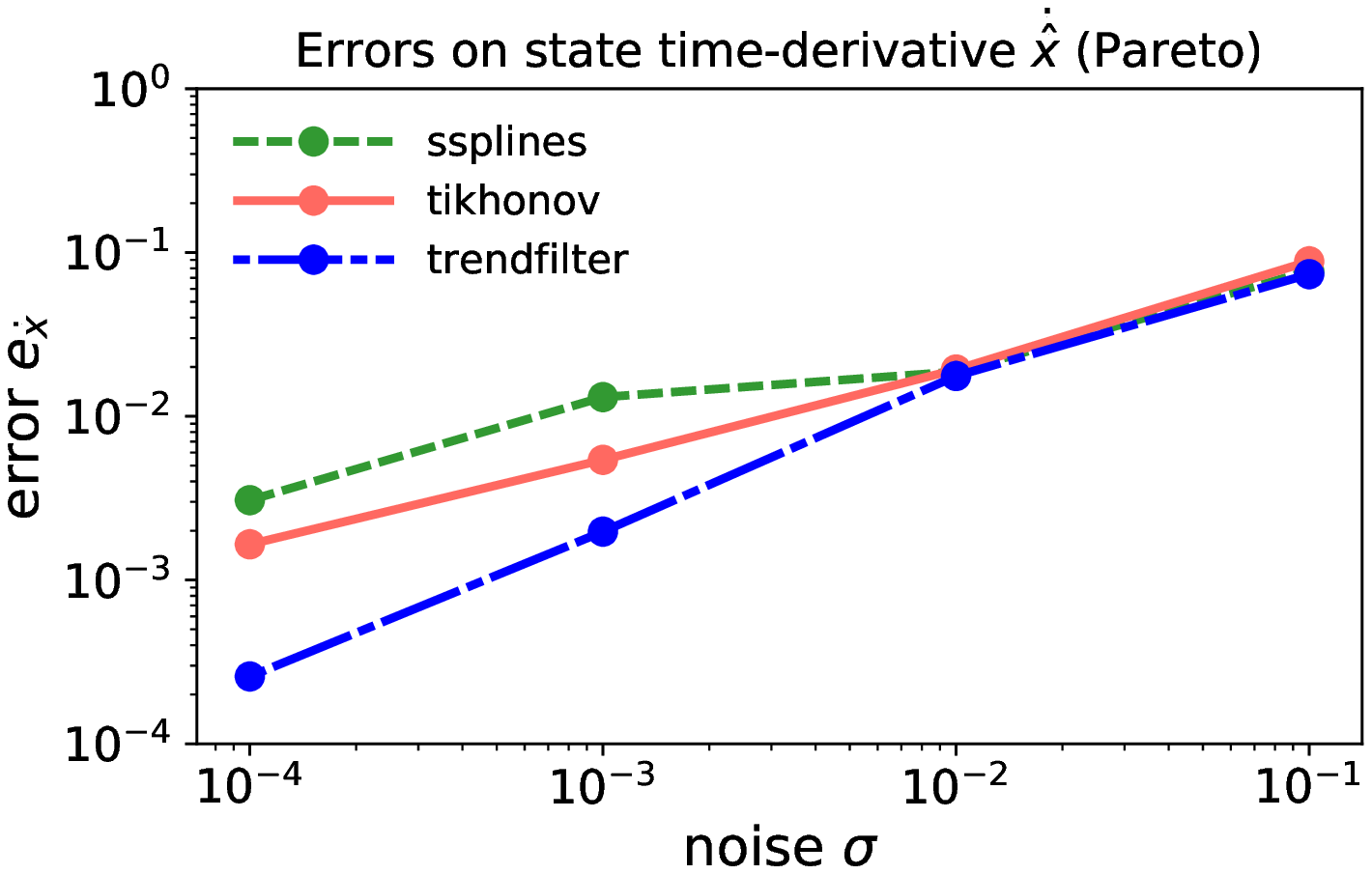}
		\caption{Comparison of the relative state (top row) and state time-derivative (bottom row) errors for the Van der Pol oscillator using local and global smoothers. The regularization parameters were computed using GCV (left column) and the Pareto curve criterion (right column).}
		\label{fig:VanderPol_filter_comparison}
	\end{figure}
	Similar observations can be made on the identification of both nonlinear oscillators. Figure~\ref{fig:wbpdn_Duff_and_VdP_filters} shows the accuracy of WBPDN for recovering the governing equation coefficients for Duffing (top panels) and Van der Pol (bottom panels). State measurements pre-processed with global filtering strategies yield similar recovery performance using WBPDN, and Pareto curves outperform GCV for estimating near optimal regularization parameters. We do not show the plots for Pareto curves and GCV functions and omit the comparison with STLS for these examples since we observed similar trends as in the Lorenz 63 case. 
	\begin{figure}[H]
		\centering
		
		\includegraphics[trim = 10 0 10 0, clip,width=0.48\textwidth]{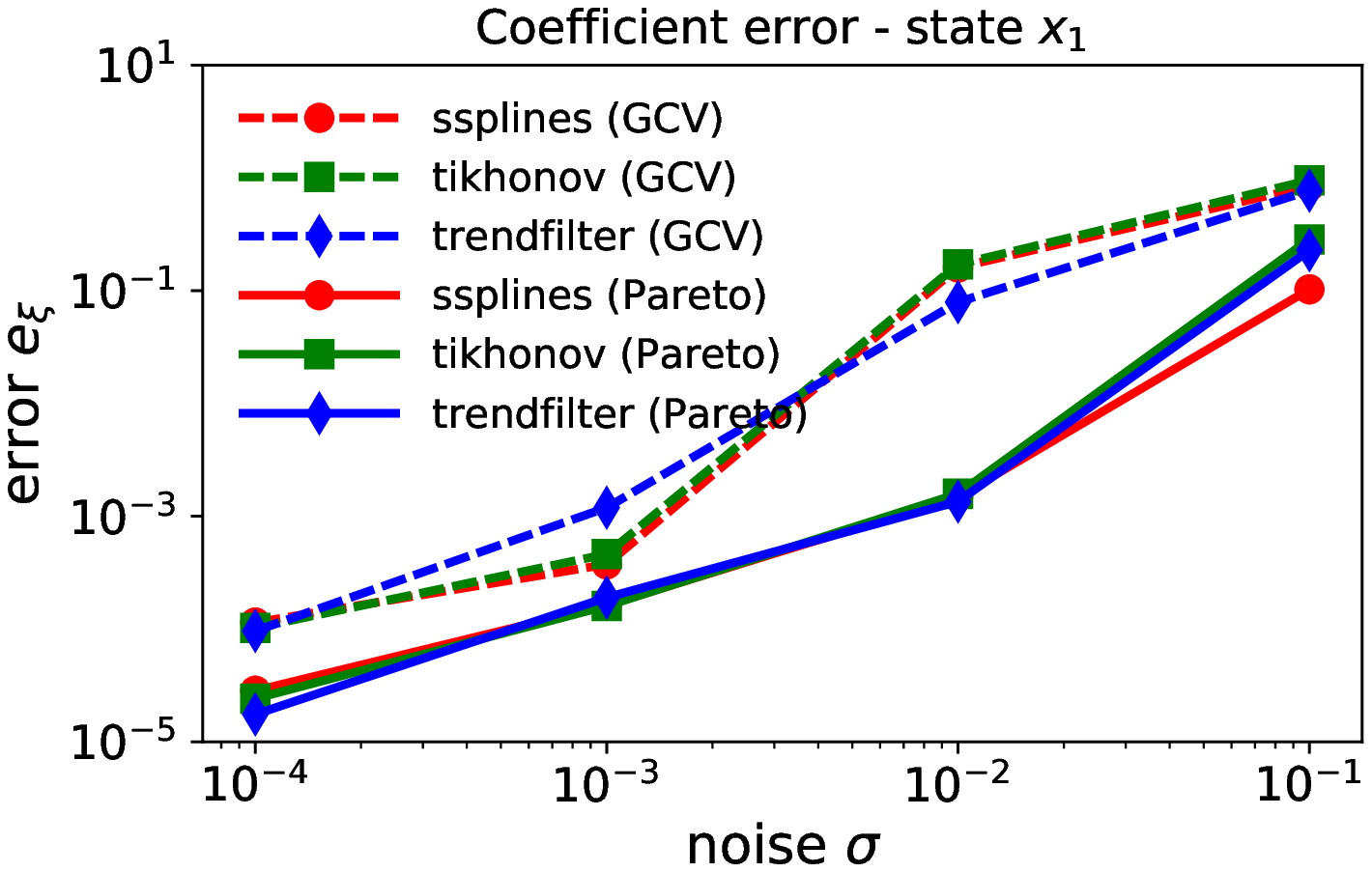}
		\includegraphics[trim = 10 0 10 0, clip,width=0.48\textwidth]{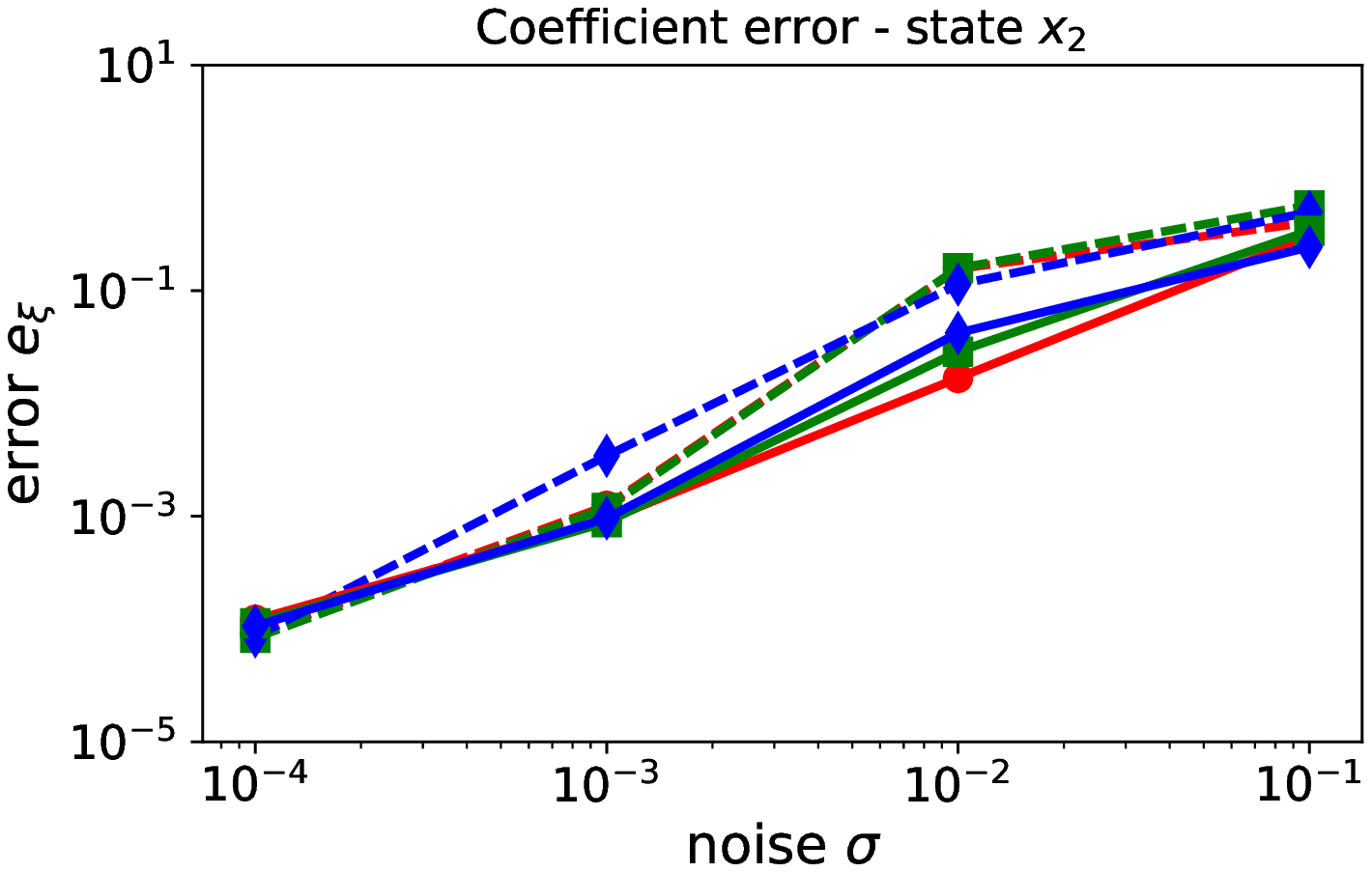}
		\includegraphics[trim = 10 0 10 0, clip,width=0.48\textwidth]{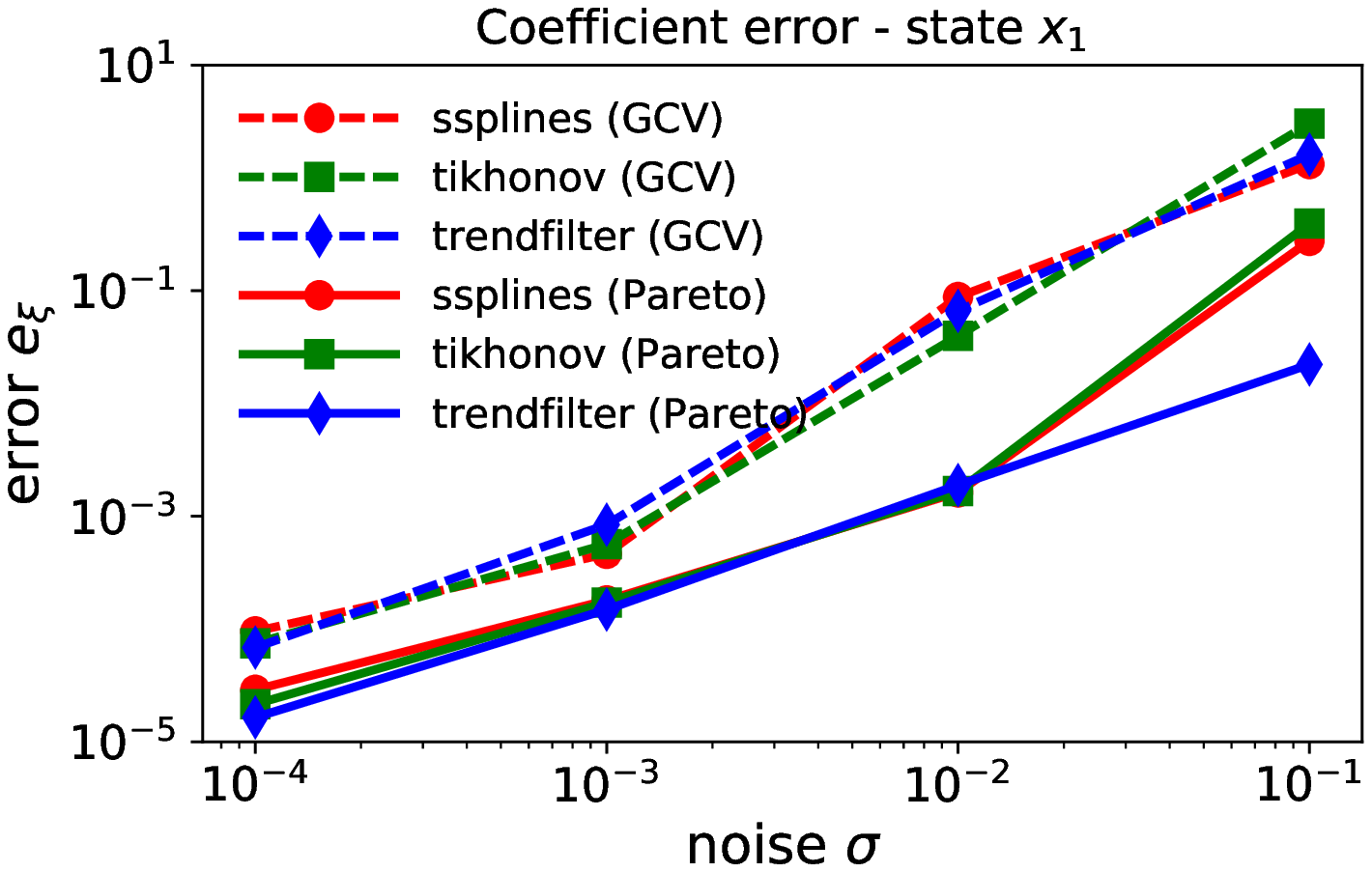}
		\includegraphics[trim = 10 0 10 0, clip,width=0.48\textwidth]{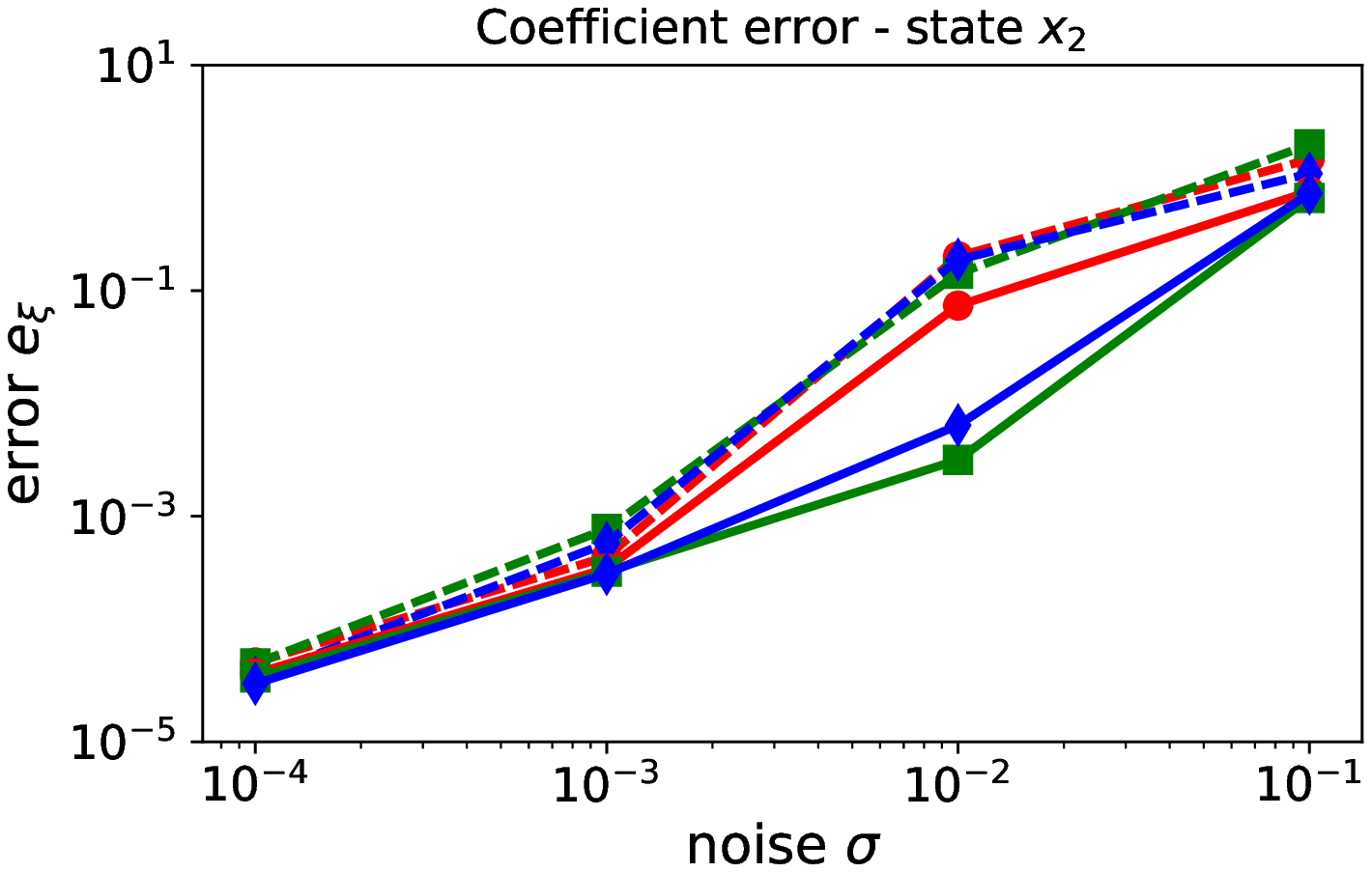}
		\caption{Relative  coefficient  errors  for  Duffing (top panels) and Van der Pol (bottom panels) oscillators via WBPDN with measurements pre-processed using smoothing splines, Tikhonov smoother and $\ell_1$-trend filtering. The dashed and solid lines correspond to the solution using GCV and Pareto curves for selecting $\lambda$, respectively.}
		\label{fig:wbpdn_Duff_and_VdP_filters}
	\end{figure}
	Finally, we report the prediction accuracy qualitatively in Fig.~\ref{fig:Duff_samples_predictions}, for Duffing, and Fig.~\ref{fig:VdP_samples_predictions}, for Van der Pol systems. We also report the prediction errors in Tab.~\ref{table:prediction_error_Duffing} and Tab.~\ref{table:prediction_error_VanderPol} up to 20 time units for the Duffing and Van der Pol, respectively. For these examples, we integrated the identified governing equations up to 20 time units, about 10 times the time span used for training. As highlighted in Remark~\ref{remark:remark4}, we removed those realizations that produced unstable dynamics for higher noise levels. As shown, the predictions match well with the exact trajectory for noise levels up to $\sigma = 0.01$, and diverges for the worst noise case $\sigma = 0.1$. It is remarkable that the recovered models capture the dynamical behavior of the Duffing and Van der Pol oscillators using a small number of discrete state samples. In the Duffing case, both the damping rate and cubic nonlinearity are well identified, except for the highest noise case. For the Van der Pol example, the identified models recover the limit cycle behavior even though the training set does not include any sample along the limit cycle trajectory.
	\begin{figure}[H]
		\centering
		\includegraphics[trim = 10 0 10 0, clip,width=0.45\textwidth]{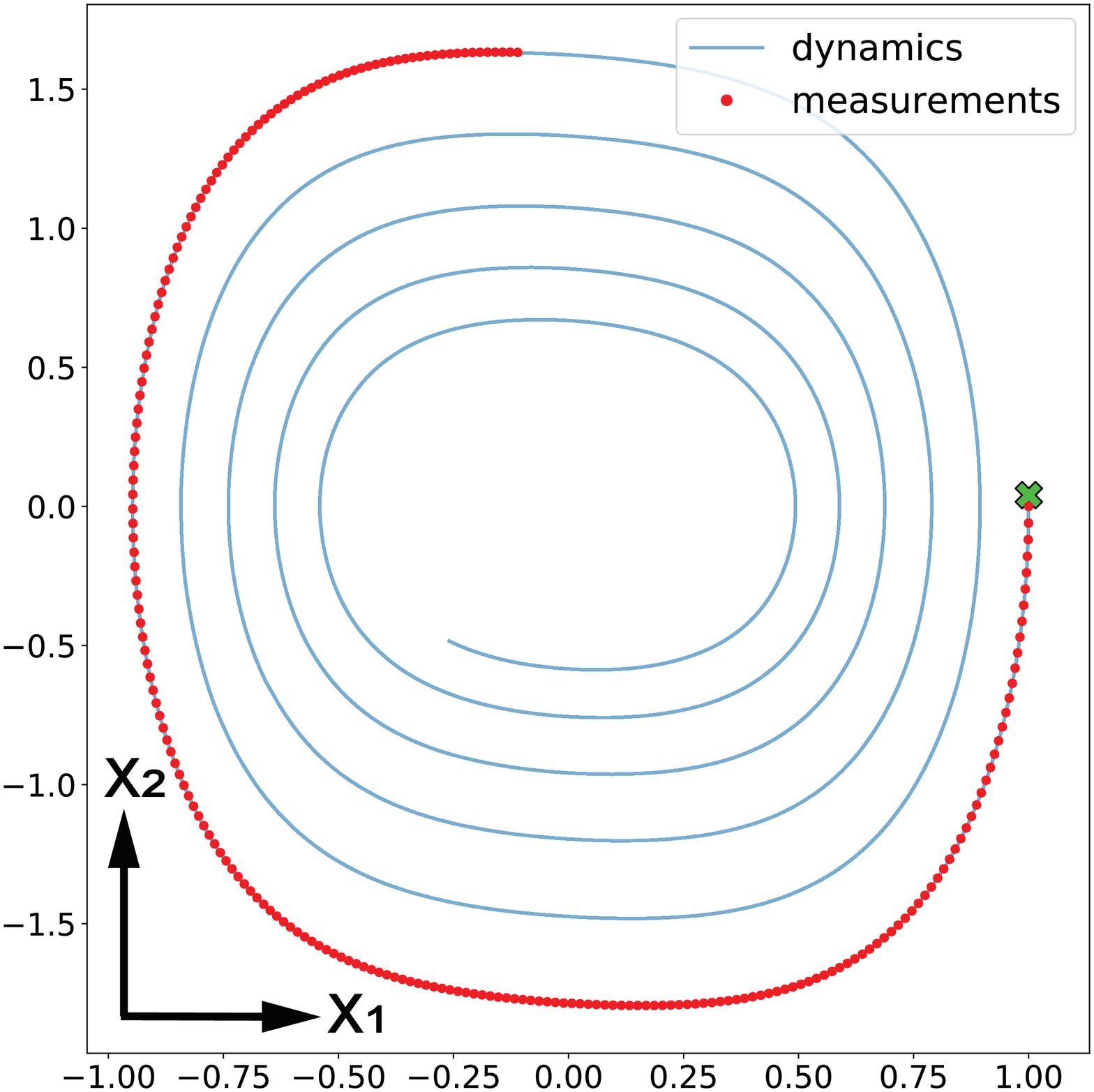}
		\includegraphics[trim = 10 0 10 0, clip,width=0.45\textwidth]{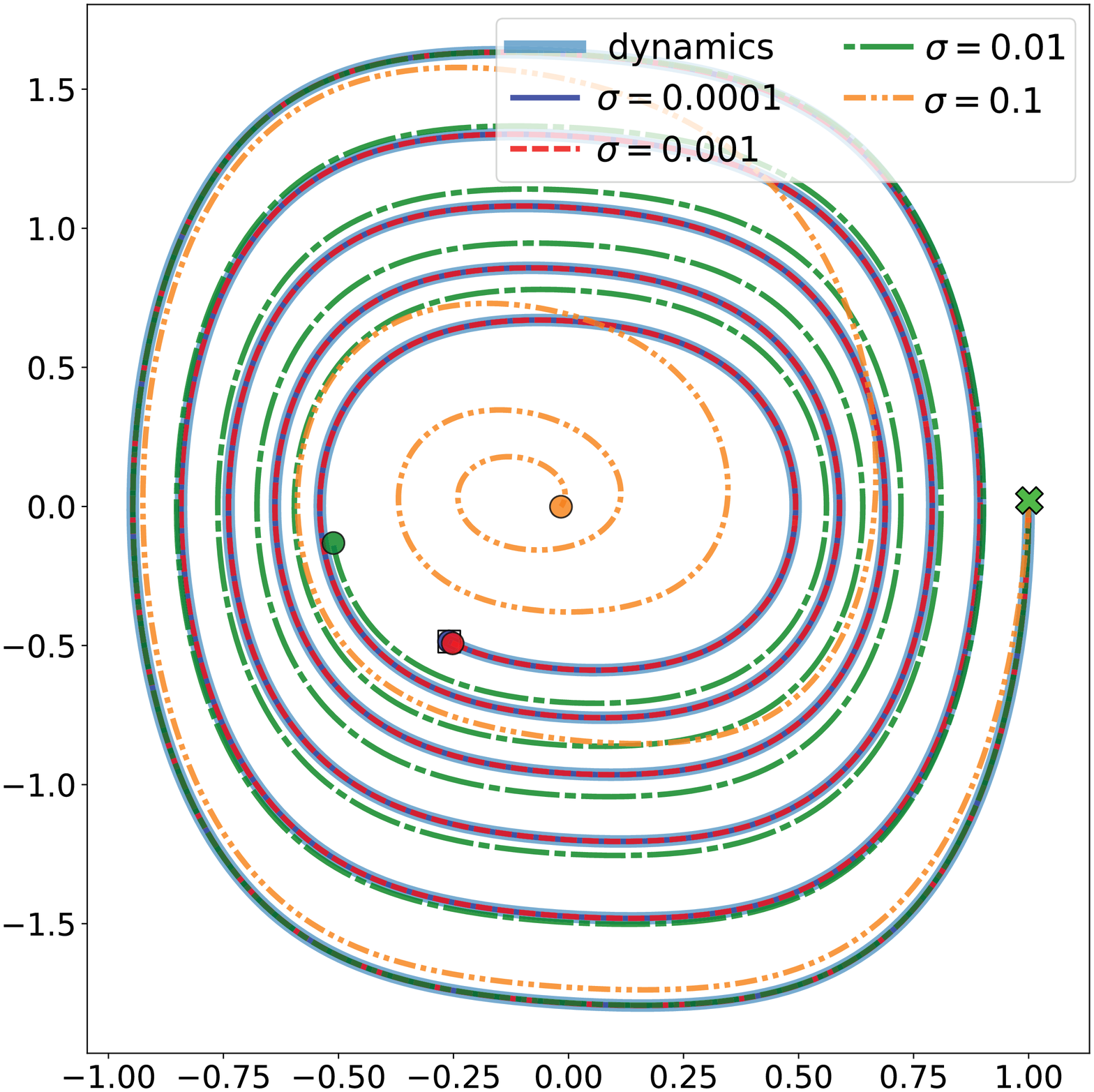}
		\caption{Predicted states for the Duffing oscillator from the recovered governing equations for different noise levels using $\ell_1$-trend filter/Pareto for filtering and WBPDN/Pareto for sparse regression. Left: The true dynamics are highlighted in blue, and discrete observations used for training are represented by the red dots. The green cross indicates the initial condition. Right: State predictions for different noise levels up to 20 time units.}
		\label{fig:Duff_samples_predictions}
	\end{figure}
	\begin{table*}
		\centering
		\caption{Mean prediction errors and their standard deviations (enclosed in parenthesis) up to 20 time units for Duffing oscillator for different noise levels using $\ell_1$-trend filter/Pareto for filtering and WBPDN/Pareto for sparse regression. For each of the recovered models from the 100 noisy trajectory realizations, we predicted the state trajectories via integrating the identified governing equations. We then computed the mean and standard deviation of the prediction errors defined in Eqn.~(\ref{eq:relative_filter_errors}).}
		\label{table:prediction_error_Duffing}
		\resizebox{\linewidth}{!}{
			\begin{tabular}{lcccc}
				\toprule
				Noise &  \multicolumn{1}{c}{$\sigma = 0.0001$} & \multicolumn{1}{c}{$\sigma = 0.001$} & \multicolumn{1}{c}{$\sigma = 0.01$} & \multicolumn{1}{c}{$\sigma = 0.1$} \\
				\midrule
				\addlinespace[10pt]
				$e_{x_1}$    & 9.06$\times 10^{-4}$ (6.96$\times 10^{-4}$) & 8.39$\times 10^{-3}$ (5.79$\times 10^{-3}$) & 3.07$\times 10^{-1}$ (3.78$\times 10^{-1}$) & 9.76$\times 10^{-1}$ (2.89$\times 10^{-1}$)\\
				$e_{x_2}$    & 7.34$\times 10^{-4}$ (5.57$\times 10^{-4}$) & 6.80$\times 10^{-3}$ (4.62$\times 10^{-3}$) & 2.76$\times 10^{-1}$ (3.55$\times 10^{-1}$) & 9.33$\times 10^{-1}$  (4.59$\times 10^{-1}$)\\
				\midrule
				$e_X$      & 7.35$\times 10^{-4}$ (5.58$\times 10^{-4}$) & 6.81$\times 10^{-3}$ (4.63$\times 10^{-3}$) & 2.76$\times 10^{-1}$ (3.54$\times 10^{-1}$) & 9.33$\times 10^{-1}$ (4.60$\times 10^{-1}$)\\
				\bottomrule
			\end{tabular}
		}
	\end{table*}
	\begin{figure}[H]
		\centering
		\includegraphics[trim = 10 0 10 0, clip,width=0.45\textwidth]{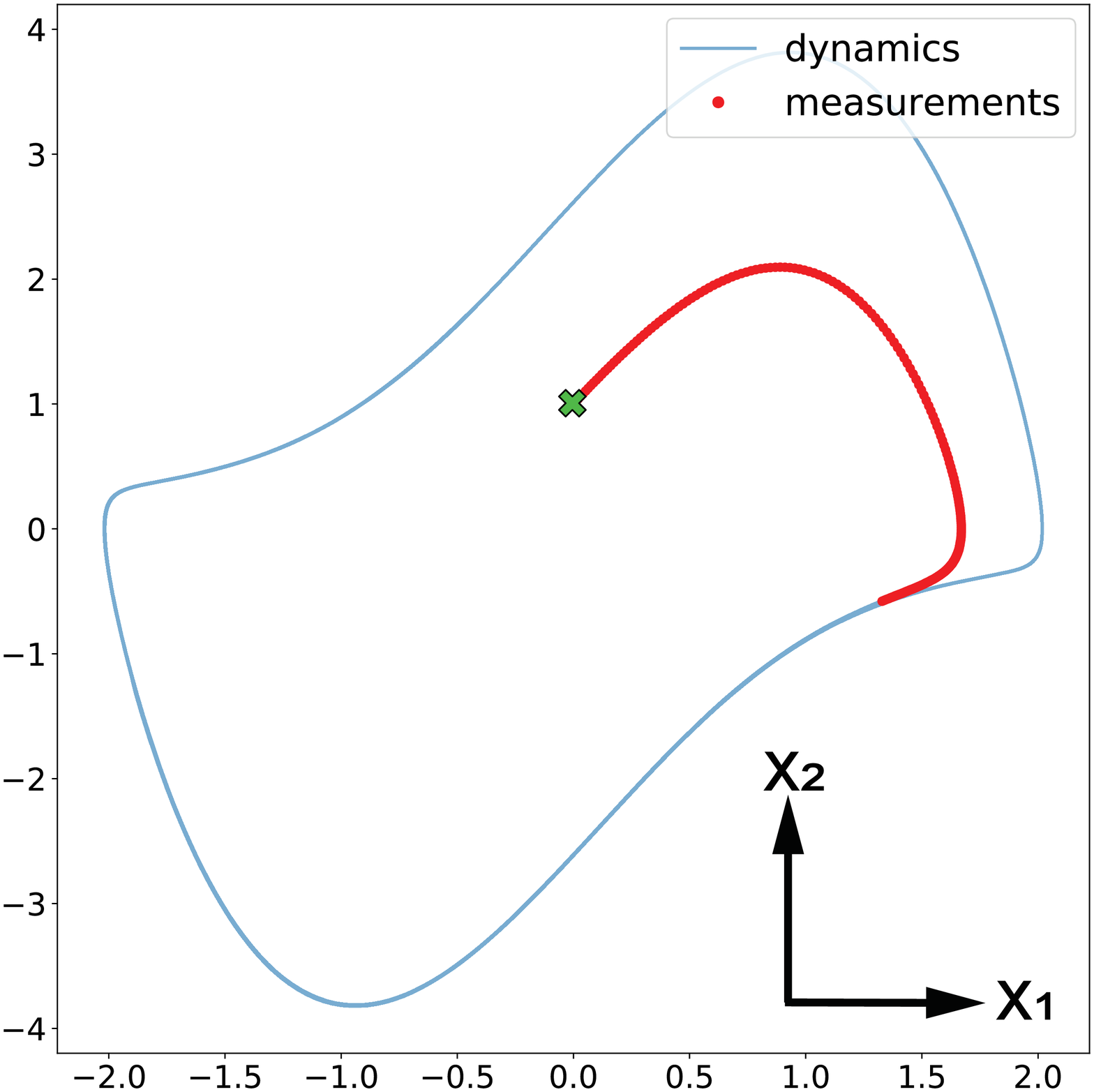}
		\includegraphics[trim = 10 0 10 0, clip,width=0.45\textwidth]{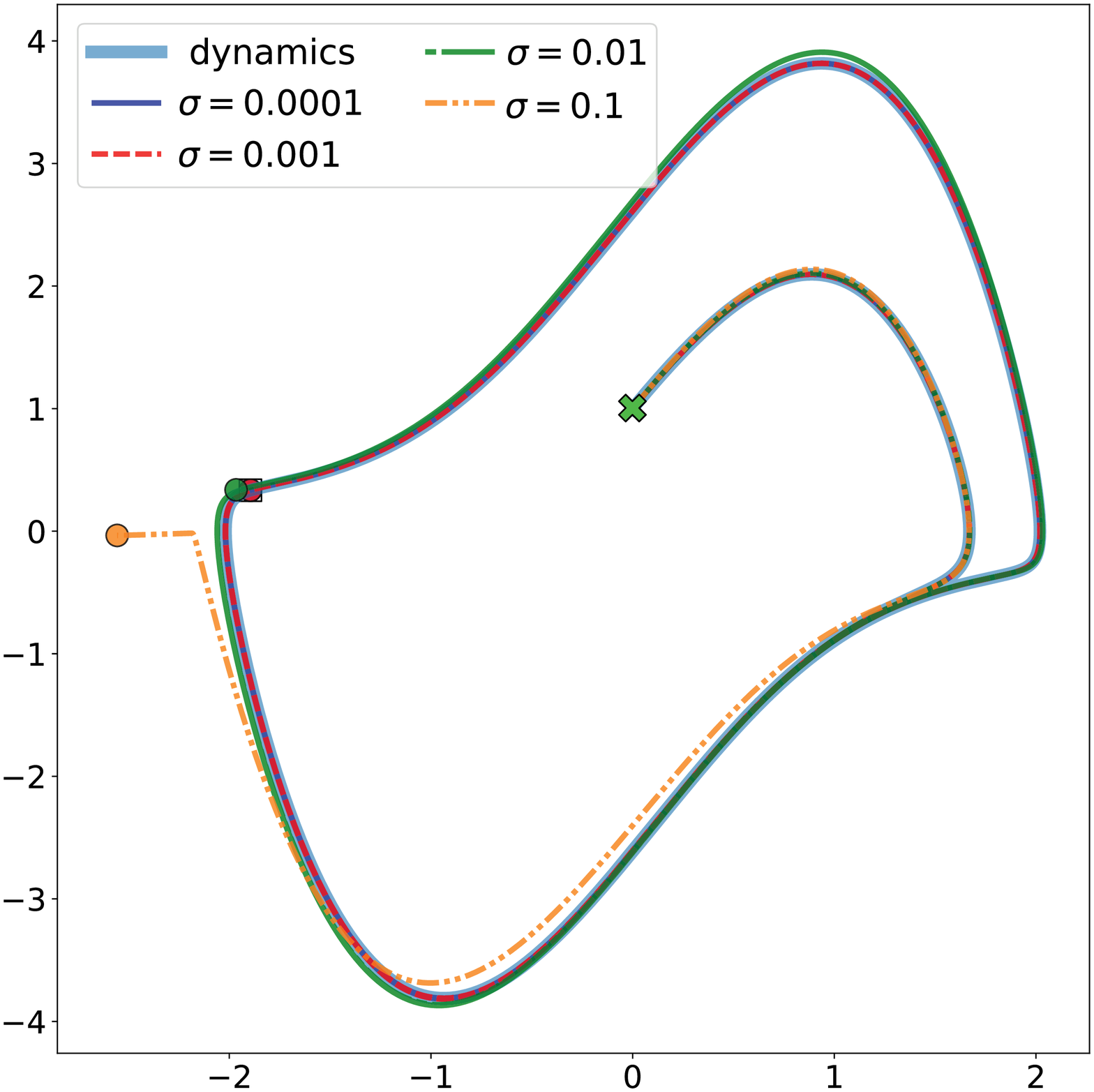}
		\caption{Predicted states for the Van der Pol oscillator from the recovered governing equations for different noise levels using $\ell_1$-trend filter/Pareto for filtering and WBPDN/Pareto for sparse regression. Left: The true dynamics are highlighted in blue, and discrete observations used for training are represented by the red dots. The green cross indicates the initial condition. Right: State predictions for different noise levels up to 20 time units.}
		\label{fig:VdP_samples_predictions}
	\end{figure}
	\begin{table*}
		\centering
		\caption{Mean prediction errors and their standard deviations (enclosed in parenthesis) up to 20 time units for Van der Pol for different noise levels using $\ell_1$-trend filter/Pareto for filtering and WBPDN/Pareto for sparse regression. For each of the recovered models from the 100 noisy trajectory realizations, we predicted the state trajectories via integrating the identified governing equations. We then computed the mean and standard deviation of the prediction errors defined in Eqn.~(\ref{eq:relative_filter_errors}).}
		\label{table:prediction_error_VanderPol}
		\resizebox{\linewidth}{!}{
			\begin{tabular}{lcccc}
				\toprule
				Noise &  \multicolumn{1}{c}{$\sigma = 0.0001$} & \multicolumn{1}{c}{$\sigma = 0.001$} & \multicolumn{1}{c}{$\sigma = 0.01$} & \multicolumn{1}{c}{$\sigma = 0.1$} \\
				\midrule
				\addlinespace[10pt]
				$e_{x_1}$    & 2.01$\times 10^{-4}$ (1.51$\times 10^{-4}$) & 1.88$\times 10^{-3}$ (1.52$\times 10^{-3}$) & 1.56$\times 10^{-2}$ (1.11$\times 10^{-2}$) & 1.62$\times 10^{0}$ (1.47$\times 10^{0}$)\\
				$e_{x_2}$    & 4.40$\times 10^{-2}$ (3.30$\times 10^{-4}$) & 4.12$\times 10^{-3}$ (3.31$\times 10^{-3}$) & 3.45$\times 10^{-2}$ (2.41$\times 10^{-2}$) & 7.46$\times 10^{-1}$  (3.68$\times 10^{-1}$)\\
				\midrule
				$e_X$      & 4.35$\times 10^{-4}$ (3.27$\times 10^{-4}$) & 4.07$\times 10^{-3}$ (3.28$\times 10^{-3}$) & 3.42$\times 10^{-2}$ (2.38$\times 10^{-2}$) & 1.64$\times 10^{0}$ (1.35$\times 10^{0}$)\\
				\bottomrule
			\end{tabular}
		}
	\end{table*}
	
	\section{Conclusion}
	\label{sec:conclusion}
	
	The motivation of this work has been to compare and investigate the performance of several filtering strategies and model selection techniques to improve the accuracy and robustness of sparse regression methods to recover governing equations of nonlinear dynamical systems from noisy state measurements. We presented several best-performing local and global noise filtering techniques for {\it a priori} state measurement denoising and estimation of state time-derivatives without stringent assumptions on the noise {\color{black}statistics}. In general, we observed that global smoothing methods along with either GCV or Pareto curves for selecting near optimal regularization parameters $\lambda$ outperform local smoothers with GCV as bandwidth selector. Of the global methods, $\ell_1$-trend filtering yields the best accuracy for state measurement denoising and estimation of state time-derivatives. However, the superior accuracy of $\ell_1$-trend filtering has not resulted in significant differences for coefficient recovery via WBPDN or STLS compared to smoothing splines and Tikhonov smoother. WBPDN produces smoother GCV functions and Pareto curves than STLS, and the selected regularization parameters are closer to the optimal ones. We conjecture that this is the reason why WBPDN yields more accurate governing equation coefficients. Similar to the findings in~\cite{cortiella2021sparse}, the corner point criterion of Pareto curves often yields better estimates than cross-validation strategies in selecting near optimal regularization parameters.
	
	We empathize that the filtering strategies and model selection techniques presented in this article can be used as guidelines for other identification algorithms that may require measurement smoothing or numerical differentiation to enhance their performance. For example, integral-based approaches, e.g.,~\cite{schaeffer2017sparse, messenger2021weakODE, messenger2021weakPDE}, which do not require numerical differentiation to recover governing equations can also benefit from the filtering strategies presented in this article.
	
	\section{Acknowledgements}
	
	This work was supported by the National Science Foundation grant CMMI-1454601.

	\bibliography{mybibfile}

	\appendix       
	
	\section*{Appendix A: Difference matrices}
	\label{app:A}
	
	Here we provide the construction of the difference matrices used in Tikhonov smoother and $\ell_1$-trend filtering. The $(k+1)$st order difference matrix, $k \geq 0$, is given by the following recursive formula~\cite{tibshirani2014adaptive}
	\begin{equation}
		\mathbf{D}_{(k+1)} = \mathbf{D}_{(1)}\mathbf{D}_{(k)} \in \R^{(m-k-1) \times m}
	\end{equation}
	where the first-order difference matrix corresponding to $k = 0$ is given by
	\begin{equation}
		\mathbf{D}_{(1)} = 
		\begin{bmatrix}
			-1 & 1 & 0 & \hdots & 0 & 0\\
			0 & -1 & 1 & \hdots & 0 & 0\\
			\vdots & \ddots & \ddots & \ddots & \ddots & \vdots\\
			0 & 0 & 0 & \hdots & -1 & 1\\
		\end{bmatrix} \in \R^{(m-1) \times m}.
	\end{equation}
	For example, the second- and third-order difference matrices read
	\begin{equation}
		\mathbf{D}_{(2)} = 
		\begin{bmatrix}
			1 & -2 & 1 & \hdots & 0 & 0\\
			0 & 1 & -2 & 1 & \hdots & 0\\
			0 & 0 & 1 & -2 & \hdots & 0\\
			0 &  & \ddots  & \ddots &  & \\
		\end{bmatrix} \in \R^{(m-2) \times m},
	\end{equation}
	and
	\begin{equation}
		\mathbf{D}_{(3)} = 
		\begin{bmatrix}
			1 & 3 & -3 & 1 & \hdots & 0\\
			0 & 1 & 3 & -3 & \hdots & 0\\
			0 & 0 & 1 & 3 & \hdots & 0\\
			0 &  & \ddots & \ddots &  & \\
		\end{bmatrix} \in \R^{(m-3) \times m}.
	\end{equation}
	
	\section*{Appendix B: Signal-to-noise ratios}
	\label{app:B}
	
	In this section, we provide the signal-to-noise ratios (SNR) of each state variable of each dynamical system for all noise levels studied in this article. The SNR, expressed in decibels, is defined for each state variable $j = 1,...,n$ as
	\begin{equation}
		(\text{SNR}_j)_{\text{dB}} =
		10~\text{log}_{10}\bigg(\frac{\sum_{k = 1}^{m}x_j(t_k)^2}{\sigma^2}\bigg).
	\end{equation}
	Tables~\ref{table:SNR_Lorenz}, \ref{table:SNR_Duffing} and \ref{table:SNR_Vanderpol} show the state variable SNR for the Lorenz 63 system, Duffing and Van der Pol oscillators, respectively.
	\begin{table}[H]
		\centering
		\caption{Signal-to-noise ratios of each state variable for the Lorenz 63 system.}
		\label{table:SNR_Lorenz} 
		\begin{tabular}{lccc}
			\addlinespace[10pt]
			&  & \multicolumn{1}{c}{SNR [dB]} & \\
			\toprule                               
			Noise &  \multicolumn{1}{c}{state $x_1$} & \multicolumn{1}{c}{state $x_2$} & \multicolumn{1}{c}{state $x_3$} \\
			\addlinespace[10pt]
			$\sigma = 0.001$ & 101.06 & 102.76 & 110.53 \\
			$\sigma = 0.01$ & 81.06 & 82.76 & 90.53 \\
			$\sigma = 0.1$ & 61.06 & 62.76 & 70.53 \\
			$\sigma = 1$ & 41.06 & 42.76 & 50.53 \\
			\bottomrule
		\end{tabular}
	\end{table}
	\begin{table}[H]
		\centering
		\caption{Signal-to-noise ratios of each state variable for the Duffing oscillator.}
		\label{table:SNR_Duffing} 
		\begin{tabular}{lcc}
			\addlinespace[10pt]
			&  \multicolumn{1}{c}{SNR [dB]} & \\
			\toprule                               
			Noise &  \multicolumn{1}{c}{state $x_1$} & \multicolumn{1}{c}{state $x_2$} \\
			\addlinespace[10pt]
			$\sigma = 0.0001$ & 99.47 & 105.85  \\
			$\sigma = 0.001$ & 79.47 & 85.85  \\
			$\sigma = 0.01$ & 59.47 & 65.85 \\
			$\sigma = 0.1$ & 39.47 & 45.85  \\
			\bottomrule
		\end{tabular}
	\end{table}
	\begin{table}[H]
		\centering
		\caption{Signal-to-noise ratios of each state variable for the Van der Pol oscillator.}
		\label{table:SNR_Vanderpol} 
		\begin{tabular}{lcc}
			\addlinespace[10pt]
			&  \multicolumn{1}{c}{SNR [dB]} & \\
			\toprule                               
			Noise &  \multicolumn{1}{c}{state $x_1$} & \multicolumn{1}{c}{state $x_2$} \\
			\addlinespace[10pt]
			$\sigma = 0.0001$ & 105.64 & 104.34  \\
			$\sigma = 0.001$ & 85.64 & 84.34  \\
			$\sigma = 0.01$ & 65.64 & 64.34 \\
			$\sigma = 0.1$ & 45.64 & 44.34  \\
			\bottomrule
		\end{tabular}
	\end{table}
	
	
	\section*{Appendix C: State, state-time derivative and coefficient error variances}
	\label{app:C}
	This section compares the variance of the relative state and state time-derivative errors defined in (\ref{eq:relative_filter_errors}) after running each of the local and global smoothing methods over 100 noisy trajectory realizations using GCV (left column) and Pareto curve criterion (right column) to automatically select the regularization parameter $\lambda$.
	
	
	\begin{figure}[H]
		\centering
		\includegraphics[trim = 0 0 0 0, clip,width=0.48\textwidth]{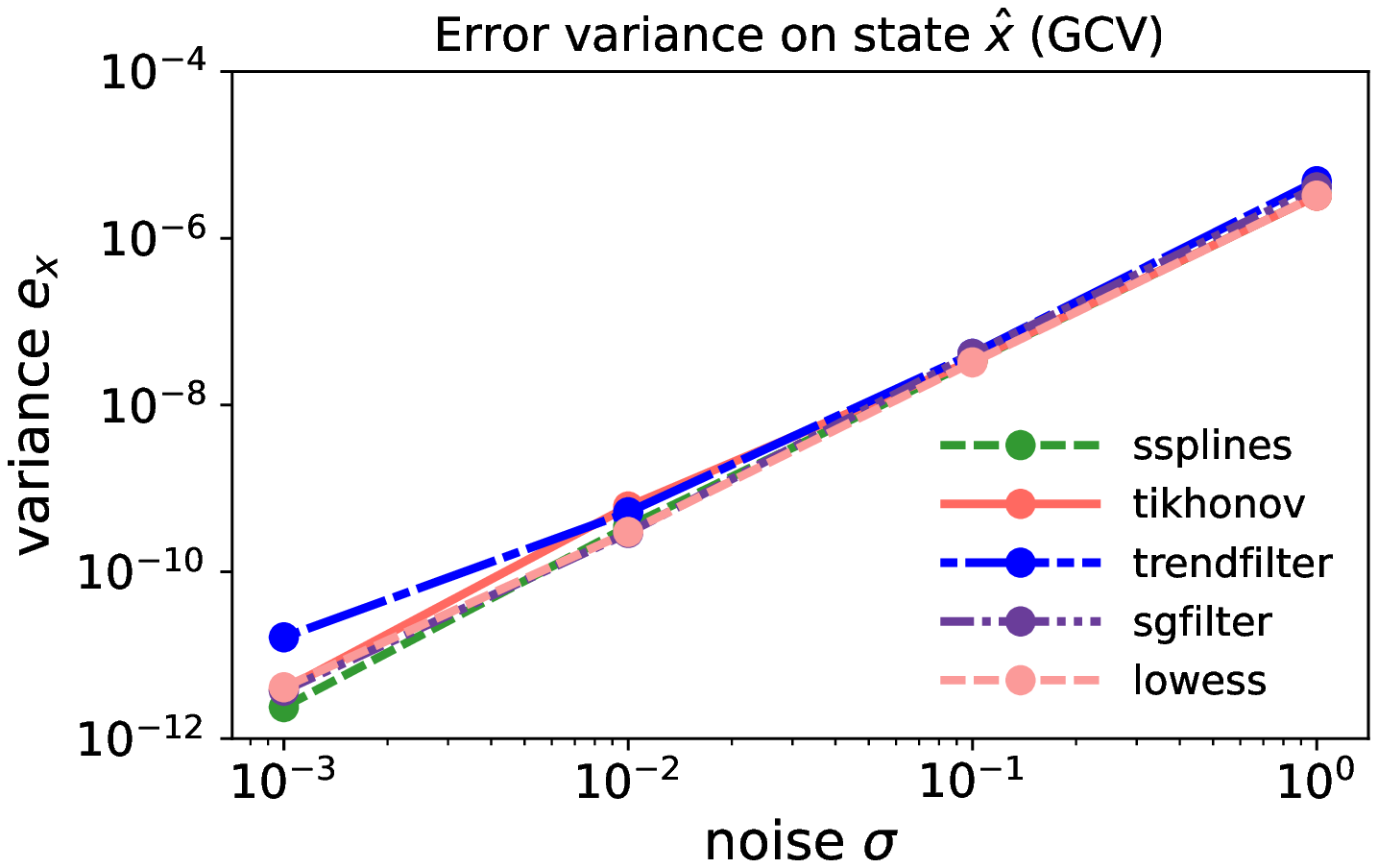}
		\includegraphics[trim = 0 0 0 0,
		clip,width=0.49\textwidth]{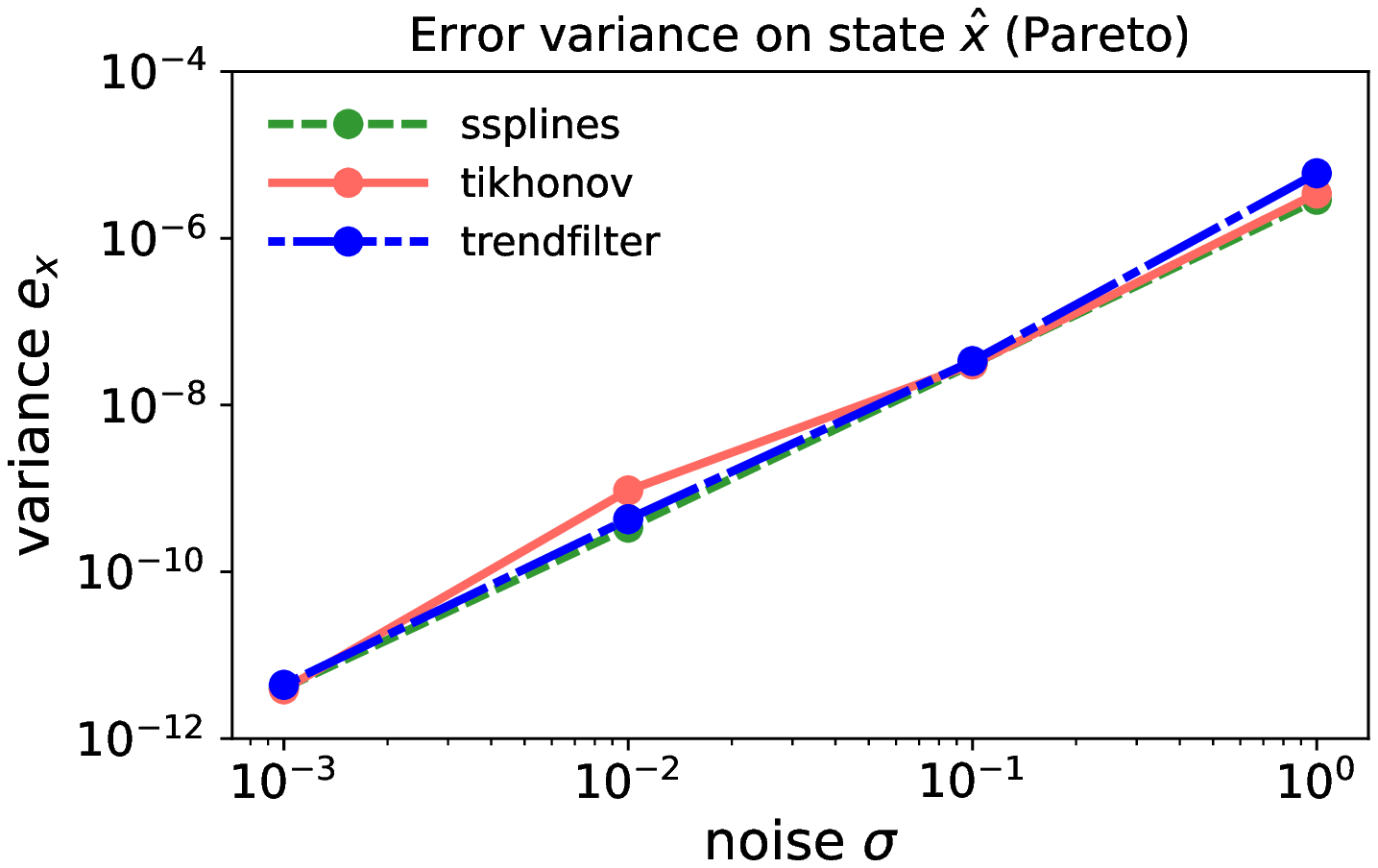}
		\includegraphics[trim = 0 0 0 0, clip,width=0.48\textwidth]{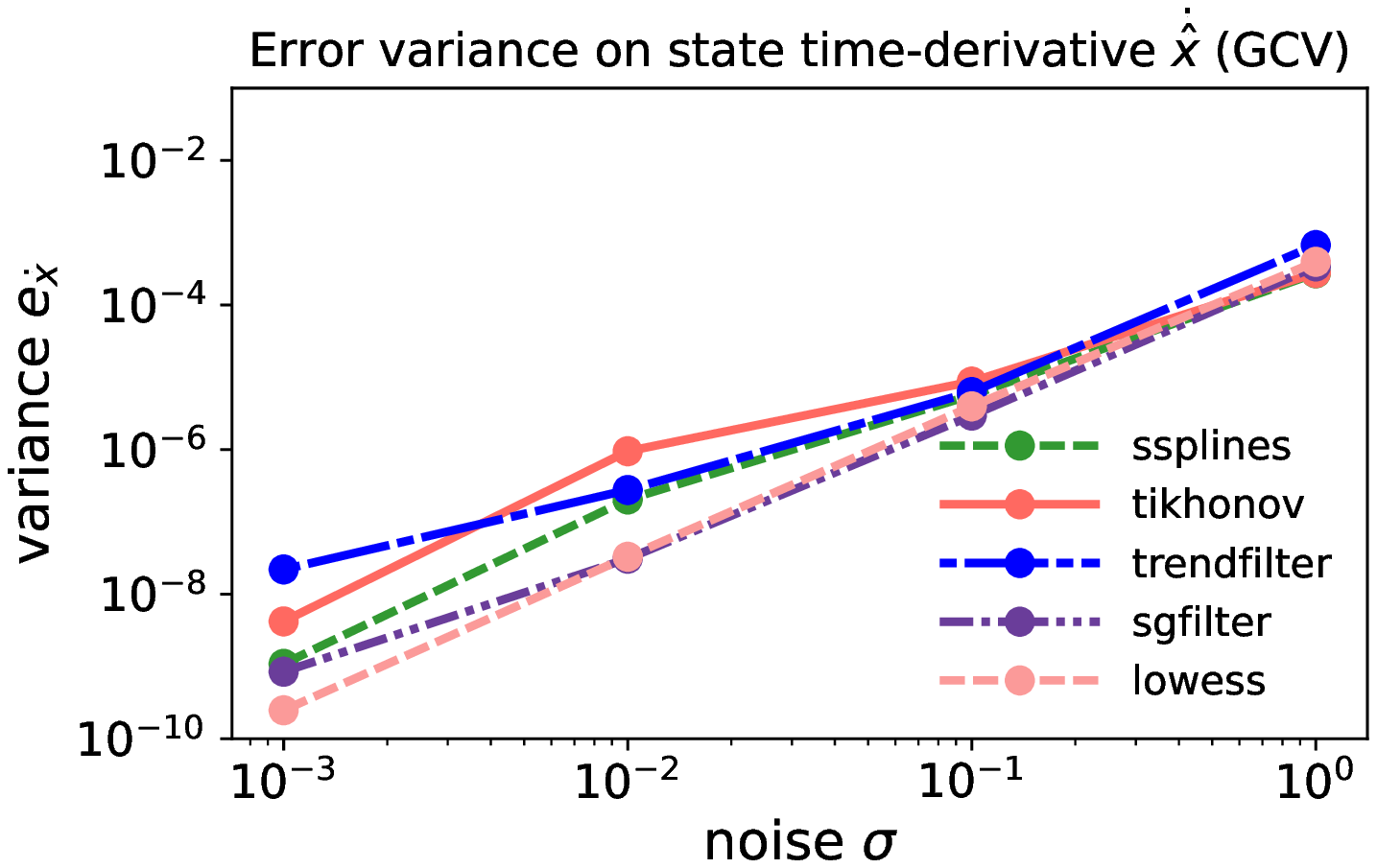}
		\includegraphics[trim = 0 0 0 0,
		clip,width=0.49\textwidth]{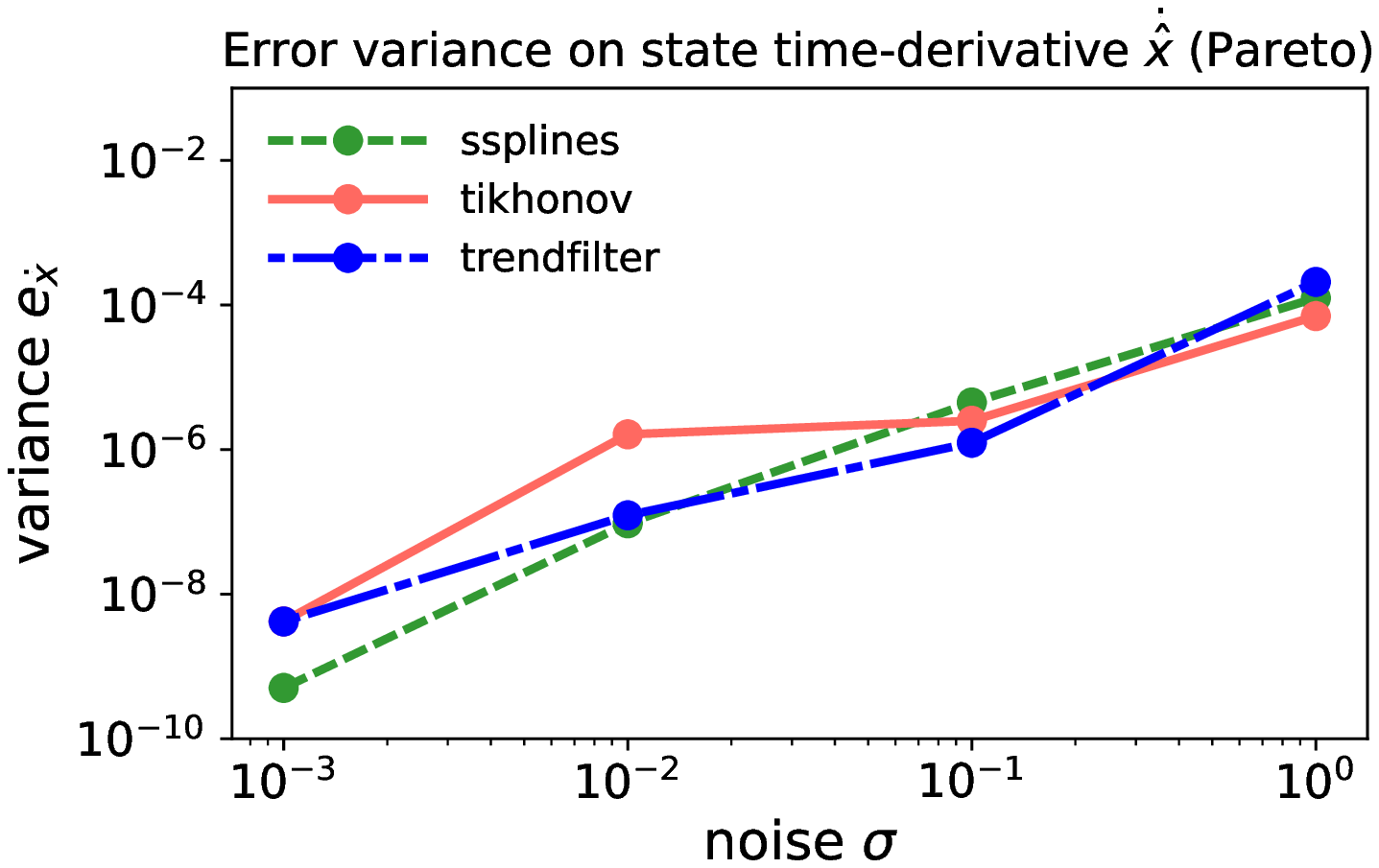}
		\caption{Variances of the relative state (top row) and state time-derivative (bottom row) errors for the Lorenz 63 system using local and global smoothers. The regularization parameters were computed using GCV (left column) and the Pareto curve criterion (right column).}
		\label{fig:Lorenz63_filter_comparison_var}
	\end{figure}
	
	\begin{figure}[H]
		\centering
		\includegraphics[trim = 10 0 10 0, clip,width=0.32\textwidth]{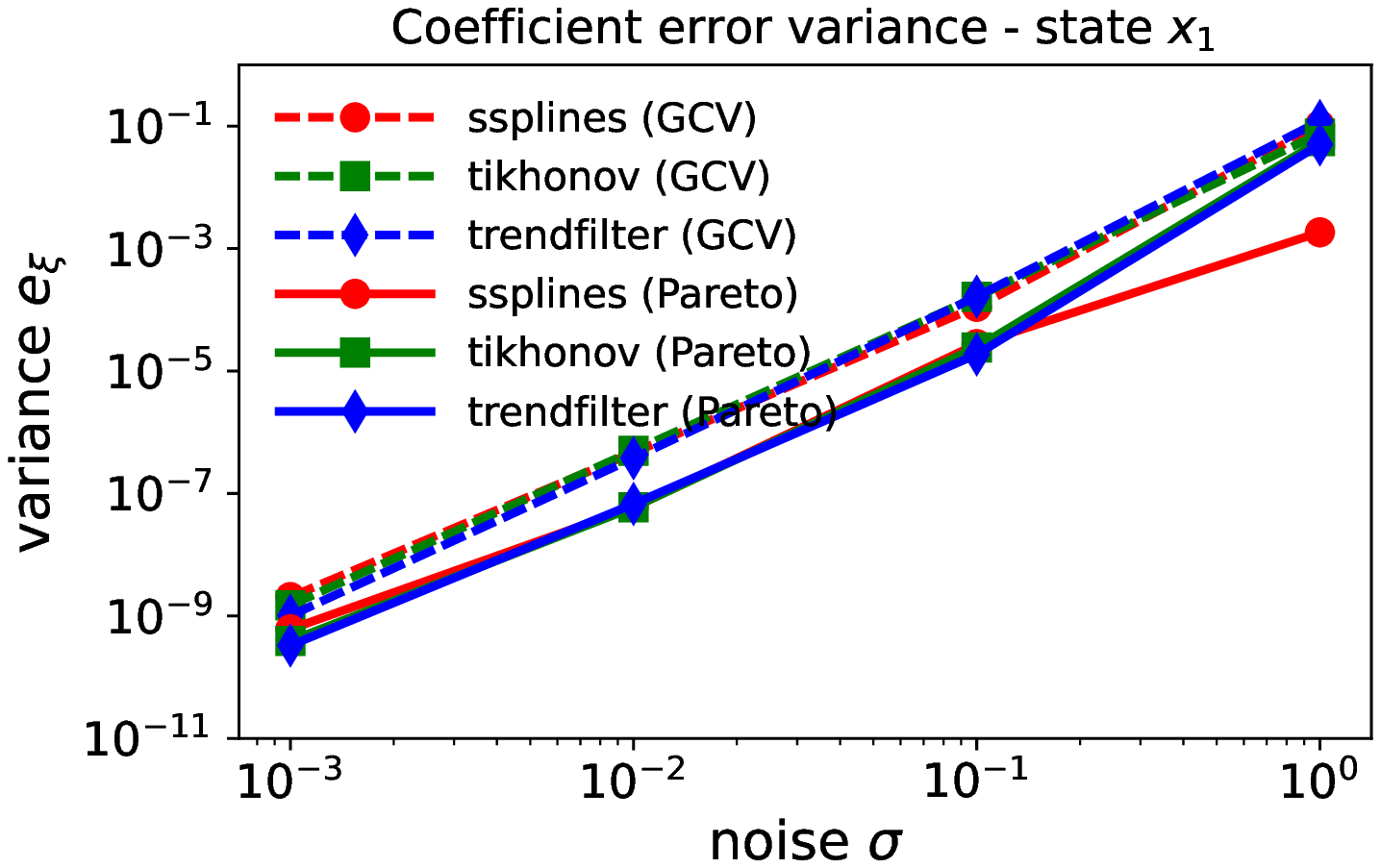}
		\includegraphics[trim = 10 0 10 0,
		clip,width=0.32\textwidth]{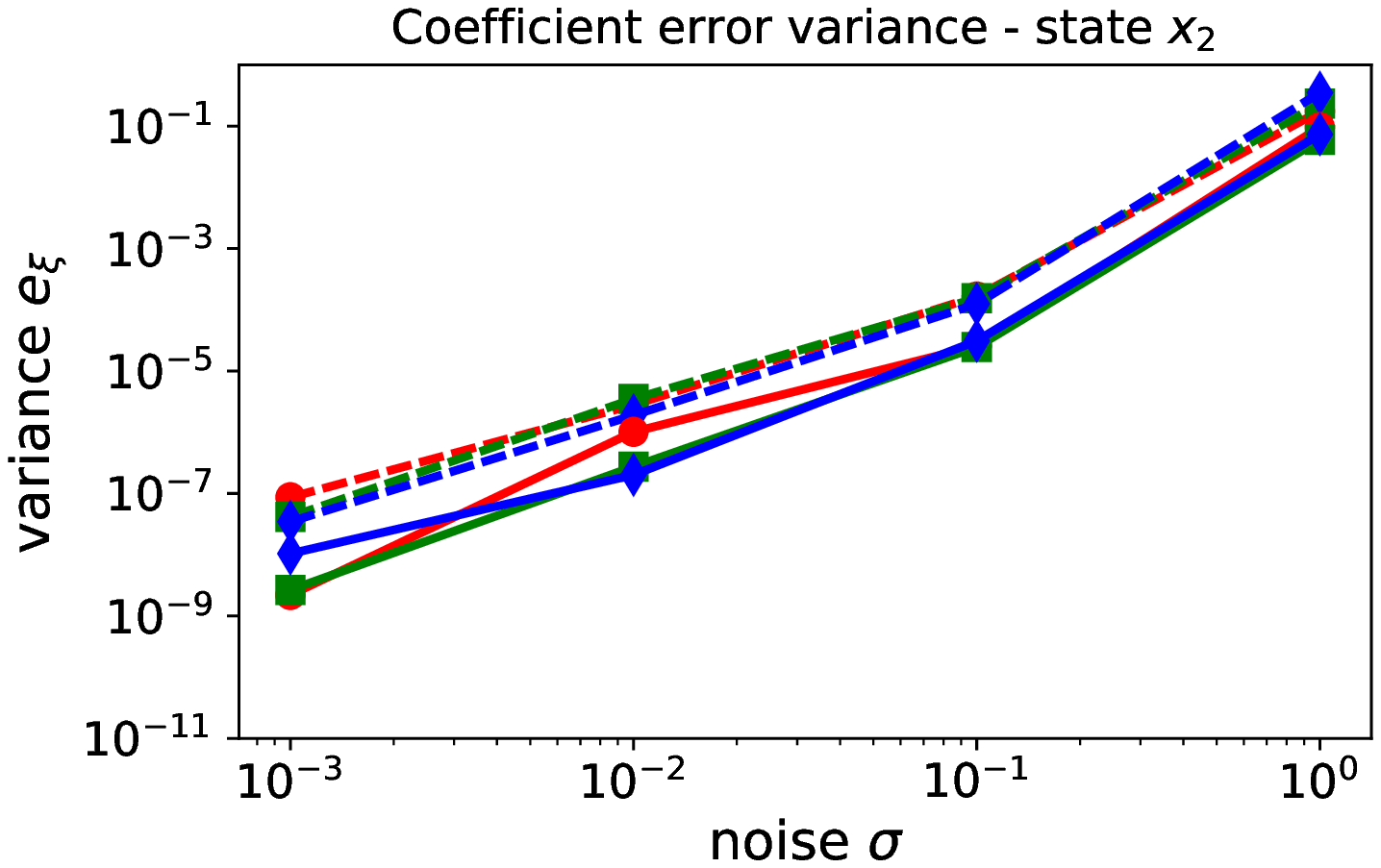}
		\includegraphics[trim = 10 0 10 0,
		clip,width=0.32\textwidth]{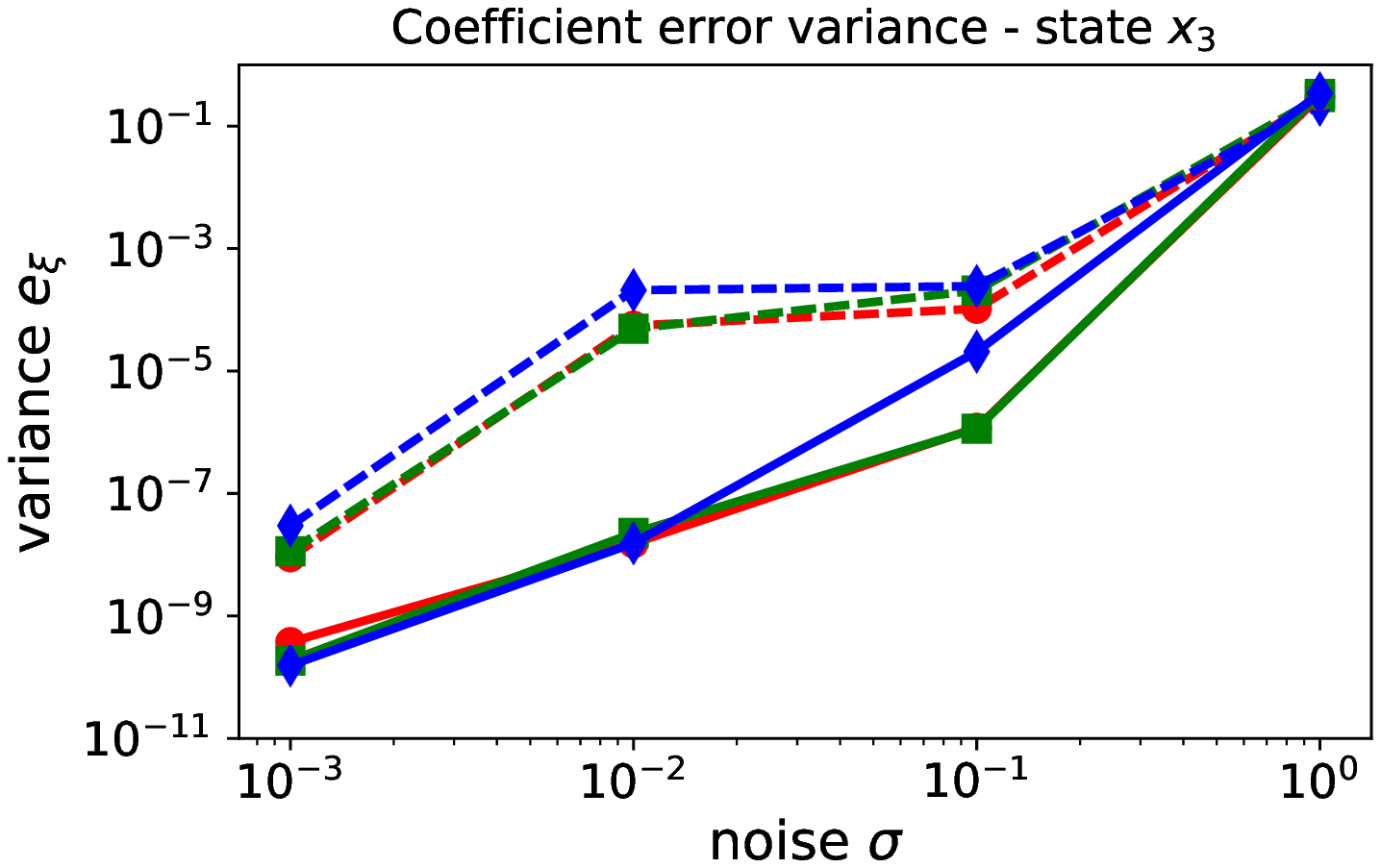}
		\caption{Variances of the relative coefficient errors for the Lorenz 63 system using global smoothers and WBPDN. The regularization parameters were computed using GCV (left column) and the Pareto curve criterion (right column).}
		\label{fig:Lorenz63_WBPDN_comparison_var}
	\end{figure}
	%
	
	\begin{figure}[H]
		\centering
		\includegraphics[trim = 0 0 00 0, clip,width=0.48\textwidth]{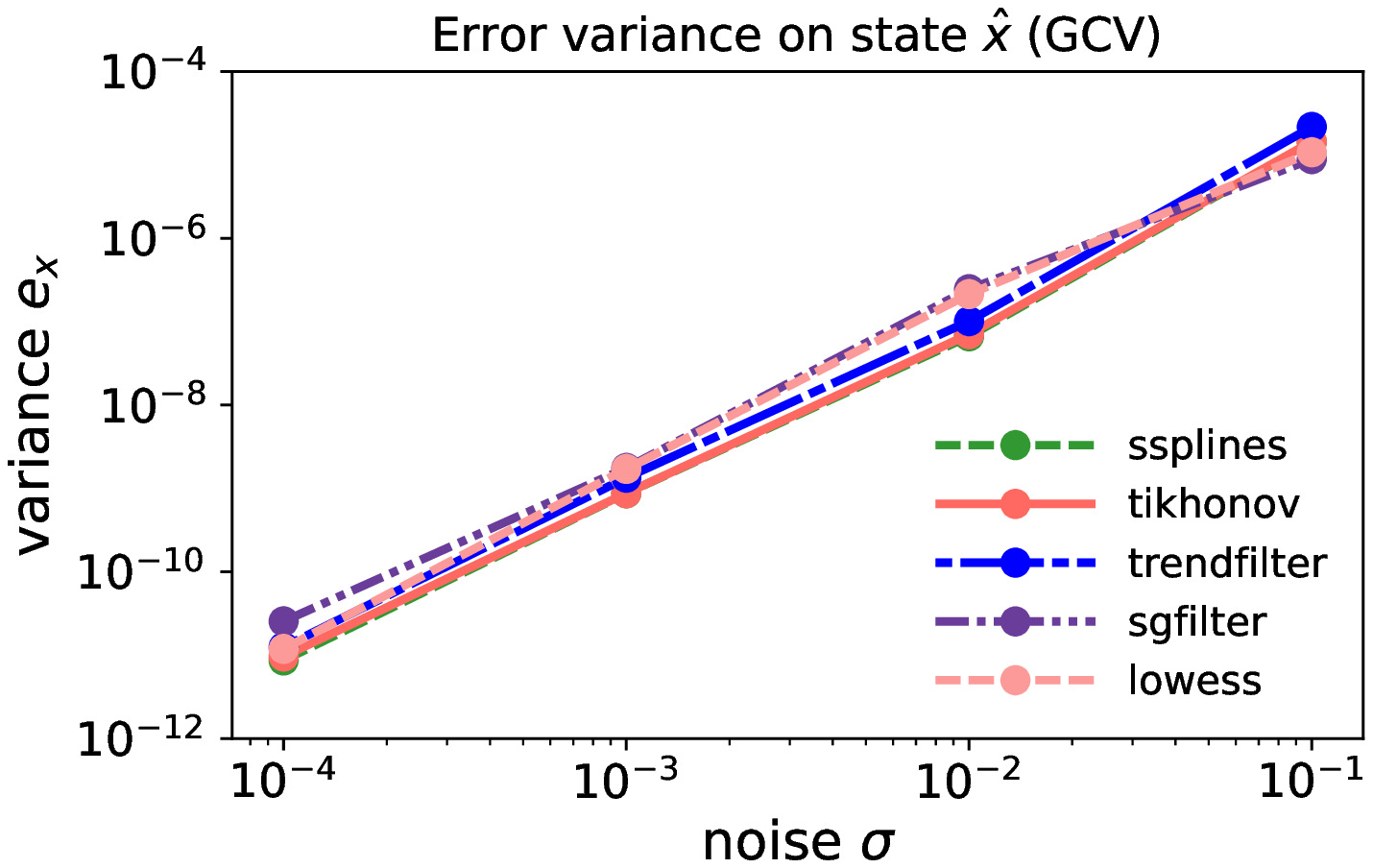}
		\includegraphics[trim = 0 0 00 0,
		clip,width=0.49\textwidth]{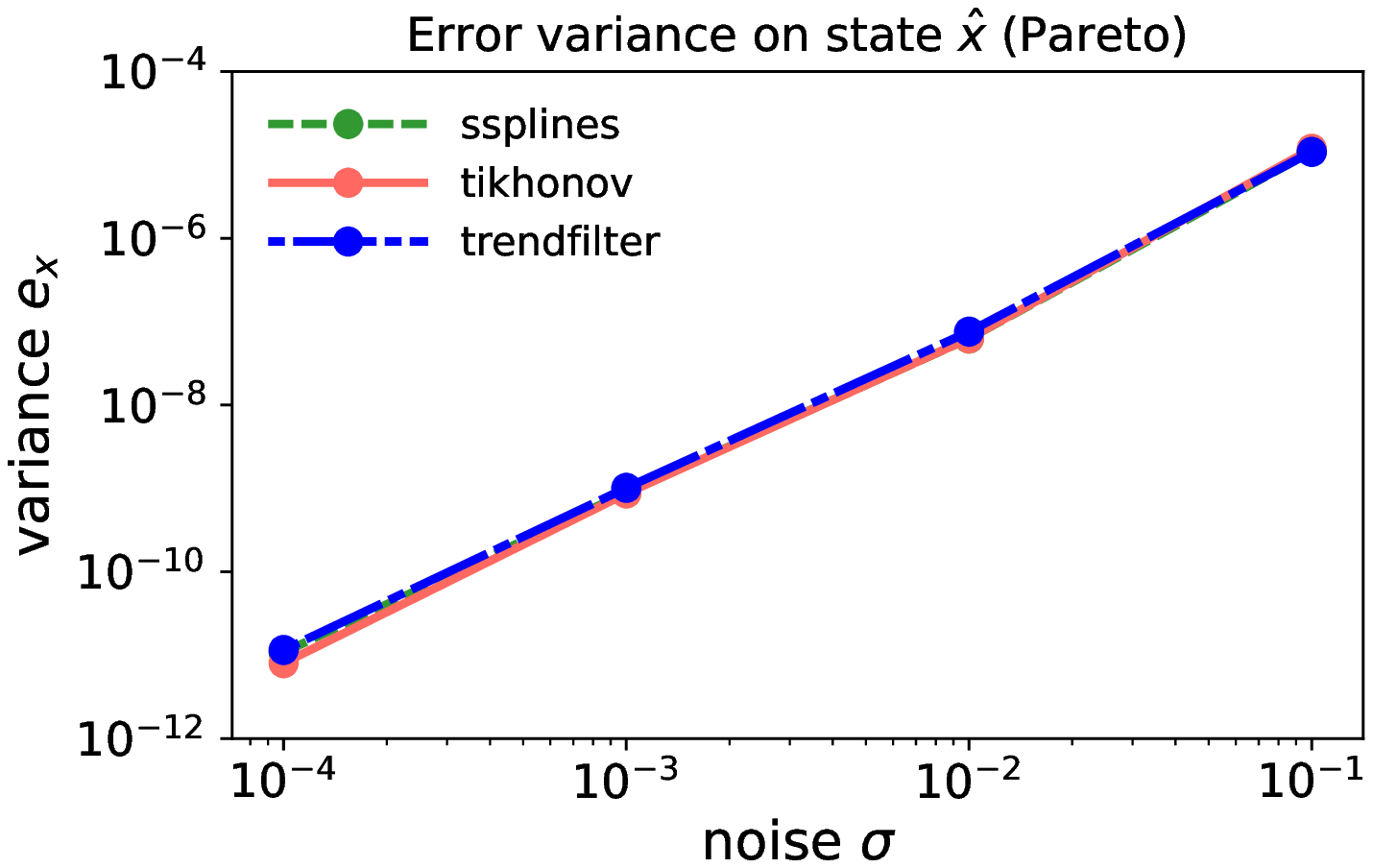}
		\includegraphics[trim = 0 0 00 0, clip,width=0.48\textwidth]{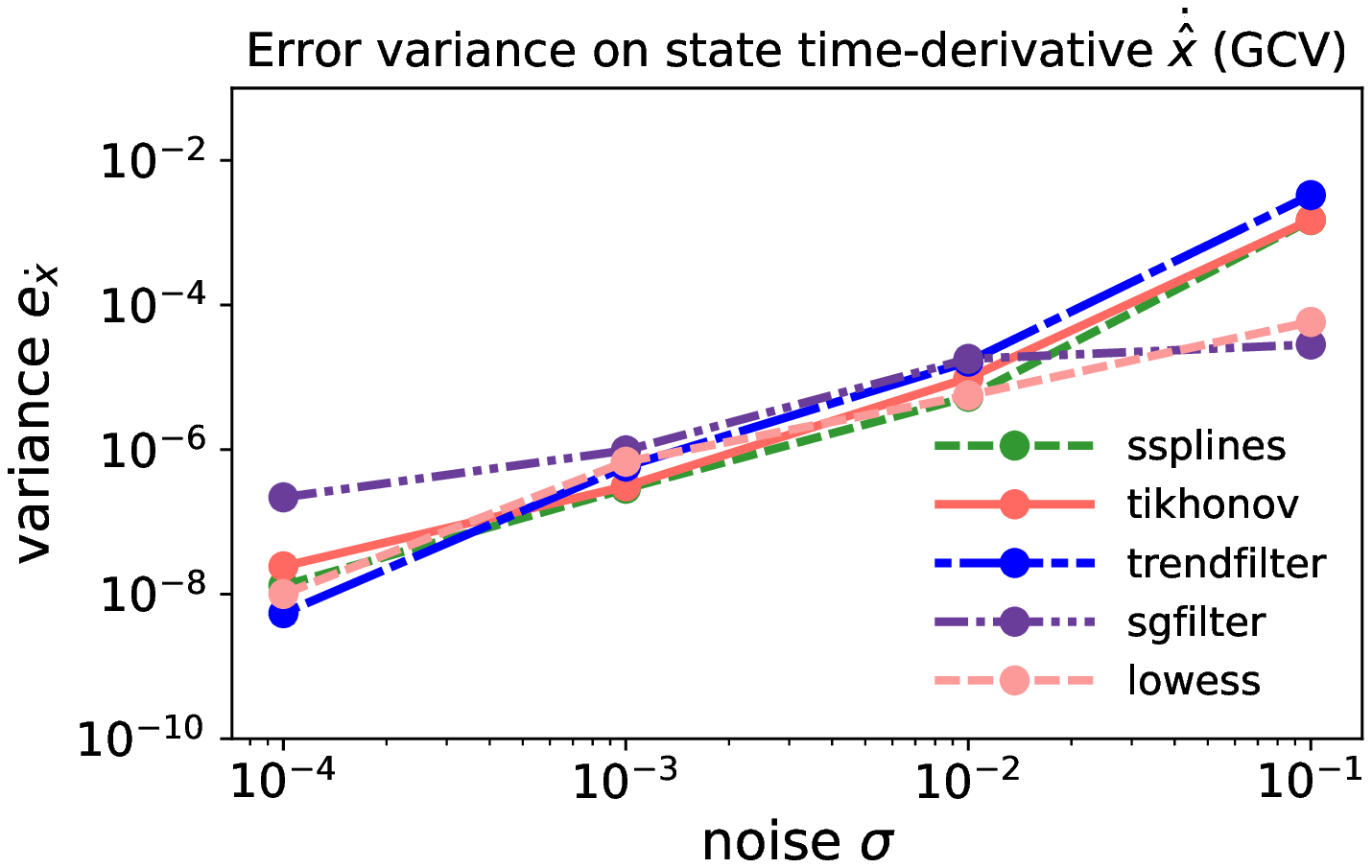}
		\includegraphics[trim = 0 0 00 0,
		clip,width=0.49\textwidth]{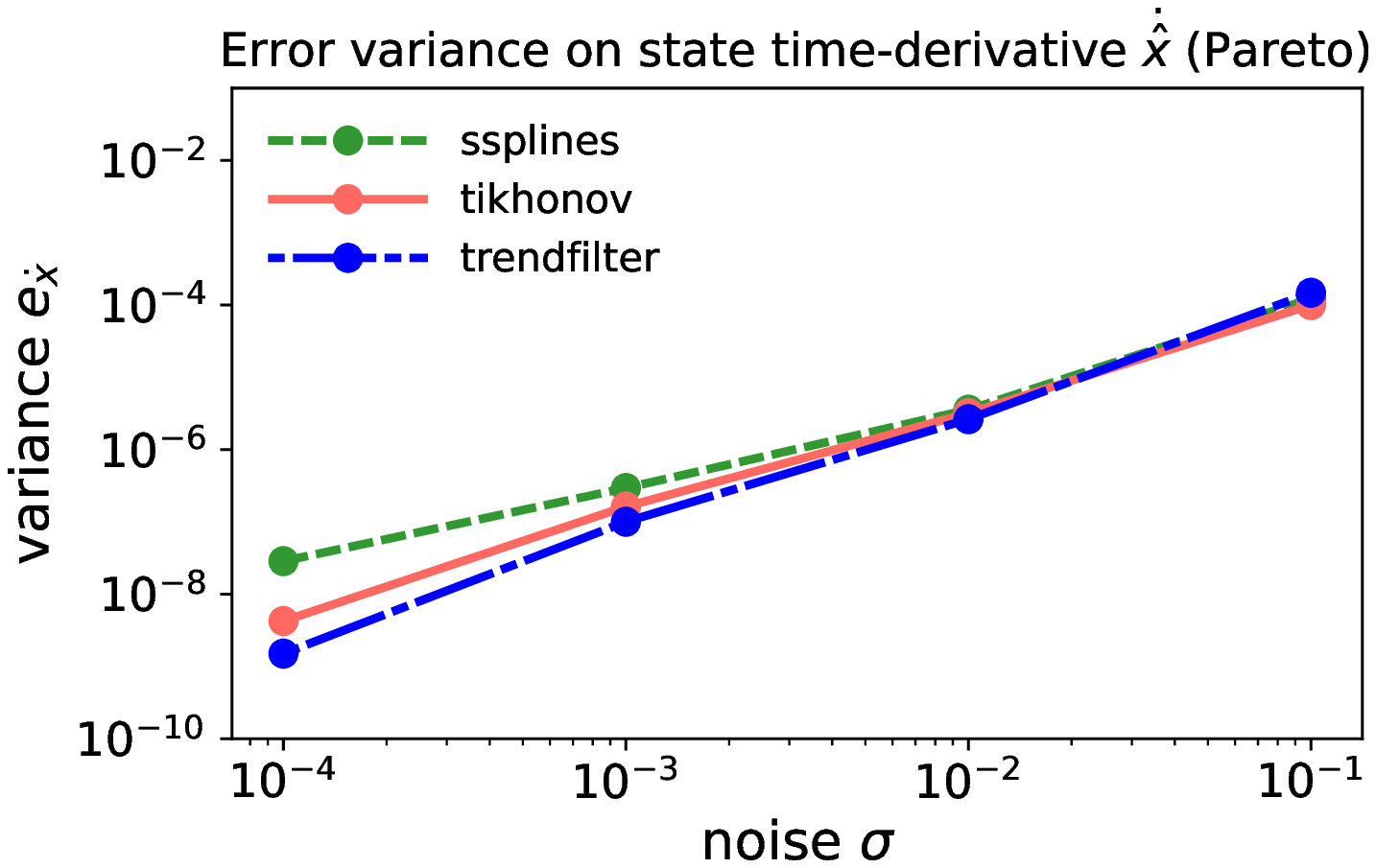}
		\caption{Variances of the relative state (top row) and state time-derivative (bottom row) errors for the Duffing oscillator using local and global smoothers. The regularization parameters were computed using GCV (left column) and the Pareto curve criterion (right column).}
		\label{fig:Duffing_filter_comparison_var}
	\end{figure}
	\begin{figure}[H]
		\centering
		\includegraphics[trim = 0 0 0 0, clip,width=0.49\textwidth]{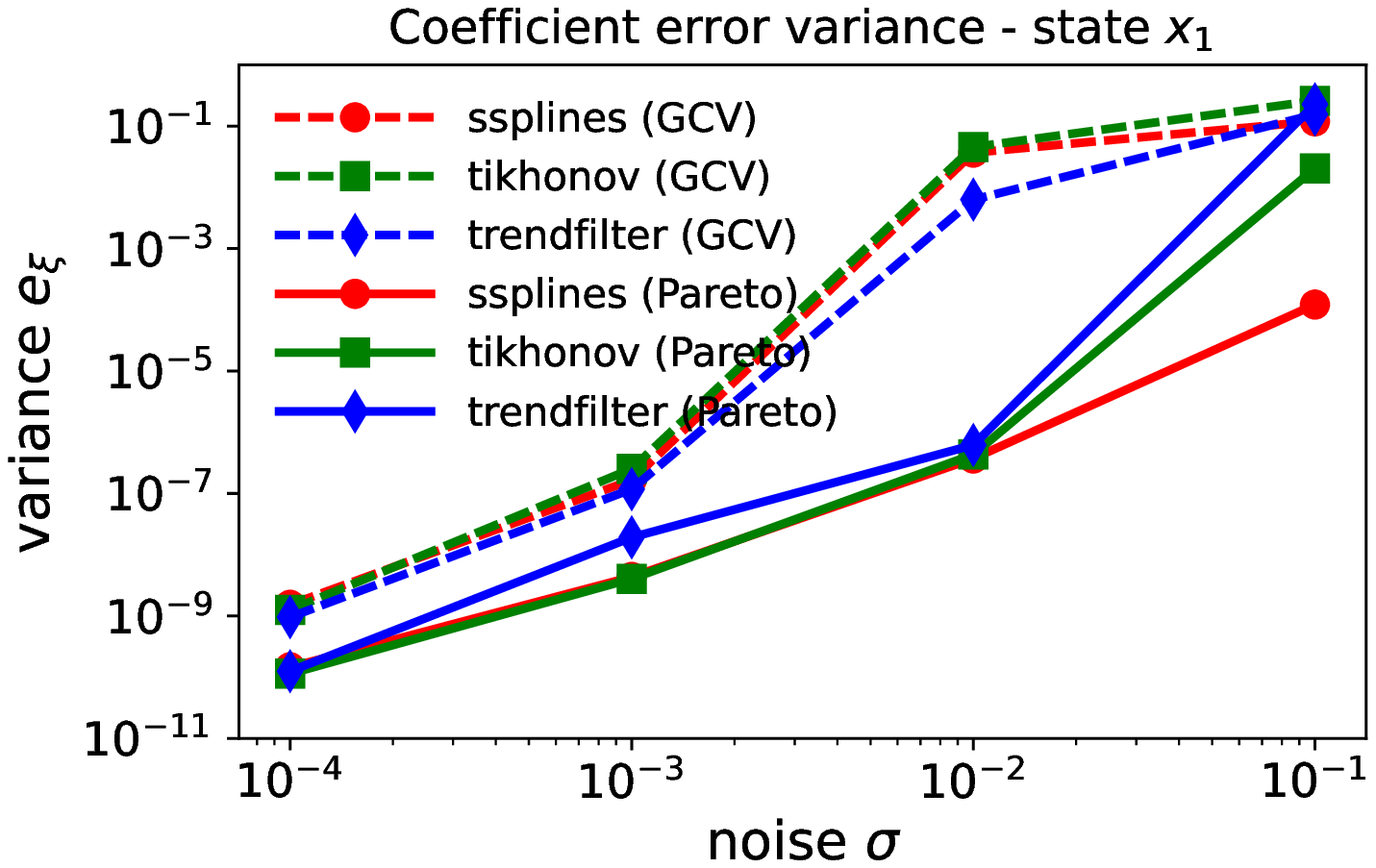}
		\includegraphics[trim = 0 0 0 0,
		clip,width=0.49\textwidth]{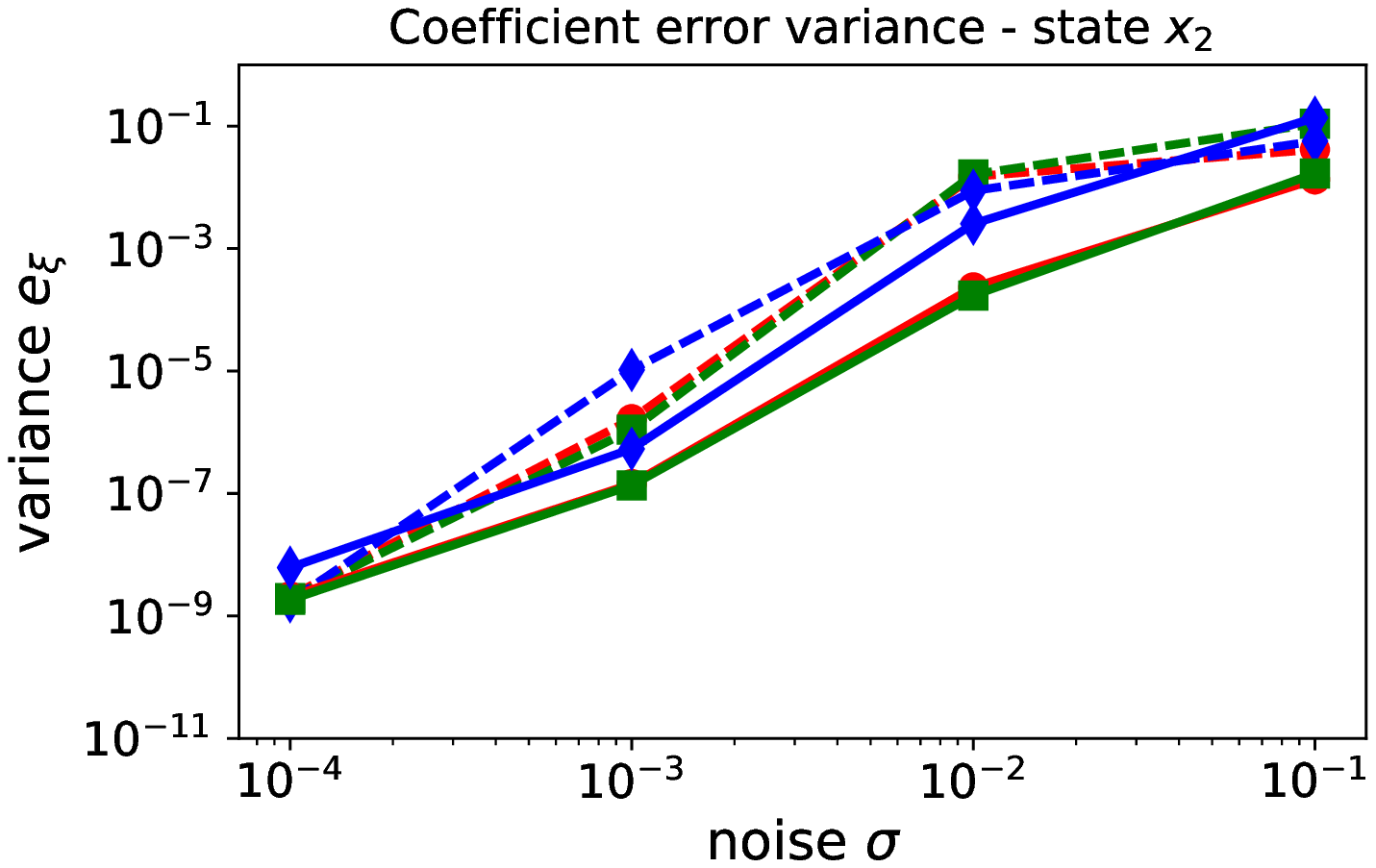}
		\caption{Variances of the relative coefficient errors for the Duffing oscillator using global smoothers and WBPDN. The regularization parameters were computed using GCV (left column) and the Pareto curve criterion (right column).}
		\label{fig:Duffing_WBPDN_comparison_var}
	\end{figure}
	%
	
	\begin{figure}[H]
		\centering
		\includegraphics[trim = 0 0 00 0, clip,width=0.48\textwidth]{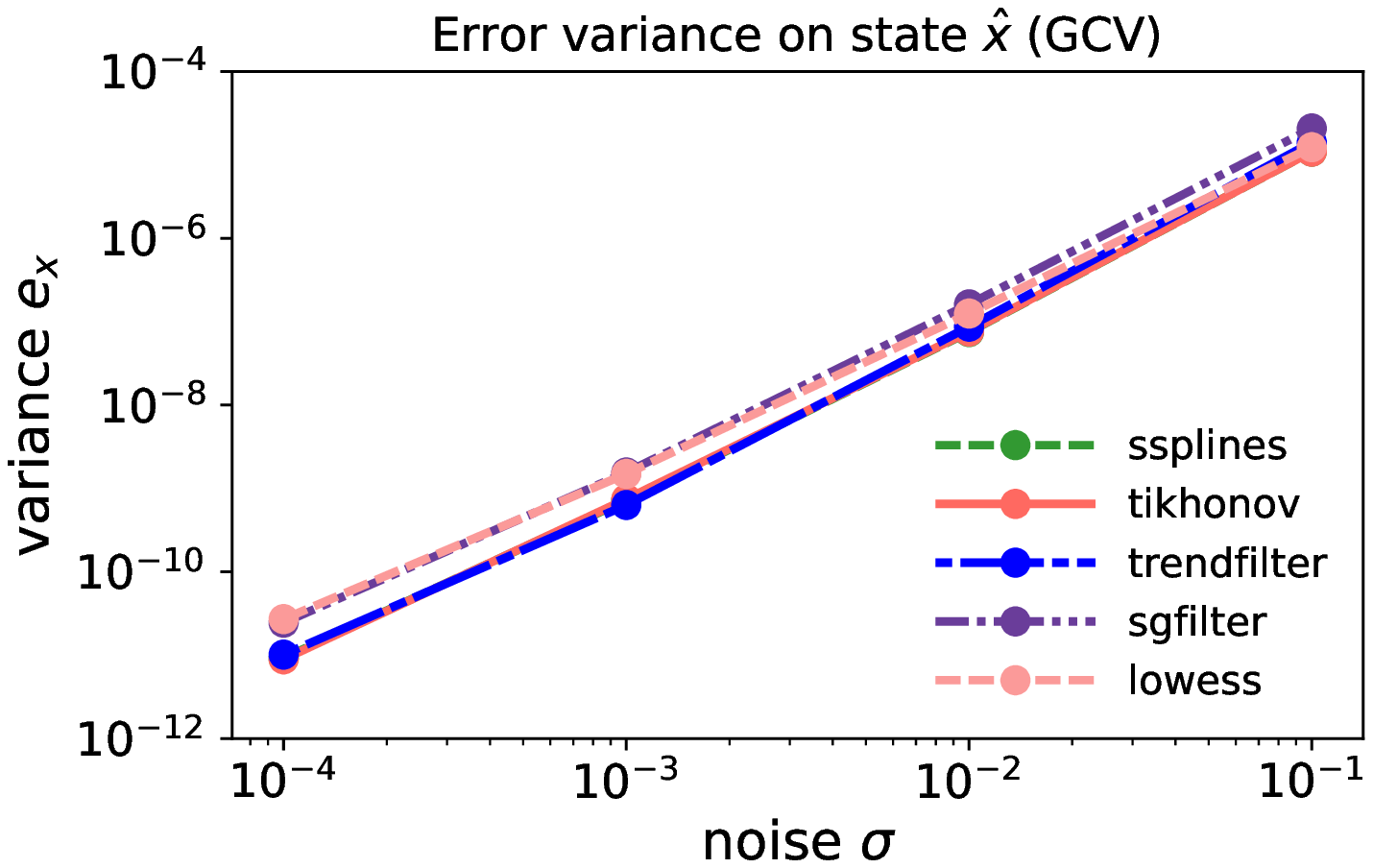}
		\includegraphics[trim = 0 0 00 0,
		clip,width=0.49\textwidth]{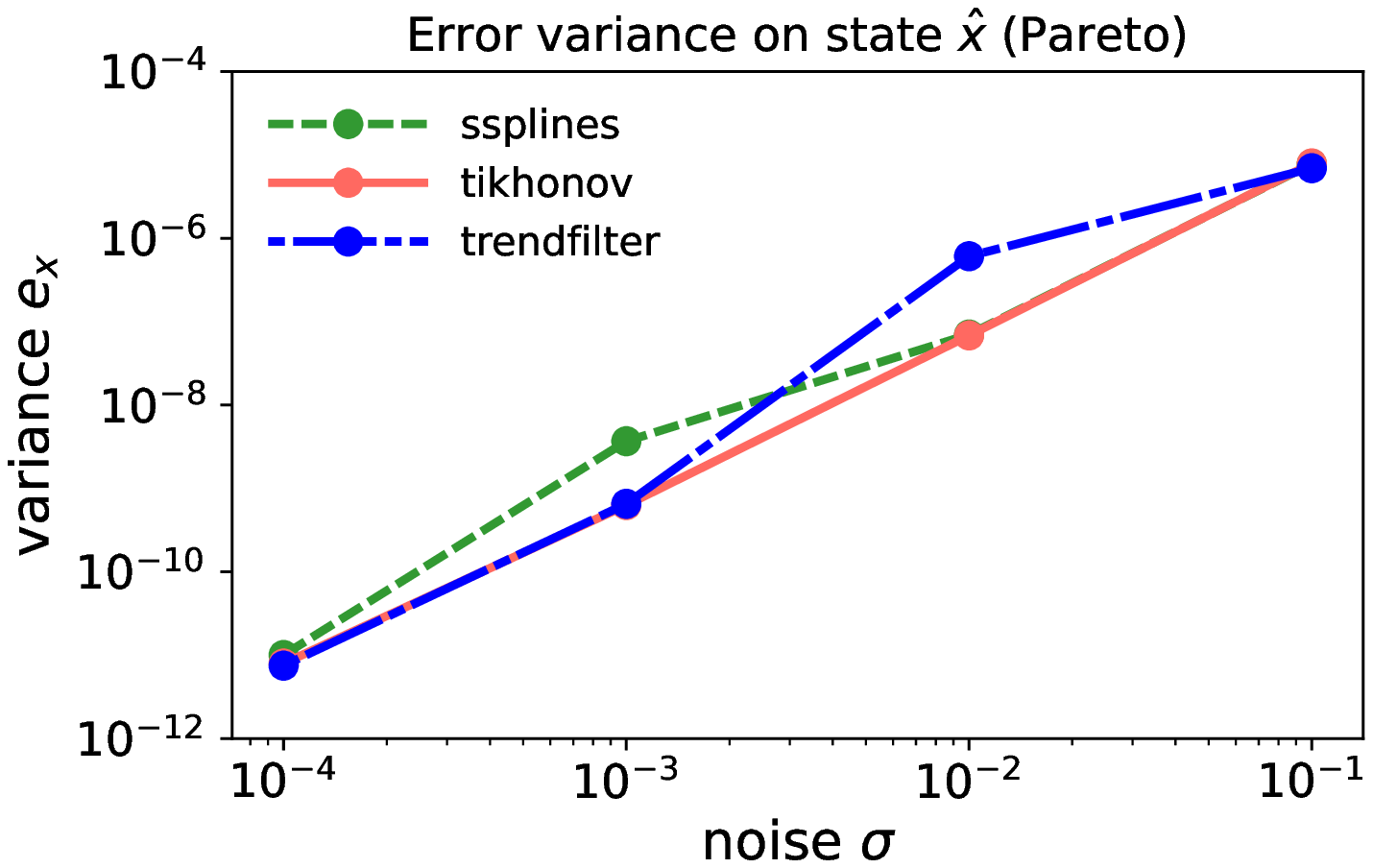}
		\includegraphics[trim = 0 0 00 0, clip,width=0.48\textwidth]{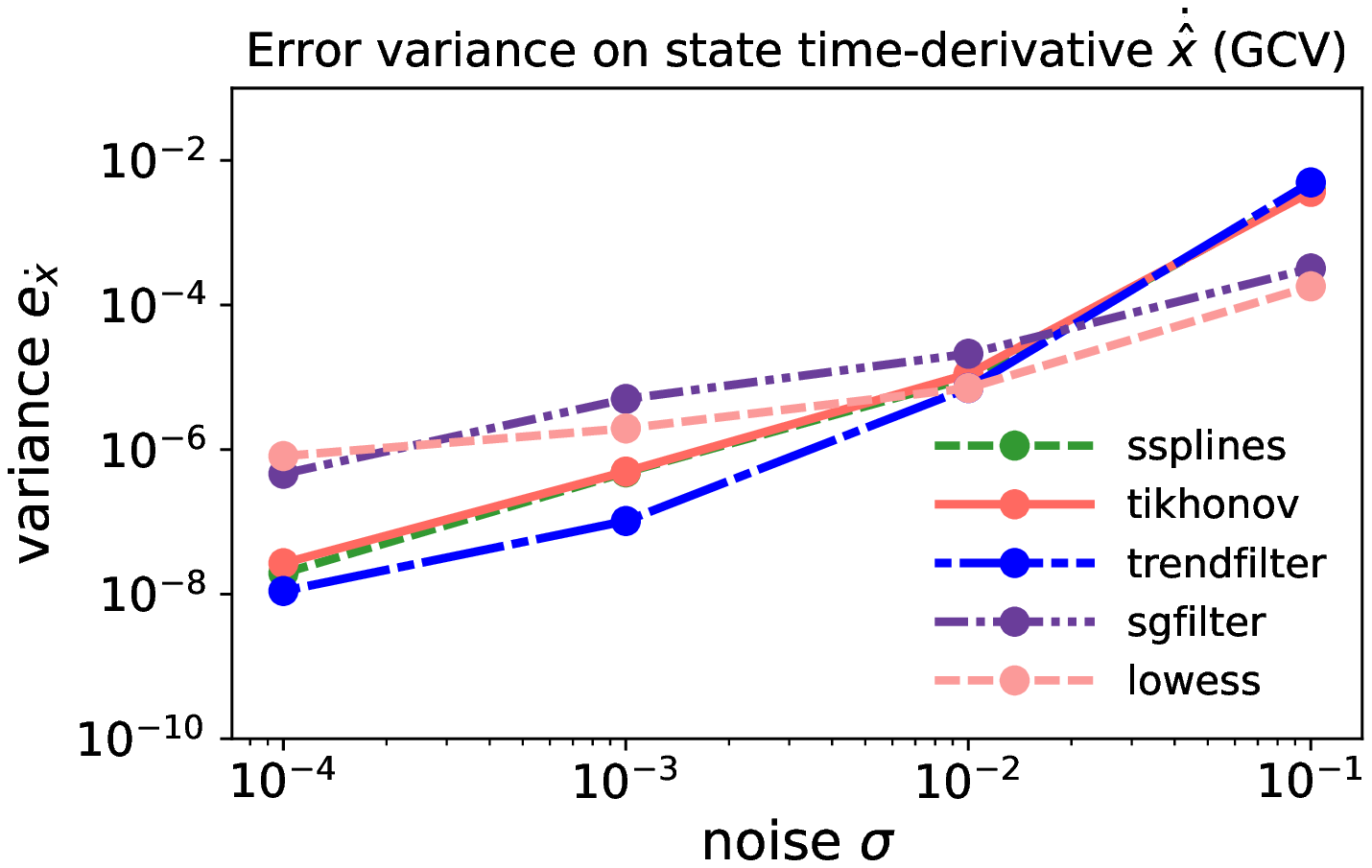}
		\includegraphics[trim = 0 0 00 0,
		clip,width=0.49\textwidth]{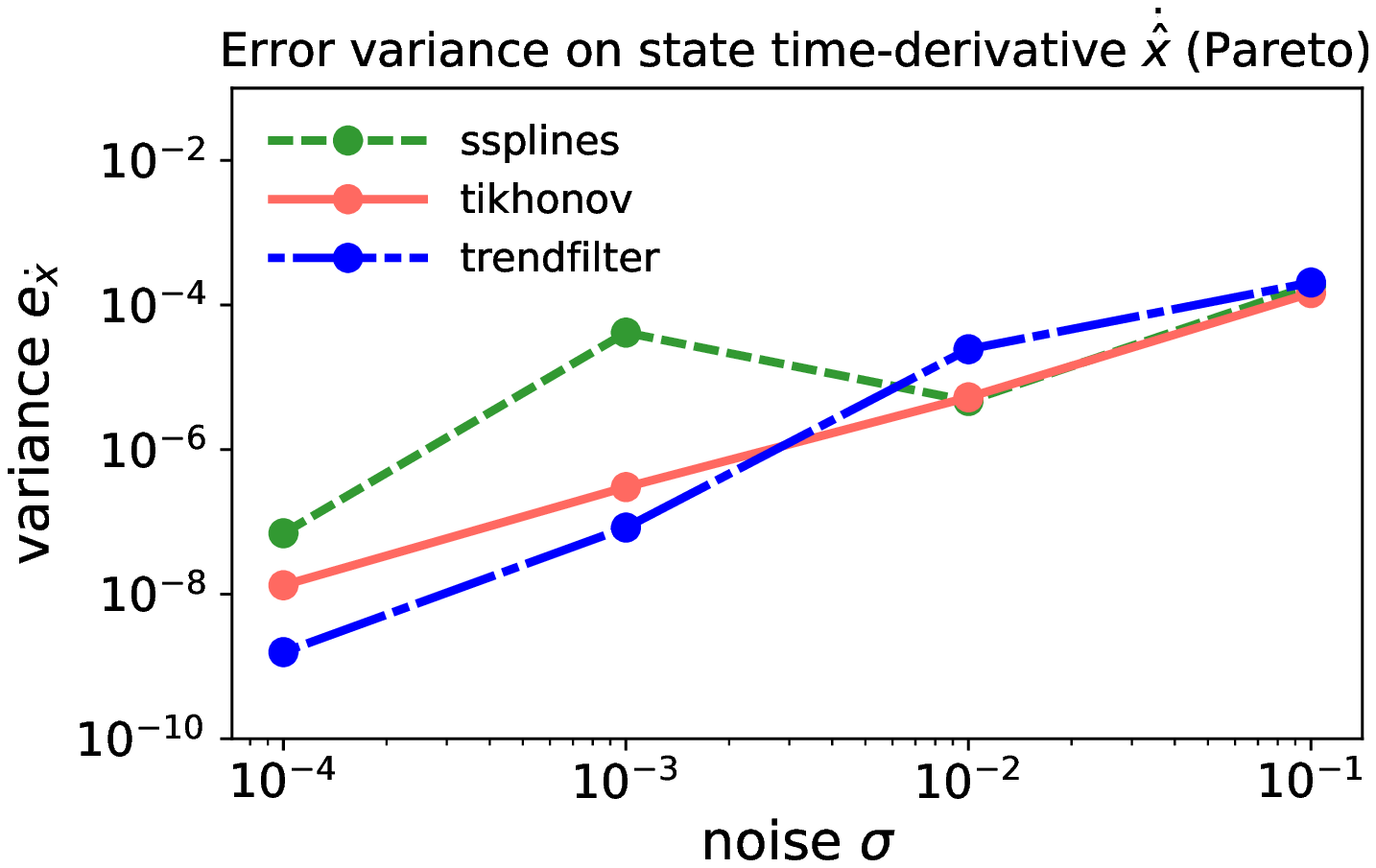}
		\caption{Variances of the relative state (top row) and state time-derivative (bottom row) errors for the Van der Pol using local and global smoothers. The regularization parameters were computed using GCV (left column) and the Pareto curve criterion (right column).}
		\label{fig:VanderPol_filter_comparison_var}
	\end{figure}
	\begin{figure}[H]
		\centering
		\includegraphics[trim = 0 0 0 0, clip,width=0.49\textwidth]{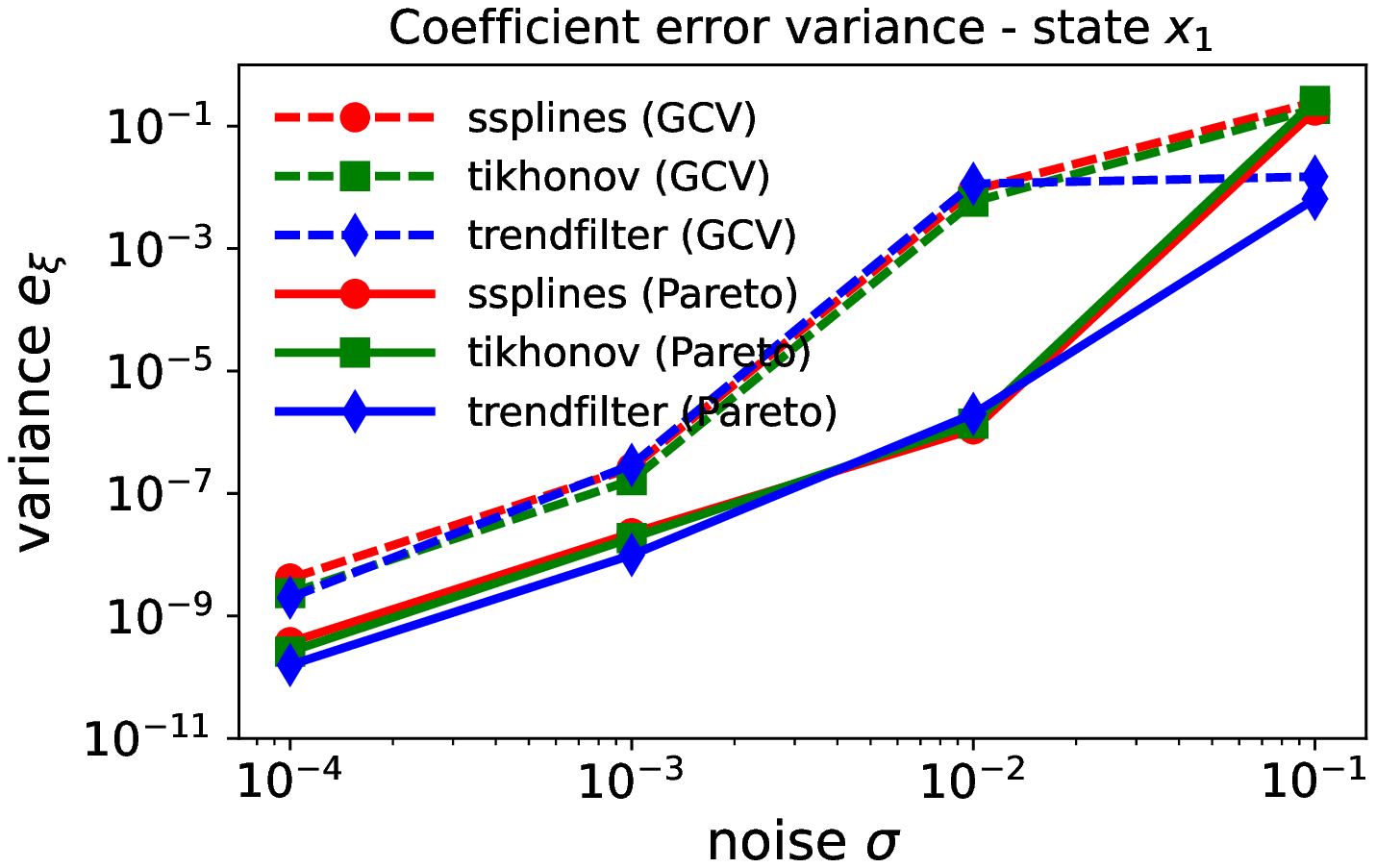}
		\includegraphics[trim = 0 0 0 0,
		clip,width=0.49\textwidth]{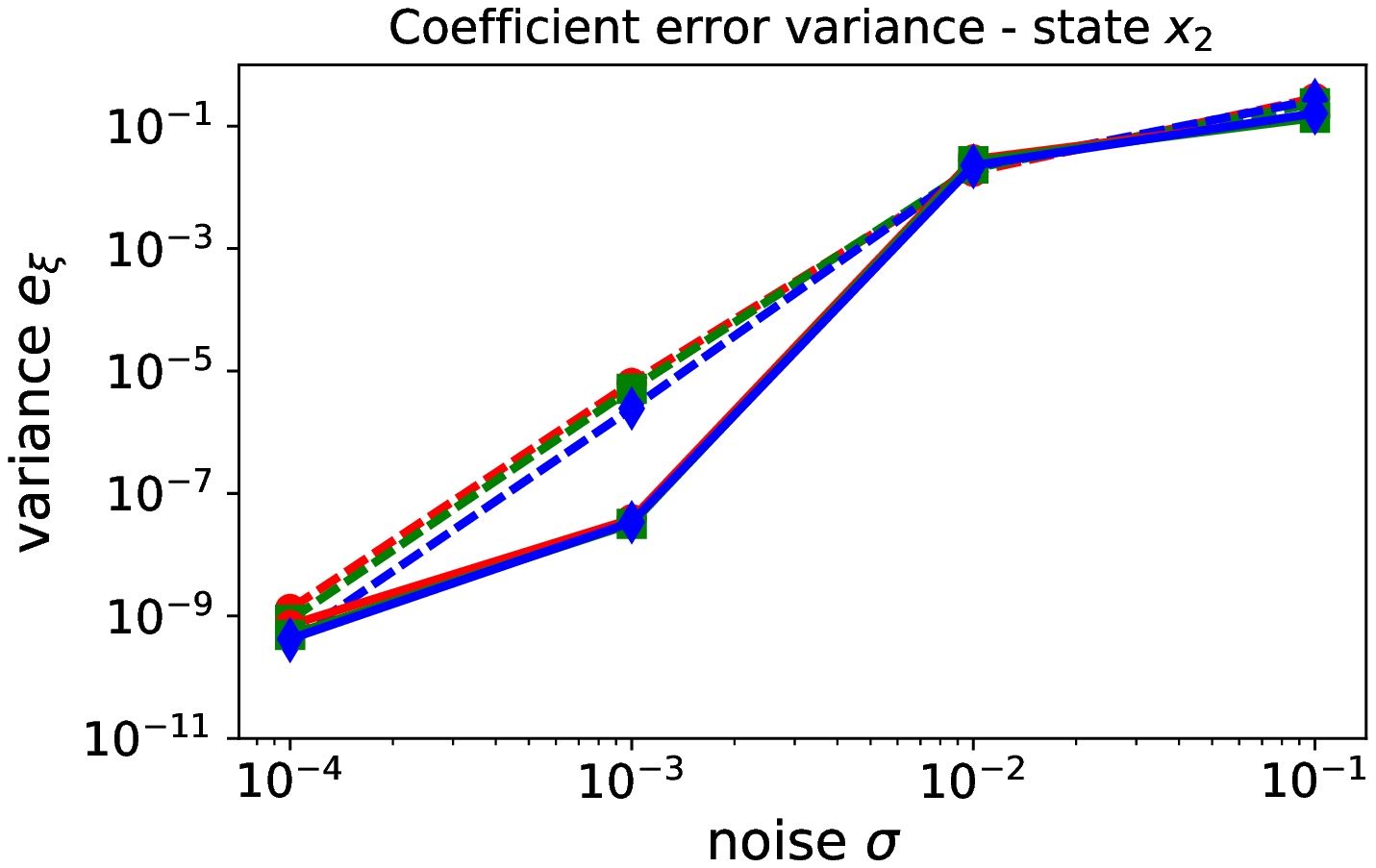}
		\caption{Variances of the relative coefficient errors for the Van der Pol oscillator using global smoothers and WBPDN. The regularization parameters were computed using GCV (left column) and the Pareto curve criterion (right column).}
		\label{fig:Vanderpol_WBPDN_comparison_var}
	\end{figure}


\section*{Appendix D: State-time derivative error performance with colored noise}
\label{app:D}

	\begin{figure}[H]
	\centering
	\includegraphics[trim = 10 0 0 0, clip,width=0.32\textwidth]{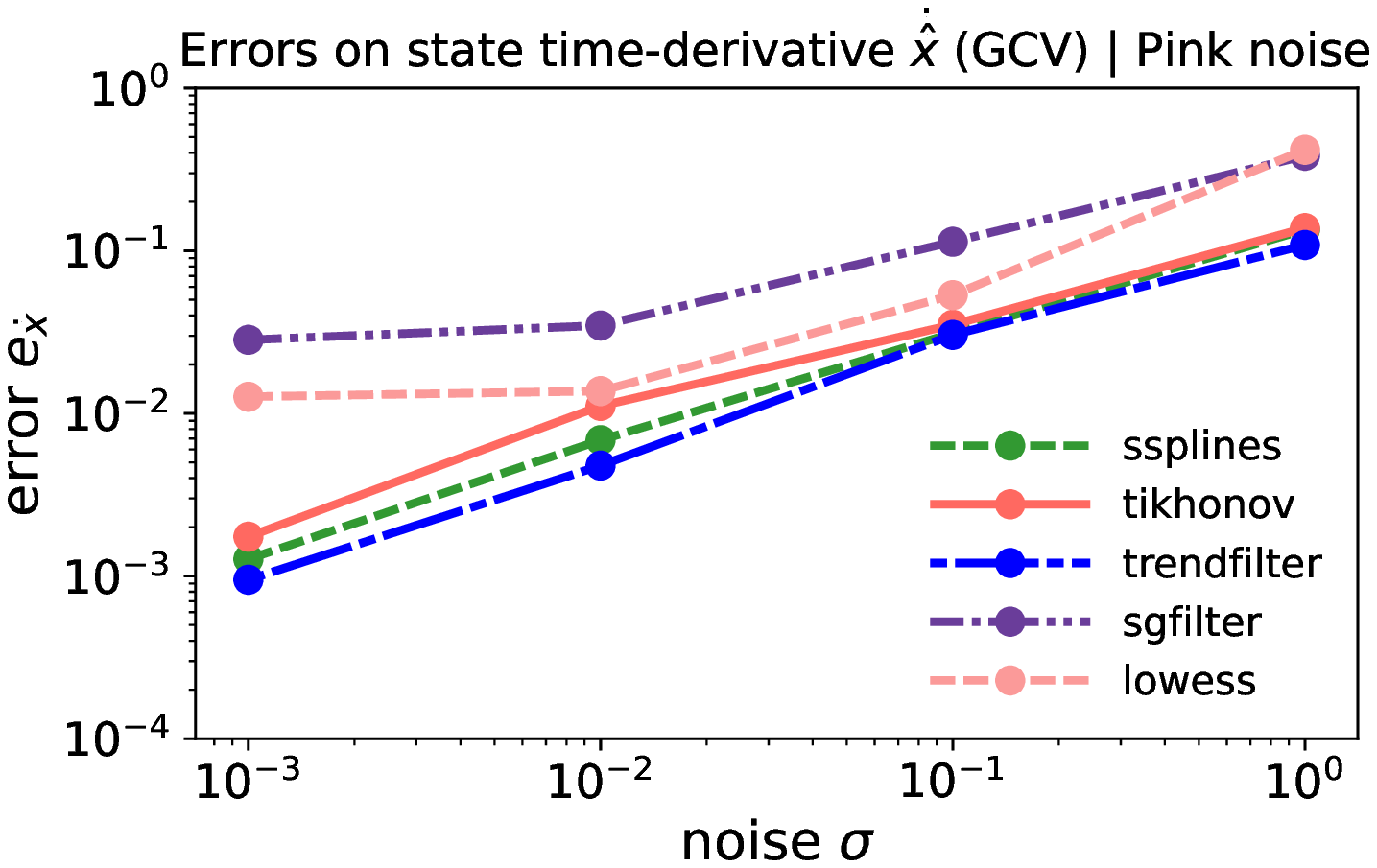}
	\includegraphics[trim = 10 0 0 0, clip,width=0.32\textwidth]{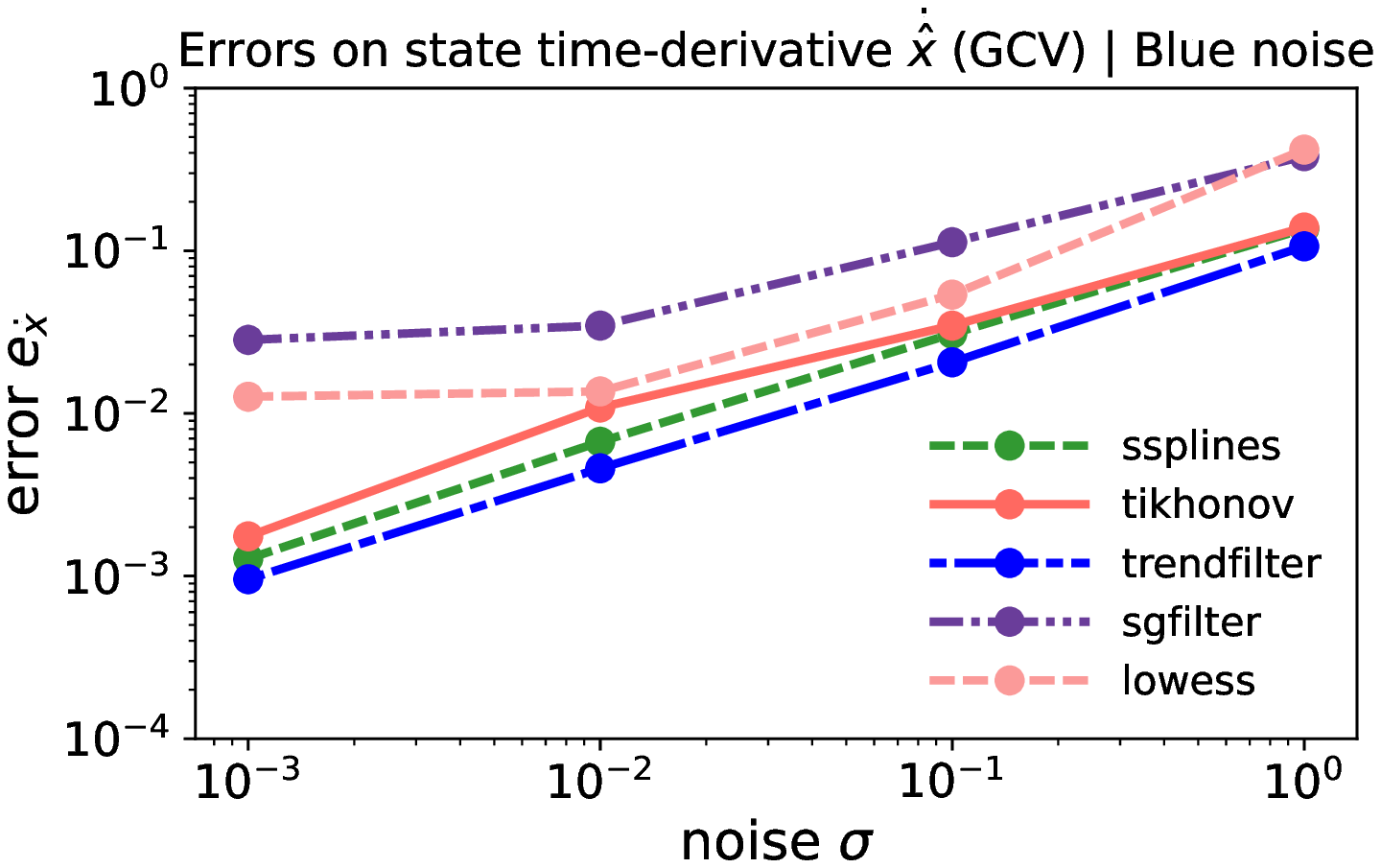}
	\includegraphics[trim = 10 0 0 0, clip,width=0.32\textwidth]{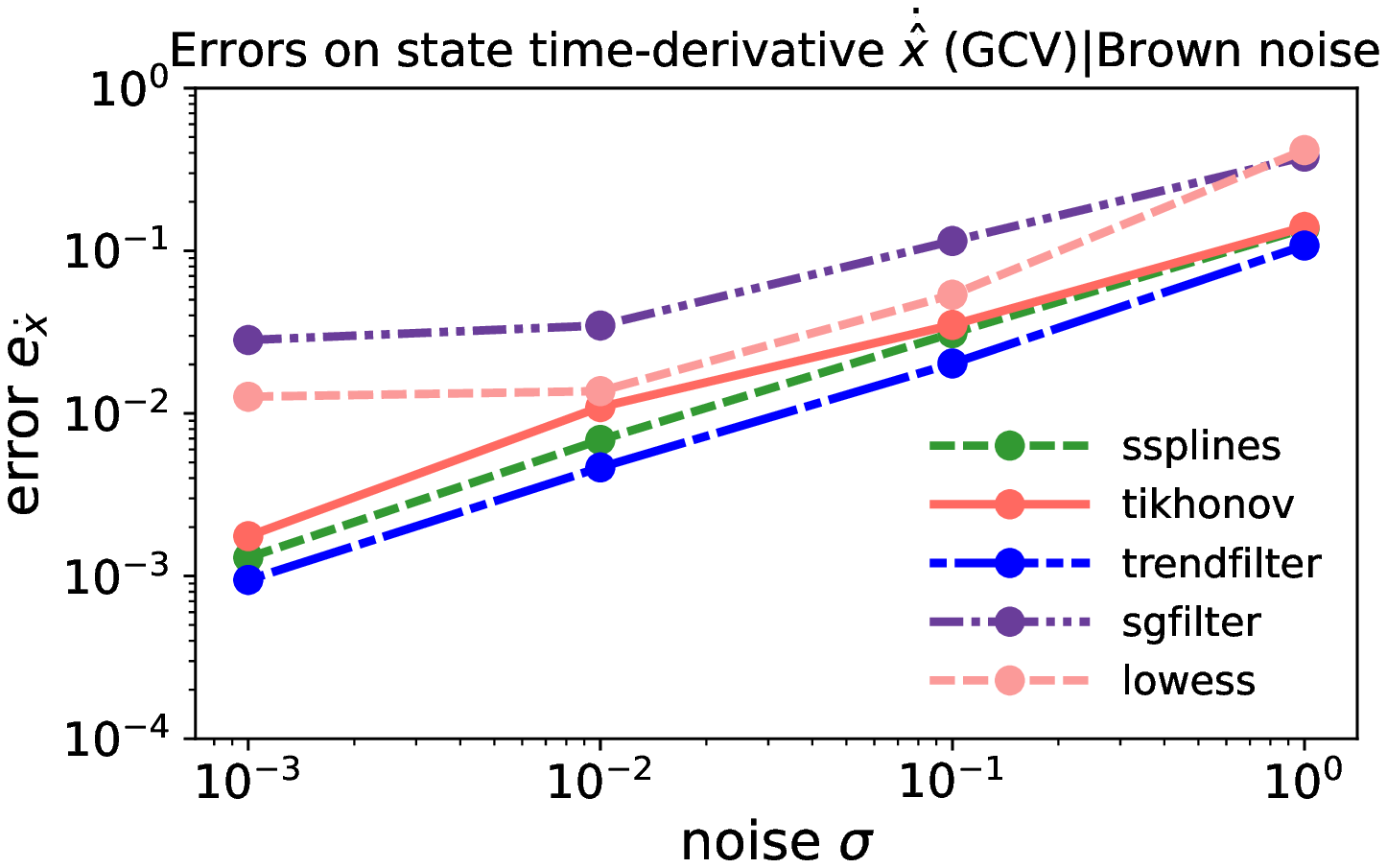}
	\includegraphics[trim = 10 0 0 0, clip,width=0.32\textwidth]{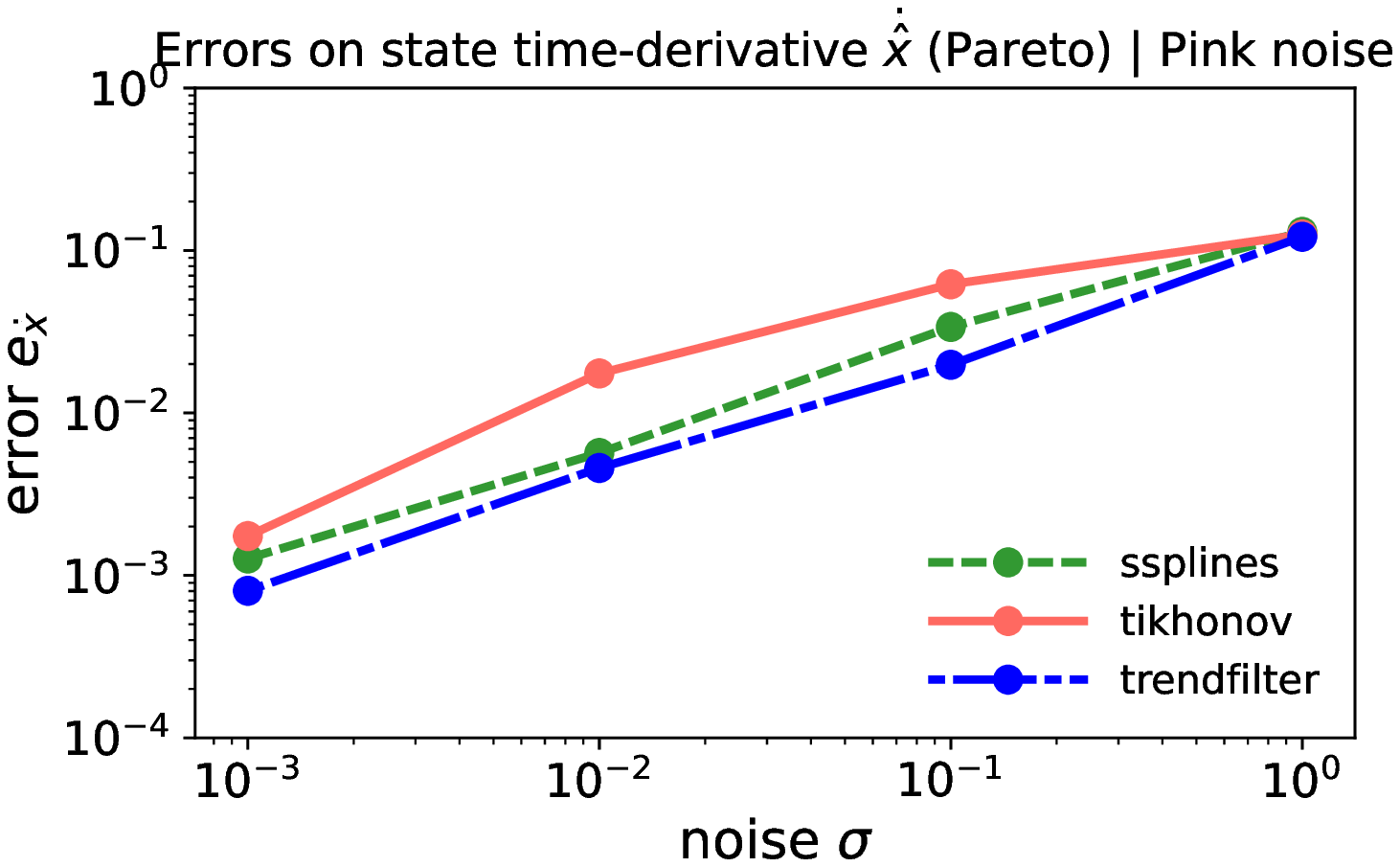}
	\includegraphics[trim = 10 0 0 0, clip,width=0.32\textwidth]{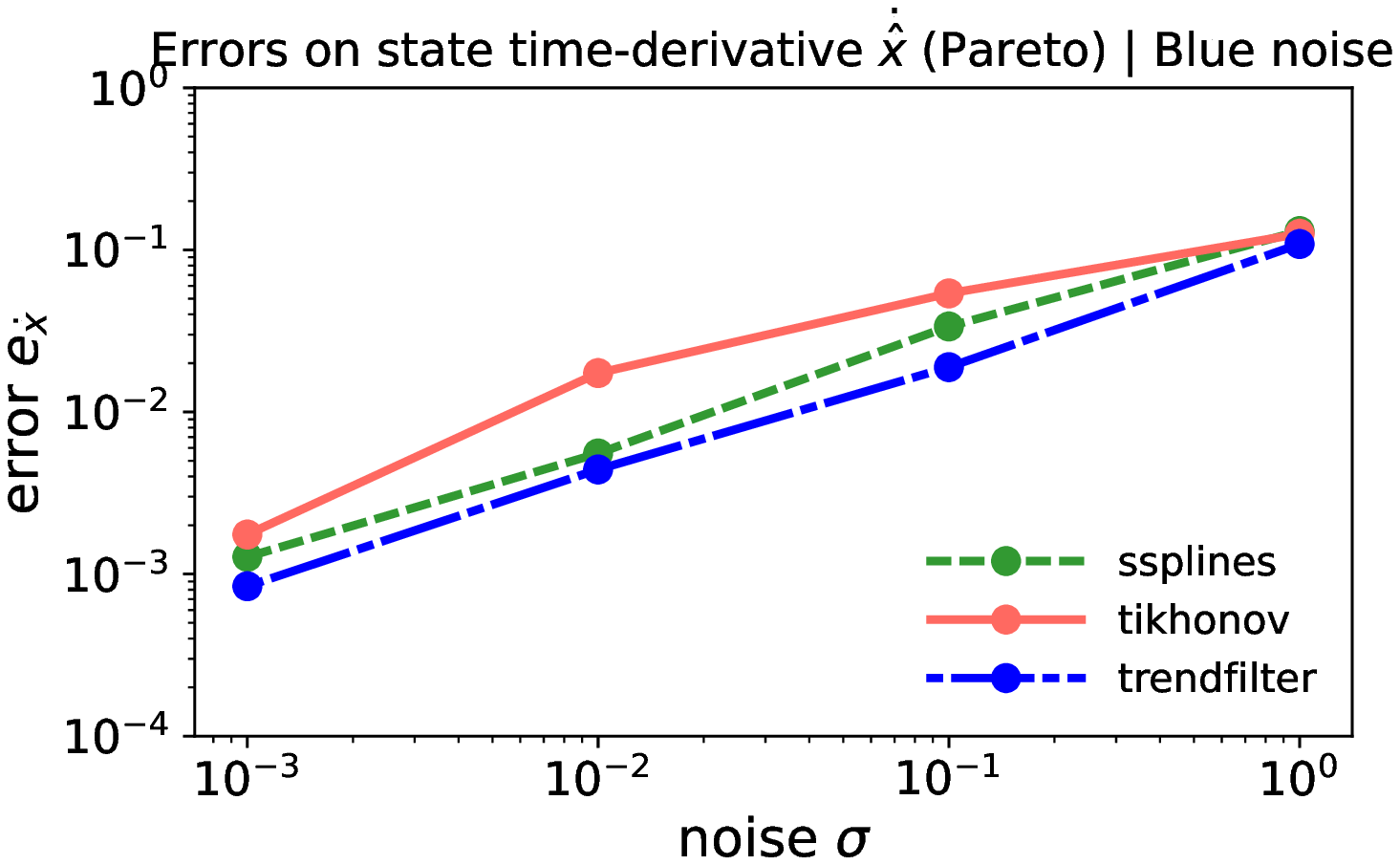}
	\includegraphics[trim = 10 0 0 0, clip,width=0.32\textwidth]{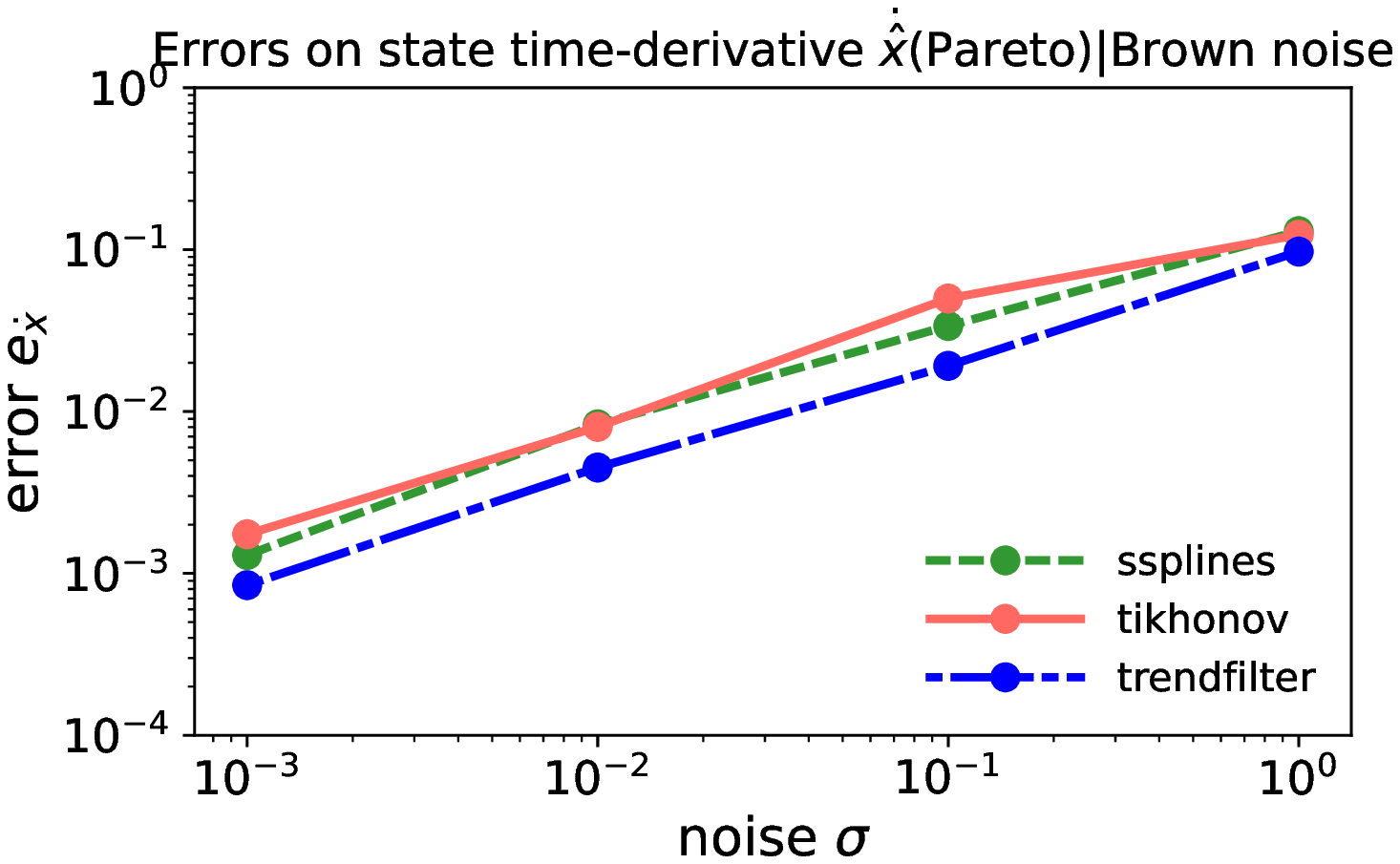}
	\caption{Comparison of the relative state time-derivative errors for the Lorenz 63 system using local and global smoothers with pink, blue and brown noise (from left to right). The regularization parameters were computed using GCV (top row) and the Pareto curve criterion (bottom row).}
	\label{fig:filter_comparison_dx_colored_noise}
\end{figure}

\end{document}